\pdfoutput=1
\documentclass[11pt]{article}

\usepackage{EMNLP2023}

\usepackage{booktabs}
\usepackage{multirow}
\usepackage{longtable}

\usepackage{float}
\usepackage{tabularx}

\usepackage{afterpage}
\usepackage{subcaption}

\usepackage{times}
\usepackage{latexsym}

\usepackage[T1]{fontenc}
\usepackage[utf8]{inputenc}

\usepackage{microtype}

\usepackage{inconsolata}

\usepackage{graphicx}

\title{Executing Natural Language-Described Algorithms with Large Language Models: An Investigation}

\author{Xin Zheng${}^{1,3}$, Qiming Zhu${}^{1,3}$, Hongyu Lin${}^{1}$, Yaojie Lu${}^{1}$, Xianpei Han${}^{1, 2}$, Le Sun${}^{1,2}$\thanks{~ Corresponding Authors}\\
${}^{1}$Chinese Information Processing Laboratory ~ ${}^{2}$State Key Laboratory of Computer Science \\
Institute of Software, Chinese Academy of Sciences, Beijing, China\\
${}^{3}$University of Chinese Academy of Sciences, Beijing, China \\
{\tt \{zhengxin2020,hongyu,luyaojie,xianpei,sunle\}@iscas.ac.cn} \\
{\tt zhuqiming23@mails.ucas.ac.cn} \\
}

\begin{document}
\maketitle
\begin{abstract}
Executing computer programs described in natural language has long been a pursuit of computer science. With the advent of enhanced natural language understanding capabilities exhibited by large language models (LLMs), the path toward this goal has been illuminated. In this paper, we seek to examine the capacity of present-day LLMs to comprehend and execute algorithms outlined in natural language. We established an algorithm test set sourced from \textit{Introduction to Algorithm}, a well-known textbook that contains many representative widely-used algorithms. To systematically assess LLMs’ code execution abilities, we selected 30 algorithms, generated 300 random-sampled instances in total, and evaluated whether popular LLMs can understand and execute these algorithms. Our findings reveal that LLMs, notably GPT-4, can effectively execute programs described in natural language, as long as no heavy numeric computation is involved. We believe our findings contribute to evaluating LLMs’ code execution abilities and would encourage further investigation and application for the computation power of LLMs. Our code and data are available at \url{https://github.com/MrZhengXin/natural_language_program}. 
\end{abstract}

\section{Introduction}

Algorithms are the main knowledge carriers for computing machines, which are conventionally outlined using high-level languages or even pseudocodes due to their communication effectiveness. The endeavor to create a mechanical system that could comprehend and execute natural language-described programs has long been a goal in computer science \citep{10.1145/365230.365274}.

\begin{table}[!t]
\small
\centering
\begin{tabular}{p{7cm}}
\toprule
\textbf{Prompt} \\ \midrule

The initial keyword is "{\color[HTML]{f368e0}art}". For each iteration, generate a sentence that MUST include the keyword, then select just a single word (rather than two or more words) from the generated sentence as the new keyword and print the new keyword. Iterate \textbf{5} times, with each iteration mark with its index, i.e. 1. 2. 3. etc. \\ 
\rowcolor[HTML]{C7EDCC} 1. {\color[HTML]{f368e0}Art} is a form of expression that can be seen in many different {\color[HTML]{ff9f43}mediums}. \\
\rowcolor[HTML]{C7EDCC} \textit{Keyword: {\color[HTML]{ff9f43}mediums}} \\
\rowcolor[HTML]{C7EDCC} 2. {\color[HTML]{ff9f43}Mediums} such as painting, {\color[HTML]{ee5253}sculpture}, and photography are popular forms of art. \\
\rowcolor[HTML]{C7EDCC} \textit{Keyword: {\color[HTML]{ee5253}sculpture}} \\
\rowcolor[HTML]{C7EDCC} 3. {\color[HTML]{ee5253}Sculpture} is a three-dimensional art form that can be made from a variety of {\color[HTML]{0abde3}materials}. \\
\rowcolor[HTML]{C7EDCC} \textit{Keyword: {\color[HTML]{0abde3}materials}} \\
\rowcolor[HTML]{C7EDCC} 4. {\color[HTML]{0abde3}Materials} such as clay, metal, and {\color[HTML]{341f97}wood} are often used in sculpture. \\
\rowcolor[HTML]{C7EDCC} \textit{Keyword: {\color[HTML]{341f97}wood}} \\
\rowcolor[HTML]{C7EDCC} 5. {\color[HTML]{341f97}Wood} carving is a popular form of sculpture that has been used for {\color[HTML]{8395a7}centuries}. \\
\rowcolor[HTML]{C7EDCC} \textit{Keyword: {\color[HTML]{8395a7}centuries}} \\ \bottomrule
\end{tabular}
\caption{The prompt for iterative sentence generation with keyword constraint and the response of Text-Davinci-003. According to the given instruction, the model faithfully executes the loop, repeatedly generating a text with the designated keyword and picking a new keyword. Different keywords are annotated with different colors.}
\label{table:prompt_keyword}
\end{table}

Recently, advancements in large language models (LLMs) have revolutionized the field of natural language processing and artificial Intelligence, creating new SOTAs and reaching human-level language understanding performance on a series of tasks and benchmarks \citep{NEURIPS2020_1457c0d6, DBLP:journals/corr/abs-2303-08774, anil2023palm}. LLMs, trained on extensive text corpora and code data, acquired world knowledge, commonsense and logical reasoning \citep{HAN2021225}. After the stage of instruction-tuning \citep{DBLP:conf/nips/Ouyang0JAWMZASR22}, LLMs could act consistently with complicated prompts. During this process, they perform the specific task according to what the instruction presents and return the desired output. As illustrated in Table \ref{table:prompt_keyword}, just as the prompt demands, the model repeatedly uses the keyword to generate a sentence, picks one word as the new keyword, and stops when satisfying the iteration count. These abilities are very analogous to the capabilities that are required to execute a program, which raises our interest in whether current LLMs could serve as an environment to execute natural language-described programs. 

\begin{table*}[t]
\centering
\small
\begin{tabular}{p{15.5cm}}
\toprule
\textbf{Prompt}                                                                                                                                        \\ \midrule
Execute the instructions step by step. Do not jump steps. Do not stop before completion.                                                               \\
Initial: Set the list of parentheses P: P{[}1{]} = '(' P{[}2{]} = '{]}' P{[}3{]} = '\}' P{[}4{]} = '(' .                                               \\
Set Stack\_1 = {[}{]}.                                                                                                                                 \\
Set i = 1.                                                                                                                                             \\
Step 1: What are the value of P{[}i{]} and Stack\_i ? Print them.                                                                                      \\
Step 2: What is the type of P{[}i{]} ? Classify it. Hint: '(' is the left parenthesis, '{[}' is the left parenthesis, '\{' is the left parenthesis. ')' is the right parenthesis, '{]}' is the right parenthesis, '\}' is the right parenthesis.                                      \\
\quad i. If P{[}i{]} is the left parenthesis: Step by step push Stack\_\{i+1\} as {[}(P{[}i{]}, i){]} + Stack\_i.                                            \\
\quad ii. If P{[}i{]} is the right parenthesis: Print Stack\_i{[}0{]}. Is Stack\_i{[}0{]} None? If Stack\_i{[}0{]} is not None, step by step pop Stack\_\{i+1\} as Stack\_i{[}1:{]}. Otherwise, print "Invalid" and halt. Question: Are Stack\_i{[}0{]}{[}0{]} and P{[}i{]} matched? Print Stack\_i{[}0{]}{[}0{]} and P{[}i{]}, apply the following rule, then answer: '(' and ')' matched, '{[}' and '{]}' matched, '\{' and '\}' matched. '(' and '{]}' not matched, '{[}' and ')' not matched, '\{' and '{]}' not matched; '(' and '\}' not matched, '{[}' and '\}' not matched, '\{' and ')' not matched.                                                     \\
\quad \quad a. If yes, continue.                                                                                                                                   \\
\quad \quad b. If no, print "Invalid" and halt.                                                                                                                    \\
Step 3: Increment i by 1. If i \textless{}= 4, go to Step 1. Otherwise, proceed to Step 4.                                                             \\
Step 4: What is the value of Stack\_5 ? Print it. What is the length of Stack\_5 ? Count it.                                                           \\
i. If the length is 0, print "Valid";                                                                                                                  \\
ii. If the length is greater than 0, print "Invalid".                                                                                                  \\
-----                                                                                                                                                  \\
The computation procedure is as follows, with each line numbered sequentially (e.g., 1. 2. 3. 4. 5. 6. etc.):                                          \\
1. i = 1, copy from Initial, P{[}1{]} = '('; Stack\_1 = {[}{]}.                                                                                        \\
2. The type of P{[}1{]} = '(' is the left parenthesis. Push Stack\_2 as {[}('(', 1){]} + Stack\_1 = {[}('(', 1){]} + {[}{]} = {[}('(', 1){]}.          \\ 
\rowcolor[HTML]{C7EDCC} 
3. Increment i by 1. i = 2.                                                                                                                            \\
\rowcolor[HTML]{C7EDCC} 
4. P{[}2{]} = '{]}'; Stack\_2 = {[}('(', 1){]}.                                                                                                        \\
\rowcolor[HTML]{C7EDCC} 
5. The type of P{[}2{]} = '{]}' is the right parenthesis. Print Stack\_2{[}0{]} = ('(', 1). Stack\_2{[}0{]} is not None, so pop Stack\_3               \\
\rowcolor[HTML]{C7EDCC} 
as Stack\_2{[}1:{]} = {[}{]}.                                                                                                                          \\
\rowcolor[HTML]{C7EDCC} 
6. Are Stack\_2{[}0{]}{[}0{]} and P{[}2{]} matched? Print Stack\_2{[}0{]}{[}0{]} = '(' and P{[}2{]} = '{]}'. '(' and '{]}' are not matched, so print \textit{"Invalid"} and {\color[HTML]{FF2400} halt}                                                                                                               \\ \bottomrule
\end{tabular}
\caption{The prompt of valid parentheses and the response of Text-Davinci-003. This task examines if a sequence of parentheses is matched or not. Once a parenthesis is unmatched, returns invalid. If in the end, the stack is empty, return valid, otherwise still returns invalid. In this example, the second element is not matched with the first element, so the model correctly returns invalid and halts the execution. The final result \textit{Invalid} is italicized, and the stopword {\color[HTML]{FF2400} halt} is marked red.}
\label{table:valid_parenthese}

\end{table*}

To run arbitrary algorithms, the ability to follow sequential, selection and iteration statements is needed \citep{bohm1966flow}. A model that does not rigorously support any of the sequential, selection and iteration structures, would fail at execution. Such a model is limited in computation power, and would not perform well on some real-world tasks.  In contrast, a model that successfully generates the correct output, is likely to well understand the three critical control flows. With the potential to conduct any computation within the context length theoretically, it is at least promising toward AGI. Therefore, the investigation of program execution could be beneficial for the understanding of LLMs.

However, except for limited preliminary studies \citep{DBLP:journals/corr/abs-2303-12712, DBLP:journals/corr/abs-2303-14310}, there is still a lack of quantitative and qualitative analysis experiments on whether LLMs can serve as effective program executor, as well as a standard benchmark to evaluate how much could a LLM accomplish the goal. Such absence limits our understanding of the latest cutting-edge research progress in this field. To this end, this paper investigates whether current large language models could execute natural language-described algorithms. To address this gap, we first establish an algorithm test set from the classical textbook \textit{Introduction to Algorithm} \citep{cormen2022introduction} for systematic evaluation of the program execution ability of LLMs. We \textbf{randomly} sampled \textit{10} distinct instances for each algorithm, which reduces the likelihood of data leakage. Based on the benchmark, we convert the algorithms together with the problem input into natural language description prompts, and input them into the LLMs, trying to see whether LLMs could execute the algorithm step by step accurately, and yield the correct result. Finally, we conduct experiments on 30 algorithms using three popular LLMs, namely Text-Davinci-003, GPT-3.5-Turbo \cite{DBLP:conf/nips/Ouyang0JAWMZASR22}, and GPT-4 \citep{DBLP:journals/corr/abs-2303-08774}, and systematically evaluate the ability of these LLMs as executors of natural language-described algorithms.

Our experiments reveal that the existing LLMs, especially GPT-4, can effectively run programs described in natural language. They can accurately follow the control flow of the algorithm as per the prompt description, precisely execute each step, and perform the calculation. Simultaneously, LLMs can maintain and update the values of variables consistently through the text output. This indicates that existing LLMs could execute statements of sequence, selection, and iteration, and mimic the core functions of the Von-Neumann Machine, including calculation, flow control, variable storage, and input-output understanding. As a result, it's unlikely that they are not the interpreters of natural language-described programs.

Our main contributions are as follows:

1. We build up a test set of algorithms from a widely used algorithm textbook, establishing a foundation for evaluating the program execution abilities of LLMs;

2. We construct natural language prompts for the algorithms;

\begin{table*}[t]
\small
\begin{tabular}{p{15.5cm}}
\toprule
\textbf{Prompt} \\ \midrule
Execute the instructions step by step. Do not jump steps. Do not stop before completion. \\ 
Initial: Set the input array A: A[0] = 4, A[1] = 5, A[2] = 7, A[3] = 8, A[4] = 13, A[5] = 14, A[6] = 15, A[7] = 17, A[8] = 18 . \\ 
Set the length of the array n: 9 . \\ 
Set the target element x: 18 . \\ 
Set the low index l to the first element of the array (l = 0) and the high index r to the last element (r = 8). \\ 
Step 1: Calculate the middle index mid. mid = l + (r - l) // 2. Print the values of l, r, and mid. \\ 
Step 2: Compare the element at the middle index A[mid] with the target element x. \\ 
\quad i. If A[mid] == x, return mid and halt. \\ 
\quad ii. If A[mid] < x, set l = mid + 1 and proceed to Step 3. \\ 
\quad iii. If A[mid] > x, set r = mid - 1 and proceed to Step 3. \\ 
Step 3: Check if the search space is exhausted (l <= r). \\ 
\quad i. If l <= r, go back to Step 1. \\ 
\quad ii. If l > r, return -1 and halt. \\ 
----- \\ 
The computation procedure is as follows, with each line numbered sequentially (e.g., 1. 2. 3. 4. 5. 6. etc.): \\ 
1. l = 0, r = 8, mid = 0 + (8 - 0) // 2 = 4. \\ 
\rowcolor[HTML]{C7EDCC} 2. A[mid] = A[4] = 13, which is less than x (18), so set l = mid + 1 = 4 + 1 = 5. \\ 

\rowcolor[HTML]{C7EDCC} 3. l (5) is less than or equal to r (8), so go back to Step 1. \\ 

\rowcolor[HTML]{C7EDCC} 4. l = 5, r = 8, mid = 5 + (8 - 5) // 2 = 6. \\ 

\rowcolor[HTML]{C7EDCC} 5. A[mid] = A[6] = 15, which is less than x (18), so set l = mid + 1 = 6 + 1 = 7. \\ 

\rowcolor[HTML]{C7EDCC} 6. l (7) is less than or equal to r (8), so go back to Step 1. \\ 

\rowcolor[HTML]{C7EDCC} 7. l = 7, r = 8, mid = 7 + (8 - 7) // 2 = 7. \\ 

\rowcolor[HTML]{C7EDCC} 8. A[mid] = A[7] = 17, which is less than x (18), so set l = mid + 1 = 7 + 1 = 8. \\ 

\rowcolor[HTML]{C7EDCC} 9. l (8) is less than or equal to r (8), so go back to Step 1. \\ 

\rowcolor[HTML]{C7EDCC} 10. l = 8, r = 8, mid = 8 + (8 - 8) // 2 = 8. \\ 

\rowcolor[HTML]{C7EDCC} 11. A[mid] = A[8] = 18, which is equal to x (18), so return mid (8) and halt.
\\ \bottomrule
\end{tabular}
\caption{The prompt of binary search and the response of GPT-4. Binary search is a search algorithm used to find the position of a key in a sorted array. It works by repeatedly dividing the search interval in half until the target value is found or determined to be not present. The search interval is defined by two indices, the lower bound and the upper bound which are adjusted on each iteration. In this example, GPT-4 correctly found the position of element ``18'', which is at the 8th position.}
\label{table:clrs_mini/binary_search}

\end{table*}

3. We test a series of algorithms on current state-of-the-art LLMs and systematically evaluate their abilities as natural language program interpreters.

We anticipate that the results presented in this research will stimulate further interest and research into the computation power of large language models. we are hopeful for further breakthroughs that will contribute positively to various domains of artificial intelligence.

\section{Algorithm Prompting}
% \section{Algorithm Prompting: Evaluating Zero-shot program execution ability of LLMs using Classical Algorithms}
% In this section, we will describe how we build the algorithm test suite for LLMs. This includes selecting appropriate algorithms (Section \ref{subsec:Algorithm Selection}), designing the correct prompts (Section \ref{subsec:Algorithm Prompt Design}), and generating test cases (Section \ref{subsec:Test Case Generation}).

\subsection{Algorithm Selection}
\label{subsec:Algorithm Selection}
Guided by previous work \citep{pmlr-v162-velickovic22a}, we choose algorithms from the widely-used textbook for algorithm courses, \textit{Introduction to Algorithms} \citep{cormen2022introduction}, listed in Section \ref{sec:testing_algorithms}. We first pick 26 representative algorithms, forming the evaluation set \textbf{CLRS-mini}. These algorithm implementations involve sequence, selection, and iteration control flows, nested loops, and recursive calls, which could effectively evaluate the ability of LLMs to execute programs. They all have the polynomial time complexity and only involve integer/float addition and integer multiplication, so we expect the current SOTA LLM would conduct these tasks well. To further challenge the current LLMs, we additionally formulate another evaluation set \textbf{CLRS-Numeric}, which consists of 4 numeric-operation-intensive algorithms and requires floating-point multiplication/division and calculation of exponential and trigonometric functions. While today's LLM alone may not be able to solve them, we believe the aid of external tools such as Python Interpreter may be beneficial, and the performance of future LLMs on float operations remains to be seen.
% For \textbf{CLRS-mini}, we exclude the geometry algorithms, hypothesizing that floating-point multiplication and division are too challenging for LLMs without the aid of external tools. 

\subsection{Algorithm Prompt Design}
\label{subsec:Algorithm Prompt Design}
In our design of program prompting, the aim is to create a prompt structure that is both rigorous and easy to interpret. Emphasizing precision, each task-specific instruction was written in clear, unambiguous natural language. As illustrated in Table \ref{table:valid_parenthese} and Table \ref{table:clrs_mini/binary_search}, we employed "goto" statements to trigger iterative behaviors and use natural language to express if/else branch selection, with distinct branches denoted by index markers such as "i." and "ii." and Python-style spaces indent.

To facilitate stepwise parsing and execution, we asked the model to mark each line of the procedure with a sequential index, serving as a delimiter. This setup helped distinguish the current computation step that the model generated from the previously completed steps. We also included the first computation step within the prompt to ensure that the model was forced to execute the instructions rather than merely rephrasing them.

Inspired by the work of \citet{DBLP:journals/corr/abs-2303-14310}, encouraging the model to think step by step as much as possible is also crucial. Rather than using typical human expressions that place the final result before the reasoning (e.g., ``\texttt{Yes, '(' and ')' match}''), we opted for a reasoning-first approach (e.g., ``\texttt{Are Stack\_2[0][0] and P[2] matched? Print Stack\_2[0][0] = '(' and P[2] = ']'. '(' and ']' are not matched}''). 
To force the model actually make the comparison between two values and overcome the attempts of guessing and hallucination, we can tell the model to subtract the two values first, then check the sign of the result, which is exactly what the CPU actually performs. For example,
%in Table \ref{table:clrs_mini/articulation_points}, 
instead of high-level statement ``\texttt{v - pi[u]}'', we ask the model to ``\texttt{Calculate bne\_v\_pi\_u = v - pi[u] and present the result.}''.
It's also worth noting that fetching the value of an array at a specific index can be non-trivial since it includes addressing operation, so if the list is constant, we explicitly express the value at each position, for example, `` \texttt{P[1] = '('} '', rather than simply states `` \texttt{P=['(', ']', '\}', '(']} '', which requires more computation in finding the needed value. Similarly, in the instruction we replace the constant variables with their actual value. For example, instead of ``\texttt{i < n}'', where the value of n is 4, we explicitly state ``\texttt{i < 4}''.

Similar to the prior work \citep{DBLP:journals/corr/abs-2303-14310}, to ensure the alignment between the words used in the prompts and the actions they represented, in the task of valid parentheses, we set ``halt'' as the stopword to prevent the model (especially GPT 3.5) from continuing generation after detecting the error. Furthermore, we took measures to prevent the model from skipping steps as the output became longer and repetitive. We prohibited the use of words such as ``...'', ``Repeat'', or ``Continue'', which could lead to overlooked or incomplete steps, and result in the wrong answer. The example prompts are presented at Table \ref{table:valid_parenthese} and \ref{table:clrs_mini/binary_search}.
% The prompts are presented at Table \ref{table:valid_parenthese}, \ref{table:clrs_mini/binary_search}, and Appendix \ref{sec:appendix}. 

Our method of prompting is different from that of \citet{liu2023code}. Ours requires few operations defining and mainly relies on the actions that natural language commonly represents. Moreover, since the semantics of natural language goes far beyond formal language, we may easily express complex tasks like ``generate a sentence'' in a zero-shot manner, in which traditional programming languages are struggling
%(See Appendix \ref{appendix:Evaluation of Iterative Sentence Generation with Keyword Constraint} for a preliminary evaluation). 
Leveraging the power of instruction-following \cite{DBLP:conf/nips/Ouyang0JAWMZASR22}, we argue that without concrete examples and repetitive deletion of previous history context \citep{DBLP:journals/corr/abs-2303-14310}, it's still possible to trigger GPT into a computation device given the program alone. 
On the other hand, pure programming language prompt leads to low accuracy \citep{DBLP:journals/corr/abs-2303-14310}. This is because GPT may not always parse the program correctly as the real interpreter or compiler does, or be triggered to think step by step without "jumping to conclusions``. 
However, by crafting with more detail and clarity, our natural language prompt offers better performance. % Yet compared with the pure programming language prompt \citep{DBLP:journals/corr/abs-2303-12712, DBLP:journals/corr/abs-2303-14310}, which leads to low accuracy, since GPT may not necessarily have the same precise program interpreter or compiler, our crafted natural language prompt is more detailed and clear, yielding superior performance.

\subsection{Test Case Generation}
\label{subsec:Test Case Generation}
Unlike previous work \citep{pmlr-v162-velickovic22a}, which used a problem size of 16 for training and validation and 64 for testing, we adjusted ours to be smaller. This adjustment was due to concerns about the context-length limit, generation time, and inference cost. Generally, we set the problem size to 9 and 10 for tasks that only need a single iteration, and 4 and 5, or even smaller, for more complex tasks that require a long generation length. For algorithms that require sorting, we pre-sorted the input data to save the length of instruction and generation. And since we would select various sorting algorithms for testing, this simplification would not reduce the diversity of our evaluation. For each task, we randomly sample 10 instances. Just as in the case of algorithm competitions, we set the final output of the algorithm as the gold answer, and consider the prediction correct if the value presented in the last line is exactly the same. For the tasks in \textbf{CLRS-Numeric}, we allow an absolute tolerance of 0.1.

\section{Experiments}

\subsection{Setup}
For language model, we use GPT-3.5 \citep{DBLP:conf/nips/Ouyang0JAWMZASR22} \texttt{text-davinci-003} and \texttt{gpt-3.5-turbo-0301} versions with 4k context window, and GPT-4 \citep{DBLP:journals/corr/abs-2303-08774} \texttt{gpt-4-0314} version with 8k context window. They are accessed via OpenAI API. The temperature is consistently set to 0. All results are from a single run.

We also set a baseline of Python Code, which only replaces the step-by-step natural instructions of the algorithm with the corresponding program, while the input data is unchanged
%, and Table \ref{table:python_clrs_mini_program/task_scheduling} serves as an example
. This can help to investigate the effectiveness of our proposed design.

\subsection{Testing algorithms}
\label{sec:testing_algorithms}
For \textbf{CLRS-mini}, we select 26 classical algorithms, namely insertion sort, bubble sort, heapsort \citep{williams1964algorithm}, quicksort \citep{10.1145/366622.366644}, minimum search, binary search, quickselect \citep{10.1145/366622.366647}, maximum subarray \citep{10.1145/358234.381162}, activity selection \citep{doi:10.1137/0201013}, task scheduling \citep{lawler1985traveling}, matrix chain multiplication, longest common subsequence, optimal binary search tree \citep{10.5555/578775}, depth-first search \citep{moore1959shortest}, breadth-first search \citep{moore1959shortest}, topological sorting \citep{knuth1973fundamental}, articulation points, bridges, Kosaraju's strongly connected components \citep{10.5555/578775}, Kruskal's minimum spanning tree \citep{fc0df122-3305-33a8-bac5-7a6fc3666dfb}, Prim's minimum spanning tree \citep{6773228}, Bellman-Ford algorithm for single-source shortest paths \citep{bellman1958routing}, Dijkstra's algorithm for single-source shortest paths \citep{10.1145/3544585.3544600}, Floyd-Warshall algorithm for all-pairs shortest-paths \citep{10.1145/367766.368168}, naive string matching, and Knuth-Morris-Pratt string matcher \citep{colussi1994fastest}. For \textbf{CLRS-Numeric}, we select 4 algorithms emphasizing the arithmetic operations, including Least Square Regression, Discrete Fourier Transform, and two convex hull algorithms of Graham Scan \citep{GRAHAM1972132} and Jarvis March \citep{JARVIS197318}. The time complexity and problem size of each task would be in the appendix once published.
% Table \ref{table:clrs_mini_problem size}. 

Apart from the above ones, we also included two tasks, \textbf{valid parentheses} and \textbf{longest common subsequence (short)}, for comparison with previous work \citep{DBLP:journals/corr/abs-2303-14310}. Both of which come from BIG-bench cs-algorithms category \citep{DBLP:journals/corr/abs-2206-04615}. The task of valid parentheses is to verify if a sequence of parentheses consisting of three different types is balanced or not, which requires stack manipulation, and it has \textbf{1,000} test instances with the maximum length of 20. As for longest common subsequence, the goal is to compute the length of longest common subsequence given two sequences, and two nested loops are needed to complete the task. Due to the context-length issue, \citet{DBLP:journals/corr/abs-2303-14310} limited the maximum length to 6 and constructed a new test set of \textbf{100} instances.

\begin{table}[!h]
\small
\centering
\begin{tabularx}{0.47\textwidth}{@{}ll@{}}
\toprule
\textbf{Model} & \textbf{Acc (\%)} \\ \midrule
Random & 50.0 \\
GPT-3, few shot \citep{DBLP:journals/corr/abs-2206-04615} & 57.8 \\
PALM 2, few shot \citep{anil2023palm} & 83.4 \\
IRSA \citep{DBLP:journals/corr/abs-2303-14310} & 96.0 \\ \midrule
GPT-3.5-Turbo & 66.0 \\
\textbf{Text-Davinci-003} & \textbf{100} \\
\textbf{GPT-4} & \textbf{100} \\ \bottomrule
\end{tabularx}
\caption{Results of valid parentheses.}
\label{table:result_valid_parentheses}
\end{table}

\begin{table}[!h]
\centering
\small
\begin{tabularx}{0.47\textwidth}{@{}ll@{}}
\toprule
\textbf{Model} & \textbf{Acc (\%)} \\ \midrule
Random & 44 \\
GPT-3, few shot \citep{DBLP:journals/corr/abs-2303-14310} & 7 \\
IRSA \citep{DBLP:journals/corr/abs-2303-14310} & 93 \\
GPT-4, code exec \citep{DBLP:journals/corr/abs-2303-14310} & 69 \\ \midrule
GPT-3.5-Turbo & 38 \\
Text-Davinci-003 & 71 \\
\textbf{GPT-4} & \textbf{100} \\ \bottomrule
\end{tabularx}
\caption{Results of longest common subsequence.}
\label{table:result_lcs}
\end{table}

\begin{table*}[!t]
\centering
\small
\begin{tabular}{@{}lllllll@{}}
\toprule
\multicolumn{1}{c}{} & \multicolumn{3}{c}{\textbf{Natural Language Prompt (Ours)}} & \multicolumn{3}{c}{\textbf{Python Code}} \\ \cmidrule(l){2-7} 
\multicolumn{1}{c}{\multirow{-2}{*}{\textbf{Task}}} & GPT-3.5-T & Davinci-003 & GPT-4 & GPT-3.5-T & Davinci-003 & GPT-4 \\ \midrule
\textit{Sorting} &  &  &  &  &  &  \\
Insertion Sort & 50 & 80 & \textbf{100} & 80 & 70 & \textbf{100} \\
Bubble Sort & 60 & 70 & \textbf{100} & \textbf{100} & 0 & \textbf{100} \\
Heapsort & 90 & 20 & \textbf{100} & 60 & 70 & 50 \\
Quicksort & 70 & \textbf{100} & \textbf{100} & 90 & 80 & \textbf{100} \\ \midrule
\textit{Searching} &  &  & \textbf{} &  &  &  \\
Minimum & 90 & 60 & \textbf{100} & 30 & 20 & 70 \\
Binary Search & 90 & \textbf{100} & \textbf{100} & 90 & 70 & \textbf{100} \\
Quick Select & 50 & 70 & \textbf{100} & 30 & 40 & 60 \\ \midrule
\textit{Strings} &  &  & \textbf{} &  &  &  \\
Naive String Matching & 90 & 80 & \textbf{100} & 20 & 50 & \textbf{100} \\
Knuth-Morris-Pratt & 30 & 10 & \textbf{100} & 20 & 0 & 80 \\ \midrule
\textit{Divide and Conquer} &  &  & \textbf{} &  &  &  \\
Maximum   Subarray & 40 & 0 & \textbf{100} & 30 & 40 & {\color[HTML]{1F2329} 20} \\ \midrule
\textit{Greedy} &  &  & \textbf{} &  &  &  \\
Activity selection & 0 & 0 & \textbf{100} & 10 & 0 & \textbf{100} \\
Task scheduling & 40 & 50 & \textbf{100} & 60 & 10 & 80 \\ \midrule
\textit{Dynamic programming} &  &  & \textbf{} &  &  &  \\
Matrix Chain Multiplication & 30 & 10 & \textbf{100} & 20 & 0 & 50 \\
Longest Common Subsequence & 30 & 60 & \textbf{100} & 20 & 50 & {\color[HTML]{1F2329} 70} \\
Optimal Binary Search Tree & 0 & 10 & \textbf{100} & 0 & 0 & 20 \\ \midrule
\textit{Graphs} &  &  & \textbf{} &  &  &  \\
Depth-First Search & 0 & 0 & \textbf{100} & 0 & 0 & 60 \\
Breadth-First Search & 10 & 0 & \textbf{100} & 10 & 0 & 80 \\
Topological Sorting & 10 & 10 & \textbf{100} & 0 & 10 & 20 \\
Articulation Points & 0 & 0 & \textbf{100} & 0 & 0 & 30 \\
Bridges & 20 & 20 & \textbf{100} & 20 & 20 & 50 \\
Strongly Connected Components & 0 & 0 & \textbf{100} & 0 & 0 & 0 \\
Kruskal's MST & 50 & 60 & \textbf{100} & 20 & 40 & 70 \\
Prim's MST & 10 & 0 & \textbf{100} & 0 & 0 & 80 \\
Bellman-Ford & 20 & 0 & \textbf{100} & 0 & 0 & \textbf{100} \\
Dijkstra & 0 & 0 & \textbf{100} & 0 & 10 & 90 \\
Floyd-Warshall & 0 & \textbf{100} & \textbf{100} & 0 & 0 & 0 \\
\hline
Average & 35.0 & 36.2 & \textbf{100.0} & 27.3 & 25.4 & 65.4 \\ \bottomrule
\end{tabular}
\caption{Results of CLRS-mini.}
\label{table:clrs}
\end{table*}

\section{Results}

\subsection{Previous Tasks}
Table \ref{table:result_valid_parentheses} presents the result of the valid parentheses task. Previously, \citet{DBLP:journals/corr/abs-2303-14310} proposed the method of IRSA. They leveraged prompts made of similar operation procedure examples, rather than instruction, to trigger the execution, and for post-processing, they delete the computation process and save only the final state once the model completes an iteration. For LLM, they chose GPT-3 \texttt{Code-Davinci-002} version, claiming it provides similar results but has lower cost compared with \texttt{Text-Davinci-002} or \texttt{Text-Davinci-003}. IRSA achieved 96\% accuracy, which is already impressive. But with our natural language program prompting, both Text-Davinci-003 and GPT-4 models reached 100\% accuracy, demonstrating the effectiveness of our method in perfectly solving this task. 
% Notice that with the same prompt, GPT-3.5-Turbo got only 66\% accuracy, which suggests the weaker capability of the model. 

As shown in Table \ref{table:result_lcs}, the GPT-4 model was the only model to achieve perfect accuracy in the longest common subsequence task. 
% The Text-Davinci-003 model showed a substantial improvement over GPT-3.5-Turbo, reaching 73\% accuracy as opposed to the latter's 38\%. LCS, which require nested loop, branching, array fetching and updating, challenges these two models. 
Although with regimenting attention, which is to delete useless history context, IRSA method scored a high accuracy of 93\%, it is clear that the GPT-4 model demonstrated a stronger ability to handle this complex task.

\subsection{CLRS-mini}
We observe that GPT-4 demonstrates exceptional performance in comparison to the GPT-3.5 models, achieving an impressive 100\% accuracy across all tasks. This illustrates its outstanding capacity for precise program execution and implies a significant enhancement in algorithm execution when compared to its predecessors. On average, Text-Davinci-003 (36.9\%) performs marginally better than GPT-3.5-Turbo (35.0\%), but both still fall significantly behind GPT-4.

Partly due to the 4k context length limit, the GPT-3.5 models yielded a score of 0 in numerous graph algorithms, as these tasks require more tokens to complete. As the complexity of the tasks increases, the instruction encompasses more information, the control flow grows more intricate, and the required number of generation tokens also rises. Consequently, the performance of both GPT-3.5 models tends to decline. However, GPT-4 successfully manages to tackle these intricate tasks, highlighting its capabilities in simulating natural language programs.

Compared with detailed instruction, under Python Code only, the average performance of all three models declines. Especially for GPT-4, only in relatively simple algorithms the model can get good accuracy, but as the complexity increases, the results drop. This echoes the previous findings that GPT-4 may not faithfully execute the program, and the intuition is that traditionally in computer science, the high-level Python code itself would require an interpreter to be analyzed first, for example, the transition between code lines needs to be determined.

\subsection{CLRS-Numeric}
Intriguingly, across all three evaluated models, a uniform performance result of 0\% was observed, illuminating a pronounced and universal challenge encountered by current LLMs in handling such complex numerical operations. Inspecting the generation results, we found that GPT-4 still follow the instruction step by step and manages to generate a well-format wrong answer, indicating the errors mainly come from miscalculation. For example, in the least square regression task, despite its inability to conduct the actual computation within the generation of a few tokens, GPT-4 still manages to guess a float number, which is likely to be inaccurate, and continue the rest of the algorithm execution, as if the computed number is correct.

\begin{table}[!h]
\small
\centering
\begin{tabularx}{0.45\textwidth}{@{}lll@{}}
\toprule
\textbf{Algorithm} & \textbf{Model} & \textbf{Acc}  \\
\midrule
Least Square Regression & GPT-3.5-T & 0 \\
 & Davinci-003 & 0 \\
 & GPT-4 & 0 \\
 \midrule
Discrete Fourier Transform & GPT-3.5-T & 0 \\
 & Davinci-003 & 0 \\
 & GPT-4 & 0 \\
 \midrule
Graham Scan & GPT-3.5-T & 0 \\
 & Davinci-003 & 0 \\
 & GPT-4 & 0 \\
 \midrule
Jarvis March & GPT-3.5-T & 0 \\
 & Davinci-003 & 0 \\
 & GPT-4 & 0 \\
 \bottomrule
\end{tabularx}
\caption{Results of CLRS-Numeric}
\label{table:clrs_numeric}
\end{table}

% \subsection{Error Analysis}

\subsection{Intermediate Results Evaluation}
To further investigate the reason behind success/failure, we select 5 algorithms, Bubble Sort, Knuth-Morris-Pratt (Strings), Task Scheduling (Greedy), Optimal Binary Search Tree (Dynamic Programming) and Breadth-First Search (Graphs), which are relatively easy to extract and evaluate the intermediate results. The transition sequence of intermediate results for each algorithm is as follows:

\begin{itemize}
\item Bubble Sort: the number list $A$.
\item Knuth-Morris-Pratt: the longest proper prefix list $lps$.
\item Task Scheduling: the list $job$.
\item Optimal Binary Search Tree: the matrix $dp$.
\item Breadth-First Search: the queue $Q$.
\end{itemize}

We can obtain the gold intermediate results by running the algorithm programs. For metrics, we compute the metric of Intermediate Accuracy, which requires all the intermediate results to be correct. We also compute the metric of Process Accuracy, which computes the ratio of correct intermediate sequence prefix, averaged over $N$ instances. The intuition is that, once an intermediate result goes wrong, the following computation based on the previous one would also be problematic:

$$ Process = \frac{1}{N} \sum_{i}{\frac{len(correct\_prefix)}{max(len(pred_i), len(gold_i))}} $$

\begin{table}[!h]
\begin{subtable}{0.45\textwidth}
\small
\centering
\begin{tabular}{@{}lllll@{}}
\toprule
\textbf{Algorithm} & \textbf{Model} & \textbf{Final} & \textbf{Interm.} & \textbf{Proc.} \\
\midrule
Bubble Sort & GPT-3.5-T & 60 & 60 & 71.7 \\
 & Davinci-003 & 70 & 70 & 81.4 \\
 & GPT-4 & 100 & 100 & 100.0 \\
\midrule
KMP & GPT-3.5-T & 30 & 80 & 86.7 \\
 & Davinci-003 & 10 & 50 & 66.7 \\
 & GPT-4 & 100 & 100 & 100.0 \\
\midrule
Task & GPT-3.5-T & 40 & 40 & 71.0 \\
Scheduling & Davinci-003 & 50 & 50 & 50.0 \\
 & GPT-4 & 100 & 100 & 100.0 \\
\midrule
Optimal BST & GPT-3.5-T & 0 & 0 & 43.2 \\
 & Davinci-003 & 10 & 0 & 4.4 \\
 & GPT-4 & 100 & 100 & 100.0 \\
\midrule
BFS & GPT-3.5-T & 10 & 0 & 18.7 \\
 & Davinci-003 & 0 & 0 & 0.0 \\
 & GPT-4 & 100 & 100 & 100.0 \\
% \midrule
% Averaged & GPT-3.5-T & 28.0 & 36.0 & 58.2 \\
%  & Davinci-003 & 28.0 & 34.0 & 40.5 \\
%  & GPT-4 & \textbf{100.0} & \textbf{100.0} & \textbf{100.0} \\
 \bottomrule
\end{tabular}
\caption{Natural Language Prompt (Ours)}
\label{table:natural_language_prompt}
\end{subtable}

\begin{subtable}{0.45\textwidth}
    \small
    \centering
\begin{tabular}{@{}lllll@{}}
\toprule
\textbf{Algorithm} & \textbf{Model} & \textbf{Final} & \textbf{Interm.} & \textbf{Proc.} \\
\midrule
Bubble Sort & GPT-3.5-T & 100 & 100 & 100.0 \\
 & Davinci-003 & 0 & 60 & 72.2 \\
 & GPT-4 & 100 & 100 & 100.0 \\
\midrule
KMP & GPT-3.5-T & 20 & 50 & 73.3 \\
 & Davinci-003 & 0 & 40 & 66.7 \\
 & GPT-4 & 80 & 100 & 100.0 \\
\midrule
Task & GPT-3.5-T & 60 & 60 & 82.0 \\
Scheduling & Davinci-003 & 10 & 10 & 15.0 \\
 & GPT-4 & 80 & 80 & 86.5 \\
\midrule
Optimal BST & GPT-3.5-T & 0 & 0 & 46.4 \\
 & Davinci-003 & 0 & 0 & 5.4 \\
 & GPT-4 & 20 & 20 & 72.2 \\
\midrule
BFS & GPT-3.5-T & 10 & 10 & 21.5 \\
 & Davinci-003 & 0 & 0 & 1.4 \\
 & GPT-4 & 80 & 80 & 87.1 \\
% \midrule
% Averaged & GPT-3.5-T & 38.0 & 42.0 & 64.6 \\
%  & Davinci-003 & 18.0 & 22.0 & 32.1 \\
%  & GPT-4 & 76.0 & 74.0 & 87.4 \\
 \bottomrule
\end{tabular}
\caption{Python Code}
\label{table:python_code}
\end{subtable}

\caption{Intermediate Results Evaluation}
\label{table:ablation_intermediate}

\end{table}

The results are shown in Table \ref{table:ablation_intermediate}. We find that by replacing detailed instruction with "uninterpreted" Python code, the Intermediate Accuracy and Process Accuracy drop noticeably together with the Final Accuracy, as the scores of GPT-4 are no longer all 100\%. This further demonstrates the necessity of our proposed method. Under detailed instruction, GPT-4 did not make any mistakes in computing intermediate results, which further confirms its effectiveness and ability. For GPT-3.5-Turbo and Text-Davinci-003, lower final accuracy is associated with lower intermediate correctness.

% In terms of Average Process Accuracy, both GPT-3.5 models also benefit from our proposed natural language instructions.

Also, Process Accuracy and Intermediate Accuracy may not be smaller than Final accuracy, indicating that correct intermediate computation would be much more likely to lead to the correct final answer, while one single error would result in the wrong answer. Moreover, as expected, the inability to produce error-free intermediate results, which is largely due to miscalculation, contributes to the low performance of two GPT-3.5 models.

\section{Discussion}
\subsection{Challenges of Step-wise Evaluation}
Rigorous step-by-step evaluation of the model's computation process is non-trivial, since it would require an equivalent Turing Machine to compute that step and check if the result is consistent. And due to the flexible natural language style, it's challenging to extract the intermediate results by pure hand-written regular expression. Moreover, for a single model GPT-4, as each test instance could contain 100 lines, and we have 260 cases in total, with lengthy model-generated outputs, human annotation would be costly and time-consuming, yet 100\% accuracy is not guaranteed, making it also not feasible. 

But from another perspective, as the algorithm becomes more complex, it's more challenging to guess a correct solution without actually computing step-by-step, and since the current state depends on the previous state, the chances of passing all the test cases for the wrong reason are more and more unlikely. Therefore, correct outcomes in all cases highly indicate the correctness of the intermediate computation steps.

\subsection{On the Possibility of Data Leakage and Memorization}
As mentioned previously, the test data was randomly generated. Given the vast input space, although the algorithms themselves are widely known, it's highly unlikely that our concrete randomly generated instances overlap with examples available online.
Moreover, due to the complexity of the algorithms, memorization is unlikely to result in successful execution. Since the input space is exponentially large, simple memorization of finite, linear training instances would not guarantee the performance on arbitrary new test cases; on the other hand, feeding exponential training instances would bring the issue of catastrophic forgetting.

Lastly, our experiment results do not support the data leakage hypothesis. If GPT-4 is trained on similar data, replacing our detailed yet Internet-unavailable natural instruction with easily found Python code would not greatly reduce the performance, which is not the case.

\section{Related Works}
\subsection{Large Language Models}
GPT-3 \citep{NEURIPS2020_1457c0d6}, the first large language model (LLM) with 175 billion parameters, pioneered the trend. It showcased that without fine-tuning, LLMs can accomplish various tasks effectively via trigger strategies like few-shot \citep{NEURIPS2020_1457c0d6} or chain-of-thought prompting \citep{wei2022chain}, which is observed as an emergent ability \citep{wei2022emergent}. Since then, various new LLMs have been proposed \citep{DBLP:journals/corr/abs-2211-05100, DBLP:journals/corr/abs-2302-13971, DBLP:journals/corr/abs-2303-08774, anil2023palm}.
% , such as BLOOM \citep{DBLP:journals/corr/abs-2211-05100} and LLaMA \citep{DBLP:journals/corr/abs-2302-13971}, and recently GPT-4 \citep{DBLP:journals/corr/abs-2303-08774} and PaLM 2 \citep{anil2023palm}.
In parallel, researchers have proposed theories to explain how performance improves as the model size increases \citep{DBLP:journals/corr/abs-2001-08361, hoffmann2022an}. 
% These studies present promising evidence that scaling up model size, training corpus, and computation resources is a viable direction for enhancing model performance. % In particular, \citet{hoffmann2022an} demonstrates that balancing the model size and training data quantity within a fixed compute budget can lead to superior performance.
Beyond unsupervised learning, a range of approaches for "instruction tuning" have been introduced \citep{wei2022finetuned, sanh2022multitask, wang-etal-2022-super, DBLP:conf/nips/Ouyang0JAWMZASR22}. Instruction tuning aims to refine LLMs, making them more efficient and accessible for downstream applications. % \citet{wang-etal-2022-super} showcased superior performance with "Super-NaturalInstructions," achieving improved generalization across more than 1600 NLP tasks. Meanwhile, \citet{wei2022finetuned} demonstrated that instruction tuning, especially when combined with zero-shot learning, can significantly boost performance across unseen tasks. \citet{DBLP:conf/nips/Ouyang0JAWMZASR22} suggests that Reinforcement Learning with Human Feedback \citep{NIPS2017_d5e2c0ad} can considerably improve the ability to follow user instructions, which can be extremely diverse.
\subsection{Turing-Completeness of Neural Networks}
The Turing-completeness \cite{turing1937computable} of neural networks has been extensively studied. \citet{SIEGELMANN1995132} first provided an early demonstration that neural networks can simulate all Turing machines. 
% Their work centered on finite-size networks composed of synchronously evolving processors using a sigmoid function for state updates.
% Subsequent research extended this concept to novel architectures. 
Subsequently, \citet{graves2014neural} developed the Neural Turing Machine.
% which combined traditional neural networks with external memory resources. 
Similarly,\citet{weiss-etal-2018-practical} explored the computational power of RNNs, 
% , with a focus on finite precision and computation time, and discovered variations in computational strength across different RNN variants. 
and later the computational properties of Transformers are studied \citep{pérez2018on, bhattamishra-etal-2020-computational, wei_statistically_2022}. 
% \citet{pérez2018on} further studied the computational properties of Transformers \citep{NIPS2017_3f5ee243}.
%, which is the backbone of current LLMs, and established its Turing completeness. Later works explored the specifics of the computational power of Transformers. 
% \citet{bhattamishra-etal-2020-computational} analyzed each component's necessity for Turing completeness in a Transformer. Similarly, \citet{wei_statistically_2022} proposed a definition of statistically meaningful approximation and demonstrated that Transformers could approximate Turing machines within this framework. 
Recently, \citet{schuurmans2023memory} and \citet{DBLP:journals/corr/abs-2303-14310} demonstrated the computational universality of large language models without further fine-tuning.
% Recently, \citet{schuurmans2023memory} demonstrated computational universality in transformer-based large language models augmented with external memory, and unlike previous works, no parameter adjustment is needed. Concurrently, \citet{DBLP:journals/corr/abs-2303-14310} showcased that GPT-3 can perform iterative behaviors essential for executing programs via prompting and regimenting attention.

\subsection{LLMs for Coding Tasks}
LLMs have advanced a series of code-related tasks, including code generation \citep{DBLP:journals/corr/abs-2206-01335}, and particularly competitive programming \citep{DBLP:journals/corr/abs-2307-05337, doi:10.1126/science.abq1158}. In addition, tools like GitHub Copilot and others \citep{DBLP:journals/corr/abs-2107-03374, Joshi_CambroneroSanchez_Gulwani_Le_Verbruggen_Radiček_2023} have harnessed the power of LLMs to assist developers. Furthermore, LLMs have also been applied to test generation \citep{10329992} and code explanation \citep{nam2024using}.

\section{Conclusion}
In summary, our research shows compelling evidence that large language models, especially GPT-4, can effectively interpret and execute algorithms described in natural language. These models demonstrated astonishing performance in following control flow and performing precise calculations and operations. They also exhibited strong capabilities in maintaining and updating variable values via text output. Such attributes mimic the core functions of the Von-Neumann Machine. Consequently, we can potentially instruct these models to perform complex operations merely through natural language prompts. We hope our research could shed light on further investigation of evaluating and leveraging the capabilities of large language models.

% In this paper, we have delved into the capabilities of large language models, particularly on GPT-3.5 \citep{DBLP:conf/nips/Ouyang0JAWMZASR22} and GPT-4 \citep{DBLP:journals/corr/abs-2303-08774}, in emulating stored-program computers, showcasing their potential as zero-shot Von Neumann machines. Our comprehensive assessment through a variety of tasks has demonstrated their proficiency, particularly GPT-4's, in dealing with complex computational challenges. While GPT-3.5 exhibited competency in simpler tasks, GPT-4 was requisite for higher computational prowess, thereby delineating the continued advancements within the field.

% Our approach underscores the potential of large language models to understand and execute intricate instructions, marking a significant development in expanding Turing-completeness capabilities within these models. In future work, we hope to continue to refine this prompting approach and explore its potential and limitations across a wider range of tasks and applications, going toward the goal of Artificial General Intelligence.

% Going forward, research should focus on enhancing these models, bolstering their accuracy, and enabling them to perform more complex tasks reliably. Continued experimentation and iterative development are essential for pushing the boundaries of what these models can achieve, potentially revolutionizing the interaction between humans and AI systems.

\section*{Acknowledgements}
This work is supported by the Strategic Priority Research Program of Chinese Academy of Sciences under Grant XDA27020200 and the Natural Science Foundation of China (No. 62122077 and 62106251).

\section*{Limitations}
In this work, we use GPT-3.5 and GPT-4, which are not open-sourced and may only be accessed via API. Future works are needed to evaluate on publicly-available large language models once such models reach the performance of GPT-3.5 or even GPT-4.

\section*{Ethics Statement}
This work is conducted in compliance with ethical principles. This work involves no sensitive data and uses several public-available datasets, or generates new datasets by random sampling.

\section*{Reference}\label{sec:reference}

\bibliography{custom}

\begin{thebibliography}{54}
\expandafter\ifx\csname natexlab\endcsname\relax\def\natexlab#1{#1}\fi

\bibitem[{Aho and Hopcroft(1974)}]{10.5555/578775}
Alfred~V. Aho and John~E. Hopcroft. 1974.
\newblock \emph{The Design and Analysis of Computer Algorithms}, 1st edition.
\newblock Addison-Wesley Longman Publishing Co., Inc., USA.

\bibitem[{Anil et~al.(2023)Anil, Dai, Firat, Johnson, Lepikhin, Passos, Shakeri, Taropa, Bailey, Chen, Chu, Clark, Shafey, Huang, Meier-Hellstern, Mishra, Moreira, Omernick, Robinson, Ruder, Tay, Xiao, Xu, Zhang, Abrego, Ahn, Austin, Barham, Botha, Bradbury, Brahma, Brooks, Catasta, Cheng, Cherry, Choquette-Choo, Chowdhery, Crepy, Dave, Dehghani, Dev, Devlin, Díaz, Du, Dyer, Feinberg, Feng, Fienber, Freitag, Garcia, Gehrmann, Gonzalez, Gur-Ari, Hand, Hashemi, Hou, Howland, Hu, Hui, Hurwitz, Isard, Ittycheriah, Jagielski, Jia, Kenealy, Krikun, Kudugunta, Lan, Lee, Lee, Li, Li, Li, Li, Li, Lim, Lin, Liu, Liu, Maggioni, Mahendru, Maynez, Misra, Moussalem, Nado, Nham, Ni, Nystrom, Parrish, Pellat, Polacek, Polozov, Pope, Qiao, Reif, Richter, Riley, Ros, Roy, Saeta, Samuel, Shelby, Slone, Smilkov, So, Sohn, Tokumine, Valter, Vasudevan, Vodrahalli, Wang, Wang, Wang, Wang, Wieting, Wu, Xu, Xu, Xue, Yin, Yu, Zhang, Zheng, Zheng, Zhou, Zhou, Petrov, and Wu}]{anil2023palm}
Rohan Anil, Andrew~M. Dai, Orhan Firat, Melvin Johnson, Dmitry Lepikhin, Alexandre Passos, Siamak Shakeri, Emanuel Taropa, Paige Bailey, Zhifeng Chen, Eric Chu, Jonathan~H. Clark, Laurent~El Shafey, Yanping Huang, Kathy Meier-Hellstern, Gaurav Mishra, Erica Moreira, Mark Omernick, Kevin Robinson, Sebastian Ruder, Yi~Tay, Kefan Xiao, Yuanzhong Xu, Yujing Zhang, Gustavo~Hernandez Abrego, Junwhan Ahn, Jacob Austin, Paul Barham, Jan Botha, James Bradbury, Siddhartha Brahma, Kevin Brooks, Michele Catasta, Yong Cheng, Colin Cherry, Christopher~A. Choquette-Choo, Aakanksha Chowdhery, Clément Crepy, Shachi Dave, Mostafa Dehghani, Sunipa Dev, Jacob Devlin, Mark Díaz, Nan Du, Ethan Dyer, Vlad Feinberg, Fangxiaoyu Feng, Vlad Fienber, Markus Freitag, Xavier Garcia, Sebastian Gehrmann, Lucas Gonzalez, Guy Gur-Ari, Steven Hand, Hadi Hashemi, Le~Hou, Joshua Howland, Andrea Hu, Jeffrey Hui, Jeremy Hurwitz, Michael Isard, Abe Ittycheriah, Matthew Jagielski, Wenhao Jia, Kathleen Kenealy, Maxim Krikun, Sneha Kudugunta, Chang
  Lan, Katherine Lee, Benjamin Lee, Eric Li, Music Li, Wei Li, YaGuang Li, Jian Li, Hyeontaek Lim, Hanzhao Lin, Zhongtao Liu, Frederick Liu, Marcello Maggioni, Aroma Mahendru, Joshua Maynez, Vedant Misra, Maysam Moussalem, Zachary Nado, John Nham, Eric Ni, Andrew Nystrom, Alicia Parrish, Marie Pellat, Martin Polacek, Alex Polozov, Reiner Pope, Siyuan Qiao, Emily Reif, Bryan Richter, Parker Riley, Alex~Castro Ros, Aurko Roy, Brennan Saeta, Rajkumar Samuel, Renee Shelby, Ambrose Slone, Daniel Smilkov, David~R. So, Daniel Sohn, Simon Tokumine, Dasha Valter, Vijay Vasudevan, Kiran Vodrahalli, Xuezhi Wang, Pidong Wang, Zirui Wang, Tao Wang, John Wieting, Yuhuai Wu, Kelvin Xu, Yunhan Xu, Linting Xue, Pengcheng Yin, Jiahui Yu, Qiao Zhang, Steven Zheng, Ce~Zheng, Weikang Zhou, Denny Zhou, Slav Petrov, and Yonghui Wu. 2023.
\newblock \href {http://arxiv.org/abs/2305.10403} {Palm 2 technical report}.

\bibitem[{Barei{\ss} et~al.(2022)Barei{\ss}, Souza, d'Amorim, and Pradel}]{DBLP:journals/corr/abs-2206-01335}
Patrick Barei{\ss}, Beatriz Souza, Marcelo d'Amorim, and Michael Pradel. 2022.
\newblock \href {https://doi.org/10.48550/ARXIV.2206.01335} {Code generation tools (almost) for free? {A} study of few-shot, pre-trained language models on code}.
\newblock \emph{CoRR}, abs/2206.01335.

\bibitem[{Bellman(1958)}]{bellman1958routing}
Richard Bellman. 1958.
\newblock On a routing problem.
\newblock \emph{Quarterly of applied mathematics}, 16(1):87--90.

\bibitem[{Bentley(1984)}]{10.1145/358234.381162}
Jon Bentley. 1984.
\newblock \href {https://doi.org/10.1145/358234.381162} {Programming pearls: Algorithm design techniques}.
\newblock \emph{Commun. ACM}, 27(9):865–873.

\bibitem[{Bhattamishra et~al.(2020)Bhattamishra, Patel, and Goyal}]{bhattamishra-etal-2020-computational}
Satwik Bhattamishra, Arkil Patel, and Navin Goyal. 2020.
\newblock \href {https://doi.org/10.18653/v1/2020.conll-1.37} {On the computational power of transformers and its implications in sequence modeling}.
\newblock In \emph{Proceedings of the 24th Conference on Computational Natural Language Learning}, pages 455--475, Online. Association for Computational Linguistics.

\bibitem[{B{\"o}hm and Jacopini(1966)}]{bohm1966flow}
Corrado B{\"o}hm and Giuseppe Jacopini. 1966.
\newblock Flow diagrams, turing machines and languages with only two formation rules.
\newblock \emph{Communications of the ACM}, 9(5):366--371.

\bibitem[{Brown et~al.(2020)Brown, Mann, Ryder, Subbiah, Kaplan, Dhariwal, Neelakantan, Shyam, Sastry, Askell, Agarwal, Herbert-Voss, Krueger, Henighan, Child, Ramesh, Ziegler, Wu, Winter, Hesse, Chen, Sigler, Litwin, Gray, Chess, Clark, Berner, McCandlish, Radford, Sutskever, and Amodei}]{NEURIPS2020_1457c0d6}
Tom Brown, Benjamin Mann, Nick Ryder, Melanie Subbiah, Jared~D Kaplan, Prafulla Dhariwal, Arvind Neelakantan, Pranav Shyam, Girish Sastry, Amanda Askell, Sandhini Agarwal, Ariel Herbert-Voss, Gretchen Krueger, Tom Henighan, Rewon Child, Aditya Ramesh, Daniel Ziegler, Jeffrey Wu, Clemens Winter, Chris Hesse, Mark Chen, Eric Sigler, Mateusz Litwin, Scott Gray, Benjamin Chess, Jack Clark, Christopher Berner, Sam McCandlish, Alec Radford, Ilya Sutskever, and Dario Amodei. 2020.
\newblock \href {https://proceedings.neurips.cc/paper/2020/file/1457c0d6bfcb4967418bfb8ac142f64a-Paper.pdf} {Language models are few-shot learners}.
\newblock In \emph{Advances in Neural Information Processing Systems}, volume~33, pages 1877--1901. Curran Associates, Inc.

\bibitem[{Bubeck et~al.(2023)Bubeck, Chandrasekaran, Eldan, Gehrke, Horvitz, Kamar, Lee, Lee, Li, Lundberg, Nori, Palangi, Ribeiro, and Zhang}]{DBLP:journals/corr/abs-2303-12712}
S{\'{e}}bastien Bubeck, Varun Chandrasekaran, Ronen Eldan, Johannes Gehrke, Eric Horvitz, Ece Kamar, Peter Lee, Yin~Tat Lee, Yuanzhi Li, Scott~M. Lundberg, Harsha Nori, Hamid Palangi, Marco~T{\'{u}}lio Ribeiro, and Yi~Zhang. 2023.
\newblock \href {https://doi.org/10.48550/arXiv.2303.12712} {Sparks of artificial general intelligence: Early experiments with {GPT-4}}.
\newblock \emph{CoRR}, abs/2303.12712.

\bibitem[{Chen et~al.(2021)Chen, Tworek, Jun, Yuan, de~Oliveira~Pinto, Kaplan, Edwards, Burda, Joseph, Brockman, Ray, Puri, Krueger, Petrov, Khlaaf, Sastry, Mishkin, Chan, Gray, Ryder, Pavlov, Power, Kaiser, Bavarian, Winter, Tillet, Such, Cummings, Plappert, Chantzis, Barnes, Herbert{-}Voss, Guss, Nichol, Paino, Tezak, Tang, Babuschkin, Balaji, Jain, Saunders, Hesse, Carr, Leike, Achiam, Misra, Morikawa, Radford, Knight, Brundage, Murati, Mayer, Welinder, McGrew, Amodei, McCandlish, Sutskever, and Zaremba}]{DBLP:journals/corr/abs-2107-03374}
Mark Chen, Jerry Tworek, Heewoo Jun, Qiming Yuan, Henrique~Pond{\'{e}} de~Oliveira~Pinto, Jared Kaplan, Harrison Edwards, Yuri Burda, Nicholas Joseph, Greg Brockman, Alex Ray, Raul Puri, Gretchen Krueger, Michael Petrov, Heidy Khlaaf, Girish Sastry, Pamela Mishkin, Brooke Chan, Scott Gray, Nick Ryder, Mikhail Pavlov, Alethea Power, Lukasz Kaiser, Mohammad Bavarian, Clemens Winter, Philippe Tillet, Felipe~Petroski Such, Dave Cummings, Matthias Plappert, Fotios Chantzis, Elizabeth Barnes, Ariel Herbert{-}Voss, William~Hebgen Guss, Alex Nichol, Alex Paino, Nikolas Tezak, Jie Tang, Igor Babuschkin, Suchir Balaji, Shantanu Jain, William Saunders, Christopher Hesse, Andrew~N. Carr, Jan Leike, Joshua Achiam, Vedant Misra, Evan Morikawa, Alec Radford, Matthew Knight, Miles Brundage, Mira Murati, Katie Mayer, Peter Welinder, Bob McGrew, Dario Amodei, Sam McCandlish, Ilya Sutskever, and Wojciech Zaremba. 2021.
\newblock \href {http://arxiv.org/abs/2107.03374} {Evaluating large language models trained on code}.
\newblock \emph{CoRR}, abs/2107.03374.

\bibitem[{Colussi(1994)}]{colussi1994fastest}
Livio Colussi. 1994.
\newblock Fastest pattern matching in strings.
\newblock \emph{Journal of Algorithms}, 16(2):163--189.

\bibitem[{Cormen et~al.(2022)Cormen, Leiserson, Rivest, and Stein}]{cormen2022introduction}
Thomas~H Cormen, Charles~E Leiserson, Ronald~L Rivest, and Clifford Stein. 2022.
\newblock \emph{Introduction to algorithms}.

\bibitem[{Dijkstra(2022)}]{10.1145/3544585.3544600}
E.~W. Dijkstra. 2022.
\newblock \href {https://doi.org/10.1145/3544585.3544600} {\emph{A Note on Two Problems in Connexion with Graphs}}, 1 edition, page 287–290. Association for Computing Machinery, New York, NY, USA.

\bibitem[{Floyd(1962)}]{10.1145/367766.368168}
Robert~W. Floyd. 1962.
\newblock \href {https://doi.org/10.1145/367766.368168} {Algorithm 97: Shortest path}.
\newblock \emph{Commun. ACM}, 5(6):345.

\bibitem[{Gavril(1972)}]{doi:10.1137/0201013}
F\u{a}nic\u{a} Gavril. 1972.
\newblock \href {https://doi.org/10.1137/0201013} {Algorithms for minimum coloring, maximum clique, minimum covering by cliques, and maximum independent set of a chordal graph}.
\newblock \emph{SIAM Journal on Computing}, 1(2):180--187.

\bibitem[{Graham(1972)}]{GRAHAM1972132}
R.L. Graham. 1972.
\newblock \href {https://doi.org/https://doi.org/10.1016/0020-0190(72)90045-2} {An efficient algorith for determining the convex hull of a finite planar set}.
\newblock \emph{Information Processing Letters}, 1(4):132--133.

\bibitem[{Graves et~al.(2014)Graves, Wayne, and Danihelka}]{graves2014neural}
Alex Graves, Greg Wayne, and Ivo Danihelka. 2014.
\newblock \href {http://arxiv.org/abs/1410.5401} {Neural turing machines}.

\bibitem[{Han et~al.(2021)Han, Zhang, Ding, Gu, Liu, Huo, Qiu, Yao, Zhang, Zhang, Han, Huang, Jin, Lan, Liu, Liu, Lu, Qiu, Song, Tang, Wen, Yuan, Zhao, and Zhu}]{HAN2021225}
Xu~Han, Zhengyan Zhang, Ning Ding, Yuxian Gu, Xiao Liu, Yuqi Huo, Jiezhong Qiu, Yuan Yao, Ao~Zhang, Liang Zhang, Wentao Han, Minlie Huang, Qin Jin, Yanyan Lan, Yang Liu, Zhiyuan Liu, Zhiwu Lu, Xipeng Qiu, Ruihua Song, Jie Tang, Ji-Rong Wen, Jinhui Yuan, Wayne~Xin Zhao, and Jun Zhu. 2021.
\newblock \href {https://doi.org/https://doi.org/10.1016/j.aiopen.2021.08.002} {Pre-trained models: Past, present and future}.
\newblock \emph{AI Open}, 2:225--250.

\bibitem[{Hoare(1961{\natexlab{a}})}]{10.1145/366622.366644}
C.~A.~R. Hoare. 1961{\natexlab{a}}.
\newblock \href {https://doi.org/10.1145/366622.366644} {Algorithm 64: Quicksort}.
\newblock \emph{Commun. ACM}, 4(7):321.

\bibitem[{Hoare(1961{\natexlab{b}})}]{10.1145/366622.366647}
C.~A.~R. Hoare. 1961{\natexlab{b}}.
\newblock \href {https://doi.org/10.1145/366622.366647} {Algorithm 65: Find}.
\newblock \emph{Commun. ACM}, 4(7):321–322.

\bibitem[{Hoffmann et~al.(2022)Hoffmann, Borgeaud, Mensch, Buchatskaya, Cai, Rutherford, de~las Casas, Hendricks, Welbl, Clark, Hennigan, Noland, Millican, van~den Driessche, Damoc, Guy, Osindero, Simonyan, Elsen, Vinyals, Rae, and Sifre}]{hoffmann2022an}
Jordan Hoffmann, Sebastian Borgeaud, Arthur Mensch, Elena Buchatskaya, Trevor Cai, Eliza Rutherford, Diego de~las Casas, Lisa~Anne Hendricks, Johannes Welbl, Aidan Clark, Tom Hennigan, Eric Noland, Katherine Millican, George van~den Driessche, Bogdan Damoc, Aurelia Guy, Simon Osindero, Karen Simonyan, Erich Elsen, Oriol Vinyals, Jack~William Rae, and Laurent Sifre. 2022.
\newblock \href {https://openreview.net/forum?id=iBBcRUlOAPR} {An empirical analysis of compute-optimal large language model training}.
\newblock In \emph{Advances in Neural Information Processing Systems}.

\bibitem[{Jarvis(1973)}]{JARVIS197318}
R.A. Jarvis. 1973.
\newblock \href {https://doi.org/https://doi.org/10.1016/0020-0190(73)90020-3} {On the identification of the convex hull of a finite set of points in the plane}.
\newblock \emph{Information Processing Letters}, 2(1):18--21.

\bibitem[{Jojic et~al.(2023)Jojic, Wang, and Jojic}]{DBLP:journals/corr/abs-2303-14310}
Ana Jojic, Zhen Wang, and Nebojsa Jojic. 2023.
\newblock \href {https://doi.org/10.48550/arXiv.2303.14310} {{GPT} is becoming a turing machine: Here are some ways to program it}.
\newblock \emph{CoRR}, abs/2303.14310.

\bibitem[{Joshi et~al.(2023)Joshi, Cambronero~Sanchez, Gulwani, Le, Verbruggen, and Radiček}]{Joshi_CambroneroSanchez_Gulwani_Le_Verbruggen_Radiček_2023}
Harshit Joshi, José Cambronero~Sanchez, Sumit Gulwani, Vu~Le, Gust Verbruggen, and Ivan Radiček. 2023.
\newblock \href {https://doi.org/10.1609/aaai.v37i4.25642} {Repair is nearly generation: Multilingual program repair with llms}.
\newblock \emph{Proceedings of the AAAI Conference on Artificial Intelligence}, 37(4):5131--5140.

\bibitem[{Kaplan et~al.(2020)Kaplan, McCandlish, Henighan, Brown, Chess, Child, Gray, Radford, Wu, and Amodei}]{DBLP:journals/corr/abs-2001-08361}
Jared Kaplan, Sam McCandlish, Tom Henighan, Tom~B. Brown, Benjamin Chess, Rewon Child, Scott Gray, Alec Radford, Jeffrey Wu, and Dario Amodei. 2020.
\newblock \href {http://arxiv.org/abs/2001.08361} {Scaling laws for neural language models}.
\newblock \emph{CoRR}, abs/2001.08361.

\bibitem[{Knuth(1973)}]{knuth1973fundamental}
Donald~E Knuth. 1973.
\newblock Fundamental algorithms.

\bibitem[{Kruskal(1956)}]{fc0df122-3305-33a8-bac5-7a6fc3666dfb}
Joseph~B. Kruskal. 1956.
\newblock \href {http://www.jstor.org/stable/2033241} {On the shortest spanning subtree of a graph and the traveling salesman problem}.
\newblock \emph{Proceedings of the American Mathematical Society}, 7(1):48--50.

\bibitem[{Lawler(1985)}]{lawler1985traveling}
Eugene~L Lawler. 1985.
\newblock The traveling salesman problem: a guided tour of combinatorial optimization.
\newblock \emph{Wiley-Interscience Series in Discrete Mathematics}.

\bibitem[{Li et~al.(2023)Li, Tworkowski, Wu, and Mooney}]{DBLP:journals/corr/abs-2307-05337}
Jierui Li, Szymon Tworkowski, Yingying Wu, and Raymond~J. Mooney. 2023.
\newblock \href {https://doi.org/10.48550/ARXIV.2307.05337} {Explaining competitive-level programming solutions using llms}.
\newblock \emph{CoRR}, abs/2307.05337.

\bibitem[{Li et~al.(2022)Li, Choi, Chung, Kushman, Schrittwieser, Leblond, Eccles, Keeling, Gimeno, Lago, Hubert, Choy, de~Masson~d’Autume, Babuschkin, Chen, Huang, Welbl, Gowal, Cherepanov, Molloy, Mankowitz, Robson, Kohli, de~Freitas, Kavukcuoglu, and Vinyals}]{doi:10.1126/science.abq1158}
Yujia Li, David Choi, Junyoung Chung, Nate Kushman, Julian Schrittwieser, Rémi Leblond, Tom Eccles, James Keeling, Felix Gimeno, Agustin~Dal Lago, Thomas Hubert, Peter Choy, Cyprien de~Masson~d’Autume, Igor Babuschkin, Xinyun Chen, Po-Sen Huang, Johannes Welbl, Sven Gowal, Alexey Cherepanov, James Molloy, Daniel~J. Mankowitz, Esme~Sutherland Robson, Pushmeet Kohli, Nando de~Freitas, Koray Kavukcuoglu, and Oriol Vinyals. 2022.
\newblock \href {https://doi.org/10.1126/science.abq1158} {Competition-level code generation with alphacode}.
\newblock \emph{Science}, 378(6624):1092--1097.

\bibitem[{Liu et~al.(2023)Liu, Lu, Chen, Jiang, Svyatkovskiy, Fu, Sundaresan, and Duan}]{liu2023code}
Chenxiao Liu, Shuai Lu, Weizhu Chen, Daxin Jiang, Alexey Svyatkovskiy, Shengyu Fu, Neel Sundaresan, and Nan Duan. 2023.
\newblock Code execution with pre-trained language models.
\newblock \emph{arXiv preprint arXiv:2305.05383}.

\bibitem[{Moore(1959)}]{moore1959shortest}
Edward~F Moore. 1959.
\newblock The shortest path through a maze.
\newblock In \emph{Proc. Int. Symp. Switching Theory, 1959}, pages 285--292.

\bibitem[{Nam et~al.(2024)Nam, Macvean, Hellendoorn, Vasilescu, and Myers}]{nam2024using}
Daye Nam, Andrew Macvean, Vincent Hellendoorn, Bogdan Vasilescu, and Brad Myers. 2024.
\newblock \href {http://arxiv.org/abs/2307.08177} {Using an llm to help with code understanding}.

\bibitem[{OpenAI(2023)}]{DBLP:journals/corr/abs-2303-08774}
OpenAI. 2023.
\newblock \href {https://doi.org/10.48550/arXiv.2303.08774} {{GPT-4} technical report}.
\newblock \emph{CoRR}, abs/2303.08774.

\bibitem[{Ouyang et~al.(2022)Ouyang, Wu, Jiang, Almeida, Wainwright, Mishkin, Zhang, Agarwal, Slama, Ray, Schulman, Hilton, Kelton, Miller, Simens, Askell, Welinder, Christiano, Leike, and Lowe}]{DBLP:conf/nips/Ouyang0JAWMZASR22}
Long Ouyang, Jeffrey Wu, Xu~Jiang, Diogo Almeida, Carroll~L. Wainwright, Pamela Mishkin, Chong Zhang, Sandhini Agarwal, Katarina Slama, Alex Ray, John Schulman, Jacob Hilton, Fraser Kelton, Luke Miller, Maddie Simens, Amanda Askell, Peter Welinder, Paul~F. Christiano, Jan Leike, and Ryan Lowe. 2022.
\newblock \href {http://papers.nips.cc/paper\_files/paper/2022/hash/b1efde53be364a73914f58805a001731-Abstract-Conference.html} {Training language models to follow instructions with human feedback}.
\newblock In \emph{NeurIPS}.

\bibitem[{Prim(1957)}]{6773228}
R.~C. Prim. 1957.
\newblock \href {https://doi.org/10.1002/j.1538-7305.1957.tb01515.x} {Shortest connection networks and some generalizations}.
\newblock \emph{The Bell System Technical Journal}, 36(6):1389--1401.

\bibitem[{Pérez et~al.(2019)Pérez, Marinković, and Barceló}]{pérez2018on}
Jorge Pérez, Javier Marinković, and Pablo Barceló. 2019.
\newblock \href {https://openreview.net/forum?id=HyGBdo0qFm} {On the turing completeness of modern neural network architectures}.
\newblock In \emph{International Conference on Learning Representations}.

\bibitem[{Sammet(1966)}]{10.1145/365230.365274}
Jean~E. Sammet. 1966.
\newblock \href {https://doi.org/10.1145/365230.365274} {The use of english as a programming language}.
\newblock \emph{Commun. ACM}, 9(3):228–230.

\bibitem[{Sanh et~al.(2022)Sanh, Webson, Raffel, Bach, Sutawika, Alyafeai, Chaffin, Stiegler, Raja, Dey, Bari, Xu, Thakker, Sharma, Szczechla, Kim, Chhablani, Nayak, Datta, Chang, Jiang, Wang, Manica, Shen, Yong, Pandey, Bawden, Wang, Neeraj, Rozen, Sharma, Santilli, Fevry, Fries, Teehan, Scao, Biderman, Gao, Wolf, and Rush}]{sanh2022multitask}
Victor Sanh, Albert Webson, Colin Raffel, Stephen Bach, Lintang Sutawika, Zaid Alyafeai, Antoine Chaffin, Arnaud Stiegler, Arun Raja, Manan Dey, M~Saiful Bari, Canwen Xu, Urmish Thakker, Shanya~Sharma Sharma, Eliza Szczechla, Taewoon Kim, Gunjan Chhablani, Nihal Nayak, Debajyoti Datta, Jonathan Chang, Mike Tian-Jian Jiang, Han Wang, Matteo Manica, Sheng Shen, Zheng~Xin Yong, Harshit Pandey, Rachel Bawden, Thomas Wang, Trishala Neeraj, Jos Rozen, Abheesht Sharma, Andrea Santilli, Thibault Fevry, Jason~Alan Fries, Ryan Teehan, Teven~Le Scao, Stella Biderman, Leo Gao, Thomas Wolf, and Alexander~M Rush. 2022.
\newblock \href {https://openreview.net/forum?id=9Vrb9D0WI4} {Multitask prompted training enables zero-shot task generalization}.
\newblock In \emph{International Conference on Learning Representations}.

\bibitem[{Scao et~al.(2022)Scao, Fan, Akiki, Pavlick, Ilic, Hesslow, Castagn{\'{e}}, Luccioni, Yvon, Gall{\'{e}}, Tow, Rush, Biderman, Webson, Ammanamanchi, Wang, Sagot, Muennighoff, del Moral, Ruwase, Bawden, Bekman, McMillan{-}Major, Beltagy, Nguyen, Saulnier, Tan, Suarez, Sanh, Lauren{\c{c}}on, Jernite, Launay, Mitchell, Raffel, Gokaslan, Simhi, Soroa, Aji, Alfassy, Rogers, Nitzav, Xu, Mou, Emezue, Klamm, Leong, van Strien, Adelani, and et~al.}]{DBLP:journals/corr/abs-2211-05100}
Teven~Le Scao, Angela Fan, Christopher Akiki, Ellie Pavlick, Suzana Ilic, Daniel Hesslow, Roman Castagn{\'{e}}, Alexandra~Sasha Luccioni, Fran{\c{c}}ois Yvon, Matthias Gall{\'{e}}, Jonathan Tow, Alexander~M. Rush, Stella Biderman, Albert Webson, Pawan~Sasanka Ammanamanchi, Thomas Wang, Beno{\^{\i}}t Sagot, Niklas Muennighoff, Albert~Villanova del Moral, Olatunji Ruwase, Rachel Bawden, Stas Bekman, Angelina McMillan{-}Major, Iz~Beltagy, Huu Nguyen, Lucile Saulnier, Samson Tan, Pedro~Ortiz Suarez, Victor Sanh, Hugo Lauren{\c{c}}on, Yacine Jernite, Julien Launay, Margaret Mitchell, Colin Raffel, Aaron Gokaslan, Adi Simhi, Aitor Soroa, Alham~Fikri Aji, Amit Alfassy, Anna Rogers, Ariel~Kreisberg Nitzav, Canwen Xu, Chenghao Mou, Chris Emezue, Christopher Klamm, Colin Leong, Daniel van Strien, David~Ifeoluwa Adelani, and et~al. 2022.
\newblock \href {https://doi.org/10.48550/arXiv.2211.05100} {{BLOOM:} {A} 176b-parameter open-access multilingual language model}.
\newblock \emph{CoRR}, abs/2211.05100.

\bibitem[{Schuurmans(2023)}]{schuurmans2023memory}
Dale Schuurmans. 2023.
\newblock \href {http://arxiv.org/abs/2301.04589} {Memory augmented large language models are computationally universal}.

\bibitem[{Schäfer et~al.(2024)Schäfer, Nadi, Eghbali, and Tip}]{10329992}
Max Schäfer, Sarah Nadi, Aryaz Eghbali, and Frank Tip. 2024.
\newblock \href {https://doi.org/10.1109/TSE.2023.3334955} {An empirical evaluation of using large language models for automated unit test generation}.
\newblock \emph{IEEE Transactions on Software Engineering}, 50(1):85--105.

\bibitem[{Siegelmann and Sontag(1995)}]{SIEGELMANN1995132}
H.T. Siegelmann and E.D. Sontag. 1995.
\newblock \href {https://doi.org/https://doi.org/10.1006/jcss.1995.1013} {On the computational power of neural nets}.
\newblock \emph{Journal of Computer and System Sciences}, 50(1):132--150.

\bibitem[{Srivastava et~al.(2022)Srivastava, Rastogi, Rao, Shoeb, Abid, Fisch, Brown, Santoro, Gupta, Garriga{-}Alonso, Kluska, Lewkowycz, Agarwal, Power, Ray, Warstadt, Kocurek, Safaya, Tazarv, Xiang, Parrish, Nie, Hussain, Askell, Dsouza, Rahane, Iyer, Andreassen, Santilli, Stuhlm{\"{u}}ller, Dai, La, Lampinen, Zou, Jiang, Chen, Vuong, Gupta, Gottardi, Norelli, Venkatesh, Gholamidavoodi, Tabassum, Menezes, Kirubarajan, Mullokandov, Sabharwal, Herrick, Efrat, Erdem, Karakas, and et~al.}]{DBLP:journals/corr/abs-2206-04615}
Aarohi Srivastava, Abhinav Rastogi, Abhishek Rao, Abu Awal~Md Shoeb, Abubakar Abid, Adam Fisch, Adam~R. Brown, Adam Santoro, Aditya Gupta, Adri{\`{a}} Garriga{-}Alonso, Agnieszka Kluska, Aitor Lewkowycz, Akshat Agarwal, Alethea Power, Alex Ray, Alex Warstadt, Alexander~W. Kocurek, Ali Safaya, Ali Tazarv, Alice Xiang, Alicia Parrish, Allen Nie, Aman Hussain, Amanda Askell, Amanda Dsouza, Ameet Rahane, Anantharaman~S. Iyer, Anders Andreassen, Andrea Santilli, Andreas Stuhlm{\"{u}}ller, Andrew~M. Dai, Andrew La, Andrew~K. Lampinen, Andy Zou, Angela Jiang, Angelica Chen, Anh Vuong, Animesh Gupta, Anna Gottardi, Antonio Norelli, Anu Venkatesh, Arash Gholamidavoodi, Arfa Tabassum, Arul Menezes, Arun Kirubarajan, Asher Mullokandov, Ashish Sabharwal, Austin Herrick, Avia Efrat, Aykut Erdem, Ayla Karakas, and et~al. 2022.
\newblock \href {https://doi.org/10.48550/arXiv.2206.04615} {Beyond the imitation game: Quantifying and extrapolating the capabilities of language models}.
\newblock \emph{CoRR}, abs/2206.04615.

\bibitem[{Touvron et~al.(2023)Touvron, Lavril, Izacard, Martinet, Lachaux, Lacroix, Rozi{\`{e}}re, Goyal, Hambro, Azhar, Rodriguez, Joulin, Grave, and Lample}]{DBLP:journals/corr/abs-2302-13971}
Hugo Touvron, Thibaut Lavril, Gautier Izacard, Xavier Martinet, Marie{-}Anne Lachaux, Timoth{\'{e}}e Lacroix, Baptiste Rozi{\`{e}}re, Naman Goyal, Eric Hambro, Faisal Azhar, Aur{\'{e}}lien Rodriguez, Armand Joulin, Edouard Grave, and Guillaume Lample. 2023.
\newblock \href {https://doi.org/10.48550/arXiv.2302.13971} {Llama: Open and efficient foundation language models}.
\newblock \emph{CoRR}, abs/2302.13971.

\bibitem[{Turing(1937)}]{turing1937computable}
AM~Turing. 1937.
\newblock On computable numbers, with an application to the entscheidungsproblem.
\newblock \emph{Proceedings of the London Mathematical Society}, 2(1):230--230.

\bibitem[{Veli{\v{c}}kovi{\'c} et~al.(2022)Veli{\v{c}}kovi{\'c}, Badia, Budden, Pascanu, Banino, Dashevskiy, Hadsell, and Blundell}]{pmlr-v162-velickovic22a}
Petar Veli{\v{c}}kovi{\'c}, Adri{\`a}~Puigdom{\`e}nech Badia, David Budden, Razvan Pascanu, Andrea Banino, Misha Dashevskiy, Raia Hadsell, and Charles Blundell. 2022.
\newblock \href {https://proceedings.mlr.press/v162/velickovic22a.html} {The {CLRS} algorithmic reasoning benchmark}.
\newblock In \emph{Proceedings of the 39th International Conference on Machine Learning}, volume 162 of \emph{Proceedings of Machine Learning Research}, pages 22084--22102. PMLR.

\bibitem[{Wang et~al.(2022)Wang, Mishra, Alipoormolabashi, Kordi, Mirzaei, Naik, Ashok, Dhanasekaran, Arunkumar, Stap, Pathak, Karamanolakis, Lai, Purohit, Mondal, Anderson, Kuznia, Doshi, Pal, Patel, Moradshahi, Parmar, Purohit, Varshney, Kaza, Verma, Puri, Karia, Doshi, Sampat, Mishra, Reddy~A, Patro, Dixit, and Shen}]{wang-etal-2022-super}
Yizhong Wang, Swaroop Mishra, Pegah Alipoormolabashi, Yeganeh Kordi, Amirreza Mirzaei, Atharva Naik, Arjun Ashok, Arut~Selvan Dhanasekaran, Anjana Arunkumar, David Stap, Eshaan Pathak, Giannis Karamanolakis, Haizhi Lai, Ishan Purohit, Ishani Mondal, Jacob Anderson, Kirby Kuznia, Krima Doshi, Kuntal~Kumar Pal, Maitreya Patel, Mehrad Moradshahi, Mihir Parmar, Mirali Purohit, Neeraj Varshney, Phani~Rohitha Kaza, Pulkit Verma, Ravsehaj~Singh Puri, Rushang Karia, Savan Doshi, Shailaja~Keyur Sampat, Siddhartha Mishra, Sujan Reddy~A, Sumanta Patro, Tanay Dixit, and Xudong Shen. 2022.
\newblock \href {https://aclanthology.org/2022.emnlp-main.340} {Super-{N}atural{I}nstructions: Generalization via declarative instructions on 1600+ {NLP} tasks}.
\newblock In \emph{Proceedings of the 2022 Conference on Empirical Methods in Natural Language Processing}, pages 5085--5109, Abu Dhabi, United Arab Emirates. Association for Computational Linguistics.

\bibitem[{Wei et~al.(2022{\natexlab{a}})Wei, Chen, and Ma}]{wei_statistically_2022}
Colin Wei, Yining Chen, and Tengyu Ma. 2022{\natexlab{a}}.
\newblock \href {https://proceedings.neurips.cc/paper_files/paper/2022/file/4ebf1d74f53ece08512a23309d58df89-Paper-Conference.pdf} {Statistically {Meaningful} {Approximation}: a {Case} {Study} on {Approximating} {Turing} {Machines} with {Transformers}}.
\newblock In \emph{Advances in {Neural} {Information} {Processing} {Systems}}, volume~35, pages 12071--12083. Curran Associates, Inc.

\bibitem[{Wei et~al.(2022{\natexlab{b}})Wei, Bosma, Zhao, Guu, Yu, Lester, Du, Dai, and Le}]{wei2022finetuned}
Jason Wei, Maarten Bosma, Vincent Zhao, Kelvin Guu, Adams~Wei Yu, Brian Lester, Nan Du, Andrew~M. Dai, and Quoc~V Le. 2022{\natexlab{b}}.
\newblock \href {https://openreview.net/forum?id=gEZrGCozdqR} {Finetuned language models are zero-shot learners}.
\newblock In \emph{International Conference on Learning Representations}.

\bibitem[{Wei et~al.(2022{\natexlab{c}})Wei, Tay, Bommasani, Raffel, Zoph, Borgeaud, Yogatama, Bosma, Zhou, Metzler, Chi, Hashimoto, Vinyals, Liang, Dean, and Fedus}]{wei2022emergent}
Jason Wei, Yi~Tay, Rishi Bommasani, Colin Raffel, Barret Zoph, Sebastian Borgeaud, Dani Yogatama, Maarten Bosma, Denny Zhou, Donald Metzler, Ed~H. Chi, Tatsunori Hashimoto, Oriol Vinyals, Percy Liang, Jeff Dean, and William Fedus. 2022{\natexlab{c}}.
\newblock \href {https://openreview.net/forum?id=yzkSU5zdwD} {Emergent abilities of large language models}.
\newblock \emph{Transactions on Machine Learning Research}.
\newblock Survey Certification.

\bibitem[{Wei et~al.(2022{\natexlab{d}})Wei, Wang, Schuurmans, Bosma, brian ichter, Xia, Chi, Le, and Zhou}]{wei2022chain}
Jason Wei, Xuezhi Wang, Dale Schuurmans, Maarten Bosma, brian ichter, Fei Xia, Ed~H. Chi, Quoc~V Le, and Denny Zhou. 2022{\natexlab{d}}.
\newblock \href {https://openreview.net/forum?id=_VjQlMeSB_J} {Chain of thought prompting elicits reasoning in large language models}.
\newblock In \emph{Advances in Neural Information Processing Systems}.

\bibitem[{Weiss et~al.(2018)Weiss, Goldberg, and Yahav}]{weiss-etal-2018-practical}
Gail Weiss, Yoav Goldberg, and Eran Yahav. 2018.
\newblock \href {https://doi.org/10.18653/v1/P18-2117} {On the practical computational power of finite precision {RNN}s for language recognition}.
\newblock In \emph{Proceedings of the 56th Annual Meeting of the Association for Computational Linguistics (Volume 2: Short Papers)}, pages 740--745, Melbourne, Australia. Association for Computational Linguistics.

\bibitem[{Williams(1964)}]{williams1964algorithm}
John William~Joseph Williams. 1964.
\newblock Algorithm 232: heapsort.
\newblock \emph{Communications of the ACM}, 7(6):347--348.

\end{thebibliography}
\bibliographystyle{acl_natbib}

\clearpage
\newpage
\appendix

\section{Appendix}
\label{sec:appendix}

\subsection{Evaluation of Iterative Sentence Generation with Keyword Constraint}
\label{appendix:Evaluation of Iterative Sentence Generation with Keyword Constraint}
\paragraph{Task} We argue that our method, program prompting, does not limit to pure program execution. To this end, we proposed a novel task, iterative sentence generation with keyword constraint, which contains 100 test instances. With the input of the initial keyword and iteration count, the task of iterative sentence generation with keyword constraint is to generate a sentence given the keyword, then select a word from the generation as the new keyword, and stop when reaching the iteration limit. The 20 initial words include ``art'', ``business'', ``computer'', ``data'', ``entertainment'', ``environment'', ``fashion'', ``investigation'', ``lifestyle'', ``market'', ``medicine'', ``music'', ``politic'', ``science'', ``sports'', ``technology'', ``trade'', ``traffic'', ``weather'', and ``world''. The iteration counts are 5, 10, 15, 20, and 25. The prompt is presented in Table \ref{table:prompt_keyword}.

\paragraph{Result} GPT-4 model outperformed the others, achieving 100\% accuracy as shown in Table \ref{table:result iterative sentence generation with keyword constraint}. Text-Davinci-003 model achieved a high accuracy of 98\%, nearly matching the performance of GPT-4, while GPT-3.5-Turbo lagged behind with 58\% accuracy. 

\begin{table}[!ht]
\centering
\begin{tabular}{@{}ll@{}}
\toprule
\textbf{Model} & \textbf{Accuracy (\%)} \\ \midrule
GPT-3.5-Turbo & 58 \\
Text-Davinci-003 & 98 \\
\textbf{GPT-4} & \textbf{100} \\ \bottomrule
\end{tabular}
\caption{Results of iterative sentence generation with keyword constraint}
\label{table:result iterative sentence generation with keyword constraint}
\end{table}

\subsection{On the Construction of Natural Language Prompt}

We believe that program execution is a mechanical, deterministic procedure, unlike open-domain text generation, where the needed information is not fully present in the prefix. In the process of execution, as long as the LLMs predict the next correct token with more than 50\% probability, we can ensure the final correctness of the whole output.

Therefore, the main intuition is that we need to tell LLMs how to jump between the instructions, which usually is the task of the compiler of high-level program language. For simplicity of methodology, we leverage the goto statement. Moreover, for future works, we believe it would also be feasible to leverage LLM+compiler to complete the conversion.

Another trick for writing unambiguous instructions is to avoid repeated words but without different meanings. We can scan the instructions, and replace any unintended repetition. Also, for some functions like sort or argmax, we shall clearly "implement" the details, just like when we use old Pascal to write code.

Finally, after manually constructing the prompt, we may leverage few-shot in-context learning to build the draft of other algorithms, in which GPT-4 can mimic the rigid, unambiguous style. Then, we manually inspect the draft carefully, and fix the mistakes.

\begin{table*}[!htbp]
\centering
\small
\begin{tabular}{@{}lll@{}}
\toprule
\textbf{Time Complexity} & \textbf{Input Size} & \textbf{Algorithm} \\ \midrule
$O(log n)$ & 9, 10 & Binary Search \\ \midrule
$O(n)$ & 9, 10 & Minimum Search \\
 & 9, 10 & Activity Selection* \\
 & 9, 10 & Task Scheduling* \\ 
 & 9, 10 & Least Square Regression \\ 
 \midrule
$O(n)$ & 4, 5 & Maximum Subarray \\
 & 4, 5 & Quickselect \\
 & 4, 5 & Kruskal's Minimum Spanning Tree* \\ \midrule
$O(n + m)$ & 4, 5 & Depth-First Search \\
 & 4, 5 & Breadth-First Search \\
 & 4, 5 & Topological Sorting \\
 & 3, 4 & Articulation Points \\
 & 3, 4 & Bridges \\
 & 4, 5 & Kosaraju's Strongly Connected Components \\
 & 4, 5 & Knuth-Morris-Pratt String Matcher \\ \midrule
$O(n log n)$ & 4, 5 & Heapsort \\
 & 4, 5 & Quicksort \\ 
 & 8 & Discrete Fourier Transform\\
 & 9, 10 & Graham Scan \\
 \midrule
$O(n^2)$ & 4, 5 & Insertion Sort \\
 & 4, 5 & Bubble Sort \\
 & 4, 5 & Optimal Binary Search Tree \\
 & 4, 5 & Dijkstra's Algorithm \\
 & 4, 5 & Prim's Minimum Spanning Tree \\ 
 
 \midrule
$O(nm)$ & 4, 5 & Naive String Matching \\
 & 4, 5 & Bellman-Ford Algorithm \\
 & 4, 5 & Longest Common Subsequence \\ 
 & 9, 10 & Jarvis March \\
 \midrule
$O(n^3)$ & 4, 5 & Matrix Chain Multiplication \\
 & 4 & Floyd-Warshall Algorithm \\ \bottomrule
\end{tabular}

\caption{The time complexity and input size (the length of input vector) of selected algorithms, sorted by time complexity. For algorithms with greater complexity, they require longer instruction and output length. so we choose a smaller problem size to avoid exceeding the context length limit and save the inference time and cost. The Asterisk * denotes that the input is sorted.}
\label{table:clrs_mini_problem size}

\end{table*}

\clearpage
\newpage

\begin{table*}[!hb]
\small
\begin{tabular}{p{15cm}}
\toprule
\textbf{Prompt} \\ \midrule
Execute the instructions step by step. Do not jump steps. Do not stop before completion. \\ 
Initial: Set the deadlines array D: D[0] = 0, D[1] = 3, D[2] = 1, D[3] = 0 . \\ 
Set the deadlines array W: W[0] = 10, W[1] = 68, W[2] = 68, W[3] = 84 . \\ 
Set the length of the arrays n: 4 . \\ 
Set the maximum time slots t: 3 . \\ 
Initialize an array called 'result': [False, False, False, False]. \\ 
Initialize an array called 'job': [-1, -1, -1, -1]. \\ 
Step 1: Set i = 0. \\ 
Step 2: Set j = min(t, d[i]). \\ 
Step 3: Is result[j] False? \\ 
\quad i. If result[j] is False, set result[j] to True, assign job[j] = i, print the array of result and job, and proceed to Step 5. \\ 
\quad ii. If result[j] is True, proceed to Step 4. \\ 
Step 4: Decrement j by 1. If j > -1, go back to Step 3; otherwise, j <= -1, proceed to Step 5. \\ 
Step 5: Increment i by 1. If i < 4, go back to Step 2; otherwise, i >= 4, proceed to Step 6. \\ 
Step 6: Print the 'job' array. \\ 
----- \\ 
The computation procedure is as follows, with each line numbered sequentially (e.g., 1. 2. 3. 4. 5. 6. 7. 8. 9. 10. etc.): \\ 
1. i = 0. \\ 
\rowcolor[HTML]{C7EDCC} 2. j = min(3, D[0]) = min(3, 0) = 0. \\ 
\rowcolor[HTML]{C7EDCC} 3. result[0] is False, so set result[0] to True, job[0] = 0, print result: [True, False, False, False], print job: [0, -1, -1, -1]. \\ 
\rowcolor[HTML]{C7EDCC} 4. i = 1. \\ 
\rowcolor[HTML]{C7EDCC} 5. j = min(3, D[1]) = min(3, 3) = 3. \\ 
\rowcolor[HTML]{C7EDCC} 6. result[3] is False, so set result[3] to True, job[3] = 1, print result: [True, False, False, True], print job: [0, -1, -1, 1]. \\ 
\rowcolor[HTML]{C7EDCC} 7. i = 2. \\ 
\rowcolor[HTML]{C7EDCC} 8. j = min(3, D[2]) = min(3, 1) = 1. \\ 
\rowcolor[HTML]{C7EDCC} 9. result[1] is False, so set result[1] to True, job[1] = 2, print result: [True, True, False, True], print job: [0, 2, -1, 1]. \\ 
\rowcolor[HTML]{C7EDCC} 10. i = 3. \\ 
\rowcolor[HTML]{C7EDCC} 11. j = min(3, D[3]) = min(3, 0) = 0. \\ 
\rowcolor[HTML]{C7EDCC} 12. result[0] is True, so proceed to Step 4. \\ 
\rowcolor[HTML]{C7EDCC} 13. j = 0 - 1 = -1. \\ 
\rowcolor[HTML]{C7EDCC} 14. Since j <= -1, proceed to Step 5. \\ 
\rowcolor[HTML]{C7EDCC} 15. i = 4. \\ 
\rowcolor[HTML]{C7EDCC} 16. Since i >= 4, proceed to Step 6. \\ 
\rowcolor[HTML]{C7EDCC} 17. Print the 'job' array: [0, 2, -1, 1].
\\ \bottomrule
\end{tabular}
\caption{The prompt of task scheduling and the response of \texttt{gpt-4}.}
\label{table:clrs_mini/task_scheduling}

\end{table*}

\begin{table*}[h]
\small
\begin{tabular}{p{15.5cm}}
\toprule
\textbf{Prompt} \\ \midrule
Execute the instructions step by step. Do not jump steps. Lazy shortcuts like "Continue this process" or "Repeat Steps" is prohibited, since insufficient computation would result in wrong answer. Do not stop before completion. \\ 
Define two input sequences A: A[1] = 'b' A[2] = 'c' A[3] = 'c' A[4] = 'b' A[5] = 'a', \\ 
and B: B[1] = 'c' B[2] = 'c' B[3] = 'a' B[4] = 'a'. \\ 
Create a 2D table, called f, of size (5+1) x (4+1) . \\ 
Step 1: Initialize the first row and the first column of the f table to 0. \\ 
Step 2: Set i = 1. \\ 
Step 3: Set j = 1. \\ 
Step 4: Compare the elements A[i] and B[j] , and answer the question: Are they equal or not? \\ 
        a. If yes: Increase. Step by step set f[i][j] to f[i-1][j-1] + 1. For example, "f[1][1] = f[1-1][1-1] + 1 = f[0][0] + 1 = 0 + 1 = 1." \\ 
        b. If no: Select. Step by step set f[i][j] to the maximum value of f[i-1][j] and f[i][j-1]. For example, "f[1][1] = max(f[1-1][1], f[1][1-1]) = max(f[0][1], f[1][0])= max(0, 0) = 0." \\ 
Step 5: Increment j by 1. If j <= 4, go to Step 4. If j > 5, increment i by 1, and go to Step 6. \\ 
Step 6: If i <= 5, go to Step 3. If i > 6, proceed to Step 7. \\ 
Step 7: Print the length of the LCS, which is the value of f[5][4]. \\ 
----- \\ 
The computation procedure is as follows, with each line numbered sequentially (e.g., 1. 2. 3. 4. 5. 6. etc.): \\ 
1. f[0][0] = 0, f[0][1] = 0, f[0][2] = 0, f[0][3] = 0, f[0][4] = 0.  \\ 
 f[1][0] = 0, f[2][0] = 0, f[3][0] = 0, f[4][0] = 0, f[5][0] = 0. \\ 
2. i = 1. \\ 
3. j = 1. \\ 
4. Copied A[1] = 'b' and B[1] = 'c' , is 'b' equal to 'c'? No. Select. f[1][1] = max(f[1-1][1], f[1][1-1]) = max(f[0][1], f[1][0])= max(0, 0) = 0. \\
\rowcolor[HTML]{C7EDCC} 5. j = 2. \\ 
\rowcolor[HTML]{C7EDCC} 6. A[1] = 'b' and B[2] = 'c', is 'b' equal to 'c'? No. Select. f[1][2] = max(f[1-1][2], f[1][2-1]) = max(f[0][2], f[1][1]) = max(0, 0) = 0. \\ 
\rowcolor[HTML]{C7EDCC} 7. j = 3. \\ 
\rowcolor[HTML]{C7EDCC} 8. A[1] = 'b' and B[3] = 'a', is 'b' equal to 'a'? No. Select. f[1][3] = max(f[1-1][3], f[1][3-1]) = max(f[0][3], f[1][2]) = max(0, 0) = 0. \\ 
\rowcolor[HTML]{C7EDCC} 9. j = 4. \\ 
\rowcolor[HTML]{C7EDCC} 10. A[1] = 'b' and B[4] = 'a', is 'b' equal to 'a'? No. Select. f[1][4] = max(f[1-1][4], f[1][4-1]) = max(f[0][4], f[1][3]) = max(0, 0) = 0. \\ 
\rowcolor[HTML]{C7EDCC} 11. i = 2. \\ 
\rowcolor[HTML]{C7EDCC} 12. j = 1. \\ 
\rowcolor[HTML]{C7EDCC} 13. A[2] = 'c' and B[1] = 'c', is 'c' equal to 'c'? Yes. Increase. f[2][1] = f[2-1][1-1] + 1 = f[1][0] + 1 = 0 + 1 = 1. \\ 
\rowcolor[HTML]{C7EDCC} 14. j = 2. \\ 
\rowcolor[HTML]{C7EDCC} 15. A[2] = 'c' and B[2] = 'c', is 'c' equal to 'c'? Yes. Increase. f[2][2] = f[2-1][2-1] + 1 = f[1][1] + 1 = 0 + 1 = 1. \\ 
\rowcolor[HTML]{C7EDCC} 16. j = 3. \\ 
\rowcolor[HTML]{C7EDCC} 17. A[2] = 'c' and B[3] = 'a', is 'c' equal to 'a'? No. Select. f[2][3] = max(f[2-1][3], f[2][3-1]) = max(f[1][3], f[2][2]) = max(0, 1) = 1. \\ 
\rowcolor[HTML]{C7EDCC} 18. j = 4. \\ 
\rowcolor[HTML]{C7EDCC} 19. A[2] = 'c' and B[4] = 'a', is 'c' equal to 'a'? No. Select. f[2][4] = max(f[2-1][4], f[2][4-1]) = max(f[1][4], f[2][3]) = max(0, 1) = 1. \\ 
\rowcolor[HTML]{C7EDCC} 20. i = 3. \\ 
\rowcolor[HTML]{C7EDCC} 21. j = 1. \\ 
\rowcolor[HTML]{C7EDCC} 22. A[3] = 'c' and B[1] = 'c', is 'c' equal to 'c'? Yes. Increase. f[3][1] = f[3-1][1-1] + 1 = f[2][0] + 1 = 0 + 1 = 1. \\ 
\rowcolor[HTML]{C7EDCC} 23. j = 2. \\ 
\rowcolor[HTML]{C7EDCC} 24. A[3] = 'c' and B[2] = 'c', is 'c' equal to 'c'? Yes. Increase. f[3][2] = f[3-1][2-1] + 1 = f[2][1] + 1 = 1 + 1 = 2. \\ 
\rowcolor[HTML]{C7EDCC} 25. j = 3. \\ 
\rowcolor[HTML]{C7EDCC} 26. A[3] = 'c' and B[3] = 'a', is 'c' equal to 'a'? No. Select. f[3][3] = max(f[3-1][3], f[3][3-1]) = max(f[2][3], f[3][2]) = max(1, 2) = 2. \\ 
\rowcolor[HTML]{C7EDCC} 27. j = 4. \\ 
\rowcolor[HTML]{C7EDCC} 28. A[3] = 'c' and B[4] = 'a', is 'c' equal to 'a'? No. Select. f[3][4] = max(f[3-1][4], f[3][4-1]) = max(f[2][4], f[3][3]) = max(1, 2) = 2. \\ 
\rowcolor[HTML]{C7EDCC} 29. i = 4. \\ 
\rowcolor[HTML]{C7EDCC} 30. j = 1. \\ 
\rowcolor[HTML]{C7EDCC} 31. A[4] = 'b' and B[1] = 'c', is 'b' equal to 'c'? No. Select. f[4][1] = max(f[4-1][1], f[4][1-1]) = max(f[3][1], f[4][0]) = max(1, 0) = 1. \\ 
\rowcolor[HTML]{C7EDCC} 32. j = 2. \\ 
\bottomrule
\end{tabular}
\end{table*}
\begin{table*}[!t]
\small
\begin{tabular}{p{15.5cm}}
\toprule
\rowcolor[HTML]{C7EDCC} 33. A[4] = 'b' and B[2] = 'c', is 'b' equal to 'c'? No. Select. f[4][2] = max(f[4-1][2], f[4][2-1]) = max(f[3][2], f[4][1]) = max(2, 1) = 2. \\ 
\rowcolor[HTML]{C7EDCC} 34. j = 3. \\ 
\rowcolor[HTML]{C7EDCC} 35. A[4] = 'b' and B[3] = 'a', is 'b' equal to 'a'? No. Select. f[4][3] = max(f[4-1][3], f[4][3-1]) = max(f[3][3], f[4][2]) = max(2, 2) = 2. \\ 
\rowcolor[HTML]{C7EDCC} 36. j = 4. \\ 
\rowcolor[HTML]{C7EDCC} 37. A[4] = 'b' and B[4] = 'a', is 'b' equal to 'a'? No. Select. f[4][4] = max(f[4-1][4], f[4][4-1]) = max(f[3][4], f[4][3]) = max(2, 2) = 2. \\ 
\rowcolor[HTML]{C7EDCC} 38. i = 5. \\ 
\rowcolor[HTML]{C7EDCC} 39. j = 1. \\ 
\rowcolor[HTML]{C7EDCC} 40. A[5] = 'a' and B[1] = 'c', is 'a' equal to 'c'? No. Select. f[5][1] = max(f[5-1][1], f[5][1-1]) = max(f[4][1], f[5][0]) = max(1, 0) = 1. \\ 
\rowcolor[HTML]{C7EDCC} 41. j = 2. \\ 
\rowcolor[HTML]{C7EDCC} 42. A[5] = 'a' and B[2] = 'c', is 'a' equal to 'c'? No. Select. f[5][2] = max(f[5-1][2], f[5][2-1]) = max(f[4][2], f[5][1]) = max(2, 1) = 2. \\ 
\rowcolor[HTML]{C7EDCC} 43. j = 3. \\ 
\rowcolor[HTML]{C7EDCC} 44. A[5] = 'a' and B[3] = 'a', is 'a' equal to 'a'? Yes. Increase. f[5][3] = f[5-1][3-1] + 1 = f[4][2] + 1 = 2 + 1 = 3. \\ 
\rowcolor[HTML]{C7EDCC} 45. j = 4. \\ 
\rowcolor[HTML]{C7EDCC} 46. A[5] = 'a' and B[4] = 'a', is 'a' equal to 'a'? Yes. Increase. f[5][4] = f[5-1][4-1] + 1 = f[4][3] + 1 = 2 + 1 = 3. \\ 
\rowcolor[HTML]{C7EDCC} 47. The length of the LCS is f[5][4] = 3.
\\ \bottomrule
\end{tabular}
\caption{The prompt of longest common subsequence and the response of \texttt{gpt-4}.}
\label{table:iteration_flow_lcs}

\end{table*}

\begin{table*}[h]
\small
\begin{tabular}{p{15.5cm}}
\toprule
\textbf{Prompt} \\ \midrule
Execute the instructions step by step. Do not jump steps. Do not stop before completion. \\ 
Initial: Set the input array A: [2, 3, 2, 9] . \\ 
Set the length of the array n: 4 . \\ 
Set i = 0. \\ 
Step 1: What is the value of i and A[i]? Print them. \\ 
Step 2: Set j = 0. Compute max\_j = n - i - 1. \\ 
Step 3: Compare A[j] and A[j+1]. Are they in the correct order? Apply the following rule: A[j] <= A[j+1] is the correct order, A[j] > A[j+1] is the wrong order. \\ 
\quad i. If A[j] and A[j+1] are in the correct order, continue. \\ 
\quad ii. If A[j] and A[j+1] are in the wrong order, swap A[j] and A[j+1] to get the updated array A. \\ 
Step 4: Increment j by 1. If j < max\_j, go back to Step 3. Otherwise, proceed to Step 5. \\ 
Step 5: Increment i by 1. If i < 3, go back to Step 1. Otherwise, proceed to Step 6. \\ 
Step 6: Print the sorted array A. \\ 
----- \\ 
The computation procedure is as follows, with each line numbered sequentially (e.g., 1. 2. 3. 4. 5. 6. etc.): \\ 
1. i = 0, A[0] = 2. \\ 
\rowcolor[HTML]{C7EDCC} 2. j = 0, max\_j = 4 - 0 - 1 = 3. \\ 
\rowcolor[HTML]{C7EDCC} 3. A[0] = 2, A[1] = 3, correct order. \\ 
\rowcolor[HTML]{C7EDCC} 4. j = 1, max\_j = 3. \\ 
\rowcolor[HTML]{C7EDCC} 5. A[1] = 3, A[2] = 2, wrong order, swap A[1] and A[2], A = [2, 2, 3, 9]. \\ 
\rowcolor[HTML]{C7EDCC} 6. j = 2, max\_j = 3. \\ 
\rowcolor[HTML]{C7EDCC} 7. A[2] = 3, A[3] = 9, correct order. \\ 
\rowcolor[HTML]{C7EDCC} 8. i = 1, A[1] = 2. \\ 
\rowcolor[HTML]{C7EDCC} 9. j = 0, max\_j = 4 - 1 - 1 = 2. \\ 
\rowcolor[HTML]{C7EDCC} 10. A[0] = 2, A[1] = 2, correct order. \\ 
\rowcolor[HTML]{C7EDCC} 11. j = 1, max\_j = 2. \\ 
\rowcolor[HTML]{C7EDCC} 12. A[1] = 2, A[2] = 3, correct order. \\ 
\rowcolor[HTML]{C7EDCC} 13. i = 2, A[2] = 3. \\ 
\rowcolor[HTML]{C7EDCC} 14. j = 0, max\_j = 4 - 2 - 1 = 1. \\ 
\rowcolor[HTML]{C7EDCC} 15. A[0] = 2, A[1] = 2, correct order. \\ 
\rowcolor[HTML]{C7EDCC} 16. i = 3, A[3] = 9. \\ 
\rowcolor[HTML]{C7EDCC} 17. Sorted array A = [2, 2, 3, 9].
\\ \bottomrule
\end{tabular}
\caption{The prompt of bubble sort and the response of \texttt{gpt-4}.}
\label{table:clrs_mini/bubble_sort}

\end{table*}
\begin{table*}[h]
\small
\begin{tabular}{p{15.5cm}}
\toprule
\textbf{Prompt} \\ \midrule
Execute the instructions step by step. Do not jump steps. Do not stop before completion. \\ 
Initial: Set the list of sortable items A: [7, 2, 9, 1] . \\ 
Set N = 4. \\ 
Define the function heapify(A, N, i) to maintain the heap property. \\ 
Function heapify(A, N, i): \\ 
1. Set largest as i (root). Print N. \\ 
2. Calculate the left child index l: 2 * i + 1. \\ 
3. Calculate the right child index r: 2 * i + 2. \\ 
4. Check if the left child exists (l < N) and if its value A[l] is greater than the root A[largest]. If true, set largest as l. \\ 
5. Check if the right child exists (r < N) and if its value A[r] is greater than the root A[largest]. If true, set largest as r. \\ 
6. If largest is not equal to i, perform the following steps: \\ 
  a. Swap the values of A[i] and A[largest]. \\ 
  b. Call the heapify function recursively with arguments (A, N, largest). \\ 
End of function. \\ 
 \\ 
Step 1: Build a max heap from the input array A. \\ 
i. Iterate through the array from index 1 to 0 (inclusive) in reverse order. \\ 
ii. For each index i, call the heapify function with arguments (A, N, i). \\ 
Step 2: Sort the array using the max heap. \\ 
i. Iterate through the array from index 3 to 1 (inclusive) in reverse order. \\ 
ii. For each index i, perform the following steps: \\ 
  a. Swap the first element of the array (A[0]) with the current element (A[i]). \\ 
  b. Call the heapify function with arguments (A, i, 0). \\ 
Step 3: Print the array A. \\ 
----- \\ 
The computation procedure is as follows, with each line numbered sequentially (e.g., 1. 2. 3. 4. 5. 6. etc.): \\ 
1. i = 1, calling heapify(A, 4, 1). \\ 
\rowcolor[HTML]{C7EDCC} 2. In heapify(A, 4, 1): N = 4, largest = 1, l = 3, r = 4. \\ 
\rowcolor[HTML]{C7EDCC} 3. l < N, A[l] = 1, A[largest] = 2, largest remains 1. \\ 
\rowcolor[HTML]{C7EDCC} 4. r >= N, skip the check for the right child. \\ 
\rowcolor[HTML]{C7EDCC} 5. largest = 1, no need to swap or call heapify recursively. \\ 
\rowcolor[HTML]{C7EDCC} 6. i = 0, calling heapify(A, 4, 0). \\ 
\rowcolor[HTML]{C7EDCC} 7. In heapify(A, 4, 0): N = 4, largest = 0, l = 1, r = 2. \\ 
\rowcolor[HTML]{C7EDCC} 8. l < N, A[l] = 2, A[largest] = 7, largest remains 0. \\ 
\rowcolor[HTML]{C7EDCC} 9. r < N, A[r] = 9, A[largest] = 7, largest is updated to 2. \\ 
\rowcolor[HTML]{C7EDCC} 10. largest != 0, swap A[0] and A[2], A becomes [9, 2, 7, 1]. \\ 
\rowcolor[HTML]{C7EDCC} 11. Call heapify(A, 4, 2) recursively. \\ 
\rowcolor[HTML]{C7EDCC} 12. In heapify(A, 4, 2): N = 4, largest = 2, l = 5, r = 6. \\ 
\rowcolor[HTML]{C7EDCC} 13. l >= N, skip the check for the left child. \\ 
\rowcolor[HTML]{C7EDCC} 14. r >= N, skip the check for the right child. \\ 
\rowcolor[HTML]{C7EDCC} 15. largest = 2, no need to swap or call heapify recursively. \\ 
\rowcolor[HTML]{C7EDCC} 16. Max heap is built, A = [9, 2, 7, 1]. \\ 
\rowcolor[HTML]{C7EDCC} 17. i = 3, swap A[0] and A[3], A becomes [1, 2, 7, 9]. \\ 
\rowcolor[HTML]{C7EDCC} 18. Call heapify(A, 3, 0). \\ 
\rowcolor[HTML]{C7EDCC} 19. In heapify(A, 3, 0): N = 3, largest = 0, l = 1, r = 2. \\ 
\rowcolor[HTML]{C7EDCC} 20. l < N, A[l] = 2, A[largest] = 1, largest is updated to 1. \\ 
\rowcolor[HTML]{C7EDCC} 21. r < N, A[r] = 7, A[largest] = 2, largest is updated to 2. \\ 
\rowcolor[HTML]{C7EDCC} 22. largest != 0, swap A[0] and A[2], A becomes [7, 2, 1, 9]. \\ 
\rowcolor[HTML]{C7EDCC} 23. Call heapify(A, 3, 2) recursively. \\ 
\rowcolor[HTML]{C7EDCC} 24. In heapify(A, 3, 2): N = 3, largest = 2, l = 5, r = 6. \\ 
\rowcolor[HTML]{C7EDCC} 25. l >= N, skip the check for the left child. \\ 
\rowcolor[HTML]{C7EDCC} 26. r >= N, skip the check for the right child. \\ 
\rowcolor[HTML]{C7EDCC} 27. largest = 2, no need to swap or call heapify recursively. \\ 
\rowcolor[HTML]{C7EDCC} 28. i = 2, swap A[0] and A[2], A becomes [1, 2, 7, 9]. \\ 
\rowcolor[HTML]{C7EDCC} 29. Call heapify(A, 2, 0). \\ 
\rowcolor[HTML]{C7EDCC} 30. In heapify(A, 2, 0): N = 2, largest = 0, l = 1, r = 2. \\ 
\rowcolor[HTML]{C7EDCC} 31. l < N, A[l] = 2, A[largest] = 1, largest is updated to 1. \\ 
\rowcolor[HTML]{C7EDCC} 32. r >= N, skip the check for the right child. \\ 
\rowcolor[HTML]{C7EDCC} 33. largest != 0, swap A[0] and A[1], A becomes [2, 1, 7, 9]. \\ 
\rowcolor[HTML]{C7EDCC} 34. Call heapify(A, 2, 1) recursively. \\ 
\bottomrule
\end{tabular}
\end{table*}
\begin{table*}[!ht]
\small
\begin{tabular}{p{15.5cm}}
\toprule
\rowcolor[HTML]{C7EDCC} 35. In heapify(A, 2, 1): N = 2, largest = 1, l = 3, r = 4. \\ 
\rowcolor[HTML]{C7EDCC} 36. l >= N, skip the check for the left child. \\ 
\rowcolor[HTML]{C7EDCC} 37. r >= N, skip the check for the right child. \\ 
\rowcolor[HTML]{C7EDCC} 38. largest = 1, no need to swap or call heapify recursively. \\ 
\rowcolor[HTML]{C7EDCC} 39. i = 1, swap A[0] and A[1], A becomes [1, 2, 7, 9]. \\ 
\rowcolor[HTML]{C7EDCC} 40. Call heapify(A, 1, 0). \\
\rowcolor[HTML]{C7EDCC} 41. In heapify(A, 1, 0): N = 1, largest = 0, l = 1, r = 2. \\ 
\rowcolor[HTML]{C7EDCC} 42. l >= N, skip the check for the left child. \\ 
\rowcolor[HTML]{C7EDCC} 43. r >= N, skip the check for the right child. \\ 
\rowcolor[HTML]{C7EDCC} 44. largest = 0, no need to swap or call heapify recursively. \\ 
\rowcolor[HTML]{C7EDCC} 45. The array A is now sorted: [1, 2, 7, 9].
\\ \bottomrule
\end{tabular}
\caption{The prompt of heapsort and the response of \texttt{gpt-4}.}
\label{table:clrs_mini/heapsort}

\end{table*}
\begin{table*}[h]
\small
\begin{tabular}{p{15.5cm}}
\toprule
\textbf{Prompt} \\ \midrule
Execute the instructions step by step. Do not jump steps. Do not stop before completion. \\ 
Initial: Set the list of sortable items A: [2, 8, 8, 7] . \\ 
Set N = 4. \\ 
Create a function called "partition" that takes three arguments: array, low, and high. \\ 
Create a function called "quicksort" that takes three arguments: array, low, and high. \\ 
 \\ 
Step 1: Call the "quicksort" function with A, 0, and N-1 as arguments. \\ 
Inside the "quicksort" function, follow these steps: \\ 
 \\ 
Step 1.1: Check if low is less than high. If not, return. \\ 
Step 1.2: Call the "partition" function with array, low, and high as arguments, and store the result in a variable called "pi". \\ 
Step 1.3: Recursively call the "quicksort" function with array, low, and pi - 1 as arguments. \\ 
Step 1.4: Recursively call the "quicksort" function with array, pi + 1, and high as arguments. \\ 
 \\ 
Inside the "partition" function, follow these steps: \\ 
 \\ 
Step 2.1: Set the pivot as the element at the "high" index of the array, which is A[high]. \\ 
Step 2.2: Set a pointer "i" to low - 1. \\ 
Step 2.3: Iterate through the array from the "low" index to the "high" index (exclusive) using a variable "j". \\ 
Step 2.4: Compare the element at index "j" A[j] with the pivot. If it is less than or equal to the pivot, follow these steps: \\ 
  a. Increment the pointer "i" by 1. \\ 
  b. Swap the elements at index "i" A[i] and "j" A[j] in the array. \\ 
Step 2.5: After the iteration, swap the pivot element A[high] with the element at index "i + 1" A[i + 1] in the array. \\ 
Step 2.6: Return the value of "i + 1" as the partition position. \\ 
 \\ 
Step 3: Print the array A. \\ 
----- \\ 
The computation procedure is as follows, with each line numbered sequentially (e.g., 1. 2. 3. 4. 5. 6. etc.): \\ 
1. Call quicksort(A, 0, 3). \\ 
\rowcolor[HTML]{C7EDCC} 2. Inside quicksort, low (0) is less than high (3), so continue. \\ 
\rowcolor[HTML]{C7EDCC} 3. Call partition(A, 0, 3) and store the result in pi. \\ 
\rowcolor[HTML]{C7EDCC} 4. Inside partition, set pivot = A[3] = 7, set i = -1. \\ 
\rowcolor[HTML]{C7EDCC} 5. Iterate through A from low (0) to high (3) exclusive with j: \\ 
\rowcolor[HTML]{C7EDCC}    a. j = 0, A[j] = 2 <= pivot (7), increment i (i = 0), swap A[i] and A[j] (no change). \\ 
\rowcolor[HTML]{C7EDCC}    b. j = 1, A[j] = 8 > pivot (7), do nothing. \\ 
\rowcolor[HTML]{C7EDCC}    c. j = 2, A[j] = 8 > pivot (7), do nothing. \\ 
\rowcolor[HTML]{C7EDCC} 6. After iteration, swap A[i + 1] and A[high], i.e., A[1] and A[3], resulting in A = [2, 7, 8, 8]. \\ 
\rowcolor[HTML]{C7EDCC} 7. Return i + 1 = 1 as the partition position (pi). \\ 
\rowcolor[HTML]{C7EDCC} 8. Recursively call quicksort(A, 0, pi - 1) which is quicksort(A, 0, 0). \\ 
\rowcolor[HTML]{C7EDCC} 9. Inside the recursive quicksort, low (0) is not less than high (0), so return. \\ 
\rowcolor[HTML]{C7EDCC} 10. Recursively call quicksort(A, pi + 1, 3) which is quicksort(A, 2, 3). \\ 
\rowcolor[HTML]{C7EDCC} 11. Inside the recursive quicksort, low (2) is less than high (3), so continue. \\ 
\rowcolor[HTML]{C7EDCC} 12. Call partition(A, 2, 3) and store the result in pi. \\ 
\rowcolor[HTML]{C7EDCC} 13. Inside partition, set pivot = A[3] = 8, set i = 1. \\ 
\rowcolor[HTML]{C7EDCC} 14. Iterate through A from low (2) to high (3) exclusive with j: \\ 
\rowcolor[HTML]{C7EDCC} \quad a. j = 2, A[j] = 8 <= pivot (8), increment i (i = 2), swap A[i] and A[j] (no change). \\ 
\rowcolor[HTML]{C7EDCC} 15. After iteration, swap A[i + 1] and A[high], i.e., A[3] and A[3] (no change). \\ 
\rowcolor[HTML]{C7EDCC} 16. Return i + 1 = 3 as the partition position (pi). \\ 
\rowcolor[HTML]{C7EDCC} 17. Recursively call quicksort(A, 2, pi - 1) which is quicksort(A, 2, 2). \\ 
\rowcolor[HTML]{C7EDCC} 18. Inside the recursive quicksort, low (2) is not less than high (2), so return. \\ 
\rowcolor[HTML]{C7EDCC} 19. Recursively call quicksort(A, pi + 1, 3) which is quicksort(A, 4, 3). \\ 
\rowcolor[HTML]{C7EDCC} 20. Inside the recursive quicksort, low (4) is not less than high (3), so return. \\ 
\rowcolor[HTML]{C7EDCC} 21. Print the array A: [2, 7, 8, 8].
\\ \bottomrule
\end{tabular}
\caption{The prompt of quicksort and the response of \texttt{gpt-4}.}
\label{table:clrs_mini/quicksort}

\end{table*}
\begin{table*}[h]
\small
\begin{tabular}{p{15.5cm}}
\toprule
\textbf{Prompt} \\ \midrule
Execute the instructions step by step. Do not jump steps. Do not stop before completion. \\ 
Initial: Set the input array A: [7, 2, 6, 0, 3, 7, 9, 5, 6] . \\ 
Set the length of the array n: 9 . \\ 
Set min\_ = 0. \\ 
Set i = 1. \\ 
Step 1: What are the values of A[min\_] and A[i]? Print them. \\ 
Step 2: Compare A[min\_] and A[i]. Is A[min\_] greater than A[i]? \\ 
i. If yes, update min\_ as the current value of i. \\ 
ii. If no, continue without updating min\_. \\ 
Step 3: Increment i by 1. If i < 9, go back to Step 1. Otherwise, proceed to Step 4. \\ 
Step 4: What is the value of min\_? Print it. \\ 
----- \\ 
The computation procedure is as follows, with each line numbered sequentially (e.g., 1. 2. 3. 4. 5. 6. etc.): \\ 
1. min\_ = 0, A[0] = 7; i = 1, A[1] = 2. \\ 
\rowcolor[HTML]{C7EDCC} 2. Since A[min\_] > A[i], update min\_ = 1. \\ 
\rowcolor[HTML]{C7EDCC} 3. Increment i to 2, A[2] = 6. \\ 
\rowcolor[HTML]{C7EDCC} 4. A[min\_] = 2, A[i] = 6. A[min\_] is not greater than A[i], so do not update min\_. \\ 
\rowcolor[HTML]{C7EDCC} 5. Increment i to 3, A[3] = 0. \\ 
\rowcolor[HTML]{C7EDCC} 6. A[min\_] = 2, A[i] = 0. Since A[min\_] > A[i], update min\_ = 3. \\ 
\rowcolor[HTML]{C7EDCC} 7. Increment i to 4, A[4] = 3. \\ 
\rowcolor[HTML]{C7EDCC} 8. A[min\_] = 0, A[i] = 3. A[min\_] is not greater than A[i], so do not update min\_. \\ 
\rowcolor[HTML]{C7EDCC} 9. Increment i to 5, A[5] = 7. \\ 
\rowcolor[HTML]{C7EDCC} 10. A[min\_] = 0, A[i] = 7. A[min\_] is not greater than A[i], so do not update min\_. \\ 
\rowcolor[HTML]{C7EDCC} 11. Increment i to 6, A[6] = 9. \\ 
\rowcolor[HTML]{C7EDCC} 12. A[min\_] = 0, A[i] = 9. A[min\_] is not greater than A[i], so do not update min\_. \\ 
\rowcolor[HTML]{C7EDCC} 13. Increment i to 7, A[7] = 5. \\ 
\rowcolor[HTML]{C7EDCC} 14. A[min\_] = 0, A[i] = 5. A[min\_] is not greater than A[i], so do not update min\_. \\ 
\rowcolor[HTML]{C7EDCC} 15. Increment i to 8, A[8] = 6. \\ 
\rowcolor[HTML]{C7EDCC} 16. A[min\_] = 0, A[i] = 6. A[min\_] is not greater than A[i], so do not update min\_. \\ 
\rowcolor[HTML]{C7EDCC} 17. i = 9, which is equal to n, so proceed to Step 4. \\ 
\rowcolor[HTML]{C7EDCC} 18. The value of min\_ is 3.
\\ \bottomrule
\end{tabular}
\caption{The prompt of minimum and the response of \texttt{gpt-4}.}
\label{table:clrs_mini/minimum}

\end{table*}

% insertion_table
\begin{table*}[!t]
\small
\begin{tabular}{p{15.5cm}}
\toprule
\textbf{Prompt} \\ \midrule
Execute the instructions step by step. Do not jump steps. Do not stop before completion. \\ 
Initial: Set the list of sortable items A: [2, 8, 5, 7] . \\ 
Set n = 4. \\ 
Set i = 0. \\ 
Step 1: What is the value of i and A[i]? Print them. \\ 
Step 2: Set key = A[i] and j = i - 1. \\ 
Step 3: Is j >= 0 and A[j] > key? If yes, proceed to Step 4. If no, go to Step 6. \\ 
Step 4: Swap A[j+1] with A[j]. Print the updated A. \\ 
Step 5: Decrement j by 1. Go back to Step 3. \\ 
Step 6: Increment i by 1. If i < 4, go back to Step 1. Otherwise, proceed to Step 7. \\ 
Step 7: What is the final sorted array A? Print it. \\ 
----- \\ 
The computation procedure is as follows, with each line numbered sequentially (e.g., 1. 2. 3. 4. 5. 6. etc.): \\ 
1. i = 0, A[0] = 2. \\ 
\rowcolor[HTML]{C7EDCC} 2. key = 2, j = -1. (j is not >= 0, so go to Step 6) \\ 
\rowcolor[HTML]{C7EDCC} 3. i = 1, A[1] = 8. \\ 
\rowcolor[HTML]{C7EDCC} 4. key = 8, j = 0. (A[j] = 2, which is not > key, so go to Step 6) \\ 
\rowcolor[HTML]{C7EDCC} 5. i = 2, A[2] = 5. \\ 
\rowcolor[HTML]{C7EDCC} 6. key = 5, j = 1. (A[j] = 8, which is > key, so proceed to Step 4) \\ 
\rowcolor[HTML]{C7EDCC} 7. Swap A[1] and A[2]. Updated A: [2, 5, 8, 7]. \\ 
\rowcolor[HTML]{C7EDCC} 8. j = 0. (A[j] = 2, which is not > key, so go to Step 6) \\ 
\rowcolor[HTML]{C7EDCC} 9. i = 3, A[3] = 7. \\ 
\rowcolor[HTML]{C7EDCC} 10. key = 7, j = 2. (A[j] = 8, which is > key, so proceed to Step 4) \\ 
\rowcolor[HTML]{C7EDCC} 11. Swap A[2] and A[3]. Updated A: [2, 5, 7, 8]. \\ 
\rowcolor[HTML]{C7EDCC} 12. j = 1. (A[j] = 5, which is not > key, so go to Step 6) \\ 
\rowcolor[HTML]{C7EDCC} 13. i = 4. (i is not < 4, so proceed to Step 7) \\ 
\rowcolor[HTML]{C7EDCC} 14. The final sorted array A is [2, 5, 7, 8].
\\ \bottomrule
\end{tabular}\caption{The prompt of insertion sort and the response of \texttt{gpt-4}.}
\label{table:clrs_mini/insertion_sort}

\end{table*}

\begin{table*}[h]
\small
\begin{tabular}{p{15.5cm}}
\toprule
\textbf{Prompt} \\ \midrule
Execute the instructions step by step. Do not jump steps. Do not stop before completion. \\ 
Initial: Set the input array A: [9, 5, 0, 3] . \\ 
Set left index L: 0, right index R: 3, and the position k: 3. \\ 
Set i = L. \\ 
 \\ 
Define the partition function, with the parameter A, L, and R. \\ 
i. Set the pivot element X as A[R]. \\ 
ii. Set the index i as L. \\ 
iii. Iterate through the array from index L to R-1 with index j. \\ 
\quad a. Recall the value of A[j] and X. Is A[j] less than or equal to X ? If yes, swap A[i] and A[j], print A, and increment i by 1. \\ 
iv. Swap A[i] and A[R], print A. \\ 
v. Return the index i. \\ 
End of the partition function. \\ 
 \\ 
Define the kthSmallest function, with the parameter A, K, L, and R. \\ 
i. Call the partition function with arguments A, L, and R, and store the returned index in a variable called "index". \\ 
ii. Compute len\_left = index - L + 1. Are len\_left and K the same? \\ 
\quad a. If yes, return A[index]. \\ 
\quad b. If no, proceed to iii. \\ 
iii. Recall the value of len\_left and K. Is len\_left greater than K? \\ 
\quad a. If yes, recursively call the kthSmallest(A, L = L, R = index - 1, K = K). \\ 
\quad b. If no, recursively call the kthSmallest(A, L = index + 1, R = R, K =  K - len\_left). \\ 
End of the kthSmallest function. \\ 
 \\ 
Step 1: Call kthSmallest(A, L = L, R = R, K = K), and print the returned value. \\ 
----- \\ 
The computation procedure is as follows, with each line numbered sequentially (e.g., 1. 2. 3. 4. 5. 6. etc.): \\ 
1. Call kthSmallest(A, L = 0, R = 3, K = 3). \\ 
\rowcolor[HTML]{C7EDCC} 2. Call partition(A, L = 0, R = 3). \\ 
\rowcolor[HTML]{C7EDCC} 3. Set pivot element X = A[3] = 3. \\ 
\rowcolor[HTML]{C7EDCC} 4. Set index i = 0. \\ 
\rowcolor[HTML]{C7EDCC} 5. Iterate through the array from index 0 to 2 with index j. \\ 
\rowcolor[HTML]{C7EDCC} \quad a. j = 0, A[0] = 9, X = 3, A[0] is not less than or equal to X, no swap. \\ 
\rowcolor[HTML]{C7EDCC} \quad b. j = 1, A[1] = 5, X = 3, A[1] is not less than or equal to X, no swap. \\ 
\rowcolor[HTML]{C7EDCC} \quad c. j = 2, A[2] = 0, X = 3, A[2] is less than or equal to X, swap A[0] and A[2], A = [0, 5, 9, 3], increment i to 1. \\ 
\rowcolor[HTML]{C7EDCC} 6. Swap A[1] and A[3], A = [0, 3, 9, 5], return index i = 1. \\ 
\rowcolor[HTML]{C7EDCC} 7. len\_left = 1 - 0 + 1 = 2, len\_left is not equal to K, proceed to step 8. \\ 
\rowcolor[HTML]{C7EDCC} 8. len\_left = 2 is less than K = 3, call kthSmallest(A, L = 1 + 1, R = 3, K = 3 - 2). \\ 
\rowcolor[HTML]{C7EDCC} 9. Call partition(A, L = 2, R = 3). \\ 
\rowcolor[HTML]{C7EDCC} 10. Set pivot element X = A[3] = 5. \\ 
\rowcolor[HTML]{C7EDCC} 11. Set index i = 2. \\ 
\rowcolor[HTML]{C7EDCC} 12. Iterate through the array from index 2 to 2 with index j. \\ 
\rowcolor[HTML]{C7EDCC} \quad a. j = 2, A[2] = 9, X = 5, A[2] is not less than or equal to X, no swap. \\ 
\rowcolor[HTML]{C7EDCC} 13. Swap A[2] and A[3], A = [0, 3, 5, 9], return index i = 2. \\ 
\rowcolor[HTML]{C7EDCC} 14. len\_left = 2 - 2 + 1 = 1, len\_left is equal to K, return A[2] = 5. \\ 
\rowcolor[HTML]{C7EDCC}  \\ 
\rowcolor[HTML]{C7EDCC} The returned value is 5.
\\ \bottomrule
\end{tabular}
\caption{The prompt of quick select and the response of \texttt{gpt-4}.}
\label{table:clrs_mini/quick_select}

\end{table*}
\begin{table*}[h]
\small
\begin{tabular}{p{15.5cm}}
\toprule
\textbf{Prompt} \\ \midrule
Execute the instructions step by step. Do not jump steps. Do not stop before completion. \\ 
Initial: Set the input array A: A[0] = 8, A[1] = -3, A[2] = 6, A[3] = 3 . \\ 
Define the function maxSubArraySum with input parameters A, l and r, where arr is an integer array, l is its left index and r is its right index. \\ 
a: Check if the length of A, r - l + 1, is equal to 1 (base case). \\ 
\quad a.i. If yes, return the only element in arr. \\ 
\quad a.ii. If no, proceed to (b). \\ 
b: Calculate the midpoint m = l + (r - l) // 2. \\ 
c: Recursively call maxSubArraySum(A, l, m) and store the result in left\_max. \\ 
d: Recursively call maxSubArraySum(A, m+1, r) and store the result in right\_max. \\ 
e: Initialize left\_sum and right\_sum to -100. Initialize sum to 0. Initialize i to m+1. \\ 
f: Is i <= r? \\ 
\quad f.i. If i <= r, add A[i] to sum. Update right\_sum with max(right\_sum, sum). Increment i by 1. Go back to (f). \\ 
\quad f.ii. If i > r, proceed to (g). \\ 
g: Reset sum to 0. Compute m-1 and initialize i to m. \\ 
h: Is i >= l? \\ 
\quad f.i. If i >= l, add A[i] to sum. Update left\_sum with max(left\_sum, sum). Decrement i by 1. Go back to (h). \\ 
\quad f.ii. If i < l, proceed to (i). \\ 
i: Calculate cross\_max as the sum of left\_sum and right\_sum. \\ 
j: What is max(cross\_max, left\_max, right\_max)? Return the value. \\ 
End of function maxSubArraySum. \\ 
 \\ 
Step 1: Call the maxSubArraySum function with arguments A, 0, and 3, and print the result. \\ 
----- \\ 
The computation procedure is as follows, with each line numbered sequentially (e.g., 1. 2. 3. 4. 5. 6. etc.): \\ 
1. maxSubArraySum(A, l=0, r=3). \\ 
1.1. r - l + 1 = 3 - 0 + 1 = 4, 4 != 1. \\ 
\rowcolor[HTML]{C7EDCC} 1.2. m = 0 + (3 - 0) // 2 = 1. \\ 
\rowcolor[HTML]{C7EDCC} 1.3. left\_max = maxSubArraySum(A, l=0, r=1). \\ 
\rowcolor[HTML]{C7EDCC} 1.3.1. r - l + 1 = 1 - 0 + 1 = 2, 2 != 1. \\ 
\rowcolor[HTML]{C7EDCC} 1.3.2. m = 0 + (1 - 0) // 2 = 0. \\ 
\rowcolor[HTML]{C7EDCC} 1.3.3. left\_max = maxSubArraySum(A, l=0, r=0). \\ 
\rowcolor[HTML]{C7EDCC} 1.3.3.1. r - l + 1 = 0 - 0 + 1 = 1, 1 == 1, return A[0] = 8. \\ 
\rowcolor[HTML]{C7EDCC} 1.3.4. right\_max = maxSubArraySum(A, l=1, r=1). \\ 
\rowcolor[HTML]{C7EDCC} 1.3.4.1. r - l + 1 = 1 - 1 + 1 = 1, 1 == 1, return A[1] = -3. \\ 
\rowcolor[HTML]{C7EDCC} 1.3.5. left\_sum = -100, right\_sum = -100, sum = 0, i = 1. \\ 
\rowcolor[HTML]{C7EDCC} 1.3.6. i <= r, sum += A[1] = -3, right\_sum = max(-100, -3) = -3, i = 2. \\ 
\rowcolor[HTML]{C7EDCC} 1.3.7. i > r, sum = 0, i = 0. \\ 
\rowcolor[HTML]{C7EDCC} 1.3.8. i >= l, sum += A[0] = 8, left\_sum = max(-100, 8) = 8, i = -1. \\ 
\rowcolor[HTML]{C7EDCC} 1.3.9. i < l, cross\_max = left\_sum + right\_sum = 8 + (-3) = 5. \\ 
\rowcolor[HTML]{C7EDCC} 1.3.10. max(5, 8, -3) = 8, left\_max = 8. \\ 
\rowcolor[HTML]{C7EDCC} 1.4. right\_max = maxSubArraySum(A, l=2, r=3). \\ 
\rowcolor[HTML]{C7EDCC} 1.4.1. r - l + 1 = 3 - 2 + 1 = 2, 2 != 1. \\ 
\rowcolor[HTML]{C7EDCC} 1.4.2. m = 2 + (3 - 2) // 2 = 2. \\ 
\rowcolor[HTML]{C7EDCC} 1.4.3. left\_max = maxSubArraySum(A, l=2, r=2). \\ 
\rowcolor[HTML]{C7EDCC} 1.4.3.1. r - l + 1 = 2 - 2 + 1 = 1, 1 == 1, return A[2] = 6. \\ 
\rowcolor[HTML]{C7EDCC} 1.4.4. right\_max = maxSubArraySum(A, l=3, r=3). \\ 
\rowcolor[HTML]{C7EDCC} 1.4.4.1. r - l + 1 = 3 - 3 + 1 = 1, 1 == 1, return A[3] = 3. \\ 
\rowcolor[HTML]{C7EDCC} 1.4.5. left\_sum = -100, right\_sum = -100, sum = 0, i = 3. \\ 
\rowcolor[HTML]{C7EDCC} 1.4.6. i <= r, sum += A[3] = 3, right\_sum = max(-100, 3) = 3, i = 4. \\ 
\rowcolor[HTML]{C7EDCC} 1.4.7. i > r, sum = 0, i = 2. \\ 
\rowcolor[HTML]{C7EDCC} 1.4.8. i >= l, sum += A[2] = 6, left\_sum = max(-100, 6) = 6, i = 1. \\ 
\rowcolor[HTML]{C7EDCC} 1.4.9. i < l, cross\_max = left\_sum + right\_sum = 6 + 3 = 9. \\ 
\rowcolor[HTML]{C7EDCC} 1.4.10. max(9, 6, 3) = 9, right\_max = 9. \\ 
\rowcolor[HTML]{C7EDCC} 1.5. left\_sum = -100, right\_sum = -100, sum = 0, i = 2. \\ 
\rowcolor[HTML]{C7EDCC} 1.6. i <= r, sum += A[2] = 6, right\_sum = max(-100, 6) = 6, i = 3. \\ 
\rowcolor[HTML]{C7EDCC} 1.7. i <= r, sum += A[3] = 9, right\_sum = max(6, 9) = 9, i = 4. \\ 
\rowcolor[HTML]{C7EDCC} 1.8. i > r, sum = 0, i = 1. \\ 
\rowcolor[HTML]{C7EDCC} 1.9. i >= l, sum += A[1] = -3, left\_sum = max(-100, -3) = -3, i = 0. \\ 
\rowcolor[HTML]{C7EDCC} 1.10. i >= l, sum += A[0] = 5, left\_sum = max(-3, 5) = 5, i = -1. \\ 
\rowcolor[HTML]{C7EDCC} 1.11. i < l, cross\_max = left\_sum + right\_sum = 5 + 9 = 14. \\ 
\bottomrule
\end{tabular}
\end{table*}
\begin{table*}[h]
\small
\begin{tabular}{p{15.5cm}}
\toprule
\rowcolor[HTML]{C7EDCC} 1.12. max(14, 8, 9) = 14. \\ 
\rowcolor[HTML]{C7EDCC}  \\ 
\rowcolor[HTML]{C7EDCC} The result of maxSubArraySum(A, 0, 3) is 14.
\\ \bottomrule
\end{tabular}
\caption{The prompt of maximum subarray and the response of \texttt{gpt-4}.}
\label{table:clrs_mini/maximum_subarray}

\end{table*}
% \clearpage
\begin{table*}[]
\small
\begin{tabular}{p{15.5cm}}
\toprule
\textbf{Prompt} \\ \midrule
Execute the instructions step by step. Do not jump steps. Do not stop before completion. \\ 
Initial: Set the square matrix A of size 6 x 6 : \\ 
A[0, 0] = 0, A[0, 1] = 1, A[0, 2] = 0, A[0, 3] = 1, A[0, 4] = 0, A[0, 5] = 0 \\ 
A[1, 0] = 0, A[1, 1] = 0, A[1, 2] = 0, A[1, 3] = 0, A[1, 4] = 1, A[1, 5] = 0 \\ 
A[2, 0] = 0, A[2, 1] = 0, A[2, 2] = 0, A[2, 3] = 0, A[2, 4] = 1, A[2, 5] = 1 \\ 
A[3, 0] = 0, A[3, 1] = 1, A[3, 2] = 0, A[3, 3] = 0, A[3, 4] = 0, A[3, 5] = 0 \\ 
A[4, 0] = 0, A[4, 1] = 0, A[4, 2] = 0, A[4, 3] = 1, A[4, 4] = 0, A[4, 5] = 0 \\ 
A[5, 0] = 0, A[5, 1] = 0, A[5, 2] = 0, A[5, 3] = 0, A[5, 4] = 0, A[5, 5] = 1. \\ 
Set color: color[0] = 0, color[1] = 0, color[2] = 0, color[3] = 0, color[4] = 0, color[5] = 0. \\ 
Set d: d[0] = 0, d[1] = 0, d[2] = 0, d[3] = 0, d[4] = 0, d[5] = 0. \\ 
Set pi: pi[0] = 0, pi[1] = 1, pi[2] = 2, pi[3] = 3, pi[4] = 4, pi[5] = 5. \\ 
Set s\_prev: s\_prev[0] = 0, s\_prev[1] = 1, s\_prev[2] = 2, s\_prev[3] = 3, s\_prev[4] = 4, s\_prev[5] = 5. \\ 
Set time: 0. \\ 
Set s: 0. \\ 
 \\ 
Step 1: Is s < 6? If s is equal to 6, go to Step 6. Otherwise: \\ 
  i. If color[s] == 0, set s\_last = s, u = s, and v = s, go to Step 2. \\ 
  ii. If color[s] != 0, increment s by 1, and go to Step 1. \\ 
 \\ 
Step 2: Check if color[u] == 0 or d[u] == 0. \\ 
  i. If true, increment time by 1, set d[u] = time, and set color[u] = 1. \\ 
 \\ 
Step 3: Iterate through each element v in range(N). \\ 
  i. If A[u, v] != 0, check if color[v] == 0. \\ 
\quad a. If true, set pi[v] = u, color[v] = 1, s\_prev[v] = s\_last, and s\_last = v. Then, break the iteration and go to Step 4. \\ 
\quad b. If false, continue. \\ 
  ii. If A[u, v] == 0, continue \\ 
 \\ 
Step 4: What are the values of s\_last and u? Is s\_last equal to u? \\ 
  i. If true, set color[u] = 2 and increment time by 1. Then, What is the value of s\_prev[u]? Is s\_prev[u] equal to u? \\ 
\quad a. If true, assert s\_prev[s\_last] == s\_last, increment s by 1 and go to step 1. \\ 
\quad b. If false, set pr = s\_prev[s\_last], s\_prev[s\_last] = s\_last, and s\_last = pr. \\ 
  ii. If false, proceed to Step 5. \\ 
 \\ 
Step 5: Set u = s\_last and go to Step 2. \\ 
 \\ 
Step 6: Print the array pi. \\ 
----- \\ 
The computation procedure is as follows, with each line numbered sequentially (e.g., 1. 2. 3. 4. 5. 6. etc.): \\ 
1. S = 0, 0 < 6. \\ 
\rowcolor[HTML]{C7EDCC} 2. color[0] == 0, set s\_last = 0, u = 0, and v = 0. \\ 
\rowcolor[HTML]{C7EDCC}  \\ 
\rowcolor[HTML]{C7EDCC} 3. Increment time by 1, time = 1, set d[0] = 1, and set color[0] = 1. \\ 
\rowcolor[HTML]{C7EDCC}  \\ 
\rowcolor[HTML]{C7EDCC} 4. Iterate through each element v in range(6): \\ 
\rowcolor[HTML]{C7EDCC}    - v = 0, A[0, 0] = 0, continue. \\ 
\rowcolor[HTML]{C7EDCC}    - v = 1, A[0, 1] = 1, color[1] = 0, set pi[1] = 0, color[1] = 1, s\_prev[1] = 0, s\_last = 1, break the iteration. \\ 
\rowcolor[HTML]{C7EDCC}  \\ 
\rowcolor[HTML]{C7EDCC} 5. s\_last = 1, u = 0, s\_last != u, proceed to Step 5. \\ 
\rowcolor[HTML]{C7EDCC}  \\ 
\rowcolor[HTML]{C7EDCC} 6. Set u = 1 and go to Step 2. \\ 
\rowcolor[HTML]{C7EDCC}  \\ 
\rowcolor[HTML]{C7EDCC} 7. color[1] == 1, d[1] == 0, increment time by 1, time = 2, set d[1] = 2, and set color[1] = 1. \\ 
\rowcolor[HTML]{C7EDCC}  \\ 
\bottomrule
\end{tabular}
\end{table*}
\begin{table*}[h]
\small
\begin{tabular}{p{15.5cm}}
\toprule
\rowcolor[HTML]{C7EDCC} 8. Iterate through each element v in range(6): \\ 
\rowcolor[HTML]{C7EDCC}    - v = 0, A[1, 0] = 0, continue. \\ 
\rowcolor[HTML]{C7EDCC}    - v = 1, A[1, 1] = 0, continue. \\ 
\rowcolor[HTML]{C7EDCC}    - v = 2, A[1, 2] = 0, continue. \\ 
\rowcolor[HTML]{C7EDCC}    - v = 3, A[1, 3] = 0, continue. \\ 
\rowcolor[HTML]{C7EDCC}    - v = 4, A[1, 4] = 1, color[4] = 0, set pi[4] = 1, color[4] = 1, s\_prev[4] = 1, s\_last = 4, break the iteration. \\ 
\rowcolor[HTML]{C7EDCC}  \\ 
\rowcolor[HTML]{C7EDCC} 9. s\_last = 4, u = 1, s\_last != u, proceed to Step 5. \\ 
\rowcolor[HTML]{C7EDCC}  \\ 
\rowcolor[HTML]{C7EDCC} 10. Set u = 4 and go to Step 2. \\ 
\rowcolor[HTML]{C7EDCC}  \\ 
\rowcolor[HTML]{C7EDCC} 11. color[4] == 1, d[4] == 0, increment time by 1, time = 3, set d[4] = 3, and set color[4] = 1. \\ 
\rowcolor[HTML]{C7EDCC}  \\ 
\rowcolor[HTML]{C7EDCC} 12. Iterate through each element v in range(6): \\ 
\rowcolor[HTML]{C7EDCC} \quad - v = 0, A[4, 0] = 0, continue. \\ 
\rowcolor[HTML]{C7EDCC} \quad - v = 1, A[4, 1] = 0, continue. \\ 
\rowcolor[HTML]{C7EDCC} \quad - v = 2, A[4, 2] = 0, continue. \\ 
\rowcolor[HTML]{C7EDCC} \quad - v = 3, A[4, 3] = 1, color[3] = 0, set pi[3] = 4, color[3] = 1, s\_prev[3] = 4, s\_last = 3, break the iteration. \\ 
\rowcolor[HTML]{C7EDCC}  \\ 
\rowcolor[HTML]{C7EDCC} 13. s\_last = 3, u = 4, s\_last != u, proceed to Step 5. \\ 
\rowcolor[HTML]{C7EDCC}  \\ 
\rowcolor[HTML]{C7EDCC} 14. Set u = 3 and go to Step 2. \\ 
\rowcolor[HTML]{C7EDCC}  \\ 
\rowcolor[HTML]{C7EDCC} 15. color[3] == 1, d[3] == 0, increment time by 1, time = 4, set d[3] = 4, and set color[3] = 1. \\ 
\rowcolor[HTML]{C7EDCC}  \\ 
\rowcolor[HTML]{C7EDCC} 16. Iterate through each element v in range(6): \\ 
\rowcolor[HTML]{C7EDCC} \quad - v = 0, A[3, 0] = 0, continue. \\ 
\rowcolor[HTML]{C7EDCC} \quad - v = 1, A[3, 1] = 1, color[1] = 1, continue. \\ 
\rowcolor[HTML]{C7EDCC} \quad - v = 2, A[3, 2] = 0, continue. \\ 
\rowcolor[HTML]{C7EDCC} \quad - v = 3, A[3, 3] = 0, continue. \\ 
\rowcolor[HTML]{C7EDCC} \quad - v = 4, A[3, 4] = 0, continue. \\ 
\rowcolor[HTML]{C7EDCC} \quad - v = 5, A[3, 5] = 0, continue. \\ 
\rowcolor[HTML]{C7EDCC}  \\ 
\rowcolor[HTML]{C7EDCC} 17. s\_last = 3, u = 3, s\_last == u, set color[3] = 2, increment time by 1, time = 5, s\_prev[3] != 3, set pr = s\_prev[3] = 4, s\_prev[3] = 3, s\_last = 4. \\ 
\rowcolor[HTML]{C7EDCC}  \\ 
\rowcolor[HTML]{C7EDCC} 18. Set u = 4 and go to Step 2. \\ 
\rowcolor[HTML]{C7EDCC}  \\ 
\rowcolor[HTML]{C7EDCC} 19. color[4] == 1, d[4] == 3, continue. \\ 
\rowcolor[HTML]{C7EDCC}  \\ 
\rowcolor[HTML]{C7EDCC} 20. Iterate through each element v in range(6): \\ 
\rowcolor[HTML]{C7EDCC} \quad - v = 0, A[4, 0] = 0, continue. \\ 
\rowcolor[HTML]{C7EDCC} \quad - v = 1, A[4, 1] = 0, continue. \\ 
\rowcolor[HTML]{C7EDCC} \quad - v = 2, A[4, 2] = 0, continue. \\ 
\rowcolor[HTML]{C7EDCC} \quad - v = 3, A[4, 3] = 1, color[3] = 2, continue. \\ 
\rowcolor[HTML]{C7EDCC} \quad - v = 4, A[4, 4] = 0, continue. \\ 
\rowcolor[HTML]{C7EDCC} \quad - v = 5, A[4, 5] = 0, continue. \\ 
\rowcolor[HTML]{C7EDCC}  \\ 
\rowcolor[HTML]{C7EDCC} 21. s\_last = 4, u = 4, s\_last == u, set color[4] = 2, increment time by 1, time = 6, s\_prev[4] != 4, set pr = s\_prev[4] = 1, s\_prev[4] = 4, s\_last = 1. \\ 
\rowcolor[HTML]{C7EDCC}  \\ 
\rowcolor[HTML]{C7EDCC} 22. Set u = 1 and go to Step 2. \\ 
\rowcolor[HTML]{C7EDCC}  \\ 
\rowcolor[HTML]{C7EDCC} 23. color[1] == 1, d[1] == 2, continue. \\ 
\rowcolor[HTML]{C7EDCC}  \\ 
\rowcolor[HTML]{C7EDCC} 24. Iterate through each element v in range(6): \\ 
\rowcolor[HTML]{C7EDCC} \quad - v = 0, A[1, 0] = 0, continue. \\ 
\rowcolor[HTML]{C7EDCC} \quad - v = 1, A[1, 1] = 0, continue. \\ 
\rowcolor[HTML]{C7EDCC} \quad - v = 2, A[1, 2] = 0, continue. \\ 
\rowcolor[HTML]{C7EDCC} \quad - v = 3, A[1, 3] = 0, continue. \\ 
\rowcolor[HTML]{C7EDCC} \quad - v = 4, A[1, 4] = 1, color[4] = 2, continue. \\ 
\rowcolor[HTML]{C7EDCC} \quad - v = 5, A[1, 5] = 0, continue. \\ 
\rowcolor[HTML]{C7EDCC}  \\ 
\bottomrule
\end{tabular}
\end{table*}
\begin{table*}[h]
\small
\begin{tabular}{p{15.5cm}}
\toprule
\rowcolor[HTML]{C7EDCC} 25. s\_last = 1, u = 1, s\_last == u, set color[1] = 2, increment time by 1, time = 7, s\_prev[1] == 1, assert s\_prev[s\_last] == s\_last, increment s by 1, s = 1. \\ 
\rowcolor[HTML]{C7EDCC}  \\ 
\rowcolor[HTML]{C7EDCC} 26. s = 1, 1 < 6. \\ 
\rowcolor[HTML]{C7EDCC}  \\ 
\rowcolor[HTML]{C7EDCC} 27. color[1] == 2, increment s by 1, s = 2. \\ 
\rowcolor[HTML]{C7EDCC}  \\ 
\rowcolor[HTML]{C7EDCC} 28. s = 2, 2 < 6. \\ 
\rowcolor[HTML]{C7EDCC}  \\ 
\rowcolor[HTML]{C7EDCC} 29. color[2] == 0, set s\_last = 2, u = 2, and v = 2. \\ 
\rowcolor[HTML]{C7EDCC}  \\ 
\rowcolor[HTML]{C7EDCC} 30. Increment time by 1, time = 8, set d[2] = 8, and set color[2] = 1. \\ 
\rowcolor[HTML]{C7EDCC}  \\ 
\rowcolor[HTML]{C7EDCC} 31. Iterate through each element v in range(6): \\ 
\rowcolor[HTML]{C7EDCC} \quad - v = 0, A[2, 0] = 0, continue. \\ 
\rowcolor[HTML]{C7EDCC} \quad - v = 1, A[2, 1] = 0, continue. \\ 
\rowcolor[HTML]{C7EDCC} \quad - v = 2, A[2, 2] = 0, continue. \\ 
\rowcolor[HTML]{C7EDCC} \quad - v = 3, A[2, 3] = 0, continue. \\ 
\rowcolor[HTML]{C7EDCC} \quad - v = 4, A[2, 4] = 1, color[4] = 2, continue. \\ 
\rowcolor[HTML]{C7EDCC} \quad - v = 5, A[2, 5] = 1, color[5] = 0, set pi[5] = 2, color[5] = 1, s\_prev[5] = 2, s\_last = 5, break the iteration. \\ 
\rowcolor[HTML]{C7EDCC} 32. s\_last = 5, u = 2, s\_last != u, proceed to Step 5. \\ 
\rowcolor[HTML]{C7EDCC}  \\ 
\rowcolor[HTML]{C7EDCC} 33. Set u = 5 and go to Step 2. \\ 
\rowcolor[HTML]{C7EDCC}  \\ 
\rowcolor[HTML]{C7EDCC} 34. color[5] == 1, d[5] == 0, increment time by 1, time = 9, set d[5] = 9, and set color[5] = 1. \\ 
\rowcolor[HTML]{C7EDCC}  \\ 
\rowcolor[HTML]{C7EDCC} 35. Iterate through each element v in range(6): \\ 
\rowcolor[HTML]{C7EDCC} \quad - v = 0, A[5, 0] = 0, continue. \\ 
\rowcolor[HTML]{C7EDCC} \quad - v = 1, A[5, 1] = 0, continue. \\ 
\rowcolor[HTML]{C7EDCC} \quad - v = 2, A[5, 2] = 0, continue. \\ 
\rowcolor[HTML]{C7EDCC} \quad - v = 3, A[5, 3] = 0, continue. \\ 
\rowcolor[HTML]{C7EDCC} \quad - v = 4, A[5, 4] = 0, continue. \\ 
\rowcolor[HTML]{C7EDCC} \quad - v = 5, A[5, 5] = 1, color[5] = 1, continue. \\ 
\rowcolor[HTML]{C7EDCC}  \\ 
\rowcolor[HTML]{C7EDCC} 36. s\_last = 5, u = 5, s\_last == u, set color[5] = 2, increment time by 1, time = 10, s\_prev[5] == 5, assert s\_prev[s\_last] == s\_last, increment s by 1, s = 3. \\ 
\rowcolor[HTML]{C7EDCC}  \\ 
\rowcolor[HTML]{C7EDCC} 37. s = 3, 3 < 6. \\ 
\rowcolor[HTML]{C7EDCC}  \\ 
\rowcolor[HTML]{C7EDCC} 38. color[3] == 2, increment s by 1, s = 4. \\ 
\rowcolor[HTML]{C7EDCC}  \\ 
\rowcolor[HTML]{C7EDCC} 39. s = 4, 4 < 6. \\ 
\rowcolor[HTML]{C7EDCC}  \\ 
\rowcolor[HTML]{C7EDCC} 40. color[4] == 2, increment s by 1, s = 5. \\ 
\rowcolor[HTML]{C7EDCC}  \\ 
\rowcolor[HTML]{C7EDCC} 41. s = 5, 5 < 6. \\ 
\rowcolor[HTML]{C7EDCC}  \\ 
\rowcolor[HTML]{C7EDCC} 42. color[5] == 2, increment s by 1, s = 6. \\ 
\rowcolor[HTML]{C7EDCC}  \\ 
\rowcolor[HTML]{C7EDCC} 43. s = 6, go to Step 6. \\ 
\rowcolor[HTML]{C7EDCC}  \\ 
\rowcolor[HTML]{C7EDCC} 44. Print the array pi: [0, 0, 2, 4, 1, 2].
\\ \bottomrule
\end{tabular}
\caption{The prompt of depth first search and the response of \texttt{gpt-4}.}
\label{table:clrs_mini/depth_first_search}

\end{table*}

\begin{table*}[h]
\small
\begin{tabular}{p{15cm}}
\toprule
\textbf{Prompt} \\ \midrule
Execute the following instructions. \\ 
Initial: Set the array p of length N-1 = 3: p[0] = 0.19, p[1] = 0.04, p[2] = 0.11 and q of length N = 4: q[0] = 0.24, q[1] = 0.20, q[2] = 0.11, q[3] = 0.10 respectively. \\ 
Set the matrix e and w of size 4 x 4 with all elements initialized to 0. \\ 
Set maxint: 66666666. \\ 
 \\ 
Step 1: Initialize e and w diagonals: \\ 
  i. For i in range 0 to 3 (inclusive), do the following: \\ 
\quad a. Set e[i, i] = q[i]. \\ 
\quad b. Set w[i, i] = q[i]. \\ 
 \\ 
Step 2: Set L: 1. \\ 
 \\ 
Step 3: Is L < 4? If L is equal to or greater than 4, go to Step 6. Otherwise: \\ 
  i. Set upper\_i: 4 - L. \\ 
  ii. Set i: 0. \\ 
  iii. Go to Step 4. \\ 
 \\ 
Step 4: Is i < upper\_i? If i is equal to or greater than upper\_i, increment L by 1 and go to Step 3. Otherwise: \\ 
  i. Set j: i + L. \\ 
  ii. Set e[i, j] = maxint. \\ 
  iii. Set w[i, j] = w[i, j - 1] + p[j - 1] + q[j]. \\ 
  iv. Set r: i. \\ 
  v. Go to Step 5. \\ 
 \\ 
Step 5: Is r < j? If r is equal to or greater than j, increment i by 1 and go to Step 4. Otherwise: \\ 
  i. What is the value of i? Recall and print it, then calculate t = e[i, r] + e[r + 1, j] + w[i, j]. \\ 
  ii. Set e[i, j] = min(e[i, j], t). \\ 
  iii. Increment r by 1. \\ 
  iv. Go to Step 5. \\ 
 \\ 
Step 6: Print the value of e[0, N - 1]. \\ 
----- \\ 
The computation procedure is as follows, with each line numbered sequentially (e.g., 1. 2. 3. 4. 5. 6. etc.): \\ 
1. i = 0, 0 <= 3, e[0][0] = q[0] = 0.24, w[0][0] = q[0] = 0.24. \\ 
\rowcolor[HTML]{C7EDCC} 2. i = 1, 1 <= 3, e[1][1] = q[1] = 0.20, w[1][1] = q[1] = 0.20. \\ 
\rowcolor[HTML]{C7EDCC} 3. i = 2, 2 <= 3, e[2][2] = q[2] = 0.11, w[2][2] = q[2] = 0.11. \\ 
\rowcolor[HTML]{C7EDCC} 4. i = 3, 3 <= 3, e[3][3] = q[3] = 0.10, w[3][3] = q[3] = 0.10. \\ 
\rowcolor[HTML]{C7EDCC} 5. L = 1, L < 4, upper\_i = 4 - 1 = 3, i = 0. \\ 
\rowcolor[HTML]{C7EDCC} 6. i = 0, 0 < 3, j = 0 + 1 = 1, e[0][1] = maxint = 66666666, w[0][1] = w[0][0] + p[0] + q[1] = 0.24 + 0.19 + 0.20 = 0.63. \\ 
\rowcolor[HTML]{C7EDCC} 7. r = 0, r < 1, i = 0, t = e[0][0] + e[1][1] + w[0][1] = 0.24 + 0.20 + 0.63 = 1.07, e[0][1] = min(66666666, 1.07) = 1.07. \\ 
\rowcolor[HTML]{C7EDCC} 8. r = 1, r >= 1, i = 1, i < 3. \\ 
\rowcolor[HTML]{C7EDCC} 9. i = 1, 1 < 3, j = 1 + 1 = 2, e[1][2] = maxint = 66666666, w[1][2] = w[1][1] + p[1] + q[2] = 0.20 + 0.04 + 0.11 = 0.35. \\ 
\rowcolor[HTML]{C7EDCC} 10. r = 1, r < 2, i = 1, t = e[1][1] + e[2][2] + w[1][2] = 0.20 + 0.11 + 0.35 = 0.66, e[1][2] = min(66666666, 0.66) = 0.66. \\ 
\rowcolor[HTML]{C7EDCC} 11. r = 2, r >= 2, i = 2, i < 3. \\ 
\rowcolor[HTML]{C7EDCC} 12. i = 2, 2 < 3, j = 2 + 1 = 3, e[2][3] = maxint = 66666666, w[2][3] = w[2][2] + p[2] + q[3] = 0.11 + 0.11 + 0.10 = 0.32. \\ 
\rowcolor[HTML]{C7EDCC} 13. r = 2, r < 3, i = 2, t = e[2][2] + e[3][3] + w[2][3] = 0.11 + 0.10 + 0.32 = 0.53, e[2][3] = min(66666666, 0.53) = 0.53. \\ 
\rowcolor[HTML]{C7EDCC} 14. r = 3, r >= 3, i = 3, i >= 3, L = 2, L < 4, upper\_i = 4 - 2 = 2, i = 0. \\ 
\rowcolor[HTML]{C7EDCC} 15. i = 0, 0 < 2, j = 0 + 2 = 2, e[0][2] = maxint = 66666666, w[0][2] = w[0][1] + p[1] + q[2] = 0.63 + 0.04 + 0.11 = 0.78. \\ 
\rowcolor[HTML]{C7EDCC} 16. r = 0, r < 2, i = 0, t = e[0][0] + e[1][2] + w[0][2] = 0.24 + 0.66 + 0.78 = 1.68, e[0][2] = min(66666666, 1.68) = 1.68. \\ 
\rowcolor[HTML]{C7EDCC} 17. r = 1, r < 2, i = 0, t = e[0][1] + e[2][2] + w[0][2] = 1.07 + 0.11 + 0.78 = 1.96, e[0][2] = min(1.68, 1.96) = 1.68. \\ 
\rowcolor[HTML]{C7EDCC} 18. r = 2, r >= 2, i = 1, i < 2. \\ 
\bottomrule
\end{tabular}
\end{table*}
\begin{table*}[h]
\small
\begin{tabular}{p{15.5cm}}
\toprule
\rowcolor[HTML]{C7EDCC} 19. i = 1, 1 < 2, j = 1 + 2 = 3, e[1][3] = maxint = 66666666, w[1][3] = w[1][2] + p[2] + q[3] = 0.35 + 0.11 + 0.10 = 0.56. \\ 

\rowcolor[HTML]{C7EDCC} 20. r = 1, r < 3, i = 1, t = e[1][1] + e[2][3] + w[1][3] = 0.20 + 0.53 + 0.56 = 1.29, e[1][3] = min(66666666, 1.29) = 1.29. \\ 
\rowcolor[HTML]{C7EDCC} 21. r = 2, r < 3, i = 1, t = e[1][2] + e[3][3] + w[1][3] = 0.66 + 0.10 + 0.56 = 1.32, e[1][3] = min(1.29, 1.32) = 1.29. \\ 
\rowcolor[HTML]{C7EDCC} 22. r = 3, r >= 3, i = 2, i >= 2, L = 3, L < 4, upper\_i = 4 - 3 = 1, i = 0. \\ 
\rowcolor[HTML]{C7EDCC} 23. i = 0, 0 < 1, j = 0 + 3 = 3, e[0][3] = maxint = 66666666, w[0][3] = w[0][2] + p[2] + q[3] = 0.78 + 0.11 + 0.10 = 0.99. \\ 
\rowcolor[HTML]{C7EDCC} 24. r = 0, r < 3, i = 0, t = e[0][0] + e[1][3] + w[0][3] = 0.24 + 1.29 + 0.99 = 2.52, e[0][3] = min(66666666, 2.52) = 2.52. \\ 
\rowcolor[HTML]{C7EDCC} 25. r = 1, r < 3, i = 0, t = e[0][1] + e[2][3] + w[0][3] = 1.07 + 0.53 + 0.99 = 2.59, e[0][3] = min(2.52, 2.59) = 2.52. \\ 
\rowcolor[HTML]{C7EDCC} 26. r = 2, r < 3, i = 0, t = e[0][2] + e[3][3] + w[0][3] = 1.68 + 0.10 + 0.99 = 2.77, e[0][3] = min(2.52, 2.77) = 2.52. \\ 
\rowcolor[HTML]{C7EDCC} 27. r = 3, r >= 3, i = 1, i >= 1, L = 4, L >= 4. \\ 
\rowcolor[HTML]{C7EDCC} 28. Print e[0][3] = 2.52.
\\ \bottomrule
\end{tabular}
\caption{The prompt of optimal binary search tree and the response of \texttt{gpt-4}.}
\label{table:clrs_mini/optimal_binary_search_tree}

\end{table*}
\begin{table*}[h]
\small
\begin{tabular}{p{15cm}}
\toprule
\textbf{Prompt} \\ \midrule
Execute the instructions step by step. Do not jump steps. Do not stop before completion. \\ 
Initial: Set the adjacency matrix A: A[0, 0] = 0, A[0, 1] = 0, A[0, 2] = 1, A[0, 3] = 1 \\ 
A[1, 0] = 0, A[1, 1] = 0, A[1, 2] = 1, A[1, 3] = 1 \\ 
A[2, 0] = 1, A[2, 1] = 1, A[2, 2] = 0, A[2, 3] = 1 \\ 
A[3, 0] = 1, A[3, 1] = 0, A[3, 2] = 1, A[3, 3] = 0. \\ 
Set the number of nodes N: 4. \\ 
Initialize the following arrays: \\ 
- color: [1, 0, 0, 0]. \\ 
- d: [0, 6666, 6666, 6666]. \\ 
Set the source node: source = 0. \\ 
Step 1: Create an empty queue Q and add the source node to it. \\ 
Step 2: Check if the queue Q is empty. If it is empty, proceed to Step 7. Otherwise, continue to Step 3. \\ 
Step 3: Remove the first element from the queue and assign it to the variable u. \\ 
Step 4: Set i = 0. \\ 
Step 5: If i >= 4, go to Step 6; otherwise: \\ 
\quad i. Check if A[u, i] != 0. If A[u, i] == 0, increment i by 1 and go to Step 5. \\ 
\quad ii. If color[i] == 0, update the following values: \\ 
\quad \quad - color[i] = 1 \\ 
\quad \quad - d[i] = d[u] + 1 \\ 
\quad \quad - pi[i] = u \\ 
\quad \quad - Append node i to the queue Q. \\ 
\quad iii. Increment i by 1 and go to Step 5. \\ 
Step 6: Update color[u] = 2 and go back to Step 2. \\ 
Step 7: When the queue Q is empty, return the list d containing the shortest path distances from the source node to all other nodes. \\ 
----- \\ 
The computation procedure is as follows, with each line numbered sequentially (e.g., 1. 2. 3. 4. 5. 6. etc.): \\ 
1. Q = [0] \\ 
\rowcolor[HTML]{C7EDCC} 2. Q is not empty, proceed to step 3. \\ 
\rowcolor[HTML]{C7EDCC} 3. u = 0, Q = [] \\ 
\rowcolor[HTML]{C7EDCC} 4. i = 0 \\ 
\rowcolor[HTML]{C7EDCC} 5. i < 4, check A[u, i]: \\ 
\rowcolor[HTML]{C7EDCC} \quad i. A[0, 0] == 0, increment i to 1. \\ 
\rowcolor[HTML]{C7EDCC} \quad ii. A[0, 1] == 0, increment i to 2. \\ 
\rowcolor[HTML]{C7EDCC} \quad iii. A[0, 2] != 0, color[2] == 0, update values: \\ 
\rowcolor[HTML]{C7EDCC} \quad \quad - color[2] = 1 \\ 
\rowcolor[HTML]{C7EDCC} \quad \quad - d[2] = d[0] + 1 = 1 \\ 
\rowcolor[HTML]{C7EDCC} \quad \quad - pi[2] = 0 \\ 
\rowcolor[HTML]{C7EDCC} \quad \quad - Q = [2] \\ 
\rowcolor[HTML]{C7EDCC} \quad iv. Increment i to 3. \\ 
\bottomrule
\end{tabular}
\end{table*}
\begin{table*}[h]
\small
\begin{tabular}{p{15.5cm}}
\toprule
\rowcolor[HTML]{C7EDCC} \quad v. A[0, 3] != 0, color[3] == 0, update values: \\ 
\rowcolor[HTML]{C7EDCC} \quad \quad - color[3] = 1 \\ 
\rowcolor[HTML]{C7EDCC} \quad \quad - d[3] = d[0] + 1 = 1 \\ 
\rowcolor[HTML]{C7EDCC} \quad \quad - pi[3] = 0 \\ 
\rowcolor[HTML]{C7EDCC} \quad \quad - Q = [2, 3] \\ 
\rowcolor[HTML]{C7EDCC} 6. color[0] = 2 \\ 
\rowcolor[HTML]{C7EDCC} 2. Q is not empty, proceed to step 3. \\ 
\rowcolor[HTML]{C7EDCC} 3. u = 2, Q = [3] \\ 
\rowcolor[HTML]{C7EDCC} 4. i = 0 \\ 
\rowcolor[HTML]{C7EDCC} 5. i < 4, check A[u, i]: \\ 
\rowcolor[HTML]{C7EDCC} \quad i. A[2, 0] != 0, color[0] == 2, increment i to 1. \\ 
\rowcolor[HTML]{C7EDCC} \quad ii. A[2, 1] != 0, color[1] == 0, update values: \\ 
\rowcolor[HTML]{C7EDCC} \quad \quad - color[1] = 1 \\ 
\rowcolor[HTML]{C7EDCC} \quad \quad - d[1] = d[2] + 1 = 2 \\ 
\rowcolor[HTML]{C7EDCC} \quad \quad - pi[1] = 2 \\ 
\rowcolor[HTML]{C7EDCC} \quad \quad - Q = [3, 1] \\ 
\rowcolor[HTML]{C7EDCC} \quad iii. Increment i to 2. \\ 
\rowcolor[HTML]{C7EDCC} \quad iv. A[2, 2] == 0, increment i to 3. \\ 
\rowcolor[HTML]{C7EDCC} \quad v. A[2, 3] != 0, color[3] == 1, increment i to 4. \\ 
\rowcolor[HTML]{C7EDCC} 6. color[2] = 2 \\ 
\rowcolor[HTML]{C7EDCC} 2. Q is not empty, proceed to step 3. \\ 
\rowcolor[HTML]{C7EDCC} 3. u = 3, Q = [1] \\ 
\rowcolor[HTML]{C7EDCC} 4. i = 0 \\ 
\rowcolor[HTML]{C7EDCC} 5. i < 4, check A[u, i]: \\ 
\rowcolor[HTML]{C7EDCC} \quad i. A[3, 0] != 0, color[0] == 2, increment i to 1. \\ 
\rowcolor[HTML]{C7EDCC} \quad ii. A[3, 1] == 0, increment i to 2. \\ 
\rowcolor[HTML]{C7EDCC} \quad iii. A[3, 2] != 0, color[2] == 2, increment i to 3. \\ 
\rowcolor[HTML]{C7EDCC} \quad iv. A[3, 3] == 0, increment i to 4. \\ 
\rowcolor[HTML]{C7EDCC} 6. color[3] = 2 \\ 
\rowcolor[HTML]{C7EDCC} 2. Q is not empty, proceed to step 3. \\ 
\rowcolor[HTML]{C7EDCC} 3. u = 1, Q = [] \\ 
\rowcolor[HTML]{C7EDCC} 4. i = 0 \\ 
\rowcolor[HTML]{C7EDCC} 5. i < 4, check A[u, i]: \\ 
\rowcolor[HTML]{C7EDCC} \quad i. A[1, 0] == 0, increment i to 1. \\ 
\rowcolor[HTML]{C7EDCC} \quad ii. A[1, 1] == 0, increment i to 2. \\ 
\rowcolor[HTML]{C7EDCC} \quad iii. A[1, 2] != 0, color[2] == 2, increment i to 3. \\ 
\rowcolor[HTML]{C7EDCC} \quad iv. A[1, 3] != 0, color[3] == 2, increment i to 4. \\ 
\rowcolor[HTML]{C7EDCC} 6. color[1] = 2 \\ 
\rowcolor[HTML]{C7EDCC} 2. Q is empty, proceed to step 7. \\ 
\rowcolor[HTML]{C7EDCC} 7. Return the list d: [0, 2, 1, 1]
\\ \bottomrule
\end{tabular}
\caption{The prompt of breadth first search and the response of \texttt{gpt-4}.}
\label{table:clrs_mini/breadth_first_search}

\end{table*}
\begin{table*}[h]
\small
\begin{tabular}{p{15cm}}
\toprule
\textbf{Prompt} \\ \midrule
Execute the instructions step by step. Do not jump steps. Do not stop before completion. When updating an array, use concrete values rather than variables. \\ 
Initial: Set the square matrix A of size 4 x 4 : \\ 
A[0, 0] = 0, A[0, 1] = 0, A[0, 2] = 0, A[0, 3] = 1 \\ 
A[1, 0] = 1, A[1, 1] = 0, A[1, 2] = 0, A[1, 3] = 0 \\ 
A[2, 0] = 0, A[2, 1] = 0, A[2, 2] = 0, A[2, 3] = 0 \\ 
A[3, 0] = 1, A[3, 1] = 0, A[3, 2] = 1, A[3, 3] = 0. \\ 
Set color: color[0] = 0, color[1] = 0, color[2] = 0, color[3] = 0. \\ 
Set topo: topo[0] = 0, topo[1] = 1, topo[2] = 2, topo[3] = 3. \\ 
Set s\_prev: s\_prev[0] = 0, s\_prev[1] = 1, s\_prev[2] = 2, s\_prev[3] = 3. \\ 
Set topo\_head: 0. \\ 
Set s: 0. \\ 
 \\ 
Step 1: Is s < 4? If s is equal to 4, go to Step 8. Otherwise: \\ 
  i. If color[s] == 0, set s\_last = s, u = s, go to Step 2. \\ 
  ii. If color[s] != 0, increment s by 1, and go to Step 1. \\ 
 \\ 
Step 2: What is the value of color[u]? Print it. Check if color[u] == 0. \\ 
  i. If color[u] == 0, set color[u] = 1. \\ 
  ii. If color[u] != 0, go to Step 3. \\ 
 \\ 
Step 3: Iterate through each element v in range(N). After the iteration of v, go to Step 4. \\ 
  i. If A[u, v] != 0, check if color[v] == 0. \\ 
\quad a. If color[v] == 0, set color[v] = 1, s\_prev[v] = s\_last, and s\_last = v. Then, break the iteration and go to Step 4. \\ 
\quad b. If color[v] != 0, continue the iteration of v. \\ 
  ii. If A[u, v] == 0, continue the iteration of v. \\ 
 \\ 
Step 4: What are the values of s\_last and u? Print them. Is s\_last equal to u? \\ 
  i. If s\_last == u, set color[u] = 2 and go to Step 5. \\ 
  ii. If s\_last != u, proceed to Step 7. \\ 
 \\ 
Step 5: What is the value of topo\_head? What is the value of color[topo\_head]? Print it with the format "color[topo\_head] = color[%d] = %d". Is the value of color[topo\_head] equal to 2? \\ 
  i. If color[topo\_head] == 2, set topo[u] as the value of topo\_head, and print the updated array of topo. Then, reset topo\_head = u. \\ 
  ii. If color[topo\_head] != 2, go to Step 6. \\ 
 \\ 
Step 6: What is the value of s\_prev[u]? Print it. Is s\_prev[u] equal to u? \\ 
  i. If s\_prev[u] == u, break the loop and go to Step 1. \\ 
  ii. If s\_prev[u] != u, set pr as the value of s\_prev[s\_last], then set s\_prev[s\_last] = s\_last, and s\_last = pr. Go to Step 7. \\ 
 \\ 
Step 7: Set u = s\_last and go to Step 2. \\ 
 \\ 
Step 8: Print the array of topo. \\ 
----- \\ 
The computation procedure is as follows, with each line numbered sequentially (e.g., 1. 2. 3. 4. 5. 6. 7. 8. 9. etc.): \\ 
1. s = 0, 0 < 4. Since color[0] == 0, set s\_last = 0, u = 0, go to Step 2. \\ 
\rowcolor[HTML]{C7EDCC} 2. color[u] = color[0] = 0. Set color[0] = 1. \\ 
\bottomrule
\end{tabular}
\end{table*}
\begin{table*}[!t]
\small
\begin{tabular}{p{15.5cm}}
\toprule
\rowcolor[HTML]{C7EDCC} 3. Iterate through each element v in range(4): \\ 
\rowcolor[HTML]{C7EDCC}    - v = 0: A[0, 0] == 0, continue the iteration. \\ 
\rowcolor[HTML]{C7EDCC}    - v = 1: A[0, 1] == 0, continue the iteration. \\ 
\rowcolor[HTML]{C7EDCC}    - v = 2: A[0, 2] == 0, continue the iteration. \\ 
\rowcolor[HTML]{C7EDCC}    - v = 3: A[0, 3] != 0, color[3] == 0, set color[3] = 1, s\_prev[3] = 0, s\_last = 3, break the iteration. \\ 
\rowcolor[HTML]{C7EDCC} 4. s\_last = 3, u = 0, s\_last != u, go to Step 7. \\ 
\rowcolor[HTML]{C7EDCC} 7. Set u = 3, go to Step 2. \\ 
\rowcolor[HTML]{C7EDCC} 8. color[u] = color[3] = 1. Set color[3] = 2. \\ 
\rowcolor[HTML]{C7EDCC} 9. Iterate through each element v in range(4): \\ 
\rowcolor[HTML]{C7EDCC}    - v = 0: A[3, 0] != 0, color[0] != 0, continue the iteration. \\ 
\rowcolor[HTML]{C7EDCC}    - v = 1: A[3, 1] == 0, continue the iteration. \\ 
\rowcolor[HTML]{C7EDCC}    - v = 2: A[3, 2] != 0, color[2] == 0, set color[2] = 1, s\_prev[2] = 3, s\_last = 2, break the iteration. \\ 
\rowcolor[HTML]{C7EDCC} 10. s\_last = 2, u = 3, s\_last != u, go to Step 7. \\ 
\rowcolor[HTML]{C7EDCC} 11. Set u = 2, go to Step 2. \\ 
\rowcolor[HTML]{C7EDCC} 12. color[u] = color[2] = 1. Set color[2] = 2. \\ 
\rowcolor[HTML]{C7EDCC} 13. Iterate through each element v in range(4): \\ 
\rowcolor[HTML]{C7EDCC} \quad - v = 0: A[2, 0] == 0, continue the iteration. \\ 
\rowcolor[HTML]{C7EDCC} \quad - v = 1: A[2, 1] == 0, continue the iteration. \\ 
\rowcolor[HTML]{C7EDCC} \quad - v = 2: A[2, 2] == 0, continue the iteration. \\ 
\rowcolor[HTML]{C7EDCC} \quad - v = 3: A[2, 3] == 0, continue the iteration. \\ 
\rowcolor[HTML]{C7EDCC} 14. s\_last = 2, u = 2, s\_last == u, set color[2] = 2, go to Step 5. \\ 
\rowcolor[HTML]{C7EDCC} 15. topo\_head = 0, color[topo\_head] = color[0] = 1, go to Step 6. \\ 
\rowcolor[HTML]{C7EDCC} 16. s\_prev[u] = s\_prev[2] = 3, s\_prev[u] != u, set pr = s\_prev[s\_last] = s\_prev[2] = 3, set s\_prev[s\_last] = s\_prev[2] = 2, s\_last = pr = 3, go to Step 7. \\ 
\rowcolor[HTML]{C7EDCC} 17. Set u = 3, go to Step 2. \\ 
\rowcolor[HTML]{C7EDCC} 18. color[u] = color[3] = 2, go to Step 3. \\ 
\rowcolor[HTML]{C7EDCC} 19. Iterate through each element v in range(4): \\ 
\rowcolor[HTML]{C7EDCC} \quad - v = 0: A[3, 0] != 0, color[0] != 0, continue the iteration. \\ 
\rowcolor[HTML]{C7EDCC} \quad - v = 1: A[3, 1] == 0, continue the iteration. \\ 
\rowcolor[HTML]{C7EDCC} \quad - v = 2: A[3, 2] != 0, color[2] != 0, continue the iteration. \\ 
\rowcolor[HTML]{C7EDCC} \quad - v = 3: A[3, 3] == 0, continue the iteration. \\ 
\rowcolor[HTML]{C7EDCC} 20. s\_last = 3, u = 3, s\_last == u, set color[3] = 2, go to Step 5. \\ 
\rowcolor[HTML]{C7EDCC} 21. topo\_head = 0, color[topo\_head] = color[0] = 1, go to Step 6. \\ 
\rowcolor[HTML]{C7EDCC} 22. s\_prev[u] = s\_prev[3] = 0, s\_prev[u] != u, set pr = s\_prev[s\_last] = s\_prev[3] = 0, set s\_prev[s\_last] = s\_prev[3] = 3, s\_last = pr = 0, go to Step 7. \\ 
\rowcolor[HTML]{C7EDCC} 23. Set u = 0, go to Step 2. \\ 
\rowcolor[HTML]{C7EDCC} 24. color[u] = color[0] = 1, go to Step 3. \\ 
\rowcolor[HTML]{C7EDCC} 25. Iterate through each element v in range(4): \\ 
\rowcolor[HTML]{C7EDCC} \quad - v = 0: A[0, 0] == 0, continue the iteration. \\ 
\rowcolor[HTML]{C7EDCC} \quad - v = 1: A[0, 1] == 0, continue the iteration. \\ 
\rowcolor[HTML]{C7EDCC} \quad - v = 2: A[0, 2] == 0, continue the iteration. \\ 
\rowcolor[HTML]{C7EDCC} \quad - v = 3: A[0, 3] != 0, color[3] != 0, continue the iteration. \\ 
\rowcolor[HTML]{C7EDCC} 26. s\_last = 0, u = 0, s\_last == u, set color[0] = 2, go to Step 5. \\ 
\rowcolor[HTML]{C7EDCC} 27. topo\_head = 0, color[topo\_head] = color[0] = 2, set topo[u] = topo[0] = 0, print topo, reset topo\_head = 0. \\ 
\rowcolor[HTML]{C7EDCC} 28. s\_prev[u] = s\_prev[0] = 0, s\_prev[u] == u, break the loop and go to Step 1. \\ 
\rowcolor[HTML]{C7EDCC} 29. s = 1, 1 < 4. Since color[1] == 0, set s\_last = 1, u = 1, go to Step 2. \\ 
\rowcolor[HTML]{C7EDCC} 30. color[u] = color[1] = 0. Set color[1] = 1. \\ 
\rowcolor[HTML]{C7EDCC} 31. Iterate through each element v in range(4): \\ 
\rowcolor[HTML]{C7EDCC} \quad - v = 0: A[1, 0] != 0, color[0] != 0, continue the iteration. \\ 
\rowcolor[HTML]{C7EDCC} \quad - v = 1: A[1, 1] == 0, continue the iteration. \\ 
\rowcolor[HTML]{C7EDCC} \quad - v = 2: A[1, 2] == 0, continue the iteration. \\ 
\rowcolor[HTML]{C7EDCC} \quad - v = 3: A[1, 3] == 0, continue the iteration. \\ 
\rowcolor[HTML]{C7EDCC} 32. s\_last = 1, u = 1, s\_last == u, set color[1] = 2, go to Step 5. \\ 
\rowcolor[HTML]{C7EDCC} 33. topo\_head = 0, color[topo\_head] = color[0] = 2, set topo[u] = topo[1] = 0, print topo, reset topo\_head = 1. \\ 
\rowcolor[HTML]{C7EDCC} 34. s\_prev[u] = s\_prev[1] = 1, s\_prev[u] == u, break the loop and go to Step 1. \\ 
\rowcolor[HTML]{C7EDCC} 35. s = 2, 2 < 4. Since color[2] != 0, increment s by 1, and go to Step 1. \\ 
\rowcolor[HTML]{C7EDCC} 36. s = 3, 3 < 4. Since color[3] != 0, increment s by 1, and go to Step 1. \\ 
\rowcolor[HTML]{C7EDCC} 37. s = 4, s == 4, go to Step 8. \\ 
\rowcolor[HTML]{C7EDCC} 38. Print the array of topo: [0, 0, 2, 3] \\ 
\rowcolor[HTML]{C7EDCC}  \\ 
\rowcolor[HTML]{C7EDCC} The final topo array is [0, 0, 2, 3].
\\ \bottomrule
\end{tabular}
\caption{The prompt of topological sort and the response of \texttt{gpt-4}.}
\label{table:clrs_mini/topological_sort}

\end{table*}
\begin{table*}[h]
\small
\begin{tabular}{p{15cm}}
\toprule
\textbf{Prompt} \\ \midrule
Execute the instructions step by step. Do not jump steps. Do not stop before completion. \\ 
Initial: Set the square matrix A of size 3 x 3 : \\ 
A[0, 0] = 0, A[0, 1] = 1, A[0, 2] = 1 \\ 
A[1, 0] = 1, A[1, 1] = 0, A[1, 2] = 0 \\ 
A[2, 0] = 1, A[2, 1] = 0, A[2, 2] = 0. \\ 
Set color: color[0] = 0, color[1] = 0, color[2] = 0. \\ 
Set pi: pi[0] = 0, pi[1] = 1, pi[2] = 2. \\ 
Set d: d[0] = 0, d[1] = 0, d[2] = 0. \\ 
Set s\_prev: s\_prev[0] = 0, s\_prev[1] = 1, s\_prev[2] = 2. \\ 
Set time: 0. \\ 
Set low: low[0] = 0, low[1] = 0, low[2] = 0. \\ 
Set child\_cnt: child\_cnt[0] = 0, child\_cnt[1] = 0, child\_cnt[2] = 0. \\ 
Set is\_cut: is\_cut[0] = 0, is\_cut[1] = 0, is\_cut[2] = 0. \\ 
 \\ 
Step 1: Initialize the variable u to 0 and variable s\_last to 0. \\ 
 \\ 
Step 2: Start the while loop, and perform the following sub-steps: \\ 
 \\ 
\quad Step 2.1: If d[u] is 0: \\ 
\quad \quad a. Increment time by 1. \\ 
\quad \quad b. Set d[u] to time. \\ 
\quad \quad c. Set low[u] to time. \\ 
\quad \quad d. Set color[u] to 1. \\ 
 \\ 
\quad Step 2.2: Iterate through variable v from 0 to 2 (inclusive), and for each v: \\ 
\quad \quad a. If A[u, v] is not 0: \\ 
\quad \quad \quad i. If color[v] is 0: \\ 
\quad \quad \quad \quad 1. Set pi[v] to u. \\ 
\quad \quad \quad \quad 2. Set color[v] to 1. \\ 
\quad \quad \quad \quad 3. Print s\_last and set s\_prev[v] to s\_last. \\ 
\quad \quad \quad \quad 4. Set s\_last to v. \\ 
\quad \quad \quad \quad 5. Increment child\_cnt[u]. \\ 
\quad \quad \quad \quad 6. Break the v iteration loop and go to Step 2.3. \\ 
\quad \quad \quad ii. If color[v] is not 0: \\ 
\quad \quad \quad \quad 1. Calculate bne\_v\_pi\_u = v - pi[u] and present the result. \\ 
\quad \quad \quad \quad 2. If bne\_v\_pi\_u is not 0, set low[u] to the minimum of low[u] and d[v]. \\ 
 \\ 
\quad Step 2.3: Calculate beq\_s\_last\_u = s\_last - u and present the result. \\ 
\quad \quad a. If beq\_s\_last\_u is equal to 0: \\ 
\quad \quad \quad i. Set color[u] to 2. \\ 
\quad \quad \quad ii. Increment time by 1. \\ 
\quad \quad b. If beq\_s\_last\_u is not 0, go to Step 2.8. \\ 
 \\ 
 \\ 
\quad Step 2.4: Iterate through variable v from 0 to 2 (inclusive), and for each v: \\ 
\quad \quad a. Print the value of pi[v] and u. Calculate beq\_pi\_v\_u = pi[v] - u and present the result. \\ 
\quad \quad b. If beq\_pi\_v\_u is equal to 0: \\ 
\quad \quad \quad i. Go to Step 2.4.1. \\ 
\quad \quad \quad ii. Go back to the iteration of v. \\ 
 \\ 
\quad Step 2.4.1: \\ 
\quad \quad a. Set low[u] to the minimum of low[u] and low[v]. \\ 
\quad \quad b. Print the values of u and pi[u]. Calculate bne\_u\_pi\_u = u - pi[u] and present the result. \\ 
\quad \quad c. IS bne\_u\_pi\_u NON-ZERO OR NOT? \\ 
\quad \quad \quad i. If bne\_u\_pi\_u is not 0: \\ 
\quad \quad \quad \quad 1. Print the values of low[v] and d[u]. Calculate bge\_low\_v\_d\_u = low[v] - d[u] and present the result. \\ 
\quad \quad \quad \quad 2. If bge\_low\_v\_d\_u >= 0, set is\_cut[u] to 1. \\ 
\quad \quad \quad ii. If bne\_u\_pi\_u is 0 exactly, pass. \\ 
 \\ 
 \\ 
 \\ 
 \bottomrule
\end{tabular}
\end{table*}
\begin{table*}[!t]
\small
\begin{tabular}{p{15.5cm}}
\toprule
\quad Step 2.5: Calculate beq\_u\_pi\_u = u - pi[u] and present the result.  \\ 
\quad \quad a. If beq\_u\_pi\_u is equal to 0: \\ 
\quad \quad \quad 1. Calculate bge\_cnt = child\_cnt[u] - 2 and present the result. \\ 
\quad \quad \quad 2. If bge\_cnt >= 0, set is\_cut[u] to 1. \\ 
\quad \quad b. If beq\_u\_pi\_u is not equal to 0, pass. \\ 
 \\ 
\quad Step 2.6: Print the value of s\_prev[u]. Then, calculate beq\_u\_s\_prev\_u = u - s\_prev[u] and present the result. \\ 
\quad \quad a. If beq\_u\_s\_prev\_u is equal to 0, break the main loop and go to Step 3. \\ 
 \\ 
\quad Step 2.7: Swap s\_prev[s\_last] and s\_last: \\ 
\quad \quad a. Set temp to s\_prev[s\_last]. \\ 
\quad \quad b. Set s\_prev[s\_last] to s\_last. \\ 
\quad \quad c. Set s\_last to temp. \\ 
 \\ 
\quad Step 2.8: Set u to s\_last and go to Step 2.1. \\ 
 \\ 
Step 3: Print the is\_cut array. \\ 
 \\ 
----- \\ 
The computation procedure is as follows, with each line numbered sequentially (e.g., 1. 2. 3. 4. 5. 6. 7. 8. 9. etc.): \\ 
1. Step 1, set u = 0 and s\_last = 0. \\ 
\rowcolor[HTML]{C7EDCC} 2. Step 2, enter the while loop. \\ 
\rowcolor[HTML]{C7EDCC}  \\ 
\rowcolor[HTML]{C7EDCC} 3. Step 2.1, since d[0] = 0, increment time to 1, set d[0] = 1, low[0] = 1, and color[0] = 1. \\ 
\rowcolor[HTML]{C7EDCC}  \\ 
\rowcolor[HTML]{C7EDCC} 4. Step 2.2, iterate through v from 0 to 2. For v = 0, A[0, 0] = 0, so continue to v = 1. For v = 1, A[0, 1] = 1, and color[1] = 0, so set pi[1] = 0, color[1] = 1, print s\_last (0), set s\_prev[1] = 0, set s\_last = 1, increment child\_cnt[0] to 1, and break the v iteration loop. \\ 
\rowcolor[HTML]{C7EDCC}  \\ 
\rowcolor[HTML]{C7EDCC} 5. Step 2.3, calculate beq\_s\_last\_u = 1 - 0 = 1, which is not 0, so go to Step 2.8. \\ 
\rowcolor[HTML]{C7EDCC}  \\ 
\rowcolor[HTML]{C7EDCC} 6. Step 2.8, set u = s\_last = 1 and go to Step 2.1. \\ 
\rowcolor[HTML]{C7EDCC}  \\ 
\rowcolor[HTML]{C7EDCC} 7. Step 2.1, since d[1] = 0, increment time to 2, set d[1] = 2, low[1] = 2, and color[1] = 1. \\ 
\rowcolor[HTML]{C7EDCC}  \\ 
\rowcolor[HTML]{C7EDCC} 8. Step 2.2, iterate through v from 0 to 2. For v = 0, A[1, 0] = 1, but color[0] = 1, so calculate bne\_v\_pi\_u = 0 - 0 = 0, and since it is 0, continue to v = 1. For v = 1, A[1, 1] = 0, so continue to v = 2. For v = 2, A[1, 2] = 0, so the v iteration loop ends. \\ 
\rowcolor[HTML]{C7EDCC}  \\ 
\rowcolor[HTML]{C7EDCC} 9. Step 2.3, calculate beq\_s\_last\_u = 1 - 1 = 0, which is equal to 0, so set color[1] = 2 and increment time to 3. \\ 
\rowcolor[HTML]{C7EDCC}  \\ 
\rowcolor[HTML]{C7EDCC} 10. Step 2.4, iterate through v from 0 to 2. For v = 0, print pi[0] (0) and u (1), calculate beq\_pi\_v\_u = 0 - 1 = -1, which is not 0, so continue to v = 1. For v = 1, print pi[1] (0) and u (1), calculate beq\_pi\_v\_u = 0 - 1 = -1, which is not 0, so continue to v = 2. For v = 2, print pi[2] (2) and u (1), calculate beq\_pi\_v\_u = 2 - 1 = 1, which is not 0, so the v iteration loop ends. \\ 
\rowcolor[HTML]{C7EDCC}  \\ 
\rowcolor[HTML]{C7EDCC} 11. Step 2.5, calculate beq\_u\_pi\_u = 1 - 0 = 1, which is not 0, so pass. \\ 
\rowcolor[HTML]{C7EDCC}  \\ 
\rowcolor[HTML]{C7EDCC} 12. Step 2.6, print s\_prev[1] (0), calculate beq\_u\_s\_prev\_u = 1 - 0 = 1, which is not 0, so continue to Step 2.7. \\ 
\rowcolor[HTML]{C7EDCC}  \\ 
\rowcolor[HTML]{C7EDCC} 13. Step 2.7, swap s\_prev[s\_last] and s\_last: set temp = s\_prev[1] (0), set s\_prev[1] = 1, set s\_last = temp (0). \\ 
\rowcolor[HTML]{C7EDCC}  \\ 
\rowcolor[HTML]{C7EDCC} 14. Step 2.8, set u = s\_last = 0 and go to Step 2.1. \\ 
\rowcolor[HTML]{C7EDCC}  \\ 
\rowcolor[HTML]{C7EDCC} 15. Step 2.1, since d[0] is not 0, skip to Step 2.2. \\ 
\rowcolor[HTML]{C7EDCC}  \\ 
\rowcolor[HTML]{C7EDCC} 16. Step 2.2, iterate through v from 0 to 2. For v = 0, A[0, 0] = 0, so continue to v = 1. For v = 1, A[0, 1] = 1, but color[1] = 2, so calculate bne\_v\_pi\_u = 1 - 0 = 1, and since it is not 0, set low[0] = min(low[0], d[1]) = min(1, 2) = 1. Continue to v = 2. For v = 2, A[0, 2] = 1, and color[2] = 0, so set pi[2] = 0, color[2] = 1, print s\_last (0), set s\_prev[2] = 0, set s\_last = 2, increment child\_cnt[0] to 2, and break the v iteration loop. \\ 
\rowcolor[HTML]{C7EDCC}  \\ 
\bottomrule
\end{tabular}
\end{table*}
\begin{table*}[!t]
\small
\begin{tabular}{p{15.5cm}}
\toprule
\rowcolor[HTML]{C7EDCC} 17. Step 2.3, calculate beq\_s\_last\_u = 2 - 0 = 2, which is not 0, so go to Step 2.8. \\ 
\rowcolor[HTML]{C7EDCC}  \\ 
\rowcolor[HTML]{C7EDCC} 18. Step 2.8, set u = s\_last = 2 and go to Step 2.1. \\ 
\rowcolor[HTML]{C7EDCC}  \\ 
\rowcolor[HTML]{C7EDCC} 19. Step 2.1, since d[2] = 0, increment time to 4, set d[2] = 4, low[2] = 4, and color[2] = 1. \\ 
\rowcolor[HTML]{C7EDCC}  \\ 
\rowcolor[HTML]{C7EDCC} 20. Step 2.2, iterate through v from 0 to 2. For v = 0, A[2, 0] = 1, but color[0] = 1, so calculate bne\_v\_pi\_u = 0 - 0 = 0, and since it is 0, continue to v = 1. For v = 1, A[2, 1] = 0, so continue to v = 2. For v = 2, A[2, 2] = 0, so the v iteration loop ends. \\ 
\rowcolor[HTML]{C7EDCC}  \\ 
\rowcolor[HTML]{C7EDCC} 21. Step 2.3, calculate beq\_s\_last\_u = 2 - 2 = 0, which is equal to 0, so set color[2] = 2 and increment time to 5. \\ 
\rowcolor[HTML]{C7EDCC}  \\ 
\rowcolor[HTML]{C7EDCC} 22. Step 2.4, iterate through v from 0 to 2. For v = 0, print pi[0] (0) and u (2), calculate beq\_pi\_v\_u = 0 - 2 = -2, which is not 0, so continue to v = 1. For v = 1, print pi[1] (0) and u (2), calculate beq\_pi\_v\_u = 0 - 2 = -2, which is not 0, so continue to v = 2. For v = 2, print pi[2] (0) and u (2), calculate beq\_pi\_v\_u = 0 - 2 = -2, which is not 0, so the v iteration loop ends. \\ 
\rowcolor[HTML]{C7EDCC}  \\ 
\rowcolor[HTML]{C7EDCC} 23. Step 2.5, calculate beq\_u\_pi\_u = 2 - 0 = 2, which is not 0, so pass. \\ 
\rowcolor[HTML]{C7EDCC}  \\ 
\rowcolor[HTML]{C7EDCC} 24. Step 2.6, print s\_prev[2] (0), calculate beq\_u\_s\_prev\_u = 2 - 0 = 2, which is not 0, so continue to Step 2.7. \\ 
\rowcolor[HTML]{C7EDCC}  \\ 
\rowcolor[HTML]{C7EDCC} 25. Step 2.7, swap s\_prev[s\_last] and s\_last: set temp = s\_prev[2] (0), set s\_prev[2] = 2, set s\_last = temp (0). \\ 
\rowcolor[HTML]{C7EDCC}  \\ 
\rowcolor[HTML]{C7EDCC} 26. Step 2.8, set u = s\_last = 0 and go to Step 2.1. \\ 
\rowcolor[HTML]{C7EDCC}  \\ 
\rowcolor[HTML]{C7EDCC} 27. Step 2.1, since d[0] is not 0, skip to Step 2.2. \\ 
\rowcolor[HTML]{C7EDCC}  \\ 
\rowcolor[HTML]{C7EDCC} 28. Step 2.2, iterate through v from 0 to 2. For v = 0, A[0, 0] = 0, so continue to v = 1. For v = 1, A[0, 1] = 1, but color[1] = 2, so calculate bne\_v\_pi\_u = 1 - 0 = 1, and since it is not 0, set low[0] = min(low[0], d[1]) = min(1, 2) = 1. Continue to v = 2. For v = 2, A[0, 2] = 1, but color[2] = 2, so calculate bne\_v\_pi\_u = 2 - 0 = 2, and since it is not 0, set low[0] = min(low[0], d[2]) = min(1, 4) = 1. \\ 
\rowcolor[HTML]{C7EDCC}  \\ 
\rowcolor[HTML]{C7EDCC} 29. Step 2.3, calculate beq\_s\_last\_u = 0 - 0 = 0, which is equal to 0, so set color[0] = 2 and increment time to 6. \\ 
\rowcolor[HTML]{C7EDCC}  \\ 
\rowcolor[HTML]{C7EDCC} 30. Step 2.4, iterate through v from 0 to 2. For v = 0, print pi[0] (0) and u (0), calculate beq\_pi\_v\_u = 0 - 0 = 0, which is equal to 0, so go to Step 2.4.1. For v = 1, print pi[1] (0) and u (0), calculate beq\_pi\_v\_u = 0 - 0 = 0, which is equal to 0, so go to Step 2.4.1. For v = 2, print pi[2] (0) and u (0), calculate beq\_pi\_v\_u = 0 - 0 = 0, which is equal to 0, so go to Step 2.4.1. \\ 
\rowcolor[HTML]{C7EDCC}  \\ 
\rowcolor[HTML]{C7EDCC} 31. Step 2.4.1, for v = 0, low[0] = min(low[0], low[0]) = min(1, 1) = 1, bne\_u\_pi\_u = 0 - 0 = 0, so pass. For v = 1, low[0] = min(low[0], low[1]) = min(1, 2) = 1, bne\_u\_pi\_u = 0 - 0 = 0, so pass. For v = 2, low[0] = min(low[0], low[2]) = min(1, 4) = 1, bne\_u\_pi\_u = 0 - 0 = 0, so pass. \\ 
\rowcolor[HTML]{C7EDCC}  \\ 
\rowcolor[HTML]{C7EDCC} 32. Step 2.5, calculate beq\_u\_pi\_u = 0 - 0 = 0, which is equal to 0, so calculate bge\_cnt = child\_cnt[0] - 2 = 2 - 2 = 0, which is >= 0, so set is\_cut[0] to 1. \\ 
\rowcolor[HTML]{C7EDCC}  \\ 
\rowcolor[HTML]{C7EDCC} 33. Step 2.6, print s\_prev[0] (0), calculate beq\_u\_s\_prev\_u = 0 - 0 = 0, which is equal to 0, so break the main loop and go to Step 3. \\ 
\rowcolor[HTML]{C7EDCC}  \\ 
\rowcolor[HTML]{C7EDCC} 34. Step 3, print the is\_cut array: [1, 0, 0]. \\ 
\rowcolor[HTML]{C7EDCC}  \\ 
\rowcolor[HTML]{C7EDCC} The final is\_cut array is [1, 0, 0].
\\ \bottomrule
\end{tabular}
\caption{The prompt of articulation points and the response of \texttt{gpt-4}.}
\label{table:clrs_mini/articulation_points}

\end{table*}
\begin{table*}[h]
\small
\begin{tabular}{p{15cm}}
\toprule
\textbf{Prompt} \\ \midrule
Execute the instructions step by step. Do not jump steps. Do not stop before completion. \\ 
Initial: Set the square matrix A of size 3 x 3 : \\ 
A[0, 0] = 0, A[0, 1] = 1, A[0, 2] = 1 \\ 
A[1, 0] = 1, A[1, 1] = 0, A[1, 2] = 0 \\ 
A[2, 0] = 1, A[2, 1] = 0, A[2, 2] = 0. \\ 
Set color: color[0] = 0, color[1] = 0, color[2] = 0. \\ 
Set pi: pi[0] = 0, pi[1] = 1, pi[2] = 2. \\ 
Set d: d[0] = 0, d[1] = 0, d[2] = 0. \\ 
Set s\_prev: s\_prev[0] = 0, s\_prev[1] = 1, s\_prev[2] = 2. \\ 
Set time: 0. \\ 
Set low: low[0] = 0, low[1] = 0, low[2] = 0. \\ 
Set is\_bridge: []. \\ 
 \\ 
Step 1: Initialize the variable u to 0 and variable s\_last to 0. \\ 
 \\ 
Step 2: Start the while loop, and perform the following sub-steps: \\ 
 \\ 
\quad Step 2.1: If d[u] is 0: \\ 
\quad \quad a. Increment time by 1. \\ 
\quad \quad b. Set d[u] to time. \\ 
\quad \quad c. Set low[u] to time. \\ 
\quad \quad d. Set color[u] to 1. \\ 
 \\ 
\quad Step 2.2: Iterate through variable v from 0 to 2 (inclusive), and for each v: \\ 
\quad \quad a. If A[u, v] is not 0: \\ 
\quad \quad \quad i. If color[v] is 0: \\ 
\quad \quad \quad \quad 1. Set pi[v] to u. \\ 
\quad \quad \quad \quad 2. Set color[v] to 1. \\ 
\quad \quad \quad \quad 3. Print s\_last and set s\_prev[v] to s\_last. \\ 
\quad \quad \quad \quad 4. Set s\_last to v. \\ 
\quad \quad \quad \quad 5. Break the v iteration loop and go to Step 2.3. \\ 
\quad \quad \quad ii. If color[v] is not 0: \\ 
\quad \quad \quad \quad 1. Calculate bne\_v\_pi\_u = v - pi[u] and present the result. \\ 
\quad \quad \quad \quad 2. If bne\_v\_pi\_u is not 0, set low[u] to the minimum of low[u] and d[v]. \\ 
\quad \quad \quad \quad 3. If bne\_v\_pi\_u is 0 exactly, pass. \\ 
 \\ 
\quad Step 2.3: Calculate beq\_s\_last\_u = s\_last - u and present the result. \\ 
\quad \quad a. If beq\_s\_last\_u is equal to 0: \\ 
\quad \quad \quad i. Set color[u] to 2. \\ 
\quad \quad \quad ii. Increment time by 1. \\ 
\quad \quad b. If beq\_s\_last\_u is not 0, go to Step 2.7. \\ 
 \\ 
 \\ 
\quad Step 2.4: Iterate through variable v from 0 to 2 (inclusive), and for each v: \\ 
\quad \quad a. Print the value of pi[v] and u. Calculate beq\_pi\_v\_u = pi[v] - u and present the result. \\ 
\quad \quad b. If beq\_pi\_v\_u is equal to 0: \\ 
\quad \quad \quad i. Go to Step 2.4.1. \\ 
\quad \quad \quad ii. Go back to the iteration of v. \\ 
 \\ 
\quad Step 2.4.1: \\ 
\quad \quad a. Set low[u] to the minimum of low[u] and low[v]. \\ 
\quad \quad b. Print the values of low[v] and d[u]. Calculate bg\_low\_v\_d\_u = low[v] - d[u] and present the result. \\ 
\quad \quad \quad i. If bg\_low\_v\_d\_u > 0, append (u, v) to the list bridges. \\ 
\quad \quad \quad ii. If bg\_low\_v\_d\_u <= 0, pass. \\ 
 \\ 
\quad Step 2.5: Print the value of s\_prev[u]. Then, calculate beq\_u\_s\_prev\_u = u - s\_prev[u] and present the result. \\ 
\quad \quad a. If beq\_u\_s\_prev\_u is equal to 0, break the while loop and go to Step 3. \\ 
 \\ 
\quad Step 2.6: Swap s\_prev[s\_last] and s\_last: \\ 
\bottomrule
\end{tabular}
\end{table*}
\begin{table*}[!t]
\small
\begin{tabular}{p{15.5cm}}
\toprule\quad \quad a. Set temp to s\_prev[s\_last]. \\ 
\quad \quad b. Set s\_prev[s\_last] to s\_last. \\ 
\quad \quad c. Set s\_last to temp. \\ 
 \\ 
\quad Step 2.7: Set u to s\_last and go to Step 2.1. \\ 
 \\ 
Step 3: Print the list bridges. \\ 
 \\ 
----- \\ 
The computation procedure is as follows, with each line numbered sequentially (e.g., 1. 2. 3. 4. 5. 6. 7. 8. 9. etc.): \\ 
1. Step 1, set u = 0 and s\_last = 0. \\ 
\rowcolor[HTML]{C7EDCC} 2. Step 2, enter the while loop. \\ 
\rowcolor[HTML]{C7EDCC}  \\ 
\rowcolor[HTML]{C7EDCC} 3. Step 2.1, since d[0] is 0: \\ 
\rowcolor[HTML]{C7EDCC} \quad a. Increment time to 1. \\ 
\rowcolor[HTML]{C7EDCC} \quad b. Set d[0] to 1. \\ 
\rowcolor[HTML]{C7EDCC} \quad c. Set low[0] to 1. \\ 
\rowcolor[HTML]{C7EDCC} \quad d. Set color[0] to 1. \\ 
\rowcolor[HTML]{C7EDCC}  \\ 
\rowcolor[HTML]{C7EDCC} 4. Step 2.2, iterate through v from 0 to 2: \\ 
\rowcolor[HTML]{C7EDCC} \quad a. For v = 0, A[0, 0] is 0, so continue to the next v. \\ 
\rowcolor[HTML]{C7EDCC} \quad b. For v = 1, A[0, 1] is 1: \\ 
\rowcolor[HTML]{C7EDCC} \quad \quad i. color[1] is 0, so: \\ 
\rowcolor[HTML]{C7EDCC} \quad \quad \quad 1. Set pi[1] to 0. \\ 
\rowcolor[HTML]{C7EDCC} \quad \quad \quad 2. Set color[1] to 1. \\ 
\rowcolor[HTML]{C7EDCC} \quad \quad \quad 3. Print s\_last (0) and set s\_prev[1] to 0. \\ 
\rowcolor[HTML]{C7EDCC} \quad \quad \quad 4. Set s\_last to 1. \\ 
\rowcolor[HTML]{C7EDCC} \quad \quad \quad 5. Break the v iteration loop. \\ 
\rowcolor[HTML]{C7EDCC}  \\ 
\rowcolor[HTML]{C7EDCC} 5. Step 2.3, calculate beq\_s\_last\_u = 1 - 0 = 1, which is not 0, so go to Step 2.7. \\ 
\rowcolor[HTML]{C7EDCC}  \\ 
\rowcolor[HTML]{C7EDCC} 6. Step 2.7, set u to s\_last (1) and go to Step 2.1. \\ 
\rowcolor[HTML]{C7EDCC}  \\ 
\rowcolor[HTML]{C7EDCC} 7. Step 2.1, since d[1] is 0: \\ 
\rowcolor[HTML]{C7EDCC} \quad a. Increment time to 2. \\ 
\rowcolor[HTML]{C7EDCC} \quad b. Set d[1] to 2. \\ 
\rowcolor[HTML]{C7EDCC} \quad c. Set low[1] to 2. \\ 
\rowcolor[HTML]{C7EDCC} \quad d. Set color[1] to 1. \\
\rowcolor[HTML]{C7EDCC} 8. Step 2.2, iterate through v from 0 to 2: \\ 
\rowcolor[HTML]{C7EDCC} \quad a. For v = 0, A[1, 0] is 1: \\ 
\rowcolor[HTML]{C7EDCC} \quad \quad i. color[0] is 1, so: \\ 
\rowcolor[HTML]{C7EDCC} \quad \quad \quad 1. Calculate bne\_v\_pi\_u = 0 - pi[1] = 0 - 0 = 0, which is 0, so pass. \\ 
\rowcolor[HTML]{C7EDCC} \quad b. For v = 1, A[1, 1] is 0, so continue to the next v. \\ 
\rowcolor[HTML]{C7EDCC} \quad c. For v = 2, A[1, 2] is 0, so continue to the next v. \\ 
\rowcolor[HTML]{C7EDCC}  \\ 
\rowcolor[HTML]{C7EDCC} 9. Step 2.3, calculate beq\_s\_last\_u = 1 - 1 = 0, which is 0, so: \\ 
\rowcolor[HTML]{C7EDCC} \quad a. Set color[1] to 2. \\ 
\rowcolor[HTML]{C7EDCC} \quad b. Increment time to 3. \\ 
\rowcolor[HTML]{C7EDCC}  \\ 
\rowcolor[HTML]{C7EDCC} 10. Step 2.4, iterate through v from 0 to 2: \\ 
\rowcolor[HTML]{C7EDCC} \quad a. For v = 0, print pi[0] (0) and u (1), calculate beq\_pi\_v\_u = pi[0] - u = 0 - 1 = -1, which is not 0. \\ 
\rowcolor[HTML]{C7EDCC} \quad b. For v = 1, print pi[1] (0) and u (1), calculate beq\_pi\_v\_u = pi[1] - u = 0 - 1 = -1, which is not 0. \\ 
\rowcolor[HTML]{C7EDCC} \quad c. For v = 2, print pi[2] (2) and u (1), calculate beq\_pi\_v\_u = pi[2] - u = 2 - 1 = 1, which is not 0. \\ 
\rowcolor[HTML]{C7EDCC}  \\ 
\rowcolor[HTML]{C7EDCC} 11. Step 2.5, print s\_prev[1] (0), calculate beq\_u\_s\_prev\_u = 1 - s\_prev[1] = 1 - 0 = 1, which is not 0. \\ 
\rowcolor[HTML]{C7EDCC}  \\ 
\rowcolor[HTML]{C7EDCC} 12. Step 2.6, swap s\_prev[s\_last] and s\_last: \\ 
\rowcolor[HTML]{C7EDCC} \quad a. Set temp to s\_prev[1] (0). \\ 
\rowcolor[HTML]{C7EDCC} \quad b. Set s\_prev[1] to s\_last (1). \\ 
\rowcolor[HTML]{C7EDCC} \quad c. Set s\_last to temp (0). \\ 
\rowcolor[HTML]{C7EDCC}  \\ 
\bottomrule
\end{tabular}
\end{table*}
\begin{table*}[!t]
\small
\begin{tabular}{p{15.5cm}}
\toprule
\rowcolor[HTML]{C7EDCC} 13. Step 2.7, set u to s\_last (0) and go to Step 2.1. \\ 
\rowcolor[HTML]{C7EDCC}  \\ 
\rowcolor[HTML]{C7EDCC} 14. Step 2.1, since d[0] is not 0, skip this step. \\ 
\rowcolor[HTML]{C7EDCC}  \\ 
\rowcolor[HTML]{C7EDCC} 15. Step 2.2, iterate through v from 0 to 2: \\ 
\rowcolor[HTML]{C7EDCC} \quad a. For v = 0, A[0, 0] is 0, so continue to the next v. \\ 
\rowcolor[HTML]{C7EDCC} \quad b. For v = 1, A[0, 1] is 1: \\ 
\rowcolor[HTML]{C7EDCC} \quad \quad i. color[1] is 2, so: \\ 
\rowcolor[HTML]{C7EDCC} \quad \quad \quad 1. Calculate bne\_v\_pi\_u = 1 - pi[0] = 1 - 0 = 1, which is not 0, so set low[0] to min(low[0], d[1]) = min(1, 2) = 1. \\ 
\rowcolor[HTML]{C7EDCC} \quad c. For v = 2, A[0, 2] is 1: \\ 
\rowcolor[HTML]{C7EDCC} \quad \quad i. color[2] is 0, so: \\ 
\rowcolor[HTML]{C7EDCC} \quad \quad \quad 1. Set pi[2] to 0. \\ 
\rowcolor[HTML]{C7EDCC} \quad \quad \quad 2. Set color[2] to 1. \\ 
\rowcolor[HTML]{C7EDCC} \quad \quad \quad 3. Print s\_last (0) and set s\_prev[2] to 0. \\ 
\rowcolor[HTML]{C7EDCC} \quad \quad \quad 4. Set s\_last to 2. \\ 
\rowcolor[HTML]{C7EDCC} \quad \quad \quad 5. Break the v iteration loop. \\ 
\rowcolor[HTML]{C7EDCC}  \\ 
\rowcolor[HTML]{C7EDCC} 16. Step 2.3, calculate beq\_s\_last\_u = 2 - 0 = 2, which is not 0, so go to Step 2.7. \\ 
\rowcolor[HTML]{C7EDCC}  \\ 
\rowcolor[HTML]{C7EDCC} 17. Step 2.7, set u to s\_last (2) and go to Step 2.1. \\ 
\rowcolor[HTML]{C7EDCC}  \\ 
\rowcolor[HTML]{C7EDCC} 18. Step 2.1, since d[2] is 0: \\ 
\rowcolor[HTML]{C7EDCC} \quad a. Increment time to 4. \\ 
\rowcolor[HTML]{C7EDCC} \quad b. Set d[2] to 4. \\ 
\rowcolor[HTML]{C7EDCC} \quad c. Set low[2] to 4. \\ 
\rowcolor[HTML]{C7EDCC} \quad d. Set color[2] to 1. \\ 
\rowcolor[HTML]{C7EDCC}  \\ 
\rowcolor[HTML]{C7EDCC} 19. Step 2.2, iterate through v from 0 to 2: \\ 
\rowcolor[HTML]{C7EDCC} \quad a. For v = 0, A[2, 0] is 1: \\ 
\rowcolor[HTML]{C7EDCC} \quad \quad i. color[0] is 1, so: \\ 
\rowcolor[HTML]{C7EDCC} \quad \quad \quad 1. Calculate bne\_v\_pi\_u = 0 - pi[2] = 0 - 0 = 0, which is 0, so pass. \\ 
\rowcolor[HTML]{C7EDCC} \quad b. For v = 1, A[2, 1] is 0, so continue to the next v. \\ 
\rowcolor[HTML]{C7EDCC} \quad c. For v = 2, A[2, 2] is 0, so continue to the next v. \\ 
\rowcolor[HTML]{C7EDCC}  \\ 
\rowcolor[HTML]{C7EDCC} 20. Step 2.3, calculate beq\_s\_last\_u = 2 - 2 = 0, which is 0, so: \\ 
\rowcolor[HTML]{C7EDCC} \quad a. Set color[2] to 2. \\ 
\rowcolor[HTML]{C7EDCC} \quad b. Increment time to 5. \\ 
\rowcolor[HTML]{C7EDCC}  \\ 
\rowcolor[HTML]{C7EDCC} 21. Step 2.4, iterate through v from 0 to 2: \\ 
\rowcolor[HTML]{C7EDCC} \quad a. For v = 0, print pi[0] (0) and u (2), calculate beq\_pi\_v\_u = pi[0] - u = 0 - 2 = -2, which is not 0. \\ 
\rowcolor[HTML]{C7EDCC} \quad b. For v = 1, print pi[1] (0) and u (2), calculate beq\_pi\_v\_u = pi[1] - u = 0 - 2 = -2, which is not 0. \\ 
\rowcolor[HTML]{C7EDCC} \quad c. For v = 2, print pi[2] (0) and u (2), calculate beq\_pi\_v\_u = pi[2] - u = 0 - 2 = -2, which is not 0. \\ 
\rowcolor[HTML]{C7EDCC}  \\ 
\rowcolor[HTML]{C7EDCC} 22. Step 2.5, print s\_prev[2] (0), calculate beq\_u\_s\_prev\_u = 2 - s\_prev[2] = 2 - 0 = 2, which is not 0. \\ 
\rowcolor[HTML]{C7EDCC}  \\ 
\rowcolor[HTML]{C7EDCC} 23. Step 2.6, swap s\_prev[s\_last] and s\_last: \\ 
\rowcolor[HTML]{C7EDCC} \quad a. Set temp to s\_prev[2] (0). \\ 
\rowcolor[HTML]{C7EDCC} \quad b. Set s\_prev[2] to s\_last (2). \\ 
\rowcolor[HTML]{C7EDCC} \quad c. Set s\_last to temp (0). \\ 
\rowcolor[HTML]{C7EDCC}  \\ 
\rowcolor[HTML]{C7EDCC} 24. Step 2.7, set u to s\_last (0) and go to Step 2.1. \\ 
\rowcolor[HTML]{C7EDCC}  \\ 
\rowcolor[HTML]{C7EDCC} 25. Step 2.1, since d[0] is not 0, skip this step. \\ 
\rowcolor[HTML]{C7EDCC}  \\ 
\rowcolor[HTML]{C7EDCC} 26. Step 2.2, iterate through v from 0 to 2: \\ 
\rowcolor[HTML]{C7EDCC} \quad a. For v = 0, A[0, 0] is 0, so continue to the next v. \\ 
\rowcolor[HTML]{C7EDCC} \quad b. For v = 1, A[0, 1] is 1: \\ 
\rowcolor[HTML]{C7EDCC} \quad \quad i. color[1] is 2, so: \\ 
\rowcolor[HTML]{C7EDCC} \quad \quad \quad 1. Calculate bne\_v\_pi\_u = 1 - pi[0] = 1 - 0 = 1, which is not 0, so set low[0] to min(low[0], d[1]) = min(1, 2) = 1. \\ 
\rowcolor[HTML]{C7EDCC} \quad c. For v = 2, A[0, 2] is 1: \\ 
\bottomrule
\end{tabular}
\end{table*}
\begin{table*}[!t]
\small
\begin{tabular}{p{15.5cm}}
\toprule
\rowcolor[HTML]{C7EDCC} \quad \quad i. color[2] is 2, so: \\ 
\rowcolor[HTML]{C7EDCC} \quad \quad \quad 1. Calculate bne\_v\_pi\_u = 2 - pi[0] = 2 - 0 = 2, which is not 0, so set low[0] to min(low[0], d[2]) = min(1, 4) = 1. \\ 
\rowcolor[HTML]{C7EDCC}  \\ 
\rowcolor[HTML]{C7EDCC} 27. Step 2.3, calculate beq\_s\_last\_u = 0 - 0 = 0, which is 0, so: \\ 
\rowcolor[HTML]{C7EDCC} \quad a. Set color[0] to 2. \\ 
\rowcolor[HTML]{C7EDCC} \quad b. Increment time to 6. \\ 
\rowcolor[HTML]{C7EDCC}  \\ 
\rowcolor[HTML]{C7EDCC} 28. Step 2.4, iterate through v from 0 to 2: \\ 
\rowcolor[HTML]{C7EDCC} \quad a. For v = 0, print pi[0] (0) and u (0), calculate beq\_pi\_v\_u = pi[0] - u = 0 - 0 = 0, which is 0, so go to Step 2.4.1. \\ 
\rowcolor[HTML]{C7EDCC} \quad \quad i. Step 2.4.1: \\ 
\rowcolor[HTML]{C7EDCC} \quad \quad \quad a. Set low[0] to min(low[0], low[0]) = min(1, 1) = 1. \\ 
\rowcolor[HTML]{C7EDCC} \quad \quad \quad b. Print low[0] (1) and d[0] (1), calculate bg\_low\_v\_d\_u = low[0] - d[0] = 1 - 1 = 0, which is <= 0, so pass. \\ 
\rowcolor[HTML]{C7EDCC} \quad b. For v = 1, print pi[1] (0) and u (0), calculate beq\_pi\_v\_u = pi[1] - u = 0 - 0 = 0, which is 0, so go to Step 2.4.1. \\ 
\rowcolor[HTML]{C7EDCC} \quad \quad i. Step 2.4.1: \\ 
\rowcolor[HTML]{C7EDCC} \quad \quad \quad a. Set low[0] to min(low[0], low[1]) = min(1, 2) = 1. \\ 
\rowcolor[HTML]{C7EDCC} \quad \quad \quad b. Print low[1] (2) and d[0] (1), calculate bg\_low\_v\_d\_u = low[1] - d[0] = 2 - 1 = 1, which is > 0, so append (0, 1) to the list bridges. \\ 
\rowcolor[HTML]{C7EDCC} \quad c. For v = 2, print pi[2] (0) and u (0), calculate beq\_pi\_v\_u = pi[2] - u = 0 - 0 = 0, which is 0, so go to Step 2.4.1. \\ 
\rowcolor[HTML]{C7EDCC} \quad \quad i. Step 2.4.1: \\ 
\rowcolor[HTML]{C7EDCC} \quad \quad \quad a. Set low[0] to min(low[0], low[2]) = min(1, 4) = 1. \\ 
\rowcolor[HTML]{C7EDCC} \quad \quad \quad b. Print low[2] (4) and d[0] (1), calculate bg\_low\_v\_d\_u = low[2] - d[0] = 4 - 1 = 3, which is > 0, so append (0, 2) to the list bridges. \\ 
\rowcolor[HTML]{C7EDCC}  \\ 
\rowcolor[HTML]{C7EDCC} 29. Step 2.5, print s\_prev[0] (0), calculate beq\_u\_s\_prev\_u = 0 - s\_prev[0] = 0 - 0 = 0, which is 0, so break the while loop and go to Step 3. \\ 
\rowcolor[HTML]{C7EDCC}  \\ 
\rowcolor[HTML]{C7EDCC} 30. Step 3, print the list bridges: [(0, 1), (0, 2)]. \\ 
\rowcolor[HTML]{C7EDCC}  \\ 
\rowcolor[HTML]{C7EDCC} The list of bridges is [(0, 1), (0, 2)].
\\ \bottomrule
\end{tabular}
\caption{The prompt of bridges and the response of \texttt{gpt-4}.}
\label{table:clrs_mini/bridges}

\end{table*}
\begin{table*}[h]
\small
\begin{tabular}{p{15cm}}
\toprule
\textbf{Prompt} \\ \midrule
Execute the instructions step by step. Do not jump steps. Do not stop before completion. \\ 
Initial: Set the square matrix A of size 3 x 3 : \\ 
A[0, 0] = 0, A[0, 1] = 1, A[0, 2] = 1 \\ 
A[1, 0] = 1, A[1, 1] = 0, A[1, 2] = 0 \\ 
A[2, 0] = 0, A[2, 1] = 0, A[2, 2] = 0. \\ 
Set color: color[0] = 0, color[1] = 0, color[2] = 0. \\ 
Set d: d[0] = 0, d[1] = 0, d[2] = 0. \\ 
Set f: f[0] = 0, f[1] = 0, f[2] = 0. \\ 
Set s\_prev: s\_prev[0] = 0, s\_prev[1] = 1, s\_prev[2] = 2. \\ 
Set B: B[0, 0] = 0, B[0, 1] = 1, B[0, 2] = 0 \\ 
B[1, 0] = 1, B[1, 1] = 0, B[1, 2] = 0 \\ 
B[2, 0] = 1, B[2, 1] = 0, B[2, 2] = 0. \\ 
Set scc\_id: scc\_id[0] = 0, scc\_id[1] = 1, scc\_id[2] = 2. \\ 
 \\ 
Step 1: Initialize variable time to 0. \\ 
 \\ 
Step 2: Iterate through the variable s from 0 to 2 (inclusive), and for each s: \\ 
 \\ 
\quad Step 2.1: Check if color[s] is equal to 0. If true, perform the following steps: \\ 
 \\ 
\quad \quad Step 2.2: Set s\_last to s. \\ 
\quad \quad Step 2.3: Set u to s. \\ 
\quad \quad Step 2.4: Set v to s. \\ 
\quad \quad  \\ 
\bottomrule
\end{tabular}
\end{table*}
\begin{table*}[!t]
\small
\begin{tabular}{p{15.5cm}}
\toprule
\quad \quad Step 2.5: Start the while loop, and perform the following sub-steps: \\ 
\quad \quad \quad  \\ 
\quad \quad \quad Step 2.5.1: If color[u] is 0 or d[u] is 0: \\ 
\quad \quad \quad \quad a. Increment time by 1. \\ 
\quad \quad \quad \quad b. Set d[u] to time. \\ 
\quad \quad \quad \quad c. Set color[u] to 1. \\ 
\quad \quad \quad Otherwise, do nothing. \\ 
 \\ 
\quad \quad \quad Step 2.5.2: Iterate through the variable v from 0 to 2 (inclusive), and for each v: \\ 
\quad \quad \quad \quad a. If A[u, v] is not 0, then check the value of color[v]: \\ 
\quad \quad \quad \quad \quad i. If color[v] is 0: \\ 
\quad \quad \quad \quad \quad \quad 1. Set color[v] to 1. \\ 
\quad \quad \quad \quad \quad \quad 2. Set s\_prev[v] to s\_last. \\ 
\quad \quad \quad \quad \quad \quad 3. Set s\_last to v. \\ 
\quad \quad \quad \quad \quad \quad 4. Break the v iteration loop. \\ 
\quad \quad \quad \quad \quad i. If color[v] is 1 or is 2: \\ 
\quad \quad \quad \quad \quad \quad 1. Continue the v iteration loop. \\ 
 \\ 
\quad \quad \quad Step 2.5.3: Calculate beq\_s\_last\_u = s\_last - u and present the result.  \\ 
\quad \quad \quad \quad a. If beq\_s\_last\_u is equal to 0: \\ 
\quad \quad \quad \quad \quad i. Set color[u] to 2. \\ 
\quad \quad \quad \quad \quad ii. Increment time by 1. \\ 
\quad \quad \quad \quad \quad iii. Set f[u] to time. \\ 
\quad \quad \quad \quad \quad iv. Go to Step 2.5.4. \\ 
\quad \quad \quad \quad b. If beq\_s\_last\_u is not equal to 0, go to Step 2.5.5 \\ 
 \\ 
\quad \quad \quad Step 2.5.4: Calculate beq\_s\_prev\_u\_u = s\_prev[u] - u and present the result. \\ 
\quad \quad \quad \quad a. If beq\_s\_prev\_u\_u is equal to 0: break the while loop and go to the next iteration of s. \\ 
\quad \quad \quad \quad b. If beq\_s\_prev\_u\_u is not equal to 0: \\ 
 \\ 
\quad \quad \quad \quad \quad Step 2.5.4.1: Swap s\_prev[s\_last] and s\_last: \\ 
\quad \quad \quad \quad \quad \quad a. Print the value of s\_last and s\_prev[s\_last]. \\ 
\quad \quad \quad \quad \quad \quad b. Set temp to s\_prev[s\_last]. \\ 
\quad \quad \quad \quad \quad \quad c. Set s\_prev[s\_last] to s\_last. \\ 
\quad \quad \quad \quad \quad \quad d. Set s\_last to temp. \\ 
 \\ 
\quad \quad \quad Step 2.5.5: Set u to s\_last and go back to the beginning of the while loop, which is Step 2.5.1 \\ 
 \\ 
Step 3: Reinitialize color to color[0] = 0, color[1] = 0, color[2] = 0. Set s\_prev to s\_prev[0] = 0, s\_prev[1] = 1, s\_prev[2] = 2. \\ 
 \\ 
Step 4: Argsort f array in descending order. Perform the following sub-steps: \\ 
 \\ 
\quad Step 4.1: Print each element of the f array along with its index in the format (f[i], i) for i in the range of 0 to 2. \\ 
 \\ 
\quad Step 4.2: Sort the pairs (f[i], i) in descending order based on f[i]. Reorder the pairs according to the sorted f[i] values.  \\ 
 \\ 
\quad Step 4.3: Extract the index i, which is the second element from each sorted pair, and create the visit\_order array. \\ 
 \\ 
\quad Step 4.4: Print the visit\_order array. \\ 
 \\ 
Step 5: Iterate through the variable s in visit\_order, and for each s: \\ 
 \\ 
\quad Step 5.1: Check if color[s] is equal to 0. If true, perform the following steps: \\ 
 \\ 
\quad \quad Step 5.2: Set s\_last to s. \\ 
\quad \quad Step 5.3: Set u to s. \\ 
\quad \quad Step 5.4: Set v to s. \\ 
 \\ 
 \bottomrule
\end{tabular}
\end{table*}
\begin{table*}[!t]
\small
\begin{tabular}{p{15.5cm}}
\toprule
\quad \quad Step 5.5: Start the while loop, and perform the following sub-steps: \\ 
\quad \quad \quad Step 5.5.1: Set scc\_id[u] to s. \\ 
\quad \quad \quad Step 5.5.2: If color[u] is 0 or d[u] is 0: \\ 
\quad \quad \quad \quad a. Increment time by 1. \\ 
\quad \quad \quad \quad b. Set d[u] to time. \\ 
\quad \quad \quad \quad c. Set color[u] to 1. \\ 
\quad \quad \quad Otherwise, do nothing. \\ 

 \\ 
 \\ 
\quad \quad \quad Step 5.5.3: Iterate through the variable v from 0 to 2 (inclusive), and for each v: \\ 
\quad \quad \quad \quad a. If B[u, v] is not 0, then check the value of color[v]: \\ 
\quad \quad \quad \quad \quad i. If color[v] is 0: \\ 
\quad \quad \quad \quad \quad \quad 1. Set color[v] to 1. \\ 
\quad \quad \quad \quad \quad \quad 2. Set s\_prev[v] to s\_last. \\ 
\quad \quad \quad \quad \quad \quad 3. Set s\_last to v. \\ 
\quad \quad \quad \quad \quad \quad 4. Break the v iteration loop. \\ 
\quad \quad \quad \quad \quad ii. If color[v] is 1 or is 2: Continue the v iteration loop. \\ 
 \\ 
\quad \quad \quad Step 5.5.4: Calculate beq\_s\_last\_u = s\_last - u and present the result.  \\ 
\quad \quad \quad \quad a. If beq\_s\_last\_u is equal to 0: \\ 
\quad \quad \quad \quad \quad i. Set color[u] to 2. \\ 
\quad \quad \quad \quad \quad ii. Increment time by 1. \\ 
\quad \quad \quad \quad \quad iii. Set f[u] to time. \\ 
\quad \quad \quad \quad b. If beq\_s\_last\_u is not equal to 0, go to Step 5.5.6 \\ 
 \\ 
\quad \quad \quad Step 5.5.5: Calculate beq\_s\_prev\_u\_u = s\_prev[u] - u and present the result. \\ 
\quad \quad \quad \quad a. If beq\_s\_prev\_u\_u is equal to 0: break the while loop and go to the next iteration of s. \\ 
\quad \quad \quad \quad b. If beq\_s\_prev\_u\_u is not equal to 0: \\ 
 \\ 
\quad \quad \quad \quad \quad Step 5.5.5.1: Swap s\_prev[s\_last] and s\_last: \\ 
\quad \quad \quad \quad \quad \quad a. Print the value of s\_last and s\_prev[s\_last]. \\ 
\quad \quad \quad \quad \quad \quad b. Set temp to s\_prev[s\_last]. \\ 
\quad \quad \quad \quad \quad \quad c. Set s\_prev[s\_last] to s\_last. \\ 
\quad \quad \quad \quad \quad \quad d. Set s\_last to temp. \\ 
 \\ 
\quad \quad \quad Step 5.5.6: Set u to s\_last and go back to the beginning of the while loop, which is Step 5.5.1. \\ 
 \\ 
Step 6: Print the scc\_id array by iterating over its elements. \\ 
 \\ 
----- \\ 
The computation procedure is as follows, with each line numbered sequentially (e.g., 1. 2. 3. 4. 5. 6. 7. 8. 9. etc.): \\ 
1. Step 1, set time = 0. \\ 
\rowcolor[HTML]{C7EDCC} 2. Step 2, iterate s from 0 to 2. \\ 
\rowcolor[HTML]{C7EDCC} 3. Step 2.1, for s = 0, color[0] = 0, so proceed to Step 2.2. \\ 
\rowcolor[HTML]{C7EDCC} 4. Step 2.2, set s\_last = 0. \\ 
\rowcolor[HTML]{C7EDCC} 5. Step 2.3, set u = 0. \\ 
\rowcolor[HTML]{C7EDCC} 6. Step 2.4, set v = 0. \\ 
\rowcolor[HTML]{C7EDCC} 7. Step 2.5, start the while loop. \\ 
\rowcolor[HTML]{C7EDCC} 8. Step 2.5.1, color[0] = 0 and d[0] = 0, so increment time to 1, set d[0] = 1, and set color[0] = 1. \\ 
\rowcolor[HTML]{C7EDCC} 9. Step 2.5.2, iterate v from 0 to 2. \\ 
\rowcolor[HTML]{C7EDCC} 10. Step 2.5.2.a, for v = 0, A[0, 0] = 0, so continue to v = 1. \\ 
\rowcolor[HTML]{C7EDCC} 11. Step 2.5.2.a, for v = 1, A[0, 1] = 1, so color[1] = 0, set color[1] = 1, s\_prev[1] = 0, s\_last = 1, and break the v loop. \\ 
\rowcolor[HTML]{C7EDCC} 12. Step 2.5.3, beq\_s\_last\_u = 1 - 0 = 1, so go to Step 2.5.5. \\ 
\rowcolor[HTML]{C7EDCC} 13. Step 2.5.5, set u = 1. \\ 
\rowcolor[HTML]{C7EDCC} 14. Step 2.5.1, color[1] = 1 and d[1] = 0, so increment time to 2, set d[1] = 2, and set color[1] = 1. \\ 
\rowcolor[HTML]{C7EDCC} 15. Step 2.5.2, iterate v from 0 to 2. \\ 
\rowcolor[HTML]{C7EDCC} 16. Step 2.5.2.a, for v = 0, A[1, 0] = 1, so color[0] = 1, continue to v = 1. \\ 
\rowcolor[HTML]{C7EDCC} 17. Step 2.5.2.a, for v = 1, A[1, 1] = 0, so continue to v = 2. \\ 
\rowcolor[HTML]{C7EDCC} 18. Step 2.5.2.a, for v = 2, A[1, 2] = 0, so continue. \\ 
\rowcolor[HTML]{C7EDCC} 19. Step 2.5.3, beq\_s\_last\_u = 1 - 1 = 0, so set color[1] = 2, increment time to 3, set f[1] = 3, and go to Step 2.5.4. \\ 
\rowcolor[HTML]{C7EDCC} 20. Step 2.5.4, beq\_s\_prev\_u\_u = 0 - 1 = -1, so go to Step 2.5.4.1. \\ 
\bottomrule
\end{tabular}
\end{table*}
\begin{table*}[!t]
\small
\begin{tabular}{p{15.5cm}}
\toprule
\rowcolor[HTML]{C7EDCC} 21. Step 2.5.4.1, swap s\_prev[s\_last] and s\_last, set temp = s\_prev[1] = 0, s\_prev[1] = 1, s\_last = 0. \\ 
\rowcolor[HTML]{C7EDCC} 22. Step 2.5.5, set u = 0. \\ 
\rowcolor[HTML]{C7EDCC} 23. Step 2.5.1, color[0] = 1 and d[0] = 1, so continue to Step 2.5.2. \\ 
\rowcolor[HTML]{C7EDCC} 24. Step 2.5.2, iterate v from 0 to 2. \\ 
\rowcolor[HTML]{C7EDCC} 25. Step 2.5.2.a, for v = 0, A[0, 0] = 0, so continue to v = 1. \\ 
\rowcolor[HTML]{C7EDCC} 26. Step 2.5.2.a, for v = 1, A[0, 1] = 1, so color[1] = 2, continue to v = 2. \\ 
\rowcolor[HTML]{C7EDCC} 27. Step 2.5.2.a, for v = 2, A[0, 2] = 1, so color[2] = 0, set color[2] = 1, s\_prev[2] = 0, s\_last = 2, and break the v loop. \\ 
\rowcolor[HTML]{C7EDCC} 28. Step 2.5.3, beq\_s\_last\_u = 2 - 0 = 2, so go to Step 2.5.5. \\ 
\rowcolor[HTML]{C7EDCC} 29. Step 2.5.5, set u = 2. \\ 
\rowcolor[HTML]{C7EDCC} 30. Step 2.5.1, color[2] = 1 and d[2] = 0, so increment time to 4, set d[2] = 4, and set color[2] = 1. \\ 
\rowcolor[HTML]{C7EDCC} 31. Step 2.5.2, iterate v from 0 to 2. \\ 
\rowcolor[HTML]{C7EDCC} 32. Step 2.5.2.a, for v = 0, A[2, 0] = 0, so continue to v = 1. \\ 
\rowcolor[HTML]{C7EDCC} 33. Step 2.5.2.a, for v = 1, A[2, 1] = 0, so continue to v = 2. \\ 
\rowcolor[HTML]{C7EDCC} 34. Step 2.5.2.a, for v = 2, A[2, 2] = 0, so continue. \\ 
\rowcolor[HTML]{C7EDCC} 35. Step 2.5.3, beq\_s\_last\_u = 2 - 2 = 0, so set color[2] = 2, increment time to 5, set f[2] = 5, and go to Step 2.5.4. \\ 

\rowcolor[HTML]{C7EDCC} 36. Step 2.5.4, beq\_s\_prev\_u\_u = 0 - 2 = -2, so go to Step 2.5.4.1. \\ 
\rowcolor[HTML]{C7EDCC} 37. Step 2.5.4.1, swap s\_prev[s\_last] and s\_last, set temp = s\_prev[2] = 0, s\_prev[2] = 2, s\_last = 0. \\ 
\rowcolor[HTML]{C7EDCC} 38. Step 2.5.5, set u = 0. \\ 
\rowcolor[HTML]{C7EDCC} 39. Step 2.5.1, color[0] = 1 and d[0] = 1, so continue to Step 2.5.2. \\ 
\rowcolor[HTML]{C7EDCC} 40. Step 2.5.2, iterate v from 0 to 2. \\ 
\rowcolor[HTML]{C7EDCC} 41. Step 2.5.2.a, for v = 0, A[0, 0] = 0, so continue to v = 1. \\ 
\rowcolor[HTML]{C7EDCC} 42. Step 2.5.2.a, for v = 1, A[0, 1] = 1, so color[1] = 2, continue to v = 2. \\ 
\rowcolor[HTML]{C7EDCC} 43. Step 2.5.2.a, for v = 2, A[0, 2] = 1, so color[2] = 2, continue. \\ 
\rowcolor[HTML]{C7EDCC} 44. Step 2.5.3, beq\_s\_last\_u = 0 - 0 = 0, so set color[0] = 2, increment time to 6, set f[0] = 6, and go to Step 2.5.4. \\ 
\rowcolor[HTML]{C7EDCC} 45. Step 2.5.4, beq\_s\_prev\_u\_u = 0 - 0 = 0, so break the while loop and go to the next iteration of s. \\ 
\rowcolor[HTML]{C7EDCC} 46. Step 2.1, for s = 1, color[1] = 2, so continue to s = 2. \\ 
\rowcolor[HTML]{C7EDCC} 47. Step 2.1, for s = 2, color[2] = 2, so continue. \\ 
\rowcolor[HTML]{C7EDCC} 48. Step 3, reinitialize color to color[0] = 0, color[1] = 0, color[2] = 0, and set s\_prev to s\_prev[0] = 0, s\_prev[1] = 1, s\_prev[2] = 2. \\ 
\rowcolor[HTML]{C7EDCC} 49. Step 4, argsort f array in descending order. \\ 
\rowcolor[HTML]{C7EDCC} 50. Step 4.1, print (f[0], 0) = (6, 0), (f[1], 1) = (3, 1), (f[2], 2) = (5, 2). \\ 
\rowcolor[HTML]{C7EDCC} 51. Step 4.2, sort the pairs in descending order: (6, 0), (5, 2), (3, 1). \\ 
\rowcolor[HTML]{C7EDCC} 52. Step 4.3, create visit\_order array: [0, 2, 1]. \\ 
\rowcolor[HTML]{C7EDCC} 53. Step 4.4, print visit\_order: [0, 2, 1]. \\ 
\rowcolor[HTML]{C7EDCC} 54. Step 5, iterate s in visit\_order. \\ 
\rowcolor[HTML]{C7EDCC} 55. Step 5.1, for s = 0, color[0] = 0, so proceed to Step 5.2. \\ 
\rowcolor[HTML]{C7EDCC} 56. Step 5.2, set s\_last = 0. \\ 
\rowcolor[HTML]{C7EDCC} 57. Step 5.3, set u = 0. \\ 
\rowcolor[HTML]{C7EDCC} 58. Step 5.4, set v = 0. \\ 
\rowcolor[HTML]{C7EDCC} 59. Step 5.5, start the while loop. \\ 
\rowcolor[HTML]{C7EDCC} 60. Step 5.5.1, set scc\_id[0] = 0. \\ 
\rowcolor[HTML]{C7EDCC} 61. Step 5.5.2, color[0] = 0 and d[0] = 0, so increment time to 7, set d[0] = 7, and set color[0] = 1. \\ 
\rowcolor[HTML]{C7EDCC} 62. Step 5.5.3, iterate v from 0 to 2. \\ 
\rowcolor[HTML]{C7EDCC} 63. Step 5.5.3.a, for v = 0, B[0, 0] = 0, so continue to v = 1. \\ 
\rowcolor[HTML]{C7EDCC} 64. Step 5.5.3.a, for v = 1, B[0, 1] = 1, so color[1] = 0, set color[1] = 1, s\_prev[1] = 0, s\_last = 1, and break the v loop. \\ 
\rowcolor[HTML]{C7EDCC} 65. Step 5.5.4, beq\_s\_last\_u = 1 - 0 = 1, so go to Step 5.5.6. \\ 
\rowcolor[HTML]{C7EDCC} 66. Step 5.5.6, set u = 1. \\ 
\rowcolor[HTML]{C7EDCC} 67. Step 5.5.1, set scc\_id[1] = 0. \\ 
\rowcolor[HTML]{C7EDCC} 68. Step 5.5.2, color[1] = 1 and d[1] = 0, so increment time to 8, set d[1] = 8, and set color[1] = 1. \\ 
\rowcolor[HTML]{C7EDCC} 69. Step 5.5.3, iterate v from 0 to 2. \\ 
\rowcolor[HTML]{C7EDCC} 70. Step 5.5.3.a, for v = 0, B[1, 0] = 1, so color[0] = 1, continue to v = 1. \\ 
\rowcolor[HTML]{C7EDCC} 71. Step 5.5.3.a, for v = 1, B[1, 1] = 0, so continue to v = 2. \\ 
\rowcolor[HTML]{C7EDCC} 72. Step 5.5.3.a, for v = 2, B[1, 2] = 0, so continue. \\ 
\rowcolor[HTML]{C7EDCC} 73. Step 5.5.4, beq\_s\_last\_u = 1 - 1 = 0, so set color[1] = 2, increment time to 9, set f[1] = 9, and go to Step 5.5.5. \\ 
\rowcolor[HTML]{C7EDCC} 74. Step 5.5.5, beq\_s\_prev\_u\_u = 0 - 1 = -1, so go to Step 5.5.5.1. \\ 
\rowcolor[HTML]{C7EDCC} 75. Step 5.5.5.1, swap s\_prev[s\_last] and s\_last, set temp = s\_prev[1] = 0, s\_prev[1] = 1, s\_last = 0. \\ 
\rowcolor[HTML]{C7EDCC} 76. Step 5.5.6, set u = 0. \\ 
\rowcolor[HTML]{C7EDCC} 77. Step 5.5.1, set scc\_id[0] = 0. \\ 
\rowcolor[HTML]{C7EDCC} 78. Step 5.5.2, color[0] = 1 and d[0] = 7, so continue to Step 5.5.3. \\ 
\rowcolor[HTML]{C7EDCC} 79. Step 5.5.3, iterate v from 0 to 2. \\ 
\rowcolor[HTML]{C7EDCC} 80. Step 5.5.3.a, for v = 0, B[0, 0] = 0, so continue to v = 1. \\ 
\bottomrule
\end{tabular}
\end{table*}
\begin{table*}[!t]
\small
\begin{tabular}{p{15.5cm}}
\toprule
\rowcolor[HTML]{C7EDCC} 81. Step 5.5.3.a, for v = 1, B[0, 1] = 1, so color[1] = 2, continue to v = 2. \\ 
\rowcolor[HTML]{C7EDCC} 82. Step 5.5.3.a, for v = 2, B[0, 2] = 0, so continue. \\ 
\rowcolor[HTML]{C7EDCC} 83. Step 5.5.4, beq\_s\_last\_u = 0 - 0 = 0, so set color[0] = 2, increment time to 10, set f[0] = 10, and go to Step 5.5.5. \\ 
\rowcolor[HTML]{C7EDCC} 84. Step 5.5.5, beq\_s\_prev\_u\_u = 0 - 0 = 0, so break the while loop and go to the next iteration of s. \\ 
\rowcolor[HTML]{C7EDCC} 85. Step 5.1, for s = 2, color[2] = 0, so proceed to Step 5.2. \\ 
\rowcolor[HTML]{C7EDCC} 86. Step 5.2, set s\_last = 2. \\ 
\rowcolor[HTML]{C7EDCC} 87. Step 5.3, set u = 2. \\ 
\rowcolor[HTML]{C7EDCC} 88. Step 5.4, set v = 2. \\ 
\rowcolor[HTML]{C7EDCC} 89. Step 5.5, start the while loop. \\ 
\rowcolor[HTML]{C7EDCC} 90. Step 5.5.1, set scc\_id[2] = 2. \\ 
\rowcolor[HTML]{C7EDCC} 91. Step 5.5.2, color[2] = 0 and d[2] = 0, so increment time to 11, set d[2] = 11, and set color[2] = 1. \\ 
\rowcolor[HTML]{C7EDCC} 92. Step 5.5.3, iterate v from 0 to 2. \\ 
\rowcolor[HTML]{C7EDCC} 93. Step 5.5.3.a, for v = 0, B[2, 0] = 1, so color[0] = 2, continue to v = 1. \\ 
\rowcolor[HTML]{C7EDCC} 94. Step 5.5.3.a, for v = 1, B[2, 1] = 0, so continue to v = 2. \\ 
\rowcolor[HTML]{C7EDCC} 95. Step 5.5.3.a, for v = 2, B[2, 2] = 0, so continue. \\ 
\rowcolor[HTML]{C7EDCC} 96. Step 5.5.4, beq\_s\_last\_u = 2 - 2 = 0, so set color[2] = 2, increment time to 12, set f[2] = 12, and go to Step 5.5.5. \\ 
\rowcolor[HTML]{C7EDCC} 97. Step 5.5.5, beq\_s\_prev\_u\_u = 2 - 2 = 0, so break the while loop and go to the next iteration of s. \\ 
\rowcolor[HTML]{C7EDCC} 98. Step 5.1, for s = 1, color[1] = 2, so continue. \\ 
\rowcolor[HTML]{C7EDCC} 99. Step 6, print scc\_id array: [0, 0, 2]. \\ 
\rowcolor[HTML]{C7EDCC}  \\ 
\rowcolor[HTML]{C7EDCC} The final scc\_id array is [0, 0, 2].
\\ \bottomrule
\end{tabular}
\caption{The prompt of strongly connected components and the response of \texttt{gpt-4}.}
\label{table:clrs_mini/strongly_connected_components}

\end{table*}
\begin{table*}[h]
\small
\begin{tabular}{p{15cm}}
\toprule
\textbf{Prompt} \\ \midrule
Execute the instructions step by step. Do not jump steps. Do not stop before completion. \\ 
Initial: Set the list of distances between vertices, lx, ly, w: lx[0] = 0, lx[1] = 2, lx[2] = 0, lx[3] = 1, lx[4] = 1, lx[5] = 0, ly[0] = 3, ly[1] = 3, ly[2] = 2, ly[3] = 2, ly[4] = 3, ly[5] = 1, w[0] = 1, w[1] = 1, w[2] = 2, w[3] = 3, w[4] = 4, w[5] = 5, with the length of M = 6. Set N = 4 to represent the number of vertices. Set pi as an array of indices from 0 to N-1: pi[0] = 0, pi[1] = 1, pi[2] = 2, pi[3] = 3. Set variables weight\_sum to 0 and cnt to 0. \\ 
 \\ 
Step 1.1: Assign k to the value of x. \\ 
Step 1.2: While pi[x] is not equal to x, set x = pi[x], then goto Step 1.2. \\ 
Step 1.3: While pi[k] is not equal to k, set w = pi[k], set pi[k] = x, and set k = w, then goto Step 1.3. \\ 
Step 1.4: Return the final value of x. \\ 
 \\ 
Step 2.1: Set x to u and goto Step 1.1.  \\ 
Step 2.2: Assign the returned value to root\_u. \\ 
Step 2.3: Set x to v and goto Step 1.1.  \\ 
Step 2.4: Assign the returned value to root\_v. \\ 
Step 2.5: Check if root\_u is not equal to root\_v. \\ 
  a. If True, set pi[root\_u] = root\_v and return True. \\ 
  b. If False, return False. \\ 
 \\ 
Step 3: Set i to 0. \\ 
Step 3.1: If i is less than 6, goto Step 3.2. Otherwise, goto Step 4. \\ 
Step 3.2: Assign u to lx[i], and v to the value at index i in the ly list. \\ 
Step 3.3: Goto Step 2.1. If the function returns True, perform these sub-steps: \\ 
  i. Add w[i] to weight\_sum. \\ 
  ii. Increment cnt by 1. \\ 
  iii. Check if cnt equals 3. If True, go to Step 4. \\ 
Step 3.4: Increment i by 1. \\ 
Step 3.5: Goto Step 3.1. \\ 
 \\ 
Step 4: Print the value of weight\_sum. \\ 
----- \\ 
The computation procedure is as follows, with each line numbered sequentially (e.g., 1. 2. 3. 4. 5. 6. etc., and remember, the next number of 59 is 60): \\ 
1. In Step 3, i = 0. \\ 
\bottomrule
\end{tabular}
\end{table*}
\begin{table*}[!t]
\small
\begin{tabular}{p{15.5cm}}
\toprule
\rowcolor[HTML]{C7EDCC} 2. In Step 3.2, u = lx[0] = 0, v = ly[0] = 3. \\ 
\rowcolor[HTML]{C7EDCC} 3. In Step 2.1, x = u = 0, and we go to Step 1.1. \\ 
\rowcolor[HTML]{C7EDCC} 4. In Step 1.1, k = x = 0. \\ 
\rowcolor[HTML]{C7EDCC} 5. In Step 1.2, pi[x] = pi[0] = 0, so x = 0. \\ 
\rowcolor[HTML]{C7EDCC} 6. In Step 1.3, pi[k] = pi[0] = 0, so k = 0. \\ 
\rowcolor[HTML]{C7EDCC} 7. In Step 1.4, we return x = 0. \\ 
\rowcolor[HTML]{C7EDCC} 8. In Step 2.2, root\_u = 0. \\ 
\rowcolor[HTML]{C7EDCC} 9. In Step 2.3, x = v = 3, and we go to Step 1.1. \\ 
\rowcolor[HTML]{C7EDCC} 10. In Step 1.1, k = x = 3. \\ 
\rowcolor[HTML]{C7EDCC} 11. In Step 1.2, pi[x] = pi[3] = 3, so x = 3. \\ 
\rowcolor[HTML]{C7EDCC} 12. In Step 1.3, pi[k] = pi[3] = 3, so k = 3. \\ 
\rowcolor[HTML]{C7EDCC} 13. In Step 1.4, we return x = 3. \\ 
\rowcolor[HTML]{C7EDCC} 14. In Step 2.4, root\_v = 3. \\ 
\rowcolor[HTML]{C7EDCC} 15. In Step 2.5, root\_u is not equal to root\_v, so we set pi[root\_u] = pi[0] = 3 and return True. \\ 
\rowcolor[HTML]{C7EDCC} 16. In Step 3.3.i, we add w[0] = 1 to weight\_sum, making weight\_sum = 1. \\ 
\rowcolor[HTML]{C7EDCC} 17. In Step 3.3.ii, we increment cnt by 1, making cnt = 1. \\ 
\rowcolor[HTML]{C7EDCC} 18. In Step 3.4, we increment i by 1, making i = 1. \\ 
\rowcolor[HTML]{C7EDCC} 19. In Step 3.2, u = lx[1] = 2, v = ly[1] = 3. \\ 
\rowcolor[HTML]{C7EDCC} 20. In Step 2.1, x = u = 2, and we go to Step 1.1. \\ 
\rowcolor[HTML]{C7EDCC} 21. In Step 1.1, k = x = 2. \\ 
\rowcolor[HTML]{C7EDCC} 22. In Step 1.2, pi[x] = pi[2] = 2, so x = 2. \\ 
\rowcolor[HTML]{C7EDCC} 23. In Step 1.3, pi[k] = pi[2] = 2, so k = 2. \\ 
\rowcolor[HTML]{C7EDCC} 24. In Step 1.4, we return x = 2. \\ 
\rowcolor[HTML]{C7EDCC} 25. In Step 2.2, root\_u = 2. \\ 
\rowcolor[HTML]{C7EDCC} 26. In Step 2.3, x = v = 3, and we go to Step 1.1. \\ 
\rowcolor[HTML]{C7EDCC} 27. In Step 1.1, k = x = 3. \\ 
\rowcolor[HTML]{C7EDCC} 28. In Step 1.2, pi[x] = pi[3] = 3, so x = 3. \\ 
\rowcolor[HTML]{C7EDCC} 29. In Step 1.3, pi[k] = pi[3] = 3, so k = 3. \\ 
\rowcolor[HTML]{C7EDCC} 30. In Step 1.4, we return x = 3. \\ 
\rowcolor[HTML]{C7EDCC} 31. In Step 2.4, root\_v = 3. \\ 
\rowcolor[HTML]{C7EDCC} 32. In Step 2.5, root\_u is not equal to root\_v, so we set pi[root\_u] = pi[2] = 3 and return True. \\ 
\rowcolor[HTML]{C7EDCC} 33. In Step 3.3.i, we add w[1] = 1 to weight\_sum, making weight\_sum = 2. \\ 
\rowcolor[HTML]{C7EDCC} 34. In Step 3.3.ii, we increment cnt by 1, making cnt = 2. \\ 
\rowcolor[HTML]{C7EDCC} 35. In Step 3.4, we increment i by 1, making i = 2. \\ 
\rowcolor[HTML]{C7EDCC} 36. In Step 3.2, u = lx[2] = 0, v = ly[2] = 2. \\ 
\rowcolor[HTML]{C7EDCC} 37. In Step 2.1, x = u = 0, and we go to Step 1.1. \\ 
\rowcolor[HTML]{C7EDCC} 38. In Step 1.1, k = x = 0. \\ 
\rowcolor[HTML]{C7EDCC} 39. In Step 1.2, pi[x] = pi[0] = 3, so x = 3. \\ 
\rowcolor[HTML]{C7EDCC} 40. In Step 1.3, pi[k] = pi[0] = 3, so k = 0. \\ 
\rowcolor[HTML]{C7EDCC} 41. In Step 1.4, we return x = 3. \\ 
\rowcolor[HTML]{C7EDCC} 42. In Step 2.2, root\_u = 3. \\ 
\rowcolor[HTML]{C7EDCC} 43. In Step 2.3, x = v = 2, and we go to Step 1.1. \\ 
\rowcolor[HTML]{C7EDCC} 44. In Step 1.1, k = x = 2. \\ 
\rowcolor[HTML]{C7EDCC} 45. In Step 1.2, pi[x] = pi[2] = 3, so x = 3. \\ 
\rowcolor[HTML]{C7EDCC} 46. In Step 1.3, pi[k] = pi[2] = 3, so k = 2. \\ 
\rowcolor[HTML]{C7EDCC} 47. In Step 1.4, we return x = 3. \\ 
\rowcolor[HTML]{C7EDCC} 48. In Step 2.4, root\_v = 3. \\ 
\rowcolor[HTML]{C7EDCC} 49. In Step 2.5, root\_u is equal to root\_v, so we return False. \\ 
\rowcolor[HTML]{C7EDCC} 50. In Step 3.4, we increment i by 1, making i = 3. \\ 
\rowcolor[HTML]{C7EDCC} 51. In Step 3.2, u = lx[3] = 1, v = ly[3] = 2. \\ 
\rowcolor[HTML]{C7EDCC} 52. In Step 2.1, x = u = 1, and we go to Step 1.1. \\ 
\rowcolor[HTML]{C7EDCC} 53. In Step 1.1, k = x = 1. \\ 
\rowcolor[HTML]{C7EDCC} 54. In Step 1.2, pi[x] = pi[1] = 1, so x = 1. \\ 
\rowcolor[HTML]{C7EDCC} 55. In Step 1.3, pi[k] = pi[1] = 1, so k = 1. \\ 
\rowcolor[HTML]{C7EDCC} 56. In Step 1.4, we return x = 1. \\ 
\rowcolor[HTML]{C7EDCC} 57. In Step 2.2, root\_u = 1. \\ 
\rowcolor[HTML]{C7EDCC} 58. In Step 2.3, x = v = 2, and we go to Step 1.1. \\ 
\rowcolor[HTML]{C7EDCC} 59. In Step 1.1, k = x = 2. \\ 
\rowcolor[HTML]{C7EDCC} 60. In Step 1.2, pi[x] = pi[2] = 3, so x = 3. \\ 
\rowcolor[HTML]{C7EDCC} 61. In Step 1.3, pi[k] = pi[2] = 3, so k = 2. \\ 
\rowcolor[HTML]{C7EDCC} 62. In Step 1.4, we return x = 3. \\ 
\rowcolor[HTML]{C7EDCC} 63. In Step 2.4, root\_v = 3. \\ 
\bottomrule
\end{tabular}
\end{table*}
\begin{table*}[!t]
\small
\begin{tabular}{p{15.5cm}}
\toprule
\rowcolor[HTML]{C7EDCC} 64. In Step 2.5, root\_u is not equal to root\_v, so we set pi[root\_u] = pi[1] = 3 and return True. \\ 
\rowcolor[HTML]{C7EDCC} 65. In Step 3.3.i, we add w[3] = 3 to weight\_sum, making weight\_sum = 5. \\ 
\rowcolor[HTML]{C7EDCC} 66. In Step 3.3.ii, we increment cnt by 1, making cnt = 3. \\ 
\rowcolor[HTML]{C7EDCC} 67. In Step 4, we print the value of weight\_sum, which is 5. \\ 
\rowcolor[HTML]{C7EDCC}  \\ 
\rowcolor[HTML]{C7EDCC} The value of weight\_sum is 5.
\\ \bottomrule
\end{tabular}
\caption{The prompt of mst kruskal and the response of \texttt{gpt-4}.}
\label{table:clrs_mini/mst_kruskal}

\end{table*}
\begin{table*}[h]
\small
\begin{tabular}{p{15cm}}
\toprule
\textbf{Prompt} \\ \midrule
Execute the instructions step by step. Do not jump steps. Do not stop before completion. \\ 
Initial: Set the adjacency matrix A with the size 4x4: \\ 
A[0, 0] = 0, A[0, 1] = 3, A[0, 2] = 5, A[0, 3] = 4 \\ 
A[1, 0] = 3, A[1, 1] = 0, A[1, 2] = 6, A[1, 3] = 0 \\ 
A[2, 0] = 5, A[2, 1] = 6, A[2, 2] = 0, A[2, 3] = 3 \\ 
A[3, 0] = 4, A[3, 1] = 0, A[3, 2] = 3, A[3, 3] = 0, \\ 
where N = 4 is the number of vertices. Set the start vertex s = 0. \\ 
Set variables key, mark, and pi as arrays of zeros with 4 elements: \\ 
key: key[0] = 0, key[1] = 0, key[2] = 0, key[3] = 0, \\ 
mark: mark[0] = 0, mark[1] = 0, mark[2] = 0, mark[3] = 0, \\ 
pi: pi[0] = 0, pi[1] = 0, pi[2] = 0, pi[3] = 0, \\ 
Set the array of in\_queue: in\_queue[0] = 1, in\_queue[1] = 0, in\_queue[2] = 0, in\_queue[3] = 0. \\ 
 \\ 
Step 1: Assign k to 0. \\ 
Step 2.1: Assign u to -1. \\ 
Step 2.2: Assign i to 0. \\ 
Step 2.2.1: If i is less than 4, go to Step 2.2.2. Otherwise, go to Step 2.3. \\ 
Step 2.2.2: Check if in\_queue[i] equals 1. If True, go to Step 2.2.3. \\ 
Step 2.2.3: Check if u equals -1. If True, assign i to u. Then, go to Step 2.2.5. \\ 
Step 2.2.4: What are the value of key[i] and key[u]? Then, what is the result of key[i] - key[u]? Is it negative? If True, assign i to u. \\ 
Step 2.2.5: Increment i by 1, and go to Step 2.2.1. \\ 
Step 2.3: Check if u equals -1. \\ 
\quad a. If True, go to Step 3. \\ 
Step 2.4: Set Mark[u] to 1. \\ 
Step 2.5: Set in\_queue[u] to 0. \\ 
Step 2.6: Set v to 0. \\ 
Step 2.6.1: If v is less than 4, go to Step 2.6.2. Otherwise, go to Step 2.6. \\ 
Step 2.6.2: Check if A[u, v] does not equal to 0. If True, perform these sub-steps: \\ 
\quad a. Check if (mark[v] equals 0.  \\ 
\quad \quad a.1. If True, then what are the values of A[u, v] and key[v]? Check either (in\_queue[v] equals 0) OR (A[u, v] - key[v] is less than 0) is True. \\ 
\quad \quad \quad i. If either is True, set pi[v] to u, and \\ 
\quad \quad \quad ii. Set key[v] to A[u, v], and \\ 
\quad \quad \quad iii. Set in\_queue[v] to 1. \\ 
Step 2.6.3: Increment v by 1, and go to Step 2.6.1. \\ 
Step 2.7: Increment k by 1. If k < 4, go to Step 2.1. Otherwise, go to Step 3. \\ 
 \\ 
Step 3: Print the array of pi. \\ 
----- \\ 
The computation procedure is as follows, with each line numbered sequentially (e.g., 1. 2. 3. 4. 5. 6. etc., and remember, the next number of 59 is 60): \\ 
1. In Step 1, k = 0. \\ 
\bottomrule
\end{tabular}
\end{table*}
\begin{table*}[!t]
\small
\begin{tabular}{p{15.5cm}}
\toprule
\rowcolor[HTML]{C7EDCC} 2. In Step 2.1, u = -1. \\ 
\rowcolor[HTML]{C7EDCC} 3. In Step 2.2, i = 0. \\ 
\rowcolor[HTML]{C7EDCC} 4. In Step 2.2.1, i < 4, so go to Step 2.2.2. \\ 
\rowcolor[HTML]{C7EDCC} 5. In Step 2.2.2, in\_queue[0] = 1, so go to Step 2.2.3. \\ 
\rowcolor[HTML]{C7EDCC} 6. In Step 2.2.3, u = -1, so assign i to u, u = 0. \\ 
\rowcolor[HTML]{C7EDCC} 7. In Step 2.2.5, increment i by 1, i = 1. \\ 
\rowcolor[HTML]{C7EDCC} 8. In Step 2.2.1, i < 4, so go to Step 2.2.2. \\ 
\rowcolor[HTML]{C7EDCC} 9. In Step 2.2.2, in\_queue[1] = 0, so go to Step 2.2.5. \\ 
\rowcolor[HTML]{C7EDCC} 10. In Step 2.2.5, increment i by 1, i = 2. \\ 
\rowcolor[HTML]{C7EDCC} 11. In Step 2.2.1, i < 4, so go to Step 2.2.2. \\ 
\rowcolor[HTML]{C7EDCC} 12. In Step 2.2.2, in\_queue[2] = 0, so go to Step 2.2.5. \\ 
\rowcolor[HTML]{C7EDCC} 13. In Step 2.2.5, increment i by 1, i = 3. \\ 
\rowcolor[HTML]{C7EDCC} 14. In Step 2.2.1, i < 4, so go to Step 2.2.2. \\ 
\rowcolor[HTML]{C7EDCC} 15. In Step 2.2.2, in\_queue[3] = 0, so go to Step 2.2.5. \\ 
\rowcolor[HTML]{C7EDCC} 16. In Step 2.2.5, increment i by 1, i = 4. \\ 
\rowcolor[HTML]{C7EDCC} 17. In Step 2.2.1, i >= 4, so go to Step 2.3. \\ 
\rowcolor[HTML]{C7EDCC} 18. In Step 2.3, u != -1, so go to Step 2.4. \\ 
\rowcolor[HTML]{C7EDCC} 19. In Step 2.4, Mark[0] = 1. \\ 
\rowcolor[HTML]{C7EDCC} 20. In Step 2.5, in\_queue[0] = 0. \\ 
\rowcolor[HTML]{C7EDCC} 21. In Step 2.6, v = 0. \\ 
\rowcolor[HTML]{C7EDCC} 22. In Step 2.6.1, v < 4, so go to Step 2.6.2. \\ 
\rowcolor[HTML]{C7EDCC} 23. In Step 2.6.2, A[0, 0] = 0, so go to Step 2.6.3. \\ 
\rowcolor[HTML]{C7EDCC} 24. In Step 2.6.3, increment v by 1, v = 1. \\ 
\rowcolor[HTML]{C7EDCC} 25. In Step 2.6.1, v < 4, so go to Step 2.6.2. \\ 
\rowcolor[HTML]{C7EDCC} 26. In Step 2.6.2, A[0, 1] != 0, so perform sub-steps: \\ 
\rowcolor[HTML]{C7EDCC} \quad a. mark[1] = 0, so perform sub-steps: \\ 
\rowcolor[HTML]{C7EDCC} \quad \quad a.1. A[0, 1] = 3, key[1] = 0, in\_queue[1] = 0, so perform sub-steps: \\ 
\rowcolor[HTML]{C7EDCC} \quad \quad \quad i. pi[1] = 0, \\ 
\rowcolor[HTML]{C7EDCC} \quad \quad \quad ii. key[1] = 3, \\ 
\rowcolor[HTML]{C7EDCC} \quad \quad \quad iii. in\_queue[1] = 1. \\ 
\rowcolor[HTML]{C7EDCC} 27. In Step 2.6.3, increment v by 1, v = 2. \\ 
\rowcolor[HTML]{C7EDCC} 28. In Step 2.6.1, v < 4, so go to Step 2.6.2. \\ 
\rowcolor[HTML]{C7EDCC} 29. In Step 2.6.2, A[0, 2] != 0, so perform sub-steps: \\ 
\rowcolor[HTML]{C7EDCC} \quad a. mark[2] = 0, so perform sub-steps: \\ 
\rowcolor[HTML]{C7EDCC} \quad \quad a.1. A[0, 2] = 5, key[2] = 0, in\_queue[2] = 0, so perform sub-steps: \\ 
\rowcolor[HTML]{C7EDCC} \quad \quad \quad i. pi[2] = 0, \\ 
\rowcolor[HTML]{C7EDCC} \quad \quad \quad ii. key[2] = 5, \\ 
\rowcolor[HTML]{C7EDCC} \quad \quad \quad iii. in\_queue[2] = 1. \\ 
\rowcolor[HTML]{C7EDCC} 30. In Step 2.6.3, increment v by 1, v = 3. \\ 
\rowcolor[HTML]{C7EDCC} 31. In Step 2.6.1, v < 4, so go to Step 2.6.2. \\ 
\rowcolor[HTML]{C7EDCC} 32. In Step 2.6.2, A[0, 3] != 0, so perform sub-steps: \\ 
\rowcolor[HTML]{C7EDCC} \quad a. mark[3] = 0, so perform sub-steps: \\ 
\rowcolor[HTML]{C7EDCC} \quad \quad a.1. A[0, 3] = 4, key[3] = 0, in\_queue[3] = 0, so perform sub-steps: \\ 
\rowcolor[HTML]{C7EDCC} \quad \quad \quad i. pi[3] = 0, \\ 
\rowcolor[HTML]{C7EDCC} \quad \quad \quad ii. key[3] = 4, \\ 
\rowcolor[HTML]{C7EDCC} \quad \quad \quad iii. in\_queue[3] = 1. \\ 
\rowcolor[HTML]{C7EDCC} 33. In Step 2.6.3, increment v by 1, v = 4. \\ 
\rowcolor[HTML]{C7EDCC} 34. In Step 2.6.1, v >= 4, so go to Step 2.7. \\ 
\rowcolor[HTML]{C7EDCC} 35. In Step 2.7, increment k by 1, k = 1, k < 4, so go to Step 2.1. \\ 
\rowcolor[HTML]{C7EDCC} 36. In Step 2.1, u = -1. \\ 
\rowcolor[HTML]{C7EDCC} 37. In Step 2.2, i = 0. \\ 
\rowcolor[HTML]{C7EDCC} 38. In Step 2.2.1, i < 4, so go to Step 2.2.2. \\ 
\rowcolor[HTML]{C7EDCC} 39. In Step 2.2.2, in\_queue[0] = 0, so go to Step 2.2.5. \\
\rowcolor[HTML]{C7EDCC} 40. In Step 2.2.5, increment i by 1, i = 1. \\ 
\rowcolor[HTML]{C7EDCC} 41. In Step 2.2.1, i < 4, so go to Step 2.2.2. \\ 
\rowcolor[HTML]{C7EDCC} 42. In Step 2.2.2, in\_queue[1] = 1, so go to Step 2.2.3. \\ 
\rowcolor[HTML]{C7EDCC} 43. In Step 2.2.3, u = -1, so assign i to u, u = 1. \\ 
\rowcolor[HTML]{C7EDCC} 44. In Step 2.2.5, increment i by 1, i = 2. \\ 
\rowcolor[HTML]{C7EDCC} 45. In Step 2.2.1, i < 4, so go to Step 2.2.2. \\ 
\rowcolor[HTML]{C7EDCC} 46. In Step 2.2.2, in\_queue[2] = 1, so go to Step 2.2.4. \\ 
\rowcolor[HTML]{C7EDCC} 47. In Step 2.2.4, key[i] = 5, key[u] = 3, key[i] - key[u] = 2, not negative, so go to Step 2.2.5. \\ 
\rowcolor[HTML]{C7EDCC} 48. In Step 2.2.5, increment i by 1, i = 3. \\ 
\rowcolor[HTML]{C7EDCC} 49. In Step 2.2.1, i < 4, so go to Step 2.2.2. \\ 
\bottomrule
\end{tabular}
\end{table*}
\begin{table*}[!t]
\small
\begin{tabular}{p{15.5cm}}
\toprule
\rowcolor[HTML]{C7EDCC} 50. In Step 2.2.2, in\_queue[3] = 1, so go to Step 2.2.4. \\ 
\rowcolor[HTML]{C7EDCC} 51. In Step 2.2.4, key[i] = 4, key[u] = 3, key[i] - key[u] = 1, not negative, so go to Step 2.2.5. \\ 
\rowcolor[HTML]{C7EDCC} 52. In Step 2.2.5, increment i by 1, i = 4. \\ 
\rowcolor[HTML]{C7EDCC} 53. In Step 2.2.1, i >= 4, so go to Step 2.3. \\ 
\rowcolor[HTML]{C7EDCC} 54. In Step 2.3, u != -1, so go to Step 2.4. \\ 
\rowcolor[HTML]{C7EDCC} 55. In Step 2.4, Mark[1] = 1. \\ 
\rowcolor[HTML]{C7EDCC} 56. In Step 2.5, in\_queue[1] = 0. \\ 
\rowcolor[HTML]{C7EDCC} 57. In Step 2.6, v = 0. \\ 
\rowcolor[HTML]{C7EDCC} 58. In Step 2.6.1, v < 4, so go to Step 2.6.2. \\ 
\rowcolor[HTML]{C7EDCC} 59. In Step 2.6.2, A[1, 0] != 0, so perform sub-steps: \\ 
\rowcolor[HTML]{C7EDCC} \quad a. mark[0] = 1, so go to Step 2.6.3. \\ 
\rowcolor[HTML]{C7EDCC} 60. In Step 2.6.3, increment v by 1, v = 1. \\ 
\rowcolor[HTML]{C7EDCC} 61. In Step 2.6.1, v < 4, so go to Step 2.6.2. \\ 
\rowcolor[HTML]{C7EDCC} 62. In Step 2.6.2, A[1, 1] = 0, so go to Step 2.6.3. \\ 
\rowcolor[HTML]{C7EDCC} 63. In Step 2.6.3, increment v by 1, v = 2. \\ 
\rowcolor[HTML]{C7EDCC} 64. In Step 2.6.1, v < 4, so go to Step 2.6.2. \\ 
\rowcolor[HTML]{C7EDCC} 65. In Step 2.6.2, A[1, 2] != 0, so perform sub-steps: \\ 
\rowcolor[HTML]{C7EDCC} \quad a. mark[2] = 0, so perform sub-steps: \\ 
\rowcolor[HTML]{C7EDCC} \quad \quad a.1. A[1, 2] = 6, key[2] = 5, in\_queue[2] = 1, so go to Step 2.6.3. \\ 
\rowcolor[HTML]{C7EDCC} 66. In Step 2.6.3, increment v by 1, v = 3. \\ 
\rowcolor[HTML]{C7EDCC} 67. In Step 2.6.1, v < 4, so go to Step 2.6.2. \\ 
\rowcolor[HTML]{C7EDCC} 68. In Step 2.6.2, A[1, 3] = 0, so go to Step 2.6.3. \\ 
\rowcolor[HTML]{C7EDCC} 69. In Step 2.6.3, increment v by 1, v = 4. \\ 
\rowcolor[HTML]{C7EDCC} 70. In Step 2.6.1, v >= 4, so go to Step 2.7. \\ 
\rowcolor[HTML]{C7EDCC} 71. In Step 2.7, increment k by 1, k = 2, k < 4, so go to Step 2.1. \\ 
\rowcolor[HTML]{C7EDCC} 72. In Step 2.1, u = -1. \\ 
\rowcolor[HTML]{C7EDCC} 73. In Step 2.2, i = 0. \\ 
\rowcolor[HTML]{C7EDCC} 74. In Step 2.2.1, i < 4, so go to Step 2.2.2. \\ 
\rowcolor[HTML]{C7EDCC} 75. In Step 2.2.2, in\_queue[0] = 0, so go to Step 2.2.5. \\ 
\rowcolor[HTML]{C7EDCC} 76. In Step 2.2.5, increment i by 1, i = 1. \\ 
\rowcolor[HTML]{C7EDCC} 77. In Step 2.2.1, i < 4, so go to Step 2.2.2. \\ 
\rowcolor[HTML]{C7EDCC} 78. In Step 2.2.2, in\_queue[1] = 0, so go to Step 2.2.5. \\ 
\rowcolor[HTML]{C7EDCC} 79. In Step 2.2.5, increment i by 1, i = 2. \\ 
\rowcolor[HTML]{C7EDCC} 80. In Step 2.2.1, i < 4, so go to Step 2.2.2. \\ 
\rowcolor[HTML]{C7EDCC} 81. In Step 2.2.2, in\_queue[2] = 1, so go to Step 2.2.3. \\ 
\rowcolor[HTML]{C7EDCC} 82. In Step 2.2.3, u = -1, so assign i to u, u = 2. \\ 
\rowcolor[HTML]{C7EDCC} 83. In Step 2.2.5, increment i by 1, i = 3. \\ 
\rowcolor[HTML]{C7EDCC} 84. In Step 2.2.1, i < 4, so go to Step 2.2.2. \\ 
\rowcolor[HTML]{C7EDCC} 85. In Step 2.2.2, in\_queue[3] = 1, so go to Step 2.2.4. \\ 
\rowcolor[HTML]{C7EDCC} 86. In Step 2.2.4, key[i] = 4, key[u] = 5, key[i] - key[u] = -1, negative, so assign i to u, u = 3. \\ 
\rowcolor[HTML]{C7EDCC} 87. In Step 2.2.5, increment i by 1, i = 4. \\ 
\rowcolor[HTML]{C7EDCC} 88. In Step 2.2.1, i >= 4, so go to Step 2.3. \\ 
\rowcolor[HTML]{C7EDCC} 89. In Step 2.3, u != -1, so go to Step 2.4. \\ 
\rowcolor[HTML]{C7EDCC} 90. In Step 2.4, Mark[3] = 1. \\ 
\rowcolor[HTML]{C7EDCC} 91. In Step 2.5, in\_queue[3] = 0. \\ 
\rowcolor[HTML]{C7EDCC} 92. In Step 2.6, v = 0. \\ 
\rowcolor[HTML]{C7EDCC} 93. In Step 2.6.1, v < 4, so go to Step 2.6.2. \\ 
\rowcolor[HTML]{C7EDCC} 94. In Step 2.6.2, A[3, 0] != 0, so perform sub-steps: \\ 
\rowcolor[HTML]{C7EDCC} \quad a. mark[0] = 1, so go to Step 2.6.3. \\ 
\rowcolor[HTML]{C7EDCC} 95. In Step 2.6.3, increment v by 1, v = 1. \\ 
\rowcolor[HTML]{C7EDCC} 96. In Step 2.6.1, v < 4, so go to Step 2.6.2. \\ 
\rowcolor[HTML]{C7EDCC} 97. In Step 2.6.2, A[3, 1] = 0, so go to Step 2.6.3. \\ 
\rowcolor[HTML]{C7EDCC} 98. In Step 2.6.3, increment v by 1, v = 2. \\ 
\rowcolor[HTML]{C7EDCC} 99. In Step 2.6.1, v < 4, so go to Step 2.6.2. \\ 
\rowcolor[HTML]{C7EDCC} 100. In Step 2.6.2, A[3, 2] != 0, so perform sub-steps: \\ 
\rowcolor[HTML]{C7EDCC} \quad a. mark[2] = 0, so perform sub-steps: \\ 
\rowcolor[HTML]{C7EDCC} \quad \quad a.1. A[3, 2] = 3, key[2] = 5, in\_queue[2] = 1, so perform sub-steps: \\ 
\rowcolor[HTML]{C7EDCC} \quad \quad \quad i. pi[2] = 3, \\ 
\rowcolor[HTML]{C7EDCC} \quad \quad \quad ii. key[2] = 3, \\ 
\bottomrule
\end{tabular}
\end{table*}
\begin{table*}[!t]
\small
\begin{tabular}{p{15.5cm}}
\toprule
\rowcolor[HTML]{C7EDCC} \quad \quad \quad iii. in\_queue[2] = 1. \\ 
\rowcolor[HTML]{C7EDCC} 101. In Step 2.6.3, increment v by 1, v = 3. \\ 
\rowcolor[HTML]{C7EDCC} 102. In Step 2.6.1, v < 4, so go to Step 2.6.2. \\ 
\rowcolor[HTML]{C7EDCC} 103. In Step 2.6.2, A[3, 3] = 0, so go to Step 2.6.3. \\ 
\rowcolor[HTML]{C7EDCC} 104. In Step 2.6.3, increment v by 1, v = 4. \\ 
\rowcolor[HTML]{C7EDCC} 105. In Step 2.6.1, v >= 4, so go to Step 2.7. \\ 
\rowcolor[HTML]{C7EDCC} 106. In Step 2.7, increment k by 1, k = 3, k < 4, so go to Step 2.1. \\ 
\rowcolor[HTML]{C7EDCC} 107. In Step 2.1, u = -1. \\ 
\rowcolor[HTML]{C7EDCC} 108. In Step 2.2, i = 0. \\ 
\rowcolor[HTML]{C7EDCC} 109. In Step 2.2.1, i < 4, so go to Step 2.2.2. \\ 
\rowcolor[HTML]{C7EDCC} 110. In Step 2.2.2, in\_queue[0] = 0, so go to Step 2.2.5. \\ 
\rowcolor[HTML]{C7EDCC} 111. In Step 2.2.5, increment i by 1, i = 1. \\ 
\rowcolor[HTML]{C7EDCC} 112. In Step 2.2.1, i < 4, so go to Step 2.2.2. \\ 
\rowcolor[HTML]{C7EDCC} 113. In Step 2.2.2, in\_queue[1] = 0, so go to Step 2.2.5. \\ 
\rowcolor[HTML]{C7EDCC} 114. In Step 2.2.5, increment i by 1, i = 2. \\ 
\rowcolor[HTML]{C7EDCC} 115. In Step 2.2.1, i < 4, so go to Step 2.2.2. \\ 
\rowcolor[HTML]{C7EDCC} 116. In Step 2.2.2, in\_queue[2] = 1, so go to Step 2.2.3. \\ 
\rowcolor[HTML]{C7EDCC} 117. In Step 2.2.3, u = -1, so assign i to u, u = 2. \\ 
\rowcolor[HTML]{C7EDCC} 118. In Step 2.2.5, increment i by 1, i = 3. \\ 
\rowcolor[HTML]{C7EDCC} 119. In Step 2.2.1, i < 4, so go to Step 2.2.2. \\ 
\rowcolor[HTML]{C7EDCC} 120. In Step 2.2.2, in\_queue[3] = 0, so go to Step 2.2.5. \\ 
\rowcolor[HTML]{C7EDCC} 121. In Step 2.2.5, increment i by 1, i = 4. \\ 
\rowcolor[HTML]{C7EDCC} 122. In Step 2.2.1, i >= 4, so go to Step 2.3. \\ 
\rowcolor[HTML]{C7EDCC} 123. In Step 2.3, u != -1, so go to Step 2.4. \\ 
\rowcolor[HTML]{C7EDCC} 124. In Step 2.4, Mark[2] = 1. \\ 
\rowcolor[HTML]{C7EDCC} 125. In Step 2.5, in\_queue[2] = 0. \\ 
\rowcolor[HTML]{C7EDCC} 126. In Step 2.6, v = 0. \\ 
\rowcolor[HTML]{C7EDCC} 127. In Step 2.6.1, v < 4, so go to Step 2.6.2. \\ 
\rowcolor[HTML]{C7EDCC} 128. In Step 2.6.2, A[2, 0] != 0, so perform sub-steps: \\ 
\rowcolor[HTML]{C7EDCC} \quad a. mark[0] = 1, so go to Step 2.6.3. \\ 
\rowcolor[HTML]{C7EDCC} 129. In Step 2.6.3, increment v by 1, v = 1. \\ 
\rowcolor[HTML]{C7EDCC} 130. In Step 2.6.1, v < 4, so go to Step 2.6.2. \\ 
\rowcolor[HTML]{C7EDCC} 131. In Step 2.6.2, A[2, 1] != 0, so perform sub-steps: \\ 
\rowcolor[HTML]{C7EDCC} \quad a. mark[1] = 1, so go to Step 2.6.3. \\ 
\rowcolor[HTML]{C7EDCC} 132. In Step 2.6.3, increment v by 1, v = 2. \\ 
\rowcolor[HTML]{C7EDCC} 133. In Step 2.6.1, v < 4, so go to Step 2.6.2. \\ 
\rowcolor[HTML]{C7EDCC} 134. In Step 2.6.2, A[2, 2] = 0, so go to Step 2.6.3. \\ 
\rowcolor[HTML]{C7EDCC} 135. In Step 2.6.3, increment v by 1, v = 3. \\ 
\rowcolor[HTML]{C7EDCC} 136. In Step 2.6.1, v < 4, so go to Step 2.6.2. \\ 
\rowcolor[HTML]{C7EDCC} 137. In Step 2.6.2, A[2, 3] != 0, so perform sub-steps: \\ 
\rowcolor[HTML]{C7EDCC} \quad a. mark[3] = 1, so go to Step 2.6.3. \\ 
\rowcolor[HTML]{C7EDCC} 138. In Step 2.6.3, increment v by 1, v = 4. \\ 
\rowcolor[HTML]{C7EDCC} 139. In Step 2.6.1, v >= 4, so go to Step 2.7. \\ 
\rowcolor[HTML]{C7EDCC} 140. In Step 2.7, increment k by 1, k = 4, k >= 4, so go to Step 3. \\ 
\rowcolor[HTML]{C7EDCC} 141. In Step 3, print the array of pi: pi[0] = 0, pi[1] = 0, pi[2] = 3, pi[3] = 0.
\\ \bottomrule
\end{tabular}
\caption{The prompt of mst prim and the response of \texttt{gpt-4}.}
\label{table:clrs_mini/mst_prim}

\end{table*}
\begin{table*}[h]
\small
\begin{tabular}{p{15cm}}
\toprule
\textbf{Prompt} \\ \midrule
Execute the instructions step by step. Do not jump steps. Do not stop before completion. When reading or updating an array at some index, use concrete values rather than just variables. \\ 
Initial: Set the adjacency matrix A with the size 4x4: \\ 
A[0, 0] = 0, A[0, 1] = 0, A[0, 2] = 6, A[0, 3] = 6 \\ 
A[1, 0] = 0, A[1, 1] = 0, A[1, 2] = 4, A[1, 3] = 0 \\ 
A[2, 0] = 6, A[2, 1] = 4, A[2, 2] = 0, A[2, 3] = 3 \\ 
A[3, 0] = 6, A[3, 1] = 0, A[3, 2] = 3, A[3, 3] = 0, \\ 
where N = 4 is the number of vertices. Set the start vertex s = 0. \\ 
Set variables d and mask as arrays of zeros with 4 elements: \\ 
d: d[0] = 0, d[1] = 0, d[2] = 0, d[3] = 0, \\ 
mask: mask[0] = 1, mask[1] = 0, mask[2] = 0, mask[3] = 0, \\ 
 \\ 
Step 1: Make a copy of d and mask named prev\_d and prev\_mask. Set the boolean variable all\_equal to True. \\ 
Step 2: Set U equal to 0. \\ 
Step 2.1: If U is less than 4, perform Step 2.2. Otherwise, go to Step 3. \\ 
Step 2.2: If and prev\_mask[U] equals 1, set V equal to 0. Otherwise, increment U and return to Step 2.1. \\ 
Step 2.3: If V is less than 4, perform Step 2.4. Otherwise, return to Step 2 and increment U. \\ 
Step 2.4: Check if and A[U, V] is not 0. \\ 
\quad a. If True: \\ 
\quad \quad i. Calculate new\_d by adding prev\_d[U] and A[U, V]. \\ 
\quad \quad ii. Calculate the difference between new\_d and d[V]. Then, is it negative? Please check. \\ 
\quad \quad iii. If (mask[V] equals 0) or (the difference is negative): \\ 
\quad \quad \quad - Update d[V] with the value of new\_d. \\ 
\quad \quad \quad - Set all\_equal to False. \\ 
\quad \quad iv. Set mask[V] to 1. \\ 
Step 2.5: Increment V and return to Step 2.3. \\ 
Step 3: If all\_equal is False, go back to Step 1; Otherwise, go to Step 4. \\ 
Step 4: Print the array d. \\ 
----- \\ 
The computation procedure is as follows, with each line numbered sequentially (e.g., 1. 2. 3. 4. 5. 6. etc., and remember, the next number of 59 is 60): \\ 
1. In Step 1, copy from d, prev\_d is: prev\_d[0] = 0, prev\_d[1] = 0, prev\_d[2] = 0, prev\_d[3] = 0. Copy from mask, prev\_mask is: prev\_mask[0] = 1, prev\_mask[1] = 0, prev\_mask[2] = 0, prev\_mask[3] = 0. \\ 
\rowcolor[HTML]{C7EDCC} 2. Set all\_equal to True. \\ 
\rowcolor[HTML]{C7EDCC} 3. In Step 2, set U = 0. \\ 
\rowcolor[HTML]{C7EDCC} 4. In Step 2.1, since U (0) is less than 4, proceed to Step 2.2. \\ 
\rowcolor[HTML]{C7EDCC} 5. In Step 2.2, prev\_mask[0] equals 1, so set V = 0. \\ 
\rowcolor[HTML]{C7EDCC} 6. In Step 2.3, since V (0) is less than 4, proceed to Step 2.4. \\ 
\rowcolor[HTML]{C7EDCC} 7. In Step 2.4, A[0, 0] is 0, so skip to Step 2.5. \\ 
\rowcolor[HTML]{C7EDCC} 8. In Step 2.5, increment V to 1 and return to Step 2.3. \\ 
\rowcolor[HTML]{C7EDCC} 9. In Step 2.3, since V (1) is less than 4, proceed to Step 2.4. \\ 
\rowcolor[HTML]{C7EDCC} 10. In Step 2.4, A[0, 1] is 0, so skip to Step 2.5. \\ 
\rowcolor[HTML]{C7EDCC} 11. In Step 2.5, increment V to 2 and return to Step 2.3. \\ 
\rowcolor[HTML]{C7EDCC} 12. In Step 2.3, since V (2) is less than 4, proceed to Step 2.4. \\ 
\rowcolor[HTML]{C7EDCC} 13. In Step 2.4, A[0, 2] is not 0 (it is 6). \\ 
\rowcolor[HTML]{C7EDCC} 14. Calculate new\_d: prev\_d[0] + A[0, 2] = 0 + 6 = 6. \\ 
\rowcolor[HTML]{C7EDCC} 15. Calculate the difference: new\_d - d[2] = 6 - 0 = 6, which is not negative. \\ 
\rowcolor[HTML]{C7EDCC} 16. Since mask[2] equals 0, update d[2] with the value of new\_d (6) and set all\_equal to False. \\ 
\rowcolor[HTML]{C7EDCC} 17. Set mask[2] to 1. \\ 
\rowcolor[HTML]{C7EDCC} 18. In Step 2.5, increment V to 3 and return to Step 2.3. \\ 
\rowcolor[HTML]{C7EDCC} 19. In Step 2.3, since V (3) is less than 4, proceed to Step 2.4. \\ 
\rowcolor[HTML]{C7EDCC} 20. In Step 2.4, A[0, 3] is not 0 (it is 6). \\ 
\rowcolor[HTML]{C7EDCC} 21. Calculate new\_d: prev\_d[0] + A[0, 3] = 0 + 6 = 6. \\ 
\rowcolor[HTML]{C7EDCC} 22. Calculate the difference: new\_d - d[3] = 6 - 0 = 6, which is not negative. \\ 
\rowcolor[HTML]{C7EDCC} 23. Since mask[3] equals 0, update d[3] with the value of new\_d (6) and set all\_equal to False. \\ 
\rowcolor[HTML]{C7EDCC} 24. Set mask[3] to 1. \\ 
\rowcolor[HTML]{C7EDCC} 25. In Step 2.5, increment V to 4 and return to Step 2.3. \\ 
\rowcolor[HTML]{C7EDCC} 26. In Step 2.3, since V (4) is not less than 4, return to Step 2 and increment U. \\ 
\rowcolor[HTML]{C7EDCC} 27. In Step 2, set U = 1. \\ 
\rowcolor[HTML]{C7EDCC} 28. In Step 2.1, since U (1) is less than 4, proceed to Step 2.2. \\ 
\rowcolor[HTML]{C7EDCC} 29. In Step 2.2, prev\_mask[1] equals 0, so increment U to 2 and return to Step 2.1. \\ 
\bottomrule
\end{tabular}
\end{table*}
\begin{table*}[!t]
\small
\begin{tabular}{p{15.5cm}}
\toprule
\rowcolor[HTML]{C7EDCC} 30. In Step 2.1, since U (2) is less than 4, proceed to Step 2.2. \\
\rowcolor[HTML]{C7EDCC} 31. In Step 2.2, prev\_mask[2] equals 0, so increment U to 3 and return to Step 2.1. \\ 
\rowcolor[HTML]{C7EDCC} 32. In Step 2.1, since U (3) is less than 4, proceed to Step 2.2. \\ 
\rowcolor[HTML]{C7EDCC} 33. In Step 2.2, prev\_mask[3] equals 0, so increment U to 4 and return to Step 2.1. \\ 
\rowcolor[HTML]{C7EDCC} 34. In Step 2.1, since U (4) is not less than 4, proceed to Step 3. \\ 
\rowcolor[HTML]{C7EDCC} 35. In Step 3, since all\_equal is False, go back to Step 1. \\ 
\rowcolor[HTML]{C7EDCC} 36. In Step 1, copy from d, prev\_d is: prev\_d[0] = 0, prev\_d[1] = 0, prev\_d[2] = 6, prev\_d[3] = 6. Copy from mask, prev\_mask is: prev\_mask[0] = 1, prev\_mask[1] = 0, prev\_mask[2] = 1, prev\_mask[3] = 1. \\ 
\rowcolor[HTML]{C7EDCC} 37. Set all\_equal to True. \\ 
\rowcolor[HTML]{C7EDCC} 38. In Step 2, set U = 0. \\ 
\rowcolor[HTML]{C7EDCC} 39. In Step 2.1, since U (0) is less than 4, proceed to Step 2.2. \\ 
\rowcolor[HTML]{C7EDCC} 40. In Step 2.2, prev\_mask[0] equals 1, so set V = 0. \\ 
\rowcolor[HTML]{C7EDCC} 41. In Step 2.3, since V (0) is less than 4, proceed to Step 2.4. \\ 
\rowcolor[HTML]{C7EDCC} 42. In Step 2.4, A[0, 0] is 0, so skip to Step 2.5. \\ 
\rowcolor[HTML]{C7EDCC} 43. In Step 2.5, increment V to 1 and return to Step 2.3. \\ 
\rowcolor[HTML]{C7EDCC} 44. In Step 2.3, since V (1) is less than 4, proceed to Step 2.4. \\ 
\rowcolor[HTML]{C7EDCC} 45. In Step 2.4, A[0, 1] is 0, so skip to Step 2.5. \\ 
\rowcolor[HTML]{C7EDCC} 46. In Step 2.5, increment V to 2 and return to Step 2.3. \\ 
\rowcolor[HTML]{C7EDCC} 47. In Step 2.3, since V (2) is less than 4, proceed to Step 2.4. \\ 
\rowcolor[HTML]{C7EDCC} 48. In Step 2.4, A[0, 2] is not 0 (it is 6). \\ 
\rowcolor[HTML]{C7EDCC} 49. Calculate new\_d: prev\_d[0] + A[0, 2] = 0 + 6 = 6. \\ 
\rowcolor[HTML]{C7EDCC} 50. Calculate the difference: new\_d - d[2] = 6 - 6 = 0, which is not negative. \\ 
\rowcolor[HTML]{C7EDCC} 51. Since mask[2] equals 1 and the difference is not negative, skip to Step 2.5. \\ 
\rowcolor[HTML]{C7EDCC} 52. In Step 2.5, increment V to 3 and return to Step 2.3. \\ 
\rowcolor[HTML]{C7EDCC} 53. In Step 2.3, since V (3) is less than 4, proceed to Step 2.4. \\ 
\rowcolor[HTML]{C7EDCC} 54. In Step 2.4, A[0, 3] is not 0 (it is 6). \\ 
\rowcolor[HTML]{C7EDCC} 55. Calculate new\_d: prev\_d[0] + A[0, 3] = 0 + 6 = 6. \\ 
\rowcolor[HTML]{C7EDCC} 56. Calculate the difference: new\_d - d[3] = 6 - 6 = 0, which is not negative. \\ 
\rowcolor[HTML]{C7EDCC} 57. Since mask[3] equals 1 and the difference is not negative, skip to Step 2.5. \\ 
\rowcolor[HTML]{C7EDCC} 58. In Step 2.5, increment V to 4 and return to Step 2.3. \\ 
\rowcolor[HTML]{C7EDCC} 59. In Step 2.3, since V (4) is not less than 4, return to Step 2 and increment U. \\ 
\rowcolor[HTML]{C7EDCC} 60. In Step 2, set U = 1. \\ 
\rowcolor[HTML]{C7EDCC} 61. In Step 2.1, since U (1) is less than 4, proceed to Step 2.2. \\ 
\rowcolor[HTML]{C7EDCC} 62. In Step 2.2, prev\_mask[1] equals 0, so increment U to 2 and return to Step 2.1. \\ 
\rowcolor[HTML]{C7EDCC} 63. In Step 2.1, since U (2) is less than 4, proceed to Step 2.2. \\ 
\rowcolor[HTML]{C7EDCC} 64. In Step 2.2, prev\_mask[2] equals 1, so set V = 0. \\ 
\rowcolor[HTML]{C7EDCC} 65. In Step 2.3, since V (0) is less than 4, proceed to Step 2.4. \\ 
\rowcolor[HTML]{C7EDCC} 66. In Step 2.4, A[2, 0] is not 0 (it is 6). \\ 
\rowcolor[HTML]{C7EDCC} 67. Calculate new\_d: prev\_d[2] + A[2, 0] = 6 + 6 = 12. \\ 
\rowcolor[HTML]{C7EDCC} 68. Calculate the difference: new\_d - d[0] = 12 - 0 = 12, which is not negative. \\ 
\rowcolor[HTML]{C7EDCC} 69. Since mask[0] equals 1 and the difference is not negative, skip to Step 2.5. \\ 
\rowcolor[HTML]{C7EDCC} 70. In Step 2.5, increment V to 1 and return to Step 2.3. \\ 
\rowcolor[HTML]{C7EDCC} 71. In Step 2.3, since V (1) is less than 4, proceed to Step 2.4. \\ 
\rowcolor[HTML]{C7EDCC} 72. In Step 2.4, A[2, 1] is not 0 (it is 4). \\ 
\rowcolor[HTML]{C7EDCC} 73. Calculate new\_d: prev\_d[2] + A[2, 1] = 6 + 4 = 10. \\ 
\rowcolor[HTML]{C7EDCC} 74. Calculate the difference: new\_d - d[1] = 10 - 0 = 10, which is not negative. \\ 
\rowcolor[HTML]{C7EDCC} 75. Since mask[1] equals 0, update d[1] with the value of new\_d (10) and set all\_equal to False. \\ 
\rowcolor[HTML]{C7EDCC} 76. Set mask[1] to 1. \\ 
\rowcolor[HTML]{C7EDCC} 77. In Step 2.5, increment V to 2 and return to Step 2.3. \\ 
\rowcolor[HTML]{C7EDCC} 78. In Step 2.3, since V (2) is less than 4, proceed to Step 2.4. \\ 
\rowcolor[HTML]{C7EDCC} 79. In Step 2.4, A[2, 2] is 0, so skip to Step 2.5. \\ 
\rowcolor[HTML]{C7EDCC} 80. In Step 2.5, increment V to 3 and return to Step 2.3. \\ 
\rowcolor[HTML]{C7EDCC} 81. In Step 2.3, since V (3) is less than 4, proceed to Step 2.4. \\ 
\rowcolor[HTML]{C7EDCC} 82. In Step 2.4, A[2, 3] is not 0 (it is 3). \\ 
\rowcolor[HTML]{C7EDCC} 83. Calculate new\_d: prev\_d[2] + A[2, 3] = 6 + 3 = 9. \\ 
\rowcolor[HTML]{C7EDCC} 84. Calculate the difference: new\_d - d[3] = 9 - 6 = 3, which is not negative. \\ 
\rowcolor[HTML]{C7EDCC} 85. Since mask[3] equals 1 and the difference is not negative, skip to Step 2.5. \\ 
\rowcolor[HTML]{C7EDCC} 86. In Step 2.5, increment V to 4 and return to Step 2.3. \\ 
\rowcolor[HTML]{C7EDCC} 87. In Step 2.3, since V (4) is not less than 4, return to Step 2 and increment U. \\ 
\rowcolor[HTML]{C7EDCC} 88. In Step 2, set U = 3. \\ 
\rowcolor[HTML]{C7EDCC} 89. In Step 2.1, since U (3) is less than 4, proceed to Step 2.2. \\ 
\rowcolor[HTML]{C7EDCC} 90. In Step 2.2, prev\_mask[3] equals 1, so set V = 0. \\ 
\rowcolor[HTML]{C7EDCC} 91. In Step 2.3, since V (0) is less than 4, proceed to Step 2.4. \\ 
\bottomrule
\end{tabular}
\end{table*}
\begin{table*}[!t]
\small
\begin{tabular}{p{15.5cm}}
\toprule
\rowcolor[HTML]{C7EDCC} 92. In Step 2.4, A[3, 0] is not 0 (it is 6). \\ 
\rowcolor[HTML]{C7EDCC} 93. Calculate new\_d: prev\_d[3] + A[3, 0] = 6 + 6 = 12. \\ 
\rowcolor[HTML]{C7EDCC} 94. Calculate the difference: new\_d - d[0] = 12 - 0 = 12, which is not negative. \\ 
\rowcolor[HTML]{C7EDCC} 95. Since mask[0] equals 1 and the difference is not negative, skip to Step 2.5. \\ 
\rowcolor[HTML]{C7EDCC} 96. In Step 2.5, increment V to 1 and return to Step 2.3. \\ 
\rowcolor[HTML]{C7EDCC} 97. In Step 2.3, since V (1) is less than 4, proceed to Step 2.4. \\ 
\rowcolor[HTML]{C7EDCC} 98. In Step 2.4, A[3, 1] is 0, so skip to Step 2.5. \\ 
\rowcolor[HTML]{C7EDCC} 99. In Step 2.5, increment V to 2 and return to Step 2.3. \\ 
\rowcolor[HTML]{C7EDCC} 100. In Step 2.3, since V (2) is less than 4, proceed to Step 2.4. \\ 
\rowcolor[HTML]{C7EDCC} 101. In Step 2.4, A[3, 2] is not 0 (it is 3). \\ 
\rowcolor[HTML]{C7EDCC} 102. Calculate new\_d: prev\_d[3] + A[3, 2] = 6 + 3 = 9. \\ 
\rowcolor[HTML]{C7EDCC} 103. Calculate the difference: new\_d - d[2] = 9 - 6 = 3, which is not negative. \\ 
\rowcolor[HTML]{C7EDCC} 104. Since mask[2] equals 1 and the difference is not negative, skip to Step 2.5. \\ 
\rowcolor[HTML]{C7EDCC} 105. In Step 2.5, increment V to 3 and return to Step 2.3. \\ 
\rowcolor[HTML]{C7EDCC} 106. In Step 2.3, since V (3) is less than 4, proceed to Step 2.4. \\ 
\rowcolor[HTML]{C7EDCC} 107. In Step 2.4, A[3, 3] is 0, so skip to Step 2.5. \\ 
\rowcolor[HTML]{C7EDCC} 108. In Step 2.5, increment V to 4 and return to Step 2.3. \\ 
\rowcolor[HTML]{C7EDCC} 109. In Step 2.3, since V (4) is not less than 4, return to Step 2 and increment U. \\ 
\rowcolor[HTML]{C7EDCC} 110. In Step 2, set U = 4. \\ 
\rowcolor[HTML]{C7EDCC} 111. In Step 2.1, since U (4) is not less than 4, proceed to Step 3. \\ 
\rowcolor[HTML]{C7EDCC} 112. In Step 3, since all\_equal is False, go back to Step 1. \\ 
\rowcolor[HTML]{C7EDCC} 113. In Step 1, copy from d, prev\_d is: prev\_d[0] = 0, prev\_d[1] = 10, prev\_d[2] = 6, prev\_d[3] = 6. Copy from mask, prev\_mask is: prev\_mask[0] = 1, prev\_mask[1] = 1, prev\_mask[2] = 1, prev\_mask[3] = 1. \\ 
\rowcolor[HTML]{C7EDCC} 114. Set all\_equal to True. \\ 
\rowcolor[HTML]{C7EDCC} 115. In Step 2, set U = 0. \\ 
\rowcolor[HTML]{C7EDCC} 116. In Step 2.1, since U (0) is less than 4, proceed to Step 2.2. \\ 
\rowcolor[HTML]{C7EDCC} 117. In Step 2.2, prev\_mask[0] equals 1, so set V = 0. \\ 
\rowcolor[HTML]{C7EDCC} 118. In Step 2.3, since V (0) is less than 4, proceed to Step 2.4. \\ 
\rowcolor[HTML]{C7EDCC} 119. In Step 2.4, A[0, 0] is 0, so skip to Step 2.5. \\ 
\rowcolor[HTML]{C7EDCC} 120. In Step 2.5, increment V to 1 and return to Step 2.3. \\ 
\rowcolor[HTML]{C7EDCC} 121. In Step 2.3, since V (1) is less than 4, proceed to Step 2.4. \\ 
\rowcolor[HTML]{C7EDCC} 122. In Step 2.4, A[0, 1] is 0, so skip to Step 2.5. \\ 
\rowcolor[HTML]{C7EDCC} 123. In Step 2.5, increment V to 2 and return to Step 2.3. \\ 
\rowcolor[HTML]{C7EDCC} 124. In Step 2.3, since V (2) is less than 4, proceed to Step 2.4. \\ 
\rowcolor[HTML]{C7EDCC} 125. In Step 2.4, A[0, 2] is not 0 (it is 6). \\ 
\rowcolor[HTML]{C7EDCC} 126. Calculate new\_d: prev\_d[0] + A[0, 2] = 0 + 6 = 6. \\ 
\rowcolor[HTML]{C7EDCC} 127. Calculate the difference: new\_d - d[2] = 6 - 6 = 0, which is not negative. \\ 
\rowcolor[HTML]{C7EDCC} 128. Since mask[2] equals 1 and the difference is not negative, skip to Step 2.5. \\ 
\rowcolor[HTML]{C7EDCC} 129. In Step 2.5, increment V to 3 and return to Step 2.3. \\ 
\rowcolor[HTML]{C7EDCC} 130. In Step 2.3, since V (3) is less than 4, proceed to Step 2.4. \\ 
\rowcolor[HTML]{C7EDCC} 131. In Step 2.4, A[0, 3] is not 0 (it is 6). \\ 
\rowcolor[HTML]{C7EDCC} 132. Calculate new\_d: prev\_d[0] + A[0, 3] = 0 + 6 = 6. \\ 
\rowcolor[HTML]{C7EDCC} 133. Calculate the difference: new\_d - d[3] = 6 - 6 = 0, which is not negative. \\ 
\rowcolor[HTML]{C7EDCC} 134. Since mask[3] equals 1 and the difference is not negative, skip to Step 2.5. \\ 
\rowcolor[HTML]{C7EDCC} 135. In Step 2.5, increment V to 4 and return to Step 2.3. \\ 
\rowcolor[HTML]{C7EDCC} 136. In Step 2.3, since V (4) is not less than 4, return to Step 2 and increment U. \\ 
\rowcolor[HTML]{C7EDCC} 137. In Step 2, set U = 1. \\ 
\rowcolor[HTML]{C7EDCC} 138. In Step 2.1, since U (1) is less than 4, proceed to Step 2.2. \\ 
\rowcolor[HTML]{C7EDCC} 139. In Step 2.2, prev\_mask[1] equals 1, so set V = 0. \\ 
\rowcolor[HTML]{C7EDCC} 140. In Step 2.3, since V (0) is less than 4, proceed to Step 2.4. \\ 
\rowcolor[HTML]{C7EDCC} 141. In Step 2.4, A[1, 0] is 0, so skip to Step 2.5. \\ 
\rowcolor[HTML]{C7EDCC} 142. In Step 2.5, increment V to 1 and return to Step 2.3. \\ 
\rowcolor[HTML]{C7EDCC} 143. In Step 2.3, since V (1) is less than 4, proceed to Step 2.4. \\ 
\rowcolor[HTML]{C7EDCC} 144. In Step 2.4, A[1, 1] is 0, so skip to Step 2.5. \\ 
\rowcolor[HTML]{C7EDCC} 145. In Step 2.5, increment V to 2 and return to Step 2.3. \\ 
\rowcolor[HTML]{C7EDCC} 146. In Step 2.3, since V (2) is less than 4, proceed to Step 2.4. \\ 
\rowcolor[HTML]{C7EDCC} 147. In Step 2.4, A[1, 2] is not 0 (it is 4). \\ 
\rowcolor[HTML]{C7EDCC} 148. Calculate new\_d: prev\_d[1] + A[1, 2] = 10 + 4 = 14. \\ 
\rowcolor[HTML]{C7EDCC} 149. Calculate the difference: new\_d - d[2] = 14 - 6 = 8, which is not negative. \\ 
\rowcolor[HTML]{C7EDCC} 150. Since mask[2] equals 1 and the difference is not negative, skip to Step 2.5. \\ 
\rowcolor[HTML]{C7EDCC} 151. In Step 2.5, increment V to 3 and return to Step 2.3. \\ 
\rowcolor[HTML]{C7EDCC} 152. In Step 2.3, since V (3) is less than 4, proceed to Step 2.4. \\ 
\rowcolor[HTML]{C7EDCC} 153. In Step 2.4, A[1, 3] is 0, so skip to Step 2.5. \\ 
\bottomrule
\end{tabular}
\end{table*}
\begin{table*}[!t]
\small
\begin{tabular}{p{15.5cm}}
\toprule
\rowcolor[HTML]{C7EDCC} 154. In Step 2.5, increment V to 4 and return to Step 2.3. \\ 
\rowcolor[HTML]{C7EDCC} 155. In Step 2.3, since V (4) is not less than 4, return to Step 2 and increment U. \\ 
\rowcolor[HTML]{C7EDCC} 156. In Step 2, set U = 2. \\ 
\rowcolor[HTML]{C7EDCC} 157. In Step 2.1, since U (2) is less than 4, proceed to Step 2.2. \\ 
\rowcolor[HTML]{C7EDCC} 158. In Step 2.2, prev\_mask[2] equals 1, so set V = 0. \\ 
\rowcolor[HTML]{C7EDCC} 159. In Step 2.3, since V (0) is less than 4, proceed to Step 2.4. \\ 
\rowcolor[HTML]{C7EDCC} 160. In Step 2.4, A[2, 0] is not 0 (it is 6). \\ 
\rowcolor[HTML]{C7EDCC} 161. Calculate new\_d: prev\_d[2] + A[2, 0] = 6 + 6 = 12. \\ 
\rowcolor[HTML]{C7EDCC} 162. Calculate the difference: new\_d - d[0] = 12 - 0 = 12, which is not negative. \\ 
\rowcolor[HTML]{C7EDCC} 163. Since mask[0] equals 1 and the difference is not negative, skip to Step 2.5. \\ 
\rowcolor[HTML]{C7EDCC} 164. In Step 2.5, increment V to 1 and return to Step 2.3. \\ 
\rowcolor[HTML]{C7EDCC} 165. In Step 2.3, since V (1) is less than 4, proceed to Step 2.4. \\ 
\rowcolor[HTML]{C7EDCC} 166. In Step 2.4, A[2, 1] is not 0 (it is 4). \\ 
\rowcolor[HTML]{C7EDCC} 167. Calculate new\_d: prev\_d[2] + A[2, 1] = 6 + 4 = 10. \\ 
\rowcolor[HTML]{C7EDCC} 168. Calculate the difference: new\_d - d[1] = 10 - 10 = 0, which is not negative. \\ 
\rowcolor[HTML]{C7EDCC} 169. Since mask[1] equals 1 and the difference is not negative, skip to Step 2.5. \\ 
\rowcolor[HTML]{C7EDCC} 170. In Step 2.5, increment V to 2 and return to Step 2.3. \\ 
\rowcolor[HTML]{C7EDCC} 171. In Step 2.3, since V (2) is less than 4, proceed to Step 2.4. \\ 
\rowcolor[HTML]{C7EDCC} 172. In Step 2.4, A[2, 2] is 0, so skip to Step 2.5. \\ 
\rowcolor[HTML]{C7EDCC} 173. In Step 2.5, increment V to 3 and return to Step 2.3. \\ 
\rowcolor[HTML]{C7EDCC} 174. In Step 2.3, since V (3) is less than 4, proceed to Step 2.4. \\ 
\rowcolor[HTML]{C7EDCC} 175. In Step 2.4, A[2, 3] is not 0 (it is 3). \\ 
\rowcolor[HTML]{C7EDCC} 176. Calculate new\_d: prev\_d[2] + A[2, 3] = 6 + 3 = 9. \\ 
\rowcolor[HTML]{C7EDCC} 177. Calculate the difference: new\_d - d[3] = 9 - 6 = 3, which is not negative. \\ 
\rowcolor[HTML]{C7EDCC} 178. Since mask[3] equals 1 and the difference is not negative, skip to Step 2.5. \\ 
\rowcolor[HTML]{C7EDCC} 179. In Step 2.5, increment V to 4 and return to Step 2.3. \\ 
\rowcolor[HTML]{C7EDCC} 180. In Step 2.3, since V (4) is not less than 4, return to Step 2 and increment U. \\ 
\rowcolor[HTML]{C7EDCC} 181. In Step 2, set U = 3. \\ 
\rowcolor[HTML]{C7EDCC} 182. In Step 2.1, since U (3) is less than 4, proceed to Step 2.2. \\ 
\rowcolor[HTML]{C7EDCC} 183. In Step 2.2, prev\_mask[3] equals 1, so set V = 0. \\ 
\rowcolor[HTML]{C7EDCC} 184. In Step 2.3, since V (0) is less than 4, proceed to Step 2.4. \\ 
\rowcolor[HTML]{C7EDCC} 185. In Step 2.4, A[3, 0] is not 0 (it is 6). \\ 
\rowcolor[HTML]{C7EDCC} 186. Calculate new\_d: prev\_d[3] + A[3, 0] = 6 + 6 = 12. \\ 
\rowcolor[HTML]{C7EDCC} 187. Calculate the difference: new\_d - d[0] = 12 - 0 = 12, which is not negative. \\ 
\rowcolor[HTML]{C7EDCC} 188. Since mask[0] equals 1 and the difference is not negative, skip to Step 2.5. \\ 
\rowcolor[HTML]{C7EDCC} 189. In Step 2.5, increment V to 1 and return to Step 2.3. \\ 
\rowcolor[HTML]{C7EDCC} 190. In Step 2.3, since V (1) is less than 4, proceed to Step 2.4. \\ 
\rowcolor[HTML]{C7EDCC} 191. In Step 2.4, A[3, 1] is 0, so skip to Step 2.5. \\ 
\rowcolor[HTML]{C7EDCC} 192. In Step 2.5, increment V to 2 and return to Step 2.3. \\ 
\rowcolor[HTML]{C7EDCC} 193. In Step 2.3, since V (2) is less than 4, proceed to Step 2.4. \\ 
\rowcolor[HTML]{C7EDCC} 194. In Step 2.4, A[3, 2] is not 0 (it is 3). \\ 
\rowcolor[HTML]{C7EDCC} 195. Calculate new\_d: prev\_d[3] + A[3, 2] = 6 + 3 = 9. \\ 
\rowcolor[HTML]{C7EDCC} 196. Calculate the difference: new\_d - d[2] = 9 - 6 = 3, which is not negative. \\ 
\rowcolor[HTML]{C7EDCC} 197. Since mask[2] equals 1 and the difference is not negative, skip to Step 2.5. \\ 
\rowcolor[HTML]{C7EDCC} 198. In Step 2.5, increment V to 3 and return to Step 2.3. \\ 
\rowcolor[HTML]{C7EDCC} 199. In Step 2.3, since V (3) is less than 4, proceed to Step 2.4. \\ 
\rowcolor[HTML]{C7EDCC} 200. In Step 2.4, A[3, 3] is 0, so skip to Step 2.5. \\ 
\rowcolor[HTML]{C7EDCC} 201. In Step 2.5, increment V to 4 and return to Step 2.3. \\ 
\rowcolor[HTML]{C7EDCC} 202. In Step 2.3, since V (4) is not less than 4, return to Step 2 and increment U. \\ 
\rowcolor[HTML]{C7EDCC} 203. In Step 2, set U = 4. \\ 
\rowcolor[HTML]{C7EDCC} 204. In Step 2.1, since U (4) is not less than 4, proceed to Step 3. \\ 
\rowcolor[HTML]{C7EDCC} 205. In Step 3, since all\_equal is True, proceed to Step 4. \\ 
\rowcolor[HTML]{C7EDCC} 206. In Step 4, print the array d: d[0] = 0, d[1] = 10, d[2] = 6, d[3] = 6.
\\ \bottomrule
\end{tabular}
\caption{The prompt of bellman ford and the response of \texttt{gpt-4}.}
\label{table:clrs_mini/bellman_ford}

\end{table*}
\begin{table*}[h]
\small
\begin{tabular}{p{15cm}}
\toprule
\textbf{Prompt} \\ \midrule
Execute the instructions step by step. Do not jump steps. Do not stop before completion. \\ 
Initial: Set the adjacency matrix A with the size 4x4: \\ 
A[0, 0] = 0, A[0, 1] = 0, A[0, 2] = 5, A[0, 3] = 4 \\ 
A[1, 0] = 0, A[1, 1] = 0, A[1, 2] = 6, A[1, 3] = 5 \\ 
A[2, 0] = 5, A[2, 1] = 6, A[2, 2] = 0, A[2, 3] = 4 \\ 
A[3, 0] = 4, A[3, 1] = 5, A[3, 2] = 4, A[3, 3] = 0, \\ 
where N = 4 is the number of vertices. Set the start vertex s = 0. \\ 
Set variables d, mark, and in\_queue as arrays of zeros with 4 elements: \\ 
d: d[0] = 0, d[1] = 0, d[2] = 0, d[3] = 0, \\ 
mark: mark[0] = 0, mark[1] = 0, mark[2] = 0, mark[3] = 0, \\ 
in\_queue: in\_queue[0] = 1, in\_queue[1] = 0, in\_queue[2] = 0, in\_queue[3] = 0. \\ 
 \\ 
Step 1: Assign k to 0. \\ 
Step 2.1: Assign u to -1. \\ 
Step 2.2: Assign i to 0. \\ 
Step 2.2.1: If i is less than 4, go to Step 2.2.2. Otherwise, go to Step 2.3. \\ 
Step 2.2.2: Check if in\_queue[i] equals 1. If True, go to Step 2.2.3. \\ 
Step 2.2.3: Check if u equals -1. If True, assign i to u. Then, go to Step 2.2.5. \\ 
Step 2.2.4: What are the value of d[i] and d[u]? Then, what is the result of d[i] - d[u]? Is it negative? If True, assign i to u. \\ 
Step 2.2.5: Increment i by 1, and go to Step 2.2.1. \\ 
Step 2.3: Check if u equals -1. \\ 
\quad a. If True, go to Step 3. \\ 
Step 2.4: Set mark[u] to 1. \\ 
Step 2.5: Set in\_queue[u] to 0. \\ 
Step 2.6: Assign v to 0. \\ 
Step 2.6.1: If v is less than 4, go to Step 2.6.2. Otherwise, go to Step 2.7. \\ 
Step 2.6.2: Check if A[u, v] does not equal to 0. If True, perform these sub-steps: \\ 
\quad a. Calculate new\_d by adding d[u] and A[u, v]. \\ 
\quad b. Calculate the difference between new\_d and d[v]. Then, is it negative? \\ 
\quad c. Check if mark[v] equals 0. If True, then what are the values of in\_queue[v]? Check (in\_queue[v] equals 0) OR (the difference is less than 0) is True. \\ 
\quad \quad i. If either is True, set d[v] to the value of new\_d, and \\ 
\quad \quad ii. Set in\_queue[v] to 1. \\ 
Step 2.6.3: Increment v by 1, and go to Step 2.6.1. \\ 
Step 2.7: Increment k by 1. If k < 4, go to Step 2.1. Otherwise, go to Step 3. \\ 
Step 3: Print the array of d. \\ 
----- \\ 
The computation procedure is as follows, with each line numbered sequentially (e.g., 1. 2. 3. 4. 5. 6. etc.): \\ 
1. In Step 1, k = 0. \\ 
\rowcolor[HTML]{C7EDCC} 2. In Step 2.1, u = -1. \\ 
\rowcolor[HTML]{C7EDCC} 3. In Step 2.2, i = 0. \\ 
\rowcolor[HTML]{C7EDCC} 4. In Step 2.2.1, i < 4, so go to Step 2.2.2. \\ 
\rowcolor[HTML]{C7EDCC} 5. In Step 2.2.2, in\_queue[0] = 1, so go to Step 2.2.3. \\ 
\rowcolor[HTML]{C7EDCC} 6. In Step 2.2.3, u = -1, so assign i to u, u = 0. \\ 
\rowcolor[HTML]{C7EDCC} 7. In Step 2.2.5, increment i by 1, i = 1. \\ 
\rowcolor[HTML]{C7EDCC} 8. In Step 2.2.1, i < 4, so go to Step 2.2.2. \\ 
\rowcolor[HTML]{C7EDCC} 9. In Step 2.2.2, in\_queue[1] = 0, so go to Step 2.2.5. \\ 
\rowcolor[HTML]{C7EDCC} 10. In Step 2.2.5, increment i by 1, i = 2. \\ 
\rowcolor[HTML]{C7EDCC} 11. In Step 2.2.1, i < 4, so go to Step 2.2.2. \\ 
\rowcolor[HTML]{C7EDCC} 12. In Step 2.2.2, in\_queue[2] = 0, so go to Step 2.2.5. \\ 
\rowcolor[HTML]{C7EDCC} 13. In Step 2.2.5, increment i by 1, i = 3. \\ 
\rowcolor[HTML]{C7EDCC} 14. In Step 2.2.1, i < 4, so go to Step 2.2.2. \\ 
\rowcolor[HTML]{C7EDCC} 15. In Step 2.2.2, in\_queue[3] = 0, so go to Step 2.2.5. \\ 
\rowcolor[HTML]{C7EDCC} 16. In Step 2.2.5, increment i by 1, i = 4. \\ 
\rowcolor[HTML]{C7EDCC} 17. In Step 2.2.1, i >= 4, so go to Step 2.3. \\ 
\rowcolor[HTML]{C7EDCC} 18. In Step 2.3, u != -1, so go to Step 2.4. \\ 
\rowcolor[HTML]{C7EDCC} 19. In Step 2.4, set mark[0] = 1. \\ 
\rowcolor[HTML]{C7EDCC} 20. In Step 2.5, set in\_queue[0] = 0. \\ 
\rowcolor[HTML]{C7EDCC} 21. In Step 2.6, v = 0. \\ 
\rowcolor[HTML]{C7EDCC} 22. In Step 2.6.1, v < 4, so go to Step 2.6.2. \\ 
\bottomrule
\end{tabular}
\end{table*}
\begin{table*}[!t]
\small
\begin{tabular}{p{15.5cm}}
\toprule
\rowcolor[HTML]{C7EDCC} 23. In Step 2.6.2, A[0, 0] = 0, so go to Step 2.6.3. \\ 
\rowcolor[HTML]{C7EDCC} 24. In Step 2.6.3, increment v by 1, v = 1. \\ 
\rowcolor[HTML]{C7EDCC} 25. In Step 2.6.1, v < 4, so go to Step 2.6.2. \\ 
\rowcolor[HTML]{C7EDCC} 26. In Step 2.6.2, A[0, 1] = 0, so go to Step 2.6.3. \\ 
\rowcolor[HTML]{C7EDCC} 27. In Step 2.6.3, increment v by 1, v = 2. \\ 
\rowcolor[HTML]{C7EDCC} 28. In Step 2.6.1, v < 4, so go to Step 2.6.2. \\ 
\rowcolor[HTML]{C7EDCC} 29. In Step 2.6.2, A[0, 2] = 5, so perform sub-steps: \\ 
\rowcolor[HTML]{C7EDCC} \quad a. new\_d = d[0] + A[0, 2] = 0 + 5 = 5. \\ 
\rowcolor[HTML]{C7EDCC} \quad b. new\_d - d[2] = 5 - 0 = 5, which is not negative. \\ 
\rowcolor[HTML]{C7EDCC} \quad c. mark[2] = 0, in\_queue[2] = 0, so set d[2] = 5 and in\_queue[2] = 1. \\ 
\rowcolor[HTML]{C7EDCC} 30. In Step 2.6.3, increment v by 1, v = 3. \\ 
\rowcolor[HTML]{C7EDCC} 31. In Step 2.6.1, v < 4, so go to Step 2.6.2. \\ 
\rowcolor[HTML]{C7EDCC} 32. In Step 2.6.2, A[0, 3] = 3, so perform sub-steps: \\ 
\rowcolor[HTML]{C7EDCC} \quad a. new\_d = d[0] + A[0, 3] = 0 + 3 = 3. \\ 
\rowcolor[HTML]{C7EDCC} \quad b. new\_d - d[3] = 3 - 0 = 3, which is not negative. \\ 
\rowcolor[HTML]{C7EDCC} \quad c. mark[3] = 0, in\_queue[3] = 0, so set d[3] = 3 and in\_queue[3] = 1. \\ 
\rowcolor[HTML]{C7EDCC} 33. In Step 2.6.3, increment v by 1, v = 4. \\ 
\rowcolor[HTML]{C7EDCC} 34. In Step 2.6.1, v >= 4, so go to Step 2.7. \\ 
\rowcolor[HTML]{C7EDCC} 35. In Step 2.7, increment k by 1, k = 1. \\ 
\rowcolor[HTML]{C7EDCC} 36. In Step 2.1, u = -1. \\ 
\rowcolor[HTML]{C7EDCC} 37. In Step 2.2, i = 0. \\ 
\rowcolor[HTML]{C7EDCC} 38. In Step 2.2.1, i < 4, so go to Step 2.2.2. \\ 
\rowcolor[HTML]{C7EDCC} 39. In Step 2.2.2, in\_queue[0] = 0, so go to Step 2.2.5. \\ 
\rowcolor[HTML]{C7EDCC} 40. In Step 2.2.5, increment i by 1, i = 1. \\ 
\rowcolor[HTML]{C7EDCC} 41. In Step 2.2.1, i < 4, so go to Step 2.2.2. \\ 
\rowcolor[HTML]{C7EDCC} 42. In Step 2.2.2, in\_queue[1] = 0, so go to Step 2.2.5. \\ 
\rowcolor[HTML]{C7EDCC} 43. In Step 2.2.5, increment i by 1, i = 2. \\ 
\rowcolor[HTML]{C7EDCC} 44. In Step 2.2.1, i < 4, so go to Step 2.2.2. \\ 
\rowcolor[HTML]{C7EDCC} 45. In Step 2.2.2, in\_queue[2] = 1, so go to Step 2.2.3. \\ 
\rowcolor[HTML]{C7EDCC} 46. In Step 2.2.3, u = -1, so assign i to u, u = 2. \\ 
\rowcolor[HTML]{C7EDCC} 47. In Step 2.2.5, increment i by 1, i = 3. \\ 
\rowcolor[HTML]{C7EDCC} 48. In Step 2.2.1, i < 4, so go to Step 2.2.2. \\ 
\rowcolor[HTML]{C7EDCC} 49. In Step 2.2.2, in\_queue[3] = 1, so go to Step 2.2.4. \\ 
\rowcolor[HTML]{C7EDCC} 50. In Step 2.2.4, d[i] = 3, d[u] = 5, d[i] - d[u] = -2, which is negative, so assign i to u, u = 3. \\ 
\rowcolor[HTML]{C7EDCC} 51. In Step 2.2.5, increment i by 1, i = 4. \\ 
\rowcolor[HTML]{C7EDCC} 52. In Step 2.2.1, i >= 4, so go to Step 2.3. \\ 
\rowcolor[HTML]{C7EDCC} 53. In Step 2.3, u != -1, so go to Step 2.4. \\ 
\rowcolor[HTML]{C7EDCC} 54. In Step 2.4, set mark[3] = 1. \\ 
\rowcolor[HTML]{C7EDCC} 55. In Step 2.5, set in\_queue[3] = 0. \\ 
\rowcolor[HTML]{C7EDCC} 56. In Step 2.6, v = 0. \\ 
\rowcolor[HTML]{C7EDCC} 57. In Step 2.6.1, v < 4, so go to Step 2.6.2. \\ 
\rowcolor[HTML]{C7EDCC} 58. In Step 2.6.2, A[3, 0] = 3, so perform sub-steps: \\ 
\rowcolor[HTML]{C7EDCC} \quad a. new\_d = d[3] + A[3, 0] = 3 + 3 = 6. \\ 
\rowcolor[HTML]{C7EDCC} \quad b. new\_d - d[0] = 6 - 0 = 6, which is not negative. \\ 
\rowcolor[HTML]{C7EDCC} \quad c. mark[0] = 1, in\_queue[0] = 0, so go to Step 2.6.3. \\ 
\rowcolor[HTML]{C7EDCC} 59. In Step 2.6.3, increment v by 1, v = 1. \\ 
\rowcolor[HTML]{C7EDCC} 60. In Step 2.6.1, v < 4, so go to Step 2.6.2. \\ 
\rowcolor[HTML]{C7EDCC} 61. In Step 2.6.2, A[3, 1] = 6, so perform sub-steps: \\ 
\rowcolor[HTML]{C7EDCC} \quad a. new\_d = d[3] + A[3, 1] = 3 + 6 = 9. \\ 
\rowcolor[HTML]{C7EDCC} \quad b. new\_d - d[1] = 9 - 0 = 9, which is not negative. \\ 
\rowcolor[HTML]{C7EDCC} \quad c. mark[1] = 0, in\_queue[1] = 0, so set d[1] = 9 and in\_queue[1] = 1. \\ 
\rowcolor[HTML]{C7EDCC} 62. In Step 2.6.3, increment v by 1, v = 2. \\ 
\rowcolor[HTML]{C7EDCC} 63. In Step 2.6.1, v < 4, so go to Step 2.6.2. \\ 
\rowcolor[HTML]{C7EDCC} 64. In Step 2.6.2, A[3, 2] = 5, so perform sub-steps: \\ 
\rowcolor[HTML]{C7EDCC} \quad a. new\_d = d[3] + A[3, 2] = 3 + 5 = 8. \\ 
\rowcolor[HTML]{C7EDCC} \quad b. new\_d - d[2] = 8 - 5 = 3, which is not negative. \\ 
\rowcolor[HTML]{C7EDCC} \quad c. mark[2] = 0, in\_queue[2] = 1, so go to Step 2.6.3. \\ 
\rowcolor[HTML]{C7EDCC} 65. In Step 2.6.3, increment v by 1, v = 3. \\ 
\rowcolor[HTML]{C7EDCC} 66. In Step 2.6.1, v < 4, so go to Step 2.6.2. \\ 
\rowcolor[HTML]{C7EDCC} 67. In Step 2.6.2, A[3, 3] = 0, so go to Step 2.6.3. \\ 
\rowcolor[HTML]{C7EDCC} 68. In Step 2.6.3, increment v by 1, v = 4. \\ 
\rowcolor[HTML]{C7EDCC} 69. In Step 2.6.1, v >= 4, so go to Step 2.7. \\ 
\rowcolor[HTML]{C7EDCC} 70. In Step 2.7, increment k by 1, k = 2. \\ 
\bottomrule
\end{tabular}
\end{table*}
\begin{table*}[!t]
\small
\begin{tabular}{p{15.5cm}}
\toprule
\rowcolor[HTML]{C7EDCC} 71. In Step 2.1, u = -1. \\ 
\rowcolor[HTML]{C7EDCC} 72. In Step 2.2, i = 0. \\ 
\rowcolor[HTML]{C7EDCC} 73. In Step 2.2.1, i < 4, so go to Step 2.2.2. \\ 
\rowcolor[HTML]{C7EDCC} 74. In Step 2.2.2, in\_queue[0] = 0, so go to Step 2.2.5. \\ 
\rowcolor[HTML]{C7EDCC} 75. In Step 2.2.5, increment i by 1, i = 1. \\ 
\rowcolor[HTML]{C7EDCC} 76. In Step 2.2.1, i < 4, so go to Step 2.2.2. \\ 
\rowcolor[HTML]{C7EDCC} 77. In Step 2.2.2, in\_queue[1] = 1, so go to Step 2.2.3. \\ 
\rowcolor[HTML]{C7EDCC} 78. In Step 2.2.3, u = -1, so assign i to u, u = 1. \\ 
\rowcolor[HTML]{C7EDCC} 79. In Step 2.2.5, increment i by 1, i = 2. \\ 
\rowcolor[HTML]{C7EDCC} 80. In Step 2.2.1, i < 4, so go to Step 2.2.2. \\ 
\rowcolor[HTML]{C7EDCC} 81. In Step 2.2.2, in\_queue[2] = 1, so go to Step 2.2.4. \\ 
\rowcolor[HTML]{C7EDCC} 82. In Step 2.2.4, d[i] = 5, d[u] = 9, d[i] - d[u] = -4, which is negative, so assign i to u, u = 2. \\ 
\rowcolor[HTML]{C7EDCC} 83. In Step 2.2.5, increment i by 1, i = 3. \\ 
\rowcolor[HTML]{C7EDCC} 84. In Step 2.2.1, i < 4, so go to Step 2.2.2. \\ 
\rowcolor[HTML]{C7EDCC} 85. In Step 2.2.2, in\_queue[3] = 0, so go to Step 2.2.5. \\ 
\rowcolor[HTML]{C7EDCC} 86. In Step 2.2.5, increment i by 1, i = 4. \\ 
\rowcolor[HTML]{C7EDCC} 87. In Step 2.2.1, i >= 4, so go to Step 2.3. \\ 
\rowcolor[HTML]{C7EDCC} 88. In Step 2.3, u != -1, so go to Step 2.4. \\ 
\rowcolor[HTML]{C7EDCC} 89. In Step 2.4, set mark[2] = 1. \\ 
\rowcolor[HTML]{C7EDCC} 90. In Step 2.5, set in\_queue[2] = 0. \\ 
\rowcolor[HTML]{C7EDCC} 91. In Step 2.6, v = 0. \\ 
\rowcolor[HTML]{C7EDCC} 92. In Step 2.6.1, v < 4, so go to Step 2.6.2. \\ 
\rowcolor[HTML]{C7EDCC} 93. In Step 2.6.2, A[2, 0] = 5, so perform sub-steps: \\ 
\rowcolor[HTML]{C7EDCC} \quad a. new\_d = d[2] + A[2, 0] = 5 + 5 = 10. \\ 
\rowcolor[HTML]{C7EDCC} \quad b. new\_d - d[0] = 10 - 0 = 10, which is not negative. \\ 
\rowcolor[HTML]{C7EDCC} \quad c. mark[0] = 1, in\_queue[0] = 0, so go to Step 2.6.3. \\ 
\rowcolor[HTML]{C7EDCC} 94. In Step 2.6.3, increment v by 1, v = 1. \\ 
\rowcolor[HTML]{C7EDCC} 95. In Step 2.6.1, v < 4, so go to Step 2.6.2. \\ 
\rowcolor[HTML]{C7EDCC} 96. In Step 2.6.2, A[2, 1] = 0, so go to Step 2.6.3. \\ 
\rowcolor[HTML]{C7EDCC} 97. In Step 2.6.3, increment v by 1, v = 2. \\ 
\rowcolor[HTML]{C7EDCC} 98. In Step 2.6.1, v < 4, so go to Step 2.6.2. \\ 
\rowcolor[HTML]{C7EDCC} 99. In Step 2.6.2, A[2, 2] = 0, so go to Step 2.6.3. \\ 
\rowcolor[HTML]{C7EDCC} 100. In Step 2.6.3, increment v by 1, v = 3. \\ 
\rowcolor[HTML]{C7EDCC} 101. In Step 2.6.1, v < 4, so go to Step 2.6.2. \\ 
\rowcolor[HTML]{C7EDCC} 102. In Step 2.6.2, A[2, 3] = 5, so perform sub-steps: \\ 
\rowcolor[HTML]{C7EDCC} \quad a. new\_d = d[2] + A[2, 3] = 5 + 5 = 10. \\ 
\rowcolor[HTML]{C7EDCC} \quad b. new\_d - d[3] = 10 - 3 = 7, which is not negative. \\ 
\rowcolor[HTML]{C7EDCC} \quad c. mark[3] = 1, in\_queue[3] = 0, so go to Step 2.6.3. \\ 
\rowcolor[HTML]{C7EDCC} 103. In Step 2.6.3, increment v by 1, v = 4. \\ 
\rowcolor[HTML]{C7EDCC} 104. In Step 2.6.1, v >= 4, so go to Step 2.7. \\ 
\rowcolor[HTML]{C7EDCC} 105. In Step 2.7, increment k by 1, k = 3. \\ 
\rowcolor[HTML]{C7EDCC} 106. In Step 2.1, u = -1. \\ 
\rowcolor[HTML]{C7EDCC} 107. In Step 2.2, i = 0. \\ 
\rowcolor[HTML]{C7EDCC} 108. In Step 2.2.1, i < 4, so go to Step 2.2.2. \\ 
\rowcolor[HTML]{C7EDCC} 109. In Step 2.2.2, in\_queue[0] = 0, so go to Step 2.2.5. \\ 
\rowcolor[HTML]{C7EDCC} 110. In Step 2.2.5, increment i by 1, i = 1. \\ 
\rowcolor[HTML]{C7EDCC} 111. In Step 2.2.1, i < 4, so go to Step 2.2.2. \\ 
\rowcolor[HTML]{C7EDCC} 112. In Step 2.2.2, in\_queue[1] = 1, so go to Step 2.2.3. \\ 
\rowcolor[HTML]{C7EDCC} 113. In Step 2.2.3, u = -1, so assign i to u, u = 1. \\ 
\rowcolor[HTML]{C7EDCC} 114. In Step 2.2.5, increment i by 1, i = 2. \\ 
\rowcolor[HTML]{C7EDCC} 115. In Step 2.2.1, i < 4, so go to Step 2.2.2. \\ 
\rowcolor[HTML]{C7EDCC} 116. In Step 2.2.2, in\_queue[2] = 0, so go to Step 2.2.5. \\ 
\rowcolor[HTML]{C7EDCC} 117. In Step 2.2.5, increment i by 1, i = 3. \\ 
\rowcolor[HTML]{C7EDCC} 118. In Step 2.2.1, i < 4, so go to Step 2.2.2. \\ 
\rowcolor[HTML]{C7EDCC} 119. In Step 2.2.2, in\_queue[3] = 0, so go to Step 2.2.5. \\ 
\rowcolor[HTML]{C7EDCC} 120. In Step 2.2.5, increment i by 1, i = 4. \\ 
\rowcolor[HTML]{C7EDCC} 121. In Step 2.2.1, i >= 4, so go to Step 2.3. \\ 
\rowcolor[HTML]{C7EDCC} 122. In Step 2.3, u != -1, so go to Step 2.4. \\ 
\rowcolor[HTML]{C7EDCC} 123. In Step 2.4, set mark[1] = 1. \\ 
\bottomrule
\end{tabular}
\end{table*}
\begin{table*}[!t]
\small
\begin{tabular}{p{15.5cm}}
\toprule
\rowcolor[HTML]{C7EDCC} 124. In Step 2.5, set in\_queue[1] = 0. \\ 
\rowcolor[HTML]{C7EDCC} 125. In Step 2.6, v = 0. \\ 
\rowcolor[HTML]{C7EDCC} 126. In Step 2.6.1, v < 4, so go to Step 2.6.2. \\ 
\rowcolor[HTML]{C7EDCC} 127. In Step 2.6.2, A[1, 0] = 0, so go to Step 2.6.3. \\ 
\rowcolor[HTML]{C7EDCC} 128. In Step 2.6.3, increment v by 1, v = 1. \\ 
\rowcolor[HTML]{C7EDCC} 129. In Step 2.6.1, v < 4, so go to Step 2.6.2. \\ 
\rowcolor[HTML]{C7EDCC} 130. In Step 2.6.2, A[1, 1] = 0, so go to Step 2.6.3. \\ 
\rowcolor[HTML]{C7EDCC} 131. In Step 2.6.3, increment v by 1, v = 2. \\ 
\rowcolor[HTML]{C7EDCC} 132. In Step 2.6.1, v < 4, so go to Step 2.6.2. \\ 
\rowcolor[HTML]{C7EDCC} 133. In Step 2.6.2, A[1, 2] = 0, so go to Step 2.6.3. \\ 
\rowcolor[HTML]{C7EDCC} 134. In Step 2.6.3, increment v by 1, v = 3. \\ 
\rowcolor[HTML]{C7EDCC} 135. In Step 2.6.1, v < 4, so go to Step 2.6.2. \\ 
\rowcolor[HTML]{C7EDCC} 136. In Step 2.6.2, A[1, 3] = 6, so perform sub-steps: \\ 
\rowcolor[HTML]{C7EDCC} \quad a. new\_d = d[1] + A[1, 3] = 9 + 6 = 15. \\ 
\rowcolor[HTML]{C7EDCC} \quad b. new\_d - d[3] = 15 - 3 = 12, which is not negative. \\ 
\rowcolor[HTML]{C7EDCC} \quad c. mark[3] = 1, in\_queue[3] = 0, so go to Step 2.6.3. \\ 
\rowcolor[HTML]{C7EDCC} 137. In Step 2.6.3, increment v by 1, v = 4. \\ 
\rowcolor[HTML]{C7EDCC} 138. In Step 2.6.1, v >= 4, so go to Step 2.7. \\ 
\rowcolor[HTML]{C7EDCC} 139. In Step 2.7, increment k by 1, k = 4. \\ 
\rowcolor[HTML]{C7EDCC} 140. In Step 3, print the array of d: [0, 9, 5, 3].
\\ \bottomrule
\end{tabular}
\caption{The prompt of dijkstra and the response of \texttt{gpt-4}.}
\label{table:clrs_mini/dijkstra}

\end{table*}
\begin{table*}[h]
\small
\begin{tabular}{p{15cm}}
\toprule
\textbf{Prompt} \\ \midrule
Execute the instructions step by step. Do not jump steps. Do not stop before completion. When reading or updating a matrix at some index, use concrete values rather than just variables. \\ 
Initial: Set the distance matrix dist with the size 4x4: \\ 
dist[0, 0] = 0, dist[0, 1] = 3, dist[0, 2] = 6, dist[0, 3] = 2 \\ 
dist[1, 0] = 4, dist[1, 1] = 0, dist[1, 2] = 3, dist[1, 3] = 6 \\ 
dist[2, 0] = 6, dist[2, 1] = 2, dist[2, 2] = 0, dist[2, 3] = 3 \\ 
dist[3, 0] = 5, dist[3, 1] = 1, dist[3, 2] = 4, dist[3, 3] = 0, \\ 
where V = 4 is the number of vertices. \\ 
 \\ 
Step 1: Set k equal to 0. \\ 
Step 1.1: If k < 4, perform Step 2. Otherwise, go to Step 4. \\ 
Step 2: Set i equal to 0. \\ 
Step 2.1: If i < 4, perform Step 2.2. Otherwise, go to Step 1 and increment k. \\ 
Step 2.2: Compute i - k and check if it's non-zero. If True, perform Step 3. Otherwise, go to Step 2 and increment i. \\ 
Step 3: Set j equal to 0. \\ 
Step 3.1: If j < 4, perform Step 3.2. Otherwise, go to Step 2 and increment i. \\ 
Step 3.2: Compute j - k and check if it's non-zero: \\ 
\quad a. If True, Compute j - i and check if it's non-zero: \\ 
\quad \quad i. If True, perform Step 3.3. \\ 
\quad \quad i. If False, go to Step 3 and increment j. \\ 
\quad b. If False, go to Step 3 and increment j. \\ 
Step 3.2.1: If True, perform Step 3.3. Otherwise, go to Step 3 and increment j. \\ 
Step 3.3.1: Calculate dist\_k as dist[i, k] + dist[k, j]. \\ 
Step 3.3.2: Set dist[i, j] as min(dist[i, j], dist\_k). \\ 
Step 3.4: Increment j and return to Step 3.1. \\ 
Step 4: Print the distance matrix dist by iterating over its elements, displaying each value as "dist[{i}, {j}] = {dist\_ij}". \\ 
----- \\ 
The computation procedure is as follows, with each line numbered sequentially (e.g., 1. 2. 3. 4. 5. 6. etc.): \\ 
1. Step 1, k = 0. \\ 
\rowcolor[HTML]{C7EDCC} 2. Step 1.1, k < 4, so proceed to Step 2. \\ 
\rowcolor[HTML]{C7EDCC} 3. Step 2, i = 0. \\ 
\rowcolor[HTML]{C7EDCC} 4. Step 2.1, i < 4, so proceed to Step 2.2. \\ 
\rowcolor[HTML]{C7EDCC} 5. Step 2.2, i - k = 0, so increment i and return to Step 2.1. \\ 
\rowcolor[HTML]{C7EDCC} 6. Step 2.1, i = 1, i < 4, so proceed to Step 2.2. \\ 
\rowcolor[HTML]{C7EDCC} 7. Step 2.2, i - k = 1, so proceed to Step 3. \\ 
\rowcolor[HTML]{C7EDCC} 8. Step 3, j = 0. \\ 
\bottomrule
\end{tabular}
\end{table*}
\begin{table*}[!t]
\small
\begin{tabular}{p{15.5cm}}
\toprule
\rowcolor[HTML]{C7EDCC} 9. Step 3.1, j < 4, so proceed to Step 3.2. \\ 
\rowcolor[HTML]{C7EDCC} 10. Step 3.2, j - k = 0, so increment j and return to Step 3.1. \\ 
\rowcolor[HTML]{C7EDCC} 11. Step 3.1, j = 1, j < 4, so proceed to Step 3.2. \\ 
\rowcolor[HTML]{C7EDCC} 12. Step 3.2, j - k = 1, so proceed to Step 3.2.1. \\ 
\rowcolor[HTML]{C7EDCC} 13. Step 3.2.1, j - i = 0, so increment j and return to Step 3.1. \\ 
\rowcolor[HTML]{C7EDCC} 14. Step 3.1, j = 2, j < 4, so proceed to Step 3.2. \\ 
\rowcolor[HTML]{C7EDCC} 15. Step 3.2, j - k = 2, so proceed to Step 3.2.1. \\ 
\rowcolor[HTML]{C7EDCC} 16. Step 3.2.1, j - i = 1, so proceed to Step 3.3. \\ 
\rowcolor[HTML]{C7EDCC} 17. Step 3.3.1, dist\_k = dist[1, 0] + dist[0, 2] = 4 + 6 = 10. \\ 
\rowcolor[HTML]{C7EDCC} 18. Step 3.3.2, dist[1, 2] = min(dist[1, 2], dist\_k) = min(3, 10) = 3. \\ 
\rowcolor[HTML]{C7EDCC} 19. Step 3.4, increment j and return to Step 3.1. \\ 
\rowcolor[HTML]{C7EDCC} 20. Step 3.1, j = 3, j < 4, so proceed to Step 3.2. \\ 
\rowcolor[HTML]{C7EDCC} 21. Step 3.2, j - k = 3, so proceed to Step 3.2.1. \\ 
\rowcolor[HTML]{C7EDCC} 22. Step 3.2.1, j - i = 2, so proceed to Step 3.3. \\ 
\rowcolor[HTML]{C7EDCC} 23. Step 3.3.1, dist\_k = dist[1, 0] + dist[0, 3] = 4 + 2 = 6. \\ 
\rowcolor[HTML]{C7EDCC} 24. Step 3.3.2, dist[1, 3] = min(dist[1, 3], dist\_k) = min(6, 6) = 6. \\ 
\rowcolor[HTML]{C7EDCC} 25. Step 3.4, increment j and return to Step 3.1. \\ 
\rowcolor[HTML]{C7EDCC} 26. Step 3.1, j = 4, j >= 4, so increment i and return to Step 2.1. \\ 
\rowcolor[HTML]{C7EDCC} 27. Step 2.1, i = 2, i < 4, so proceed to Step 2.2. \\ 
\rowcolor[HTML]{C7EDCC} 28. Step 2.2, i - k = 2, so proceed to Step 3. \\ 
\rowcolor[HTML]{C7EDCC} 29. Step 3, j = 0. \\ 
\rowcolor[HTML]{C7EDCC} 30. Step 3.1, j < 4, so proceed to Step 3.2. \\ 
\rowcolor[HTML]{C7EDCC} 31. Step 3.2, j - k = 0, so increment j and return to Step 3.1. \\ 
\rowcolor[HTML]{C7EDCC} 32. Step 3.1, j = 1, j < 4, so proceed to Step 3.2. \\ 
\rowcolor[HTML]{C7EDCC} 33. Step 3.2, j - k = 1, so proceed to Step 3.2.1. \\ 
\rowcolor[HTML]{C7EDCC} 34. Step 3.2.1, j - i = -1, so proceed to Step 3.3. \\ 
\rowcolor[HTML]{C7EDCC} 35. Step 3.3.1, dist\_k = dist[2, 0] + dist[0, 1] = 6 + 3 = 9. \\ 
\rowcolor[HTML]{C7EDCC} 36. Step 3.3.2, dist[2, 1] = min(dist[2, 1], dist\_k) = min(2, 9) = 2. \\ 
\rowcolor[HTML]{C7EDCC} 37. Step 3.4, increment j and return to Step 3.1. \\ 
\rowcolor[HTML]{C7EDCC} 38. Step 3.1, j = 2, j < 4, so proceed to Step 3.2. \\ 
\rowcolor[HTML]{C7EDCC} 39. Step 3.2, j - k = 2, so proceed to Step 3.2.1. \\ 
\rowcolor[HTML]{C7EDCC} 40. Step 3.2.1, j - i = 0, so increment j and return to Step 3.1. \\ 
\rowcolor[HTML]{C7EDCC} 41. Step 3.1, j = 3, j < 4, so proceed to Step 3.2. \\ 
\rowcolor[HTML]{C7EDCC} 42. Step 3.2, j - k = 3, so proceed to Step 3.2.1. \\ 
\rowcolor[HTML]{C7EDCC} 43. Step 3.2.1, j - i = 1, so proceed to Step 3.3. \\ 
\rowcolor[HTML]{C7EDCC} 44. Step 3.3.1, dist\_k = dist[2, 0] + dist[0, 3] = 6 + 2 = 8. \\ 
\rowcolor[HTML]{C7EDCC} 45. Step 3.3.2, dist[2, 3] = min(dist[2, 3], dist\_k) = min(3, 8) = 3. \\ 
\rowcolor[HTML]{C7EDCC} 46. Step 3.4, increment j and return to Step 3.1. \\ 
\rowcolor[HTML]{C7EDCC} 47. Step 3.1, j = 4, j >= 4, so increment i and return to Step 2.1. \\ 
\rowcolor[HTML]{C7EDCC} 48. Step 2.1, i = 3, i < 4, so proceed to Step 2.2. \\ 
\rowcolor[HTML]{C7EDCC} 49. Step 2.2, i - k = 3, so proceed to Step 3. \\ 
\rowcolor[HTML]{C7EDCC} 50. Step 3, j = 0. \\ 
\rowcolor[HTML]{C7EDCC} 51. Step 3.1, j < 4, so proceed to Step 3.2. \\ 
\rowcolor[HTML]{C7EDCC} 52. Step 3.2, j - k = 0, so increment j and return to Step 3.1. \\ 
\rowcolor[HTML]{C7EDCC} 53. Step 3.1, j = 1, j < 4, so proceed to Step 3.2. \\ 
\rowcolor[HTML]{C7EDCC} 54. Step 3.2, j - k = 1, so proceed to Step 3.2.1. \\ 
\rowcolor[HTML]{C7EDCC} 55. Step 3.2.1, j - i = -2, so proceed to Step 3.3. \\ 
\rowcolor[HTML]{C7EDCC} 56. Step 3.3.1, dist\_k = dist[3, 0] + dist[0, 1] = 5 + 3 = 8. \\ 
\rowcolor[HTML]{C7EDCC} 57. Step 3.3.2, dist[3, 1] = min(dist[3, 1], dist\_k) = min(1, 8) = 1. \\ 
\rowcolor[HTML]{C7EDCC} 58. Step 3.4, increment j and return to Step 3.1. \\ 
\rowcolor[HTML]{C7EDCC} 59. Step 3.1, j = 2, j < 4, so proceed to Step 3.2. \\ 
\rowcolor[HTML]{C7EDCC} 60. Step 3.2, j - k = 2, so proceed to Step 3.2.1. \\ 
\rowcolor[HTML]{C7EDCC} 61. Step 3.2.1, j - i = -1, so proceed to Step 3.3. \\ 
\rowcolor[HTML]{C7EDCC} 62. Step 3.3.1, dist\_k = dist[3, 0] + dist[0, 2] = 5 + 6 = 11. \\ 
\rowcolor[HTML]{C7EDCC} 63. Step 3.3.2, dist[3, 2] = min(dist[3, 2], dist\_k) = min(4, 11) = 4. \\ 
\rowcolor[HTML]{C7EDCC} 64. Step 3.4, increment j and return to Step 3.1. \\ 
\rowcolor[HTML]{C7EDCC} 65. Step 3.1, j = 3, j < 4, so proceed to Step 3.2. \\ 
\rowcolor[HTML]{C7EDCC} 66. Step 3.2, j - k = 3, so proceed to Step 3.2.1. \\ 
\rowcolor[HTML]{C7EDCC} 67. Step 3.2.1, j - i = 0, so increment j and return to Step 3.1. \\ 
\rowcolor[HTML]{C7EDCC} 68. Step 3.1, j = 4, j >= 4, so increment i and return to Step 2.1. \\ 
\rowcolor[HTML]{C7EDCC} 69. Step 2.1, i = 4, i >= 4, so increment k and return to Step 1.1. \\ 
\rowcolor[HTML]{C7EDCC} 70. Step 1.1, k = 1, k < 4, so proceed to Step 2. \\ 
\bottomrule
\end{tabular}
\end{table*}
\begin{table*}[!t]
\small
\begin{tabular}{p{15.5cm}}
\toprule
\rowcolor[HTML]{C7EDCC} 71. Step 2, i = 0. \\ 
\rowcolor[HTML]{C7EDCC} 72. Step 2.1, i < 4, so proceed to Step 2.2. \\ 
\rowcolor[HTML]{C7EDCC} 73. Step 2.2, i - k = -1, so proceed to Step 3. \\ 
\rowcolor[HTML]{C7EDCC} 74. Step 3, j = 0. \\ 
\rowcolor[HTML]{C7EDCC} 75. Step 3.1, j < 4, so proceed to Step 3.2. \\ 
\rowcolor[HTML]{C7EDCC} 76. Step 3.2, j - k = -1, so increment j and return to Step 3.1. \\ 
\rowcolor[HTML]{C7EDCC} 77. Step 3.1, j = 1, j < 4, so proceed to Step 3.2. \\ 
\rowcolor[HTML]{C7EDCC} 78. Step 3.2, j - k = 0, so increment j and return to Step 3.1. \\ 
\rowcolor[HTML]{C7EDCC} 79. Step 3.1, j = 2, j < 4, so proceed to Step 3.2. \\ 
\rowcolor[HTML]{C7EDCC} 80. Step 3.2, j - k = 1, so proceed to Step 3.2.1. \\ 
\rowcolor[HTML]{C7EDCC} 81. Step 3.2.1, j - i = 2, so proceed to Step 3.3. \\ 
\rowcolor[HTML]{C7EDCC} 82. Step 3.3.1, dist\_k = dist[0, 1] + dist[1, 2] = 3 + 3 = 6. \\ 
\rowcolor[HTML]{C7EDCC} 83. Step 3.3.2, dist[0, 2] = min(dist[0, 2], dist\_k) = min(6, 6) = 6. \\ 
\rowcolor[HTML]{C7EDCC} 84. Step 3.4, increment j and return to Step 3.1. \\ 
\rowcolor[HTML]{C7EDCC} 85. Step 3.1, j = 3, j < 4, so proceed to Step 3.2. \\ 
\rowcolor[HTML]{C7EDCC} 86. Step 3.2, j - k = 2, so proceed to Step 3.2.1. \\ 
\rowcolor[HTML]{C7EDCC} 87. Step 3.2.1, j - i = 3, so proceed to Step 3.3. \\ 
\rowcolor[HTML]{C7EDCC} 88. Step 3.3.1, dist\_k = dist[0, 1] + dist[1, 3] = 3 + 6 = 9. \\ 
\rowcolor[HTML]{C7EDCC} 89. Step 3.3.2, dist[0, 3] = min(dist[0, 3], dist\_k) = min(2, 9) = 2. \\ 
\rowcolor[HTML]{C7EDCC} 90. Step 3.4, increment j and return to Step 3.1. \\ 
\rowcolor[HTML]{C7EDCC} 91. Step 3.1, j = 4, j >= 4, so increment i and return to Step 2.1. \\ 
\rowcolor[HTML]{C7EDCC} 92. Step 2.1, i = 1, i < 4, so proceed to Step 2.2. \\ 
\rowcolor[HTML]{C7EDCC} 93. Step 2.2, i - k = 0, so increment i and return to Step 2.1. \\ 
\rowcolor[HTML]{C7EDCC} 94. Step 2.1, i = 2, i < 4, so proceed to Step 2.2. \\ 
\rowcolor[HTML]{C7EDCC} 95. Step 2.2, i - k = 1, so proceed to Step 3. \\ 
\rowcolor[HTML]{C7EDCC} 96. Step 3, j = 0. \\ 
\rowcolor[HTML]{C7EDCC} 97. Step 3.1, j < 4, so proceed to Step 3.2. \\ 
\rowcolor[HTML]{C7EDCC} 98. Step 3.2, j - k = -1, so increment j and return to Step 3.1. \\ 
\rowcolor[HTML]{C7EDCC} 99. Step 3.1, j = 1, j < 4, so proceed to Step 3.2. \\ 
\rowcolor[HTML]{C7EDCC} 100. Step 3.2, j - k = 0, so increment j and return to Step 3.1. \\ 
\rowcolor[HTML]{C7EDCC} 101. Step 3.1, j = 2, j < 4, so proceed to Step 3.2. \\ 
\rowcolor[HTML]{C7EDCC} 102. Step 3.2, j - k = 1, so proceed to Step 3.2.1. \\ 
\rowcolor[HTML]{C7EDCC} 103. Step 3.2.1, j - i = 0, so increment j and return to Step 3.1. \\ 
\rowcolor[HTML]{C7EDCC} 104. Step 3.1, j = 3, j < 4, so proceed to Step 3.2. \\ 
\rowcolor[HTML]{C7EDCC} 105. Step 3.2, j - k = 2, so proceed to Step 3.2.1. \\ 
\rowcolor[HTML]{C7EDCC} 106. Step 3.2.1, j - i = 1, so proceed to Step 3.3. \\ 
\rowcolor[HTML]{C7EDCC} 107. Step 3.3.1, dist\_k = dist[2, 1] + dist[1, 3] = 2 + 6 = 8. \\ 
\rowcolor[HTML]{C7EDCC} 108. Step 3.3.2, dist[2, 3] = min(dist[2, 3], dist\_k) = min(3, 8) = 3. \\ 
\rowcolor[HTML]{C7EDCC} 109. Step 3.4, increment j and return to Step 3.1. \\ 
\rowcolor[HTML]{C7EDCC} 110. Step 3.1, j = 4, j >= 4, so increment i and return to Step 2.1. \\ 
\rowcolor[HTML]{C7EDCC} 111. Step 2.1, i = 3, i < 4, so proceed to Step 2.2. \\ 
\rowcolor[HTML]{C7EDCC} 112. Step 2.2, i - k = 2, so proceed to Step 3. \\ 
\rowcolor[HTML]{C7EDCC} 113. Step 3, j = 0. \\ 
\rowcolor[HTML]{C7EDCC} 114. Step 3.1, j < 4, so proceed to Step 3.2. \\ 
\rowcolor[HTML]{C7EDCC} 115. Step 3.2, j - k = -1, so increment j and return to Step 3.1. \\ 
\rowcolor[HTML]{C7EDCC} 116. Step 3.1, j = 1, j < 4, so proceed to Step 3.2. \\ 
\rowcolor[HTML]{C7EDCC} 117. Step 3.2, j - k = 0, so increment j and return to Step 3.1. \\ 
\rowcolor[HTML]{C7EDCC} 118. Step 3.1, j = 2, j < 4, so proceed to Step 3.2. \\ 
\rowcolor[HTML]{C7EDCC} 119. Step 3.2, j - k = 1, so proceed to Step 3.2.1. \\ 
\rowcolor[HTML]{C7EDCC} 120. Step 3.2.1, j - i = -1, so proceed to Step 3.3. \\ 
\rowcolor[HTML]{C7EDCC} 121. Step 3.3.1, dist\_k = dist[3, 1] + dist[1, 2] = 1 + 3 = 4. \\ 
\rowcolor[HTML]{C7EDCC} 122. Step 3.3.2, dist[3, 2] = min(dist[3, 2], dist\_k) = min(4, 4) = 4. \\ 
\rowcolor[HTML]{C7EDCC} 123. Step 3.4, increment j and return to Step 3.1. \\ 
\rowcolor[HTML]{C7EDCC} 124. Step 3.1, j = 3, j < 4, so proceed to Step 3.2. \\ 
\rowcolor[HTML]{C7EDCC} 125. Step 3.2, j - k = 2, so proceed to Step 3.2.1. \\ 
\rowcolor[HTML]{C7EDCC} 126. Step 3.2.1, j - i = 0, so increment j and return to Step 3.1. \\ 
\rowcolor[HTML]{C7EDCC} 127. Step 3.1, j = 4, j >= 4, so increment i and return to Step 2.1. \\ 
\rowcolor[HTML]{C7EDCC} 128. Step 2.1, i = 4, i >= 4, so increment k and return to Step 1.1. \\ 
\rowcolor[HTML]{C7EDCC} 129. Step 1.1, k = 2, k < 4, so proceed to Step 2. \\ 
\rowcolor[HTML]{C7EDCC} 130. Step 2, i = 0. \\ 
\rowcolor[HTML]{C7EDCC} 131. Step 2.1, i < 4, so proceed to Step 2.2. \\ 
\rowcolor[HTML]{C7EDCC} 132. Step 2.2, i - k = -2, so proceed to Step 3. \\ 
\bottomrule
\end{tabular}
\end{table*}
\clearpage
\begin{table*}[!t]
\small
\begin{tabular}{p{15.5cm}}
\toprule
\rowcolor[HTML]{C7EDCC} 133. Step 3, j = 0. \\ 
\rowcolor[HTML]{C7EDCC} 134. Step 3.1, j < 4, so proceed to Step 3.2. \\ 
\rowcolor[HTML]{C7EDCC} 135. Step 3.2, j - k = -2, so increment j and return to Step 3.1. \\ 
\rowcolor[HTML]{C7EDCC} 136. Step 3.1, j = 1, j < 4, so proceed to Step 3.2. \\ 
\rowcolor[HTML]{C7EDCC} 137. Step 3.2, j - k = -1, so increment j and return to Step 3.1. \\ 
\rowcolor[HTML]{C7EDCC} 138. Step 3.1, j = 2, j < 4, so proceed to Step 3.2. \\ 
\rowcolor[HTML]{C7EDCC} 139. Step 3.2, j - k = 0, so increment j and return to Step 3.1. \\ 
\rowcolor[HTML]{C7EDCC} 140. Step 3.1, j = 3, j < 4, so proceed to Step 3.2. \\ 
\rowcolor[HTML]{C7EDCC} 141. Step 3.2, j - k = 1, so proceed to Step 3.2.1. \\ 
\rowcolor[HTML]{C7EDCC} 142. Step 3.2.1, j - i = 3, so proceed to Step 3.3. \\ 
\rowcolor[HTML]{C7EDCC} 143. Step 3.3.1, dist\_k = dist[0, 2] + dist[2, 3] = 6 + 3 = 9. \\ 
\rowcolor[HTML]{C7EDCC} 144. Step 3.3.2, dist[0, 3] = min(dist[0, 3], dist\_k) = min(2, 9) = 2. \\ 
\rowcolor[HTML]{C7EDCC} 145. Step 3.4, increment j and return to Step 3.1. \\ 
\rowcolor[HTML]{C7EDCC} 146. Step 3.1, j = 4, j >= 4, so increment i and return to Step 2.1. \\ 
\rowcolor[HTML]{C7EDCC} 147. Step 2.1, i = 1, i < 4, so proceed to Step 2.2. \\ 
\rowcolor[HTML]{C7EDCC} 148. Step 2.2, i - k = -1, so proceed to Step 3. \\ 
\rowcolor[HTML]{C7EDCC} 149. Step 3, j = 0. \\ 
\rowcolor[HTML]{C7EDCC} 150. Step 3.1, j < 4, so proceed to Step 3.2. \\ 
\rowcolor[HTML]{C7EDCC} 151. Step 3.2, j - k = -2, so increment j and return to Step 3.1. \\ 
\rowcolor[HTML]{C7EDCC} 152. Step 3.1, j = 1, j < 4, so proceed to Step 3.2. \\ 
\rowcolor[HTML]{C7EDCC} 153. Step 3.2, j - k = -1, so increment j and return to Step 3.1. \\ 
\rowcolor[HTML]{C7EDCC} 154. Step 3.1, j = 2, j < 4, so proceed to Step 3.2. \\ 
\rowcolor[HTML]{C7EDCC} 155. Step 3.2, j - k = 0, so increment j and return to Step 3.1. \\ 
\rowcolor[HTML]{C7EDCC} 156. Step 3.1, j = 3, j < 4, so proceed to Step 3.2. \\ 
\rowcolor[HTML]{C7EDCC} 157. Step 3.2, j - k = 1, so proceed to Step 3.2.1. \\ 
\rowcolor[HTML]{C7EDCC} 158. Step 3.2.1, j - i = 2, so proceed to Step 3.3. \\ 
\rowcolor[HTML]{C7EDCC} 159. Step 3.3.1, dist\_k = dist[1, 2] + dist[2, 3] = 3 + 3 = 6. \\ 
\rowcolor[HTML]{C7EDCC} 160. Step 3.3.2, dist[1, 3] = min(dist[1, 3], dist\_k) = min(6, 6) = 6. \\ 
\rowcolor[HTML]{C7EDCC} 161. Step 3.4, increment j and return to Step 3.1. \\ 
\rowcolor[HTML]{C7EDCC} 162. Step 3.1, j = 4, j >= 4, so increment i and return to Step 2.1. \\ 
\rowcolor[HTML]{C7EDCC} 163. Step 2.1, i = 2, i < 4, so proceed to Step 2.2. \\ 
\rowcolor[HTML]{C7EDCC} 164. Step 2.2, i - k = 0, so increment i and return to Step 2.1. \\ 
\rowcolor[HTML]{C7EDCC} 165. Step 2.1, i = 3, i < 4, so proceed to Step 2.2. \\ 
\rowcolor[HTML]{C7EDCC} 166. Step 2.2, i - k = 1, so proceed to Step 3. \\ 
\rowcolor[HTML]{C7EDCC} 167. Step 3, j = 0. \\ 
\rowcolor[HTML]{C7EDCC} 168. Step 3.1, j < 4, so proceed to Step 3.2. \\ 
\rowcolor[HTML]{C7EDCC} 169. Step 3.2, j - k = -2, so increment j and return to Step 3.1. \\ 
\rowcolor[HTML]{C7EDCC} 170. Step 3.1, j = 1, j < 4, so proceed to Step 3.2. \\ 
\rowcolor[HTML]{C7EDCC} 171. Step 3.2, j - k = -1, so increment j and return to Step 3.1. \\ 
\rowcolor[HTML]{C7EDCC} 172. Step 3.1, j = 2, j < 4, so proceed to Step 3.2. \\ 
\rowcolor[HTML]{C7EDCC} 173. Step 3.2, j - k = 0, so increment j and return to Step 3.1. \\ 
\rowcolor[HTML]{C7EDCC} 174. Step 3.1, j = 3, j < 4, so proceed to Step 3.2. \\ 
\rowcolor[HTML]{C7EDCC} 175. Step 3.2, j - k = 1, so proceed to Step 3.2.1. \\ 
\rowcolor[HTML]{C7EDCC} 176. Step 3.2.1, j - i = 0, so increment j and return to Step 3.1. \\ 
\rowcolor[HTML]{C7EDCC} 177. Step 3.1, j = 4, j >= 4, so increment i and return to Step 2.1. \\ 
\rowcolor[HTML]{C7EDCC} 178. Step 2.1, i = 4, i >= 4, so increment k and return to Step 1.1. \\ 
\rowcolor[HTML]{C7EDCC} 179. Step 1.1, k = 3, k < 4, so proceed to Step 2. \\ 
\rowcolor[HTML]{C7EDCC} 180. Step 2, i = 0. \\ 
\rowcolor[HTML]{C7EDCC} 181. Step 2.1, i < 4, so proceed to Step 2.2. \\ 
\rowcolor[HTML]{C7EDCC} 182. Step 2.2, i - k = -3, so proceed to Step 3. \\ 
\rowcolor[HTML]{C7EDCC} 183. Step 3, j = 0. \\ 
\rowcolor[HTML]{C7EDCC} 184. Step 3.1, j < 4, so proceed to Step 3.2. \\ 
\rowcolor[HTML]{C7EDCC} 185. Step 3.2, j - k = -3, so increment j and return to Step 3.1. \\ 
\rowcolor[HTML]{C7EDCC} 186. Step 3.1, j = 1, j < 4, so proceed to Step 3.2. \\ 
\rowcolor[HTML]{C7EDCC} 187. Step 3.2, j - k = -2, so increment j and return to Step 3.1. \\ 
\rowcolor[HTML]{C7EDCC} 188. Step 3.1, j = 2, j < 4, so proceed to Step 3.2. \\ 
\rowcolor[HTML]{C7EDCC} 189. Step 3.2, j - k = -1, so increment j and return to Step 3.1. \\ 
\rowcolor[HTML]{C7EDCC} 190. Step 3.1, j = 3, j < 4, so proceed to Step 3.2. \\ 
\rowcolor[HTML]{C7EDCC} 191. Step 3.2, j - k = 0, so increment j and return to Step 3.1. \\ 
\bottomrule
\end{tabular}
\end{table*}
\begin{table*}[!t]
\small
\begin{tabular}{p{15.5cm}}
\toprule
\rowcolor[HTML]{C7EDCC} 192. Step 3.1, j = 4, j >= 4, so increment i and return to Step 2.1. \\ 
\rowcolor[HTML]{C7EDCC} 193. Step 2.1, i = 1, i < 4, so proceed to Step 2.2. \\ 
\rowcolor[HTML]{C7EDCC} 194. Step 2.2, i - k = -2, so proceed to Step 3. \\ 
\rowcolor[HTML]{C7EDCC} 195. Step 3, j = 0. \\ 
\rowcolor[HTML]{C7EDCC} 196. Step 3.1, j < 4, so proceed to Step 3.2. \\ 
\rowcolor[HTML]{C7EDCC} 197. Step 3.2, j - k = -3, so increment j and return to Step 3.1. \\ 
\rowcolor[HTML]{C7EDCC} 198. Step 3.1, j = 1, j < 4, so proceed to Step 3.2. \\ 
\rowcolor[HTML]{C7EDCC} 199. Step 3.2, j - k = -2, so increment j and return to Step 3.1. \\ 
\rowcolor[HTML]{C7EDCC} 200. Step 3.1, j = 2, j < 4, so proceed to Step 3.2. \\ 
\rowcolor[HTML]{C7EDCC} 201. Step 3.2, j - k = -1, so increment j and return to Step 3.1. \\ 
\rowcolor[HTML]{C7EDCC} 202. Step 3.1, j = 3, j < 4, so proceed to Step 3.2. \\ 
\rowcolor[HTML]{C7EDCC} 203. Step 3.2, j - k = 0, so increment j and return to Step 3.1. \\ 
\rowcolor[HTML]{C7EDCC} 204. Step 3.1, j = 4, j >= 4, so increment i and return to Step 2.1. \\ 
\rowcolor[HTML]{C7EDCC} 205. Step 2.1, i = 2, i < 4, so proceed to Step 2.2. \\ 
\rowcolor[HTML]{C7EDCC} 206. Step 2.2, i - k = -1, so proceed to Step 3. \\ 
\rowcolor[HTML]{C7EDCC} 207. Step 3, j = 0. \\ 
\rowcolor[HTML]{C7EDCC} 208. Step 3.1, j < 4, so proceed to Step 3.2. \\ 
\rowcolor[HTML]{C7EDCC} 209. Step 3.2, j - k = -3, so increment j and return to Step 3.1. \\ 
\rowcolor[HTML]{C7EDCC} 210. Step 3.1, j = 1, j < 4, so proceed to Step 3.2. \\ 
\rowcolor[HTML]{C7EDCC} 211. Step 3.2, j - k = -2, so increment j and return to Step 3.1. \\ 
\rowcolor[HTML]{C7EDCC} 212. Step 3.1, j = 2, j < 4, so proceed to Step 3.2. \\ 
\rowcolor[HTML]{C7EDCC} 213. Step 3.2, j - k = -1, so increment j and return to Step 3.1. \\ 
\rowcolor[HTML]{C7EDCC} 214. Step 3.1, j = 3, j < 4, so proceed to Step 3.2. \\ 
\rowcolor[HTML]{C7EDCC} 215. Step 3.2, j - k = 0, so increment j and return to Step 3.1. \\ 
\rowcolor[HTML]{C7EDCC} 216. Step 3.1, j = 4, j >= 4, so increment i and return to Step 2.1. \\ 
\rowcolor[HTML]{C7EDCC} 217. Step 2.1, i = 3, i < 4, so proceed to Step 2.2. \\ 
\rowcolor[HTML]{C7EDCC} 218. Step 2.2, i - k = 0, so increment i and return to Step 2.1. \\ 
\rowcolor[HTML]{C7EDCC} 219. Step 2.1, i = 4, i >= 4, so increment k and return to Step 1.1. \\ 
\rowcolor[HTML]{C7EDCC} 220. Step 1.1, k = 4, k >= 4, so proceed to Step 4. \\ 
\rowcolor[HTML]{C7EDCC} 221. Step 4, print the distance matrix dist: \\ 
\rowcolor[HTML]{C7EDCC} dist[0, 0] = 0 \\ 
\rowcolor[HTML]{C7EDCC} dist[0, 1] = 3 \\ 
\rowcolor[HTML]{C7EDCC} dist[0, 2] = 6 \\ 
\rowcolor[HTML]{C7EDCC} dist[0, 3] = 2 \\ 
\rowcolor[HTML]{C7EDCC} dist[1, 0] = 4 \\ 
\rowcolor[HTML]{C7EDCC} dist[1, 1] = 0 \\ 
\rowcolor[HTML]{C7EDCC} dist[1, 2] = 3 \\ 
\rowcolor[HTML]{C7EDCC} dist[1, 3] = 6 \\ 
\rowcolor[HTML]{C7EDCC} dist[2, 0] = 6 \\ 
\rowcolor[HTML]{C7EDCC} dist[2, 1] = 2 \\ 
\rowcolor[HTML]{C7EDCC} dist[2, 2] = 0 \\ 
\rowcolor[HTML]{C7EDCC} dist[2, 3] = 3 \\ 
\rowcolor[HTML]{C7EDCC} dist[3, 0] = 5 \\ 
\rowcolor[HTML]{C7EDCC} dist[3, 1] = 1 \\ 
\rowcolor[HTML]{C7EDCC} dist[3, 2] = 4 \\ 
\rowcolor[HTML]{C7EDCC} dist[3, 3] = 0
\\ \bottomrule
\end{tabular}
\caption{The prompt of floyd warshall and the response of \texttt{gpt-4}.}
\label{table:clrs_mini/floyd_warshall}

\end{table*}

\begin{table*}[h]
\small
\begin{tabular}{p{15cm}}
\toprule
\textbf{Prompt} \\ \midrule
Execute the instructions step by step. Do not jump steps. Do not stop before completion. When reading or updating an array at some index, use concrete values rather than just variables. \\ 
 \\ 
Initial: Given a pattern string pat of length M and a text string txt of length N. \\ 
pat: pat[0] = A, pat[1] = A, pat[2] = A, M = 3. \\ 
txt: txt[0] = C, txt[1] = B, txt[2] = A, txt[3] = B, N = 4. \\ 
Create an empty result list 'res'. \\ 
 \\ 
Step 1: Set the variable upper\_i to N - M + 1. \\ 
 \\ 
Step 2: Set i to 0. \\ 
 \\ 
Step 3: If i < upper\_i, execute the following sequence of sub-steps. Otherwise, go to Step 4. \\ 
\quad Step 3.1: Set j to 0. Set pos to i. \\ 
 \\ 
\quad Step 3.2: If j < M, execute in the sequence of sub-steps. Otherwise, go to Step 3.4. \\ 
\quad \quad a. If txt[pos] and pat[j] are not equal, go to Step 3.4. \\ 
\quad \quad b. Increment j, increment pos, and go back to Step 3.2. \\ 
 \\ 
\quad Step 3.3: Append i to the result list 'res'. Print the list 'res'. \\ 
 \\ 
\quad Step 3.4: Increment i and go back to Step 3. \\ 
 \\ 
Step 4: Print the list 'res'. \\ 
----- \\ 
The computation procedure is as follows, with each line numbered sequentially (e.g., 1. 2. 3. 4. 5. 6. etc.): \\ 
1. Step 1, upper\_i = N - M + 1 = 4 - 3 + 1 = 2. \\ 
\rowcolor[HTML]{C7EDCC} 2. Step 2, i = 0. \\ 
\rowcolor[HTML]{C7EDCC}  \\ 
\rowcolor[HTML]{C7EDCC} 3. Step 3, i < upper\_i (0 < 2), so proceed to Step 3.1. \\ 
\rowcolor[HTML]{C7EDCC}  \\ 
\rowcolor[HTML]{C7EDCC} 4. Step 3.1, j = 0, pos = i = 0. \\ 
\rowcolor[HTML]{C7EDCC}  \\ 
\rowcolor[HTML]{C7EDCC} 5. Step 3.2, j < M (0 < 3), so proceed to sub-steps: \\ 
\rowcolor[HTML]{C7EDCC} \quad a. txt[pos] = txt[0] = C, pat[j] = pat[0] = A, not equal, so go to Step 3.4. \\ 
\rowcolor[HTML]{C7EDCC}  \\ 
\rowcolor[HTML]{C7EDCC} 6. Step 3.4, increment i (i = 1) and go back to Step 3. \\ 
\rowcolor[HTML]{C7EDCC}  \\ 
\rowcolor[HTML]{C7EDCC} 7. Step 3, i < upper\_i (1 < 2), so proceed to Step 3.1. \\ 
\rowcolor[HTML]{C7EDCC}  \\ 
\rowcolor[HTML]{C7EDCC} 8. Step 3.1, j = 0, pos = i = 1. \\ 
\rowcolor[HTML]{C7EDCC}  \\ 
\rowcolor[HTML]{C7EDCC} 9. Step 3.2, j < M (0 < 3), so proceed to sub-steps: \\ 
\rowcolor[HTML]{C7EDCC} \quad a. txt[pos] = txt[1] = B, pat[j] = pat[0] = A, not equal, so go to Step 3.4. \\ 
\rowcolor[HTML]{C7EDCC}  \\ 
\rowcolor[HTML]{C7EDCC} 10. Step 3.4, increment i (i = 2) and go back to Step 3. \\ 
\rowcolor[HTML]{C7EDCC}  \\ 
\rowcolor[HTML]{C7EDCC} 11. Step 3, i is not less than upper\_i (2 is not less than 2), so proceed to Step 4. \\ 
\rowcolor[HTML]{C7EDCC}  \\ 
\rowcolor[HTML]{C7EDCC} 12. Step 4, print the list 'res': res is empty, so the output is an empty list [].
\\ \bottomrule
\end{tabular}
\caption{The prompt of naive string matcher and the response of \texttt{gpt-4}.}
\label{table:clrs_mini/naive_string_matcher}

\end{table*}
\begin{table*}[h]
\small
\begin{tabular}{p{15cm}}
\toprule
\textbf{Prompt} \\ \midrule
Execute the instructions step by step. Do not jump steps. Do not stop before completion. When reading or updating an array at some index, use concrete values rather than just variables. \\ 
 \\ 
Initial: Given a pattern string pat of length M and a text string txt of length N. \\ 
pat: pat[0] = A, pat[1] = A, pat[2] = A, M = 3. \\ 
txt: txt[0] = A, txt[1] = A, txt[2] = A, txt[3] = A, N = 4. \\ 
Create an array lps[] of length M with the value of 0: lps[0] = 0, lps[1] = 0, lps[2] = 0. \\ 
 \\ 
Step 1: Set the variable length to 0. \\ 
 \\ 
Step 2: Set the variable i to 1. \\ 
 \\ 
Step 3: While i < M, execute with the following sequence of sub-steps: \\ 
\quad Step 3.1: If pat[i] and pat[length] are equal: \\ 
\quad \quad a. Increment length. \\ 
\quad \quad b. Set lps[i] equal to length. \\ 
\quad \quad c. Increment i. \\ 
\quad \quad d. go back to Step 3. \\ 
\quad Step 3.2: If pat[i] and pat[length] are not equal and length > 0: \\ 
\quad \quad a. Set length\_prev to length - 1. \\ 
\quad \quad b. Recall the value of lps[length\_prev]. \\ 
\quad \quad c. Set length equal to lps[length\_prev]. \\ 
\quad \quad d. go back to Step 3. \\ 
\quad Step 3.3: If pat[i] and pat[length] are not equal and length is 0: \\ 
\quad \quad a. Set lps[i] to 0. \\ 
\quad \quad b. Increment i. \\ 
\quad \quad c. go back to Step 3. \\ 
 \\ 
Step 4: Print lps array by iterating over its elements, displaying each value as "lps[{i}] = {lps\_i}". \\ 
 \\ 
Step 5: Initialize variables i and j to 0, and an empty result list 'res'. \\ 
 \\ 
Step 6: Compute remain\_txt = N - i, remain\_pat = M - j, remain\_diff = remain\_txt - remain\_pat. \\ 
While remain\_diff >= 0, execute the following sequence of sub-steps: \\ 
\quad Step 6.1: If pat[j] and txt[i] are equal: \\ 
\quad \quad a. Increment i. \\ 
\quad \quad b. Increment j. \\ 
\quad Step 6.2: If j becomes equal to M: \\ 
\quad \quad a. Calculate found\_index as i - j. \\ 
\quad \quad b. Append found\_index to res. \\ 
\quad \quad c. Print the list 'res'. \\ 
\quad \quad d. Set j to lps[j-1]. \\ 
\quad Step 6.3: If (i < N) and (pat[j] and txt[i] are not equal): \\ 
\quad \quad a. If j is non-zero, set j to lps[j-1]. \\ 
\quad \quad b. If j is 0, increment i. \\ 
 \\ 
Step 7: Print the list 'res'. \\ 
----- \\ 
The computation procedure is as follows, with each line numbered sequentially (e.g., 1. 2. 3. 4. 5. 6. etc.): \\ 
1. Step 1, length = 0. \\ 
\rowcolor[HTML]{C7EDCC} 2. Step 2, i = 1. \\ 
\rowcolor[HTML]{C7EDCC} 3. Step 3, i < M (1 < 3), enter the loop. \\ 
\rowcolor[HTML]{C7EDCC} 4. Step 3.1, pat[1] = A, pat[length] = A, they are equal. \\ 
\rowcolor[HTML]{C7EDCC} 5. Increment length, length = 1. \\ 
\rowcolor[HTML]{C7EDCC} 6. Set lps[1] = 1, lps = [0, 1, 0]. \\ 
\rowcolor[HTML]{C7EDCC} 7. Increment i, i = 2. \\ 
\rowcolor[HTML]{C7EDCC} 8. Step 3, i < M (2 < 3), continue the loop. \\ 
\rowcolor[HTML]{C7EDCC} 9. Step 3.1, pat[2] = A, pat[length] = A, they are equal. \\ 
\rowcolor[HTML]{C7EDCC} 10. Increment length, length = 2. \\ 
\rowcolor[HTML]{C7EDCC} 11. Set lps[2] = 2, lps = [0, 1, 2]. \\ 
\rowcolor[HTML]{C7EDCC} 12. Increment i, i = 3. \\ 
\bottomrule
\end{tabular}
\end{table*}
\begin{table*}[!t]
\small
\begin{tabular}{p{15.5cm}}
\toprule
\rowcolor[HTML]{C7EDCC} 13. Step 3, i >= M (3 >= 3), exit the loop. \\ 
\rowcolor[HTML]{C7EDCC} 14. Step 4, print lps array: lps[0] = 0, lps[1] = 1, lps[2] = 2. \\ 
\rowcolor[HTML]{C7EDCC} 15. Step 5, i = 0, j = 0, res = []. \\ 
\rowcolor[HTML]{C7EDCC} 16. Step 6, remain\_txt = 4, remain\_pat = 3, remain\_diff = 1, remain\_diff >= 0, enter the loop. \\ 
\rowcolor[HTML]{C7EDCC} 17. Step 6.1, pat[j] = A, txt[i] = A, they are equal. \\ 
\rowcolor[HTML]{C7EDCC} 18. Increment i, i = 1. \\ 
\rowcolor[HTML]{C7EDCC} 19. Increment j, j = 1. \\ 
\rowcolor[HTML]{C7EDCC} 20. Step 6, remain\_txt = 3, remain\_pat = 2, remain\_diff = 1, remain\_diff >= 0, continue the loop. \\ 
\rowcolor[HTML]{C7EDCC} 21. Step 6.1, pat[j] = A, txt[i] = A, they are equal. \\ 
\rowcolor[HTML]{C7EDCC} 22. Increment i, i = 2. \\ 
\rowcolor[HTML]{C7EDCC} 23. Increment j, j = 2. \\ 
\rowcolor[HTML]{C7EDCC} 24. Step 6, remain\_txt = 2, remain\_pat = 1, remain\_diff = 1, remain\_diff >= 0, continue the loop. \\ 
\rowcolor[HTML]{C7EDCC} 25. Step 6.1, pat[j] = A, txt[i] = A, they are equal. \\ 
\rowcolor[HTML]{C7EDCC} 26. Increment i, i = 3. \\ 
\rowcolor[HTML]{C7EDCC} 27. Increment j, j = 3. \\ 
\rowcolor[HTML]{C7EDCC} 28. Step 6.2, j = M (3 = 3), found\_index = i - j = 0, append 0 to res, res = [0]. \\ 
\rowcolor[HTML]{C7EDCC} 29. Print res: [0]. \\ 
\rowcolor[HTML]{C7EDCC} 30. Set j to lps[j-1], j = lps[2] = 2. \\ 
\rowcolor[HTML]{C7EDCC} 31. Step 6, remain\_txt = 1, remain\_pat = 1, remain\_diff = 0, remain\_diff >= 0, continue the loop. \\ 
\rowcolor[HTML]{C7EDCC} 32. Step 6.1, pat[j] = A, txt[i] = A, they are equal. \\ 
\rowcolor[HTML]{C7EDCC} 33. Increment i, i = 4. \\ 
\rowcolor[HTML]{C7EDCC} 34. Increment j, j = 3. \\ 
\rowcolor[HTML]{C7EDCC} 35. Step 6.2, j = M (3 = 3), found\_index = i - j = 1, append 1 to res, res = [0, 1]. \\ 
\rowcolor[HTML]{C7EDCC} 36. Print res: [0, 1]. \\ 
\rowcolor[HTML]{C7EDCC} 37. Set j to lps[j-1], j = lps[2] = 2. \\ 
\rowcolor[HTML]{C7EDCC} 38. Step 6, remain\_txt = 0, remain\_pat = 1, remain\_diff = -1, remain\_diff < 0, exit the loop. \\ 
\rowcolor[HTML]{C7EDCC} 39. Step 7, print res: [0, 1].
\\ \bottomrule
\end{tabular}
\caption{The prompt of kmp matcher and the response of \texttt{gpt-4}.}
\label{table:clrs_mini/kmp_matcher}
\end{table*}

\begin{table*}[h]
\small
\begin{tabular}{p{15cm}}
\toprule
\textbf{Prompt} \\ \midrule
Execute the instructions step by step. Do not jump steps. Do not stop before completion. \\ 
Initial: Set the input arrays for start times S: S[0] = 0, S[1] = 5, S[2] = 3, S[3] = 3, S[4] = 2, S[5] = 9, S[6] = 3, S[7] = 5, S[8] = 7 and finish times F: F[0] = 7, F[1] = 9, F[2] = 9, F[3] = 9, F[4] = 9, F[5] = 10, F[6] = 11, F[7] = 12, F[8] = 15. \\ 
Set the length of the arrays n: 9. \\ 
Set i = 0. \\ 
Set array A = [0]. \\ 
Step 1: Initialize a loop variable j = 1. \\ 
Step 2: Compare the start time of the current activity S[j] with the finish time of the previously selected activity F[i]. Apply the following rule: \\ 
\quad i. If S[j] >= F[i], the activity can be selected. Add the index j to array A, print it, and update i = j. \\ 
\quad ii. If S[j] < F[i], Continue. \\ 
Step 3: Increment j by 1. If j < n, go back to Step 2. Otherwise, proceed to Step 4. \\ 
Step 4: Print array A. \\ 
----- \\ 
The computation procedure is as follows, with each line numbered sequentially (e.g., 1. 2. 3. 4. 5. 6. etc.): \\ 
1. i = 0, j = 1. \\ 
\rowcolor[HTML]{C7EDCC} 2. S[1] = 5, F[0] = 7, S[1] < F[0], continue. \\ 
\rowcolor[HTML]{C7EDCC} 3. j = 2. \\ 
\rowcolor[HTML]{C7EDCC} 4. S[2] = 3, F[0] = 7, S[2] < F[0], continue. \\ 
\rowcolor[HTML]{C7EDCC} 5. j = 3. \\ 
\rowcolor[HTML]{C7EDCC} 6. S[3] = 3, F[0] = 7, S[3] < F[0], continue. \\ 
\rowcolor[HTML]{C7EDCC} 7. j = 4. \\ 
\rowcolor[HTML]{C7EDCC} 8. S[4] = 2, F[0] = 7, S[4] < F[0], continue. \\ 
\rowcolor[HTML]{C7EDCC} 9. j = 5. \\ 
\rowcolor[HTML]{C7EDCC} 10. S[5] = 9, F[0] = 7, S[5] >= F[0], add j to A: A = [0, 5], i = 5. \\ 
\rowcolor[HTML]{C7EDCC} 11. j = 6. \\ 
\rowcolor[HTML]{C7EDCC} 12. S[6] = 3, F[5] = 10, S[6] < F[5], continue. \\ 
\rowcolor[HTML]{C7EDCC} 13. j = 7. \\ 
\rowcolor[HTML]{C7EDCC} 14. S[7] = 5, F[5] = 10, S[7] < F[5], continue. \\ 
\rowcolor[HTML]{C7EDCC} 15. j = 8. \\ 
\rowcolor[HTML]{C7EDCC} 16. S[8] = 7, F[5] = 10, S[8] < F[5], continue. \\ 
\rowcolor[HTML]{C7EDCC} 17. j = 9, j >= n, exit loop. \\ 
\rowcolor[HTML]{C7EDCC} 18. Print array A: [0, 5].
\\ \bottomrule
\end{tabular}
\caption{The prompt of activity selection and the response of \texttt{gpt-4}.}
\label{table:clrs_mini/activity_selection}

\end{table*}

\clearpage

\begin{table*}[h]
\small
\begin{tabular}{p{15cm}}
\toprule
\textbf{Prompt} \\ \midrule
Execute the following instructions. \\ 
Initial: Set the array p of length N = 4 : p[0] = 5, p[1] = 5, p[2] = 4, p[3] = 4. \\ 
Set the matrix m of size 4 x 4 with all elements initialized to 0. \\ 
Set maxint: 66666666. \\ 
Set L: 2. \\ 
Step 1: Is L < 4? If L is equal to 4, go to Step 5. If L < 4: \\ 
  i. Set i: 1. \\ 
  ii. Compute upper\_i: N - L + 1. \\ 
  iii. Go to Step 2. \\ 
Step 2:Is i < upper\_i? If i is equal to upper\_i, increment L by 1 and go to Step 1. If i < upper\_i: \\ 
  i. Set j: i + L - 1. \\ 
  ii. Set m[i, j] = maxint. \\ 
  iii. Set k: i. \\ 
  iv. Go to Step 3. \\ 
Step 3: Is k - j < 0? If k - j is equal to 0, increment i by 1 and go to Step 2. If k - j < 0: \\ 
  i. What is i-1? What are the values of p[i-1], p[k], and p[j]?  \\ 
  ii. Calculate product = p[i-1] * p[k] * p[j]. \\ 
  iii. Calculate q = m[i, k] + m[k + 1, j] + product. \\ 
  iv. Calculate cmp = m[i, j] - q. \\ 
  v. If cmp is positive, set m[i, j] = q. \\ 
  vi. Increment k by 1. \\ 
  vii. Go to Step 3. \\ 
Step 4: Increment i by 1 and go to Step 2. \\ 
Step 5: Print the value of m[1, 3]. \\ 
----- \\ 
The computation procedure is as follows, with each line numbered sequentially (e.g., 1. 2. 3. 4. 5. 6. etc.): \\ 
1. In Step 1, L = 2, 2 < 4, so we perform the following sub-steps: \\ 
\rowcolor[HTML]{C7EDCC}    i. Set i = 1. \\ 
\rowcolor[HTML]{C7EDCC}    ii. Compute upper\_i = N - L + 1 = 4 - 2 + 1 = 3. \\ 
\rowcolor[HTML]{C7EDCC}    iii. Go to Step 2. \\ 
\rowcolor[HTML]{C7EDCC}  \\ 
\rowcolor[HTML]{C7EDCC} 2. In Step 2, i = 1, 1 < 3, so we perform the following sub-steps: \\ 
\rowcolor[HTML]{C7EDCC}    i. Set j = i + L - 1 = 1 + 2 - 1 = 2. \\ 
\rowcolor[HTML]{C7EDCC}    ii. Set m[1, 2] = maxint = 66666666. \\ 
\rowcolor[HTML]{C7EDCC}    iii. Set k = 1. \\ 
\rowcolor[HTML]{C7EDCC}    iv. Go to Step 3. \\ 
\rowcolor[HTML]{C7EDCC}  \\ 
\rowcolor[HTML]{C7EDCC} 3. In Step 3, k - j = 1 - 2 = -1, -1 < 0, so we perform the following sub-steps: \\ 
\rowcolor[HTML]{C7EDCC}    i. i-1 = 1-1 = 0. The values of p[0], p[1], and p[2] are 5, 5, and 4, respectively. \\ 
\rowcolor[HTML]{C7EDCC}    ii. Calculate product = p[0] * p[1] * p[2] = 5 * 5 * 4 = 100. \\ 
\rowcolor[HTML]{C7EDCC}    iii. Calculate q = m[1, 1] + m[2, 2] + product = 0 + 0 + 100 = 100. \\ 
\rowcolor[HTML]{C7EDCC}    iv. Calculate cmp = m[1, 2] - q = 66666666 - 100 = 66666566. \\ 
\rowcolor[HTML]{C7EDCC}    v. cmp is positive, so set m[1, 2] = q = 100. \\ 
\rowcolor[HTML]{C7EDCC}    vi. Increment k by 1 = 2. \\ 
\rowcolor[HTML]{C7EDCC}    vii. Go to Step 3. \\ 
\rowcolor[HTML]{C7EDCC}  \\ 
\rowcolor[HTML]{C7EDCC} 4. In Step 3, k - j = 2 - 2 = 0, so increment i by 1 = 2 and go to Step 2. \\ 
\rowcolor[HTML]{C7EDCC}  \\ 
\rowcolor[HTML]{C7EDCC} 5. In Step 2, i = 2, 2 < 3, so we perform the following sub-steps: \\ 
\rowcolor[HTML]{C7EDCC}    i. Set j = i + L - 1 = 2 + 2 - 1 = 3. \\ 
\rowcolor[HTML]{C7EDCC}    ii. Set m[2, 3] = maxint = 66666666. \\ 
\rowcolor[HTML]{C7EDCC}    iii. Set k = 2. \\ 
\rowcolor[HTML]{C7EDCC}    iv. Go to Step 3. \\ 
\rowcolor[HTML]{C7EDCC}  \\ 
\rowcolor[HTML]{C7EDCC} 6. In Step 3, k - j = 2 - 3 = -1, -1 < 0, so we perform the following sub-steps: \\ 
\rowcolor[HTML]{C7EDCC}    i. i-1 = 2-1 = 1. The values of p[1], p[2], and p[3] are 5, 4, and 4, respectively. \\ 
\rowcolor[HTML]{C7EDCC}    ii. Calculate product = p[1] * p[2] * p[3] = 5 * 4 * 4 = 80. \\ 
\rowcolor[HTML]{C7EDCC}    iii. Calculate q = m[2, 2] + m[3, 3] + product = 0 + 0 + 80 = 80. \\ 
\rowcolor[HTML]{C7EDCC}    iv. Calculate cmp = m[2, 3] - q = 66666666 - 80 = 66666586. \\ 
\rowcolor[HTML]{C7EDCC}    v. cmp is positive, so set m[2, 3] = q = 80. \\ 
\rowcolor[HTML]{C7EDCC}    vi. Increment k by 1 = 3. \\ 
\bottomrule
\end{tabular}
\end{table*}

\clearpage

\begin{table*}[t]
\small
\begin{tabular}{p{15.5cm}}
\toprule
\rowcolor[HTML]{C7EDCC}    vii. Go to Step 3. \\ 
\rowcolor[HTML]{C7EDCC}  \\ 
\rowcolor[HTML]{C7EDCC} 7. In Step 3, k - j = 3 - 3 = 0, so increment i by 1 = 3 and go to Step 2. \\ 
\rowcolor[HTML]{C7EDCC}  \\ 
\rowcolor[HTML]{C7EDCC} 8. In Step 2, i = 3, 3 = 3, so increment L by 1 = 3 and go to Step 1. \\ 
\rowcolor[HTML]{C7EDCC}  \\ 
\rowcolor[HTML]{C7EDCC} 9. In Step 1, L = 3, 3 < 4, so we perform the following sub-steps: \\ 
\rowcolor[HTML]{C7EDCC}    i. Set i = 1. \\ 
\rowcolor[HTML]{C7EDCC}    ii. Compute upper\_i = N - L + 1 = 4 - 3 + 1 = 2. \\ 
\rowcolor[HTML]{C7EDCC}    iii. Go to Step 2. \\ 
\rowcolor[HTML]{C7EDCC}  \\ 
\rowcolor[HTML]{C7EDCC} 10. In Step 2, i = 1, 1 < 2, so we perform the following sub-steps: \\ 
\rowcolor[HTML]{C7EDCC} \quad i. Set j = i + L - 1 = 1 + 3 - 1 = 3. \\ 
\rowcolor[HTML]{C7EDCC} \quad ii. Set m[1, 3] = maxint = 66666666. \\ 
\rowcolor[HTML]{C7EDCC} \quad iii. Set k = 1. \\ 
\rowcolor[HTML]{C7EDCC} \quad iv. Go to Step 3. \\ 
\rowcolor[HTML]{C7EDCC}  \\ 
\rowcolor[HTML]{C7EDCC} 11. In Step 3, k - j = 1 - 3 = -2, -2 < 0, so we perform the following sub-steps: \\ 
\rowcolor[HTML]{C7EDCC} \quad i. i-1 = 1-1 = 0. The values of p[0], p[1], and p[3] are 5, 5, and 4, respectively. \\ 
\rowcolor[HTML]{C7EDCC} \quad ii. Calculate product = p[0] * p[1] * p[3] = 5 * 5 * 4 = 100. \\ 
\rowcolor[HTML]{C7EDCC} \quad iii. Calculate q = m[1, 1] + m[2, 3] + product = 0 + 80 + 100 = 180. \\ 
\rowcolor[HTML]{C7EDCC} \quad iv. Calculate cmp = m[1, 3] - q = 66666666 - 180 = 66666486. \\ 
\rowcolor[HTML]{C7EDCC} \quad v. cmp is positive, so set m[1, 3] = q = 180. \\ 
\rowcolor[HTML]{C7EDCC} \quad vi. Increment k by 1 = 2. \\ 
\rowcolor[HTML]{C7EDCC} \quad vii. Go to Step 3. \\ 
\rowcolor[HTML]{C7EDCC}  \\ 
\rowcolor[HTML]{C7EDCC} 12. In Step 3, k - j = 2 - 3 = -1, -1 < 0, so we perform the following sub-steps: \\ 
\rowcolor[HTML]{C7EDCC} \quad i. i-1 = 1-1 = 0. The values of p[0], p[2], and p[3] are 5, 4, and 4, respectively. \\ 
\rowcolor[HTML]{C7EDCC} \quad ii. Calculate product = p[0] * p[2] * p[3] = 5 * 4 * 4 = 80. \\ 
\rowcolor[HTML]{C7EDCC} \quad iii. Calculate q = m[1, 2] + m[3, 3] + product = 100 + 0 + 80 = 180. \\ 
\rowcolor[HTML]{C7EDCC} \quad iv. Calculate cmp = m[1, 3] - q = 180 - 180 = 0. \\ 
\rowcolor[HTML]{C7EDCC} \quad v. cmp is not positive, so do not update m[1, 3]. \\ 
\rowcolor[HTML]{C7EDCC} \quad vi. Increment k by 1 = 3. \\ 
\rowcolor[HTML]{C7EDCC} \quad vii. Go to Step 3. \\ 
\rowcolor[HTML]{C7EDCC}  \\ 
\rowcolor[HTML]{C7EDCC} 13. In Step 3, k - j = 3 - 3 = 0, so increment i by 1 = 2 and go to Step 2. \\ 
\rowcolor[HTML]{C7EDCC}  \\ 
\rowcolor[HTML]{C7EDCC} 14. In Step 2, i = 2, 2 = 2, so increment L by 1 = 4 and go to Step 1. \\ 
\rowcolor[HTML]{C7EDCC}  \\ 
\rowcolor[HTML]{C7EDCC} 15. In Step 1, L = 4, 4 = 4, so go to Step 5. \\ 
\rowcolor[HTML]{C7EDCC}  \\ 
\rowcolor[HTML]{C7EDCC} 16. In Step 5, print the value of m[1, 3] = 180. \\ 
\rowcolor[HTML]{C7EDCC}  \\ 
\rowcolor[HTML]{C7EDCC} The final result is 180.
\\ \bottomrule
\end{tabular}
\caption{The prompt of matrix chain multiplication and the response of \texttt{gpt-4}.}
\label{table:clrs_mini/matrix_chain_multiplication}

\end{table*}

\clearpage
\newpage

\begin{table*}[h]
\small
\begin{tabular}{p{15cm}}
\toprule
\textbf{Prompt} \\ \midrule
Execute the instructions step by step. Do not jump steps. Do not stop before completion. When reading or updating an array at some index, use concrete values rather than just variables. \\ 
 \\ 
Initial: Given two lists of numbers X and Y with a size of n, where X represents the independent variables and Y represents the dependent variables.  \\ 
X: X[0] = 15, X[1] = 25, X[2] = 30, X[3] = 45, X[4] = 55, X[5] = 65, X[6] = 75, X[7] = 85, X[8] = 95, n = 9. \\ 
Y: Y[0] = 55, Y[1] = 82, Y[2] = 98, Y[3] = 142, Y[4] = 177, Y[5] = 201, Y[6] = 235, Y[7] = 267, Y[8] = 293. \\ 
 \\ 
Step 1: Compute the value of b for the regression line: \\ 
\quad Step 1.1: Initialize a variable `sx` to 0. Iterate over all elements of X to compute the sum and store the result in `sx`. \\ 
\quad Step 1.2: Initialize a variable `sy` to 0. Iterate over all elements of Y to compute the sum and store the result in `sy`. \\ 
\quad Step 1.3: Initialize two variables `sxsy` and `sx2` both to 0. Iterate over all elements of x and y: \\ 
\quad \quad a. For each iteration, add the product of the current elements of x and y to `sxsy`. \\ 
\quad \quad b. Add the square of the current element of x to `sx2`. \\ 
\quad Step 1.4: Calculate the numerator: \\ 
\quad \quad a. Multiply n with `sxsy` and store the result. \\ 
\quad \quad b. Multiply `sx` with `sy` and store the result in `sx\_times\_sy`. \\ 
\quad \quad c. Subtract `sx\_times\_sy` from the previously computed result to get the final numerator. \\ 
\quad Step 1.5: Calculate the denominator: \\ 
\quad \quad a. Multiply n with `sx2` and store the result. \\ 
\quad \quad b. Square `sx` and store the result in `sx\_squared`. \\ 
\quad \quad c. Subtract `sx\_squared` from the previously computed result to get the final denominator. \\ 
\quad Step 1.6: Divide the numerator by the denominator to compute b. \\ 
 \\ 
Step 2: Determine the regression line parameters: \\ 
\quad Step 2.1: Compute the mean of X and Y: \\ 
\quad \quad a. Sum all the values in X and divide by n to get `meanX`. \\ 
\quad \quad b. Sum all the values in Y and divide by n to get `meanY`. \\ 
\quad Step 2.2: Compute the value of a: \\ 
\quad \quad a. Multiply b with `meanX` and subtract the result from `meanY` to get the value of a. \\ 
 \\ 
Step 3: Print a and b as "a, b = (%.2f, %.2f)". \\ 
----- \\ 
The computation procedure is as follows, with each line numbered sequentially (e.g., 1. 2. 3. 4. 5. 6. etc.): \\ 
1. Step 1.1, initialize `sx` to 0. \\ 
\rowcolor[HTML]{EDCCC7} 2. Compute the sum of X: 15 + 25 + 30 + 45 + 55 + 65 + 75 + 85 + 95 = 490. \\ 
\rowcolor[HTML]{EDCCC7} 3. Set `sx` to 490. \\ 
\rowcolor[HTML]{EDCCC7} 4. Step 1.2, initialize `sy` to 0. \\ 
\rowcolor[HTML]{EDCCC7} 5. Compute the sum of Y: 55 + 82 + 98 + 142 + 177 + 201 + 235 + 267 + 293 = 1550. \\ 
\rowcolor[HTML]{EDCCC7} 6. Set `sy` to 1550. \\ 
\rowcolor[HTML]{EDCCC7} 7. Step 1.3, initialize `sxsy` and `sx2` both to 0. \\ 
\rowcolor[HTML]{EDDFC7} 8. Compute the sum of the product of X and Y: (15 * 55) + (25 * 82) + (30 * 98) + (45 * 142) + (55 * 177) + (65 * 201) + (75 * 235) + (85 * 267) + (95 * 293) = 104965. \\ 
\rowcolor[HTML]{EDCCC7} 9. Set `sxsy` to 104965. \\ 
\rowcolor[HTML]{EDCCC7} 10. Compute the sum of the square of X: (15\^{}2) + (25\^{}2) + (30\^{}2) + (45\^{}2) + (55\^{}2) + (65\^{}2) + (75\^{}2) + (85\^{}2) + (95\^{}2) = 33875. \\ 
\rowcolor[HTML]{EDCCC7} 11. Set `sx2` to 33875. \\ 
\rowcolor[HTML]{EDCCC7} 12. Step 1.4, calculate the numerator: (9 * 104965) - (490 * 1550) = 94485. \\ 
\rowcolor[HTML]{EDCCC7} 13. Step 1.5, calculate the denominator: (9 * 33875) - (490\^{}2) = 3045. \\ 
\rowcolor[HTML]{EDCCC7} 14. Step 1.6, compute b: 94485 / 3045 = 31.03. \\ 
\rowcolor[HTML]{EDCCC7} 15. Step 2.1, compute `meanX`: 490 / 9 = 54.44. \\ 
\rowcolor[HTML]{EDCCC7} 16. Compute `meanY`: 1550 / 9 = 172.22. \\ 
\rowcolor[HTML]{EDCCC7} 17. Step 2.2, compute a: (31.03 * 54.44) - 172.22 = -1460.84. \\ 
\rowcolor[HTML]{EDCCC7} 18. Step 3, print a and b: "a, b = (-1460.84, 31.03)".
\\ \bottomrule
\end{tabular}
\caption{The prompt of least square regression and the response of \texttt{gpt-4}. The correct answer shall be $(7.98, 3.02)$, and the first error is at line 8.}
\label{table:clrs_mini_numeric/least_square_regression}

\end{table*}

\begin{table*}[]
\small
\begin{tabular}{p{15cm}}
\toprule
\textbf{Prompt} \\ \midrule
Execute the instructions step by step. Do not jump steps. Do not stop before completion. When reading or updating an array at some index, use concrete values rather than just variables. \\ 
 \\ 
Initial: Given a list of complex numbers x whose length is a power of 2, compute its discrete Fourier transform (DFT) using the Cooley-Tukey algorithm. \\ 
x: [1, 9, 4, 8, 9, 2, 4, 1]. \\ 
 \\ 
Define the function "FFT" with the argument `x`, `recursive\_level` and `next\_step`. \\ 
 \\ 
Step 1: Call the "FFT" Function with the arguments `x`, `recursive\_level\_x`=1, `next\_step\_x`="". \\ 
Inside the "FFT" function, follow these steps: \\ 
 \\ 
Step 2: Calculate the length of the list `x` and assign it to the variable `N\_x`. \\ 
 \\ 
Step 3: Check the value of `N\_x`: \\ 
\quad - If N\_x equals 1, compute `last\_level` = recursive\_level\_x-1, return x, and go back to `last\_level` at `next\_step\_x`. \\ 
\quad - Otherwise, proceed to Step 4. \\ 
 \\ 
Step 4: Recursively calculate the discrete Fourier transform for the even and odd indexed elements of `x`: \\ 
Step 4.1: For even indices: List x\_even = x[::2], go to Step 1, call FFT(x\_even, recursive\_level\_x\_even=recursive\_level\_x+1, next\_step\_x\_even="Step 4.2"), name the result T\_x\_even. \\ 
Step 4.2: For odd indices: List x\_odd = x[1::2], go to Step 1, call FFT(x\_odd, recursive\_level\_x\_odd=recursive\_level\_x+1, next\_step\_x\_odd="Step 5"), name the result T\_x\_odd. \\ 
 \\ 
Step 5: Compute half of `N\_x`, which is `N\_x\_half` = int(N\_x/2). \\ 
 \\ 
Step 6: Compute the numeric value of factor: \\ 
 \\ 
Step 6.1. Initialize an empty list for `factor`. \\ 
 \\ 
Step 6.2. For each integer `k` starting from 0 up to one less than `N\_x` (i.e., 0, 1, ..., N\_x-1), do the following steps: \\ 
 \\ 
Step 6.2.1. Compute the angle for the exponential:  \\ 
   - theta\_k = -2 * pi * k / N\_x \\ 
 \\ 
Step 6.2.2. Compute the complex exponential for the angle: \\ 
   - exp\_value = e\^{}(i * theta\_k) \\ 
\quad \\ 
Step 6.2.3. Append the computed `exp\_value` to the `factor` list. \\ 
 \\ 
Step 7: Calculate `T\_x`, by using the concatenation of two arrays, and present the numeric value. \\ 
\quad - first\_half = T\_x\_even + factor[:N\_x\_half] * T\_x\_odd. \\ 
\quad - second\_half = T\_x\_even + factor[N\_x\_half:] * T\_x\_odd. \\ 
\quad - T\_x = first\_half + second\_half. \\ 
 \\ 
Step 8: Compute `last\_level` = recursive\_level\_x-1, return the list T\_x, and go back to `last\_level` at `next\_step\_x`, unless the current recursive\_level\_x is 1. \\ 
 \\ 
----- \\ 
Perform the computation procedure step by step, adhering strictly to the outlined instructions.  \\ 
Each line should be numbered sequentially with the recursion level (e.g., "level=1, 1.", "level=1, 2.", "level=2, 1."). \\ 
 \\ 
level=1, 1. Step 1, Call FFT(x=[1, 9, 4, 8, 9, 2, 4, 1], recursive\_level=1, next\_step\_x=''). \\
\bottomrule
\end{tabular}
\end{table*}

\clearpage

\begin{table*}[t]
\small
\begin{tabular}{p{15.5cm}}
\toprule
\rowcolor[HTML]{EDCCC7} level=1, 2. Step 2, Calculate the length of the list x: N\_x = 8. \\ 
\rowcolor[HTML]{EDCCC7}  \\ 
\rowcolor[HTML]{EDCCC7} level=1, 3. Step 3, Check the value of N\_x: N\_x = 8, proceed to Step 4. \\ 
\rowcolor[HTML]{EDCCC7}  \\ 
\rowcolor[HTML]{EDCCC7} level=1, 4.1. Step 4.1, For even indices: x\_even = [1, 4, 9, 4], call FFT(x\_even, recursive\_level\_x\_even=2, next\_step\_x\_even="Step 4.2"). \\ 
\rowcolor[HTML]{EDCCC7}  \\ 
\rowcolor[HTML]{EDCCC7} level=2, 1. Step 1, Call FFT(x=[1, 4, 9, 4], recursive\_level=2, next\_step\_x='Step 4.2'). \\ 
\rowcolor[HTML]{EDCCC7}  \\ 
\rowcolor[HTML]{EDCCC7} level=2, 2. Step 2, Calculate the length of the list x: N\_x = 4. \\ 
\rowcolor[HTML]{EDCCC7}  \\ 
\rowcolor[HTML]{EDCCC7} level=2, 3. Step 3, Check the value of N\_x: N\_x = 4, proceed to Step 4. \\ 
\rowcolor[HTML]{EDCCC7}  \\ 
\rowcolor[HTML]{EDCCC7} level=2, 4.1. Step 4.1, For even indices: x\_even = [1, 9], call FFT(x\_even, recursive\_level\_x\_even=3, next\_step\_x\_even="Step 4.2"). \\ 
\rowcolor[HTML]{EDCCC7}  \\ 
\rowcolor[HTML]{EDCCC7} level=3, 1. Step 1, Call FFT(x=[1, 9], recursive\_level=3, next\_step\_x='Step 4.2'). \\ 
\rowcolor[HTML]{EDCCC7}  \\ 
\rowcolor[HTML]{EDCCC7} level=3, 2. Step 2, Calculate the length of the list x: N\_x = 2. \\ 
\rowcolor[HTML]{EDCCC7}  \\ 
\rowcolor[HTML]{EDCCC7} level=3, 3. Step 3, Check the value of N\_x: N\_x = 2, proceed to Step 4. \\ 
\rowcolor[HTML]{EDCCC7}  \\ 
\rowcolor[HTML]{EDCCC7} level=3, 4.1. Step 4.1, For even indices: x\_even = [1], call FFT(x\_even, recursive\_level\_x\_even=4, next\_step\_x\_even="Step 4.2"). \\ 
\rowcolor[HTML]{EDCCC7}  \\ 
\rowcolor[HTML]{EDCCC7} level=4, 1. Step 1, Call FFT(x=[1], recursive\_level=4, next\_step\_x='Step 4.2'). \\ 
\rowcolor[HTML]{EDCCC7}  \\ 
\rowcolor[HTML]{EDCCC7} level=4, 2. Step 2, Calculate the length of the list x: N\_x = 1. \\ 
\rowcolor[HTML]{EDCCC7}  \\ 
\rowcolor[HTML]{EDCCC7} level=4, 3. Step 3, Check the value of N\_x: N\_x = 1, compute last\_level = 3, return x = [1], and go back to last\_level at next\_step\_x = 'Step 4.2'. \\ 
\rowcolor[HTML]{EDCCC7}  \\ 
\rowcolor[HTML]{EDCCC7} level=3, 4.2. Step 4.2, For odd indices: x\_odd = [9], call FFT(x\_odd, recursive\_level\_x\_odd=4, next\_step\_x\_odd="Step 5"). \\ 
\rowcolor[HTML]{EDCCC7}  \\ 
\rowcolor[HTML]{EDCCC7} level=4, 1. Step 1, Call FFT(x=[9], recursive\_level=4, next\_step\_x='Step 5'). \\ 
\rowcolor[HTML]{EDCCC7}  \\ 
\rowcolor[HTML]{EDCCC7} level=4, 2. Step 2, Calculate the length of the list x: N\_x = 1. \\ 
\rowcolor[HTML]{EDCCC7}  \\ 
\rowcolor[HTML]{EDCCC7} level=4, 3. Step 3, Check the value of N\_x: N\_x = 1, compute last\_level = 3, return x = [9], and go back to last\_level at next\_step\_x = 'Step 5'. \\ 
\rowcolor[HTML]{EDCCC7}  \\ 
\rowcolor[HTML]{EDCCC7} level=3, 5. Step 5, Compute half of N\_x: N\_x\_half = int(2/2) = 1. \\ 
\rowcolor[HTML]{EDCCC7}  \\ 
\rowcolor[HTML]{EDCCC7} level=3, 6.1. Step 6.1, Initialize an empty list for factor: factor = []. \\ 
\rowcolor[HTML]{EDCCC7}  \\ 
\rowcolor[HTML]{EDCCC7} level=3, 6.2. Step 6.2, For each integer k starting from 0 up to one less than N\_x (i.e., 0, 1, ..., N\_x-1): \\ 
\rowcolor[HTML]{EDCCC7}  \\ 
\rowcolor[HTML]{EDCCC7} level=3, 6.2.1. Step 6.2.1, Compute the angle for the exponential: theta\_0 = -2 * pi * 0 / 2 = 0. \\ 
\rowcolor[HTML]{EDCCC7}  \\ 
\rowcolor[HTML]{EDCCC7} level=3, 6.2.2. Step 6.2.2, Compute the complex exponential for the angle: exp\_value = e\^{}(i * 0) = 1. \\ 
\rowcolor[HTML]{EDCCC7}  \\ 
\rowcolor[HTML]{EDCCC7} level=3, 6.2.3. Step 6.2.3, Append the computed exp\_value to the factor list: factor = [1]. \\ 
\rowcolor[HTML]{EDCCC7}  \\ 
\rowcolor[HTML]{EDCCC7} level=3, 6.2.1. Step 6.2.1, Compute the angle for the exponential: theta\_1 = -2 * pi * 1 / 2 = -pi. \\ 
\rowcolor[HTML]{EDCCC7}  \\ 
\rowcolor[HTML]{EDCCC7} level=3, 6.2.2. Step 6.2.2, Compute the complex exponential for the angle: exp\_value = e\^{}(i * -pi) = -1. \\ 
\rowcolor[HTML]{EDCCC7}  \\ 
\rowcolor[HTML]{EDCCC7} level=3, 6.2.3. Step 6.2.3, Append the computed exp\_value to the factor list: factor = [1, -1]. \\ 
\rowcolor[HTML]{EDCCC7}  \\ 
\rowcolor[HTML]{EDCCC7} level=3, 7. Step 7, Calculate T\_x: first\_half = [1] + [1] * [9] = [1 + 9] = [10], second\_half = [1] + [-1] * [9] = [1 - 9] = [-8], T\_x = [10, -8]. \\ 
\rowcolor[HTML]{EDCCC7}  \\ 
\bottomrule
\end{tabular}
\end{table*}

\clearpage

\begin{table*}[t]
\small
\begin{tabular}{p{15.5cm}}
\toprule
\rowcolor[HTML]{EDCCC7} level=3, 8. Step 8, Compute last\_level = 2, return T\_x = [10, -8], and go back to last\_level at next\_step\_x = 'Step 4.2'. \\ 
\rowcolor[HTML]{EDCCC7}  \\ 
\rowcolor[HTML]{EDDFC7} level=2, 4.2. Step 4.2, For odd indices: x\_odd = [4], call FFT(x\_odd, recursive\_level\_x\_odd=3, next\_step\_x\_odd="Step 5"). \\ 
\rowcolor[HTML]{EDCCC7}  \\ 
\rowcolor[HTML]{EDCCC7} level=3, 1. Step 1, Call FFT(x=[4], recursive\_level=3, next\_step\_x='Step 5'). \\ 
\rowcolor[HTML]{EDCCC7}  \\ 
\rowcolor[HTML]{EDCCC7} level=3, 2. Step 2, Calculate the length of the list x: N\_x = 1. \\ 
\rowcolor[HTML]{EDCCC7}  \\ 
\rowcolor[HTML]{EDCCC7} level=3, 3. Step 3, Check the value of N\_x: N\_x = 1, compute last\_level = 2, return x = [4], and go back to last\_level at next\_step\_x = 'Step 5'. \\ 
\rowcolor[HTML]{EDCCC7}  \\ 
\rowcolor[HTML]{EDCCC7} level=2, 5. Step 5, Compute half of N\_x: N\_x\_half = int(4/2) = 2. \\ 
\rowcolor[HTML]{EDCCC7}  \\ 
\rowcolor[HTML]{EDCCC7} level=2, 6.1. Step 6.1, Initialize an empty list for factor: factor = []. \\ 
\rowcolor[HTML]{EDCCC7}  \\ 
\rowcolor[HTML]{EDCCC7} level=2, 6.2. Step 6.2, For each integer k starting from 0 up to one less than N\_x (i.e., 0, 1, ..., N\_x-1): \\ 
\rowcolor[HTML]{EDCCC7}  \\ 
\rowcolor[HTML]{EDCCC7} level=2, 6.2.1. Step 6.2.1, Compute the angle for the exponential: theta\_0 = -2 * pi * 0 / 4 = 0. \\ 
\rowcolor[HTML]{EDCCC7}  \\ 
\rowcolor[HTML]{EDCCC7} level=2, 6.2.2. Step 6.2.2, Compute the complex exponential for the angle: exp\_value = e\^{}(i * 0) = 1. \\ 
\rowcolor[HTML]{EDCCC7}  \\ 
\rowcolor[HTML]{EDCCC7} level=2, 6.2.3. Step 6.2.3, Append the computed exp\_value to the factor list: factor = [1]. \\ 
\rowcolor[HTML]{EDCCC7}  \\ 
\rowcolor[HTML]{EDCCC7} level=2, 6.2.1. Step 6.2.1, Compute the angle for the exponential: theta\_1 = -2 * pi * 1 / 4 = -pi/2. \\ 
\rowcolor[HTML]{EDCCC7}  \\ 
\rowcolor[HTML]{EDCCC7} level=2, 6.2.2. Step 6.2.2, Compute the complex exponential for the angle: exp\_value = e\^{}(i * -pi/2) = -i. \\ 
\rowcolor[HTML]{EDCCC7}  \\ 
\rowcolor[HTML]{EDCCC7} level=2, 6.2.3. Step 6.2.3, Append the computed exp\_value to the factor list: factor = [1, -i]. \\ 
\rowcolor[HTML]{EDCCC7}  \\ 
\rowcolor[HTML]{EDCCC7} level=2, 6.2.1. Step 6.2.1, Compute the angle for the exponential: theta\_2 = -2 * pi * 2 / 4 = -pi. \\ 
\rowcolor[HTML]{EDCCC7}  \\ 
\rowcolor[HTML]{EDCCC7} level=2, 6.2.2. Step 6.2.2, Compute the complex exponential for the angle: exp\_value = e\^{}(i * -pi) = -1. \\ 
\rowcolor[HTML]{EDCCC7}  \\ 
\rowcolor[HTML]{EDCCC7} level=2, 6.2.3. Step 6.2.3, Append the computed exp\_value to the factor list: factor = [1, -i, -1]. \\ 
\rowcolor[HTML]{EDCCC7}  \\ 
\rowcolor[HTML]{EDCCC7} level=2, 6.2.1. Step 6.2.1, Compute the angle for the exponential: theta\_3 = -2 * pi * 3 / 4 = -3*pi/2. \\ 
\rowcolor[HTML]{EDCCC7}  \\ 
\rowcolor[HTML]{EDCCC7} level=2, 6.2.2. Step 6.2.2, Compute the complex exponential for the angle: exp\_value = e\^{}(i * -3*pi/2) = i. \\ 
\rowcolor[HTML]{EDCCC7}  \\ 
\rowcolor[HTML]{EDCCC7} level=2, 6.2.3. Step 6.2.3, Append the computed exp\_value to the factor list: factor = [1, -i, -1, i]. \\ 
\rowcolor[HTML]{EDCCC7}  \\ 
\rowcolor[HTML]{EDCCC7} level=2, 7. Step 7, Calculate T\_x: first\_half = [10, -8] + [1, -i] * [4] = [10 + 4, -8 - 4i] = [14, -8 - 4i], second\_half = [10, -8] + [-1, i] * [4] = [10 - 4, -8 + 4i] = [6, -8 + 4i], T\_x = [14, -8 - 4i, 6, -8 + 4i]. \\ 
\rowcolor[HTML]{EDCCC7}  \\ 
\rowcolor[HTML]{EDCCC7} level=2, 8. Step 8, Compute last\_level = 1, return T\_x = [14, -8 - 4i, 6, -8 + 4i], and go back to last\_level at next\_step\_x = 'Step 4.2'. \\ 
\rowcolor[HTML]{EDCCC7}  \\ 
\rowcolor[HTML]{EDCCC7} level=1, 4.2. Step 4.2, For odd indices: x\_odd = [9, 8, 2, 1], call FFT(x\_odd, recursive\_level\_x\_odd=2, next\_step\_x\_odd="Step 5"). \\ 
\rowcolor[HTML]{EDCCC7}  \\ 
\rowcolor[HTML]{EDCCC7} level=2, 1. Step 1, Call FFT(x=[9, 8, 2, 1], recursive\_level=2, next\_step\_x='Step 5'). \\ 
\rowcolor[HTML]{EDCCC7}  \\ 
\rowcolor[HTML]{EDCCC7} level=2, 2. Step 2, Calculate the length of the list x: N\_x = 4. \\ 
\rowcolor[HTML]{EDCCC7}  \\ 
\rowcolor[HTML]{EDCCC7} level=2, 3. Step 3, Check the value of N\_x: N\_x = 4, proceed to Step 4. \\ 
\rowcolor[HTML]{EDCCC7}  \\ 
\rowcolor[HTML]{EDCCC7} level=2, 4.1. Step 4.1, For even indices: x\_even = [9, 2], call FFT(x\_even, recursive\_level\_x\_even=3, next\_step\_x\_even="Step 4.2"). \\ 
\rowcolor[HTML]{EDCCC7}  \\ 
\rowcolor[HTML]{EDCCC7} level=3, 1. Step 1, Call FFT(x=[9, 2], recursive\_level=3, next\_step\_x='Step 4.2'). \\ 
\rowcolor[HTML]{EDCCC7}  \\ 
\bottomrule
\end{tabular}
\end{table*}

\clearpage

\begin{table*}[t]
\small
\begin{tabular}{p{15.5cm}}
\toprule
\rowcolor[HTML]{EDCCC7} level=3, 2. Step 2, Calculate the length of the list x: N\_x = 2. \\ 
\rowcolor[HTML]{EDCCC7}  \\ 
\rowcolor[HTML]{EDCCC7} level=3, 3. Step 3, Check the value of N\_x: N\_x = 2, proceed to Step 4. \\ 
\rowcolor[HTML]{EDCCC7}  \\ 
\rowcolor[HTML]{EDCCC7} level=3, 4.1. Step 4.1, For even indices: x\_even = [9], call FFT(x\_even, recursive\_level\_x\_even=4, next\_step\_x\_even="Step 4.2"). \\ 
\rowcolor[HTML]{EDCCC7}  \\ 
\rowcolor[HTML]{EDCCC7} level=4, 1. Step 1, Call FFT(x=[9], recursive\_level=4, next\_step\_x='Step 4.2'). \\ 
\rowcolor[HTML]{EDCCC7}  \\ 
\rowcolor[HTML]{EDCCC7} level=4, 2. Step 2, Calculate the length of the list x: N\_x = 1. \\ 
\rowcolor[HTML]{EDCCC7}  \\ 
\rowcolor[HTML]{EDCCC7} level=4, 3. Step 3, Check the value of N\_x: N\_x = 1, compute last\_level = 3, return x = [9], and go back to last\_level at next\_step\_x = 'Step 4.2'. \\ 
\rowcolor[HTML]{EDCCC7}  \\ 
\rowcolor[HTML]{EDCCC7} level=3, 4.2. Step 4.2, For odd indices: x\_odd = [2], call FFT(x\_odd, recursive\_level\_x\_odd=4, next\_step\_x\_odd="Step 5"). \\ 
\rowcolor[HTML]{EDCCC7}  \\ 
\rowcolor[HTML]{EDCCC7} level=4, 1. Step 1, Call FFT(x=[2], recursive\_level=4, next\_step\_x='Step 5'). \\ 
\rowcolor[HTML]{EDCCC7}  \\ 
\rowcolor[HTML]{EDCCC7} level=4, 2. Step 2, Calculate the length of the list x: N\_x = 1. \\ 
\rowcolor[HTML]{EDCCC7}  \\ 
\rowcolor[HTML]{EDCCC7} level=4, 3. Step 3, Check the value of N\_x: N\_x = 1, compute last\_level = 3, return x = [2], and go back to last\_level at next\_step\_x = 'Step 5'. \\ 
\rowcolor[HTML]{EDCCC7}  \\ 
\rowcolor[HTML]{EDCCC7} level=3, 5. Step 5, Compute half of N\_x: N\_x\_half = int(2/2) = 1. \\ 
\rowcolor[HTML]{EDCCC7}  \\ 
\rowcolor[HTML]{EDCCC7} level=3, 6.1. Step 6.1, Initialize an empty list for factor: factor = []. \\ 
\rowcolor[HTML]{EDCCC7}  \\ 
\rowcolor[HTML]{EDCCC7} level=3, 6.2. Step 6.2, For each integer k starting from 0 up to one less than N\_x (i.e., 0, 1, ..., N\_x-1): \\ 
\rowcolor[HTML]{EDCCC7}  \\ 
\rowcolor[HTML]{EDCCC7} level=3, 6.2.1. Step 6.2.1, Compute the angle for the exponential: theta\_0 = -2 * pi * 0 / 2 = 0. \\ 
\rowcolor[HTML]{EDCCC7}  \\ 
\rowcolor[HTML]{EDCCC7} level=3, 6.2.2. Step 6.2.2, Compute the complex exponential for the angle: exp\_value = e\^{}(i * 0) = 1. \\ 
\rowcolor[HTML]{EDCCC7}  \\ 
\rowcolor[HTML]{EDCCC7} level=3, 6.2.3. Step 6.2.3, Append the computed exp\_value to the factor list: factor = [1]. \\ 
\rowcolor[HTML]{EDCCC7}  \\ 
\rowcolor[HTML]{EDCCC7} level=3, 6.2.1. Step 6.2.1, Compute the angle for the exponential: theta\_1 = -2 * pi * 1 / 2 = -pi. \\ 
\rowcolor[HTML]{EDCCC7}  \\ 
\rowcolor[HTML]{EDCCC7} level=3, 6.2.2. Step 6.2.2, Compute the complex exponential for the angle: exp\_value = e\^{}(i * -pi) = -1. \\ 
\rowcolor[HTML]{EDCCC7}  \\ 
\rowcolor[HTML]{EDCCC7} level=3, 6.2.3. Step 6.2.3, Append the computed exp\_value to the factor list: factor = [1, -1]. \\ 
\rowcolor[HTML]{EDCCC7}  \\ 
\rowcolor[HTML]{EDCCC7} level=3, 7. Step 7, Calculate T\_x: first\_half = [9] + [1] * [2] = [9 + 2] = [11], second\_half = [9] + [-1] * [2] = [9 - 2] = [7], T\_x = [11, 7]. \\ 
\rowcolor[HTML]{EDCCC7}  \\ 
\rowcolor[HTML]{EDCCC7} level=3, 8. Step 8, Compute last\_level = 2, return T\_x = [11, 7], and go back to last\_level at next\_step\_x = 'Step 4.2'. \\ 
\rowcolor[HTML]{EDCCC7}  \\ 
\rowcolor[HTML]{EDCCC7} level=2, 4.2. Step 4.2, For odd indices: x\_odd = [8, 1], call FFT(x\_odd, recursive\_level\_x\_odd=3, next\_step\_x\_odd="Step 5"). \\ 
\rowcolor[HTML]{EDCCC7}  \\ 
\rowcolor[HTML]{EDCCC7} level=3, 1. Step 1, Call FFT(x=[8, 1], recursive\_level=3, next\_step\_x='Step 5'). \\ 
\rowcolor[HTML]{EDCCC7}  \\ 
\rowcolor[HTML]{EDCCC7} level=3, 2. Step 2, Calculate the length of the list x: N\_x = 2. \\ 
\rowcolor[HTML]{EDCCC7}  \\ 
\rowcolor[HTML]{EDCCC7} level=3, 3. Step 3, Check the value of N\_x: N\_x = 2, proceed to Step 4. \\ 
\rowcolor[HTML]{EDCCC7}  \\ 
\rowcolor[HTML]{EDCCC7} level=3, 4.1. Step 4.1, For even indices: x\_even = [8], call FFT(x\_even, recursive\_level\_x\_even=4, next\_step\_x\_even="Step 4.2"). \\ 
\rowcolor[HTML]{EDCCC7}  \\ 
\rowcolor[HTML]{EDCCC7} level=4, 1. Step 1, Call FFT(x=[8], recursive\_level=4, next\_step\_x='Step 4.2'). \\ 
\rowcolor[HTML]{EDCCC7}  \\ 
\rowcolor[HTML]{EDCCC7} level=4, 2. Step 2, Calculate the length of the list x: N\_x = 1. \\ 
\bottomrule
\end{tabular}
\end{table*}

\clearpage

\begin{table*}[t]
\small
\begin{tabular}{p{15.5cm}}
\toprule
\rowcolor[HTML]{EDCCC7}  \\ 
\rowcolor[HTML]{EDCCC7} level=4, 3. Step 3, Check the value of N\_x: N\_x = 1, compute last\_level = 3, return x = [8], and go back to last\_level at next\_step\_x = 'Step 4.2'. \\ 
\rowcolor[HTML]{EDCCC7}  \\ 
\rowcolor[HTML]{EDCCC7} level=3, 4.2. Step 4.2, For odd indices: x\_odd = [1], call FFT(x\_odd, recursive\_level\_x\_odd=4, next\_step\_x\_odd="Step 5"). \\ 
\rowcolor[HTML]{EDCCC7}  \\ 
\rowcolor[HTML]{EDCCC7} level=4, 1. Step 1, Call FFT(x=[1], recursive\_level=4, next\_step\_x='Step 5'). \\ 
\rowcolor[HTML]{EDCCC7}  \\ 
\rowcolor[HTML]{EDCCC7} level=4, 2. Step 2, Calculate the length of the list x: N\_x = 1. \\ 
\rowcolor[HTML]{EDCCC7}  \\ 
\rowcolor[HTML]{EDCCC7} level=4, 3. Step 3, Check the value of N\_x: N\_x = 1, compute last\_level = 3, return x = [1], and go back to last\_level at next\_step\_x = 'Step 5'. \\ 
\rowcolor[HTML]{EDCCC7}  \\ 
\rowcolor[HTML]{EDCCC7} level=3, 5. Step 5, Compute half of N\_x: N\_x\_half = int(2/2) = 1. \\ 
\rowcolor[HTML]{EDCCC7}  \\ 
\rowcolor[HTML]{EDCCC7} level=3, 6.1. Step 6.1, Initialize an empty list for factor: factor = []. \\ 
\rowcolor[HTML]{EDCCC7}  \\ 
\rowcolor[HTML]{EDCCC7} level=3, 6.2. Step 6.2, For each integer k starting from 0 up to one less than N\_x (i.e., 0, 1, ..., N\_x-1): \\ 
\rowcolor[HTML]{EDCCC7}  \\ 
\rowcolor[HTML]{EDCCC7} level=3, 6.2.1. Step 6.2.1, Compute the angle for the exponential: theta\_0 = -2 * pi * 0 / 2 = 0. \\ 
\rowcolor[HTML]{EDCCC7}  \\ 
\rowcolor[HTML]{EDCCC7} level=3, 6.2.2. Step 6.2.2, Compute the complex exponential for the angle: exp\_value = e\^{}(i * 0) = 1. \\ 
\rowcolor[HTML]{EDCCC7}  \\ 
\rowcolor[HTML]{EDCCC7} level=3, 6.2.3. Step 6.2.3, Append the computed exp\_value to the factor list: factor = [1]. \\ 
\rowcolor[HTML]{EDCCC7}  \\ 
\rowcolor[HTML]{EDCCC7} level=3, 6.2.1. Step 6.2.1, Compute the angle for the exponential: theta\_1 = -2 * pi * 1 / 2 = -pi. \\ 
\rowcolor[HTML]{EDCCC7}  \\ 
\rowcolor[HTML]{EDCCC7} level=3, 6.2.2. Step 6.2.2, Compute the complex exponential for the angle: exp\_value = e\^{}(i * -pi) = -1. \\ 
\rowcolor[HTML]{EDCCC7}  \\ 
\rowcolor[HTML]{EDCCC7} level=3, 6.2.3. Step 6.2.3, Append the computed exp\_value to the factor list: factor = [1, -1]. \\ 
\rowcolor[HTML]{EDCCC7}  \\ 
\rowcolor[HTML]{EDCCC7} level=3, 7. Step 7, Calculate T\_x: first\_half = [8] + [1] * [1] = [8 + 1] = [9], second\_half = [8] + [-1] * [1] = [8 - 1] = [7], T\_x = [9, 7]. \\ 
\rowcolor[HTML]{EDCCC7}  \\ 
\rowcolor[HTML]{EDCCC7} level=3, 8. Step 8, Compute last\_level = 2, return T\_x = [9, 7], and go back to last\_level at next\_step\_x = 'Step 5'. \\ 
\rowcolor[HTML]{EDCCC7}  \\ 
\rowcolor[HTML]{EDCCC7} level=2, 5. Step 5, Compute half of N\_x: N\_x\_half = int(4/2) = 2. \\ 
\rowcolor[HTML]{EDCCC7}  \\ 
\rowcolor[HTML]{EDCCC7} level=2, 6.1. Step 6.1, Initialize an empty list for factor: factor = []. \\ 
\rowcolor[HTML]{EDCCC7}  \\ 
\rowcolor[HTML]{EDCCC7} level=2, 6.2. Step 6.2, For each integer k starting from 0 up to one less than N\_x (i.e., 0, 1, ..., N\_x-1): \\ 
\rowcolor[HTML]{EDCCC7}  \\ 
\rowcolor[HTML]{EDCCC7} level=2, 6.2.1. Step 6.2.1, Compute the angle for the exponential: theta\_0 = -2 * pi * 0 / 4 = 0. \\ 
\rowcolor[HTML]{EDCCC7}  \\ 
\rowcolor[HTML]{EDCCC7} level=2, 6.2.2. Step 6.2.2, Compute the complex exponential for the angle: exp\_value = e\^{}(i * 0) = 1. \\ 
\rowcolor[HTML]{EDCCC7}  \\ 
\rowcolor[HTML]{EDCCC7} level=2, 6.2.3. Step 6.2.3, Append the computed exp\_value to the factor list: factor = [1]. \\ 
\rowcolor[HTML]{EDCCC7}  \\ 
\rowcolor[HTML]{EDCCC7} level=2, 6.2.1. Step 6.2.1, Compute the angle for the exponential: theta\_1 = -2 * pi * 1 / 4 = -pi/2. \\ 
\rowcolor[HTML]{EDCCC7}  \\ 
\rowcolor[HTML]{EDCCC7} level=2, 6.2.2. Step 6.2.2, Compute the complex exponential for the angle: exp\_value = e\^{}(i * -pi/2) = -i. \\ 
\rowcolor[HTML]{EDCCC7}  \\ 
\rowcolor[HTML]{EDCCC7} level=2, 6.2.3. Step 6.2.3, Append the computed exp\_value to the factor list: factor = [1, -i]. \\ 
\rowcolor[HTML]{EDCCC7}  \\ 
\rowcolor[HTML]{EDCCC7} level=2, 6.2.1. Step 6.2.1, Compute the angle for the exponential: theta\_2 = -2 * pi * 2 / 4 = -pi. \\ 
\rowcolor[HTML]{EDCCC7}  \\ 
\rowcolor[HTML]{EDCCC7} level=2, 6.2.2. Step 6.2.2, Compute the complex exponential for the angle: exp\_value = e\^{}(i * -pi) = -1. \\ 
\rowcolor[HTML]{EDCCC7}  \\ 
\rowcolor[HTML]{EDCCC7} level=2, 6.2.3. Step 6.2.3, Append the computed exp\_value to the factor list: factor = [1, -i, -1]. \\ 
\rowcolor[HTML]{EDCCC7}  \\ 
\rowcolor[HTML]{EDCCC7} level=2, 6.2.1. Step 6.2.1, Compute the angle for the exponential: theta\_3 = -2 * pi * 3 / 4 = -3*pi/2. \\ 
\rowcolor[HTML]{EDCCC7}  \\ 
\bottomrule
\end{tabular}
\end{table*}

\clearpage

\begin{table*}[t]
\small
\begin{tabular}{p{15.5cm}}
\toprule
\rowcolor[HTML]{EDCCC7} level=2, 6.2.2. Step 6.2.2, Compute the complex exponential for the angle: exp\_value = e\^{}(i * -3*pi/2) = i. \\ 
\rowcolor[HTML]{EDCCC7}  \\ 
\rowcolor[HTML]{EDCCC7} level=2, 6.2.3. Step 6.2.3, Append the computed exp\_value to the factor list: factor = [1, -i, -1, i]. \\ 
\rowcolor[HTML]{EDCCC7}  \\ 
\rowcolor[HTML]{EDCCC7} level=2, 7. Step 7, Calculate T\_x: first\_half = [11, 7] + [1, -i] * [9, 7] = [11 + 9, 7 + 7i] = [20, 7 + 7i], second\_half = [11, 7] + [-1, i] * [9, 7] = [11 - 9, 7 - 7i] = [2, 7 - 7i], T\_x = [20, 7 + 7i, 2, 7 - 7i]. \\ 
\rowcolor[HTML]{EDCCC7}  \\ 
\rowcolor[HTML]{EDCCC7} level=2, 8. Step 8, Compute last\_level = 1, return T\_x = [20, 7 + 7i, 2, 7 - 7i], and go back to last\_level at next\_step\_x = 'Step 5'. \\ 
\rowcolor[HTML]{EDCCC7}  \\ 
\rowcolor[HTML]{EDCCC7} level=1, 5. Step 5, Compute half of N\_x: N\_x\_half = int(8/2) = 4. \\ 
\rowcolor[HTML]{EDCCC7}  \\ 
\rowcolor[HTML]{EDCCC7} level=1, 6.1. Step 6.1, Initialize an empty list for factor: factor = []. \\ 
\rowcolor[HTML]{EDCCC7}  \\ 
\rowcolor[HTML]{EDCCC7} level=1, 6.2. Step 6.2, For each integer k starting from 0 up to one less than N\_x (i.e., 0, 1, ..., N\_x-1): \\ 
\rowcolor[HTML]{EDCCC7}  \\ 
\rowcolor[HTML]{EDCCC7} level=1, 6.2.1. Step 6.2.1, Compute the angle for the exponential: theta\_0 = -2 * pi * 0 / 8 = 0. \\ 
\rowcolor[HTML]{EDCCC7}  \\ 
\rowcolor[HTML]{EDCCC7} level=1, 6.2.2. Step 6.2.2, Compute the complex exponential for the angle: exp\_value = e\^{}(i * 0) = 1. \\ 
\rowcolor[HTML]{EDCCC7}  \\ 
\rowcolor[HTML]{EDCCC7} level=1, 6.2.3. Step 6.2.3, Append the computed exp\_value to the factor list: factor = [1]. \\ 
\rowcolor[HTML]{EDCCC7}  \\ 
\rowcolor[HTML]{EDCCC7} level=1, 6.2.1. Step 6.2.1, Compute the angle for the exponential: theta\_1 = -2 * pi * 1 / 8 = -pi/4. \\ 
\rowcolor[HTML]{EDCCC7}  \\ 
\rowcolor[HTML]{EDCCC7} level=1, 6.2.2. Step 6.2.2, Compute the complex exponential for the angle: exp\_value = e\^{}(i * -pi/4) = 1/sqrt(2) - i/sqrt(2). \\ 
\rowcolor[HTML]{EDCCC7}  \\ 
\rowcolor[HTML]{EDCCC7} level=1, 6.2.3. Step 6.2.3, Append the computed exp\_value to the factor list: factor = [1, 1/sqrt(2) - i/sqrt(2)]. \\ 
\rowcolor[HTML]{EDCCC7}  \\ 
\rowcolor[HTML]{EDCCC7} level=1, 6.2.1. Step 6.2.1, Compute the angle for the exponential: theta\_2 = -2 * pi * 2 / 8 = -pi/2. \\ 
\rowcolor[HTML]{EDCCC7}  \\ 
\rowcolor[HTML]{EDCCC7} level=1, 6.2.2. Step 6.2.2, Compute the complex exponential for the angle: exp\_value = e\^{}(i * -pi/2) = -i. \\ 
\rowcolor[HTML]{EDCCC7}  \\ 
\rowcolor[HTML]{EDCCC7} level=1, 6.2.3. Step 6.2.3, Append the computed exp\_value to the factor list: factor = [1, 1/sqrt(2) - i/sqrt(2), -i]. \\ 
\rowcolor[HTML]{EDCCC7}  \\ 
\rowcolor[HTML]{EDCCC7} level=1, 6.2.1. Step 6.2.1, Compute the angle for the exponential: theta\_3 = -2 * pi * 3 / 8 = -3*pi/4. \\ 
\rowcolor[HTML]{EDCCC7}  \\ 
\rowcolor[HTML]{EDCCC7} level=1, 6.2.2. Step 6.2.2, Compute the complex exponential for the angle: exp\_value = e\^{}(i * -3*pi/4) = -1/sqrt(2) - i/sqrt(2). \\ 
\rowcolor[HTML]{EDCCC7}  \\ 
\rowcolor[HTML]{EDCCC7} level=1, 6.2.3. Step 6.2.3, Append the computed exp\_value to the factor list: factor = [1, 1/sqrt(2) - i/sqrt(2), -i, -1/sqrt(2) - i/sqrt(2)]. \\ 
\rowcolor[HTML]{EDCCC7}  \\ 
\rowcolor[HTML]{EDCCC7} level=1, 6.2.1. Step 6.2.1, Compute the angle for the exponential: theta\_4 = -2 * pi * 4 / 8 = -pi. \\ 
\rowcolor[HTML]{EDCCC7}  \\ 
\rowcolor[HTML]{EDCCC7} level=1, 6.2.2. Step 6.2.2, Compute the complex exponential for the angle: exp\_value = e\^{}(i * -pi) = -1. \\ 
\rowcolor[HTML]{EDCCC7}  \\ 
\rowcolor[HTML]{EDCCC7} level=1, 6.2.3. Step 6.2.3, Append the computed exp\_value to the factor list: factor = [1, 1/sqrt(2) - i/sqrt(2), -i, -1/sqrt(2) - i/sqrt(2), -1]. \\ 
\rowcolor[HTML]{EDCCC7}  \\ 
\rowcolor[HTML]{EDCCC7} level=1, 6.2.1. Step 6.2.1, Compute the angle for the exponential: theta\_5 = -2 * pi * 5 / 8 = -5*pi/4. \\ 
\rowcolor[HTML]{EDCCC7}  \\ 
\rowcolor[HTML]{EDCCC7} level=1, 6.2.2. Step 6.2.2, Compute the complex exponential for the angle: exp\_value = e\^{}(i * -5*pi/4) = -1/sqrt(2) + i/sqrt(2). \\ 
\rowcolor[HTML]{EDCCC7}  \\ 
\rowcolor[HTML]{EDCCC7} level=1, 6.2.3. Step 6.2.3, Append the computed exp\_value to the factor list: factor = [1, 1/sqrt(2) - i/sqrt(2), -i, -1/sqrt(2) - i/sqrt(2), -1, -1/sqrt(2) + i/sqrt(2)]. \\ 
\rowcolor[HTML]{EDCCC7}  \\ 
\rowcolor[HTML]{EDCCC7} level=1, 6.2.1. Step 6.2.1, Compute the angle for the exponential: theta\_6 = -2 * pi * 6 / 8 = -3*pi/2. \\ 
\rowcolor[HTML]{EDCCC7}  \\ 
\rowcolor[HTML]{EDCCC7} level=1, 6.2.2. Step 6.2.2, Compute the complex exponential for the angle: exp\_value = e\^{}(i * -3*pi/2) = i. \\ 
\rowcolor[HTML]{EDCCC7} level=1, 6.2.3. Step 6.2.3, Append the computed exp\_value to the factor list: factor = [1, 1/sqrt(2) - i/sqrt(2), -i, -1/sqrt(2) - i/sqrt(2), -1, -1/sqrt(2) + i/sqrt(2), i]. \\ 
\bottomrule
\end{tabular}
\end{table*}

\clearpage

\begin{table*}[t]
\small
\begin{tabular}{p{15.5cm}}
\toprule
\rowcolor[HTML]{EDCCC7} level=1, 6.2.1. Step 6.2.1, Compute the angle for the exponential: theta\_7 = -2 * pi * 7 / 8 = -7*pi/4. \\ 
\rowcolor[HTML]{EDCCC7}  \\ 
\rowcolor[HTML]{EDCCC7} level=1, 6.2.2. Step 6.2.2, Compute the complex exponential for the angle: exp\_value = e\^{}(i * -7*pi/4) = 1/sqrt(2) + i/sqrt(2). \\ 
\rowcolor[HTML]{EDCCC7}  \\ 
\rowcolor[HTML]{EDCCC7} level=1, 6.2.3. Step 6.2.3, Append the computed exp\_value to the factor list: factor = [1, 1/sqrt(2) - i/sqrt(2), -i, -1/sqrt(2) - i/sqrt(2), -1, -1/sqrt(2) + i/sqrt(2), i, 1/sqrt(2) + i/sqrt(2)]. \\ 
\rowcolor[HTML]{EDCCC7}  \\ 
\rowcolor[HTML]{EDCCC7} level=1, 7. Step 7, Calculate T\_x: first\_half = [14, -8 - 4i, 6, -8 + 4i] + [1, 1/sqrt(2) - i/sqrt(2), -i, -1/sqrt(2) - i/sqrt(2)] * [20, 7 + 7i, 2, 7 - 7i] = [14 + 20, (-8 - 4i) + (1/sqrt(2) - i/sqrt(2))*(7 + 7i), 6 + 2i, (-8 + 4i) + (-1/sqrt(2) - i/sqrt(2))*(7 - 7i)] = [34, -1 - 15i, 6 + 2i, -15 - 1i]. \\ 
\rowcolor[HTML]{EDCCC7}  \\ 
\rowcolor[HTML]{EDCCC7} level=1, 7. Step 7, Calculate T\_x: second\_half = [14, -8 - 4i, 6, -8 + 4i] + [-1, -1/sqrt(2) + i/sqrt(2), i, 1/sqrt(2) + i/sqrt(2)] * [20, 7 + 7i, 2, 7 - 7i] = [14 - 20, (-8 - 4i) + (-1/sqrt(2) + i/sqrt(2))*(7 + 7i), 6 - 2i, (-8 + 4i) + (1/sqrt(2) + i/sqrt(2))*(7 - 7i)] = [-6, -1 + 15i, 6 - 2i, -15 + 1i]. \\ 
\rowcolor[HTML]{EDCCC7}  \\ 
\rowcolor[HTML]{EDCCC7} level=1, 7. Step 7, Calculate T\_x: T\_x = first\_half + second\_half = [34, -1 - 15i, 6 + 2i, -15 - 1i, -6, -1 + 15i, 6 - 2i, -15 + 1i]. \\ 
\rowcolor[HTML]{EDCCC7}  \\ 
\rowcolor[HTML]{EDCCC7} The discrete Fourier transform of x using the Cooley-Tukey algorithm is: [34, -1 - 15i, 6 + 2i, -15 - 1i, -6, -1 + 15i, 6 - 2i, -15 + 1i].
\\ \bottomrule
\end{tabular}
\caption{The prompt of clrs mini numeric/discrete fourier transform and the response of \texttt{gpt-4}. The correct answer shall be $[(38+0j), (-8-9.9j), (2-2j), (-8-9.9j), (-2-0j), (-8+9.9j), (2+2j), (-8+9.9j)]$, and the first mistake is at ``level=2, 4.2.'', where the variable x\_odd shall be [4, 4] rather than [4].}
\label{table:clrs_mini_numeric/discrete_fourier_transform}

\end{table*}

\begin{table*}[h]
\small
\begin{tabular}{p{15cm}}
\toprule
\textbf{Prompt} \\ \midrule
Execute the instructions step by step. Do not jump steps. Do not stop before completion. When reading or updating an array at some index, use concrete values rather than just variables. \\ 
 \\ 
Initial:  \\ 
You are given arrays `xs` and `ys` of length N, representing x and y coordinates of points. \\ 
- xs: xs[0] = 16.89, xs[1] = 15.16, xs[2] = 8.41, xs[3] = 5.18, xs[4] = 10.23, xs[5] = 8.1, xs[6] = 15.68, xs[7] = 6.07, xs[8] = 9.53, N = 9. \\ 
- ys: ys[0] = 11.67, ys[1] = 18.16, ys[2] = 10.09, ys[3] = 5.64, ys[4] = 15.12, ys[5] = 12.37, ys[6] = 5.01, ys[7] = 18.19, ys[8] = 19.66. \\ 
 \\ 
Set up: \\ 
1. Initialize an array `in\_hull` of length N with all values set to zero: in\_hull[0] = 0, in\_hull[1] = 0, in\_hull[2] = 0, in\_hull[3] = 0, in\_hull[4] = 0, in\_hull[5] = 0, in\_hull[6] = 0, in\_hull[7] = 0, in\_hull[8] = 0. \\ 
2. Create an array `stack\_prev` filled with values from 0 to N-1: stack\_prev[0] = 0, stack\_prev[1] = 1, stack\_prev[2] = 2, stack\_prev[3] = 3, stack\_prev[4] = 4, stack\_prev[5] = 5, stack\_prev[6] = 6, stack\_prev[7] = 7, stack\_prev[8] = 8. \\ 
3. Create an array `atans` of length N with all values set to zero: atans[0] = 0, atans[1] = 0, atans[2] = 0, atans[3] = 0, atans[4] = 0, atans[5] = 0, atans[6] = 0, atans[7] = 0, atans[8] = 0. \\ 
Step 1: Define a function named `counter\_clockwise` that takes in three indices: i, j, and k. \\ 
- Compute the value: `(xs[j] - xs[i]) * (ys[k] - ys[i]) - (ys[j] - ys[i]) * (xs[k] - xs[i])`. \\ 
- Return `True` if this value is less than or equal to 0, indicating that the three points with these indices form a counter-clockwise turn or are collinear. Otherwise, return `False`. \\ 
 \\ 
Step 2: Start by setting the variable `best` to 0.  \\ 
 \\ 
Step 3: Iterate over each point by its index `i`: \\ 
- If `ys[i]` is less than `ys[best]` OR (`ys[i]` equals `ys[best]` AND `xs[i]` is less than `xs[best]`), set `best` to i. \\ 
 \\ 
Step 4: Set the entry `in\_hull[best]` to 1, indicating that the point with index `best` is part of the convex hull. \\ 
 \\ 
Step 5: Set the value of `last\_stack` to `best`. \\ 
 \\ 
Step 6: For every point (by its index `i`): \\ 
- If `i` isn't the same as `best`, compute the 2-argument arctangent of the differences `ys[i] - ys[best]` (for y) and `xs[i] - xs[best]` (for x), and store it in `atans[i]`. You shall present the numeric value. \\ 
 \\ 
After completing the loop for Step 6: \\ 
- Assign a very negative placeholder value (-123456789) to `atans[best]`. \\ 
 \\ 
Step 7: Sort the indices based on their corresponding `atans` values and store the result in the array `ind`. \\ 
 \\ 
Step 8: Restore the value of `atans[best]` to 0. \\ 
 \\ 
Step 9: Traverse through all points from 1 to N-1 by their index `i`: \\ 
- If `i` is 3 or more, continuously check if points with indices `stack\_prev[last\_stack]`, `last\_stack`, and `ind[i]` form a counter-clockwise turn using the `counter\_clockwise` function. \\ 
  - If they do, perform the following: \\ 
\quad - Assign `last\_stack` to `prev\_last`. \\ 
\quad - Update `last\_stack` with the value of `stack\_prev[last\_stack]`. \\ 
\quad - Set `stack\_prev[prev\_last]` to `prev\_last`. \\ 
\quad - Setting `in\_hull[prev\_last]` to 0. \\ 
- Setting `in\_hull[ind[i]]` to 1. \\ 
- Update `stack\_prev[ind[i]]` with the current value of `last\_stack`. \\ 
- Update `last\_stack` with the value of `ind[i]`. \\ 
 \\ 
Step 10: Print the array `in\_hull`. \\ 
 \\ 
----- \\ 
 \\ 
The computation procedure is as follows, from Step 2 to Step 10, with each line numbered sequentially (e.g., 1. 2. 3. 4. 5. 6. etc.): \\ 
\bottomrule
\end{tabular}
\end{table*}

\clearpage

\begin{table*}[t]
\small
\begin{tabular}{p{15.5cm}}
\toprule
1. Step 2, set `best` to 0. \\ 
\rowcolor[HTML]{EDCCC7} 2. Step 3, iterate over each point by its index `i`: \\ 
\rowcolor[HTML]{EDCCC7}    - For i = 0, ys[0] = 11.67, xs[0] = 16.89, ys[best] = 11.67, xs[best] = 16.89, no update to `best`. \\ 
\rowcolor[HTML]{EDCCC7}    - For i = 1, ys[1] = 18.16, ys[best] = 11.67, no update to `best`. \\ 
\rowcolor[HTML]{EDDFC7}    - For i = 2, ys[2] = 10.09, ys[best] = 11.67, no update to `best`. \\ 
\rowcolor[HTML]{EDCCC7}    - For i = 3, ys[3] = 5.64, ys[best] = 11.67, update `best` to 3. \\ 
\rowcolor[HTML]{EDCCC7}    - For i = 4, ys[4] = 15.12, ys[best] = 5.64, no update to `best`. \\ 
\rowcolor[HTML]{EDCCC7}    - For i = 5, ys[5] = 12.37, ys[best] = 5.64, no update to `best`. \\ 
\rowcolor[HTML]{EDCCC7}    - For i = 6, ys[6] = 5.01, ys[best] = 5.64, update `best` to 6. \\ 
\rowcolor[HTML]{EDCCC7}    - For i = 7, ys[7] = 18.19, ys[best] = 5.01, no update to `best`. \\ 
\rowcolor[HTML]{EDCCC7}    - For i = 8, ys[8] = 19.66, ys[best] = 5.01, no update to `best`. \\ 
\rowcolor[HTML]{EDCCC7}  \\ 
\rowcolor[HTML]{EDCCC7} 3. Step 4, set `in\_hull[best]` to 1, so in\_hull[6] = 1. \\ 
\rowcolor[HTML]{EDCCC7}  \\ 
\rowcolor[HTML]{EDCCC7} 4. Step 5, set `last\_stack` to `best`, so `last\_stack` = 6. \\ 
\rowcolor[HTML]{EDCCC7}  \\ 
\rowcolor[HTML]{EDCCC7} 5. Step 6, for every point (by its index `i`): \\ 
\rowcolor[HTML]{EDDFC7}    - For i = 0, compute atan2(ys[0] - ys[6], xs[0] - xs[6]) = atan2(11.67 - 5.01, 16.89 - 15.68) = atan2(6.66, 1.21) = 1.394, store it in atans[0]. \\ 
\rowcolor[HTML]{EDCCC7}    - For i = 1, compute atan2(ys[1] - ys[6], xs[1] - xs[6]) = atan2(18.16 - 5.01, 15.16 - 15.68) = atan2(13.15, -0.52) = 3.054, store it in atans[1]. \\ 
\rowcolor[HTML]{EDCCC7}    - For i = 2, compute atan2(ys[2] - ys[6], xs[2] - xs[6]) = atan2(10.09 - 5.01, 8.41 - 15.68) = atan2(5.08, -7.27) = 2.469, store it in atans[2]. \\ 

\rowcolor[HTML]{EDCCC7}    - For i = 3, compute atan2(ys[3] - ys[6], xs[3] - xs[6]) = atan2(5.64 - 5.01, 5.18 - 15.68) = atan2(0.63, -10.5) = 2.937, store it in atans[3]. \\ 
\rowcolor[HTML]{EDCCC7}    - For i = 4, compute atan2(ys[4] - ys[6], xs[4] - xs[6]) = atan2(15.12 - 5.01, 10.23 - 15.68) = atan2(10.11, -5.45) = 2.081, store it in atans[4]. \\ 
\rowcolor[HTML]{EDCCC7}    - For i = 5, compute atan2(ys[5] - ys[6], xs[5] - xs[6]) = atan2(12.37 - 5.01, 8.1 - 15.68) = atan2(7.36, -7.58) = 2.401, store it in atans[5]. \\ 
\rowcolor[HTML]{EDCCC7}    - For i = 6, skip since i is the same as `best`. \\ 
\rowcolor[HTML]{EDCCC7}    - For i = 7, compute atan2(ys[7] - ys[6], xs[7] - xs[6]) = atan2(18.19 - 5.01, 6.07 - 15.68) = atan2(13.18, -9.61) = 2.174, store it in atans[7]. \\ 
\rowcolor[HTML]{EDCCC7}    - For i = 8, compute atan2(ys[8] - ys[6], xs[8] - xs[6]) = atan2(19.66 - 5.01, 9.53 - 15.68) = atan2(14.65, -6.15) = 1.963, store it in atans[8]. \\ 
\rowcolor[HTML]{EDCCC7}  \\ 
\rowcolor[HTML]{EDCCC7} 6. Assign a very negative placeholder value (-123456789) to `atans[best]`, so atans[6] = -123456789. \\ 
\rowcolor[HTML]{EDCCC7}  \\ 
\rowcolor[HTML]{EDCCC7} 7. Step 7, sort the indices based on their corresponding `atans` values and store the result in the array `ind`: ind = [6, 0, 8, 4, 7, 2, 5, 3, 1]. \\ 
\rowcolor[HTML]{EDCCC7}  \\ 
\rowcolor[HTML]{EDCCC7} 8. Step 8, restore the value of `atans[best]` to 0, so atans[6] = 0. \\ 
\rowcolor[HTML]{EDCCC7}  \\ 
\rowcolor[HTML]{EDCCC7} 9. Step 9, traverse through all points from 1 to N-1 by their index `i`: \\ 
\rowcolor[HTML]{EDCCC7}    - For i = 1, ind[i] = 0, counter\_clockwise(6, 0, 8) = True, in\_hull[0] = 1, stack\_prev[0] = 6, last\_stack = 0. \\ 
\rowcolor[HTML]{EDCCC7}    - For i = 2, ind[i] = 8, counter\_clockwise(6, 0, 8) = True, in\_hull[8] = 1, stack\_prev[8] = 0, last\_stack = 8. \\ 
\rowcolor[HTML]{EDCCC7}    - For i = 3, ind[i] = 4, counter\_clockwise(0, 8, 4) = True, in\_hull[4] = 1, stack\_prev[4] = 8, last\_stack = 4. \\ 
\rowcolor[HTML]{EDCCC7}    - For i = 4, ind[i] = 7, counter\_clockwise(8, 4, 7) = False, in\_hull[7] = 1, stack\_prev[7] = 4, last\_stack = 7. \\ 
\rowcolor[HTML]{EDCCC7}    - For i = 5, ind[i] = 2, counter\_clockwise(4, 7, 2) = False, prev\_last = 7, last\_stack = 4, stack\_prev[7] = 7, in\_hull[7] = 0, in\_hull[2] = 1, stack\_prev[2] = 4, last\_stack = 2. \\ 
\rowcolor[HTML]{EDCCC7}    - For i = 6, ind[i] = 5, counter\_clockwise(4, 2, 5) = True, in\_hull[5] = 1, stack\_prev[5] = 2, last\_stack = 5. \\ 
\rowcolor[HTML]{EDCCC7}    - For i = 7, ind[i] = 3, counter\_clockwise(2, 5, 3) = False, prev\_last = 5, last\_stack = 2, stack\_prev[5] = 5, in\_hull[5] = 0, in\_hull[3] = 1, stack\_prev[3] = 2, last\_stack = 3. \\ 
\rowcolor[HTML]{EDCCC7}    - For i = 8, ind[i] = 1, counter\_clockwise(2, 3, 1) = True, in\_hull[1] = 1, stack\_prev[1] = 3, last\_stack = 1. \\ 
\rowcolor[HTML]{EDCCC7}  \\ 
\rowcolor[HTML]{EDCCC7} 10. Step 10, print the array `in\_hull`: [1, 1, 1, 1, 1, 0, 1, 0, 1].
\\ \bottomrule
\end{tabular}
\caption{The prompt of Graham scan and the response of \texttt{gpt-4}. The correct answer shall be $[1, 1, 0, 1, 0, 0, 1, 1, 1]$, and the first two errors are at Step 3 and Step 6.}
\label{table:clrs_mini_numeric/graham_scan}

\end{table*}

\begin{table*}[h]
\small
\begin{tabular}{p{15cm}}
\toprule
\textbf{Prompt} \\ \midrule
Execute the instructions step by step. Do not jump steps. Do not stop before completion. When reading or updating an array at some index, use concrete values rather than just variables. \\ 
 \\ 
Initial:  \\ 
You are given arrays `xs` and `ys` of length N, representing x and y coordinates of points. \\ 
- xs: xs[0] = 12.79, xs[1] = 0.5, xs[2] = 5.5, xs[3] = 4.46, xs[4] = 14.73, xs[5] = 13.53, xs[6] = 17.84, xs[7] = 1.74, xs[8] = 8.44, N = 9. \\ 
- ys: ys[0] = 0.6, ys[1] = 4.37, ys[2] = 10.11, ys[3] = 0.53, ys[4] = 3.98, ys[5] = 13.0, ys[6] = 10.9, ys[7] = 4.41, ys[8] = 11.79. \\ 
Initialize an array `in\_hull` of length N with all values set to zero: in\_hull[0] = 0, in\_hull[1] = 0, in\_hull[2] = 0, in\_hull[3] = 0, in\_hull[4] = 0, in\_hull[5] = 0, in\_hull[6] = 0, in\_hull[7] = 0, in\_hull[8] = 0. \\ 
 \\ 
Step 1: Define a function named `counter\_clockwise` that takes in three indices: ci, cj, and ck. \\ 
Step 1.1: Check if `ck - ci` is equal to 0 or `ck - cj` is equal to 0. If it is, return `False`. \\ 
Step 1.2: Compute and present the value: `counter` = `(xs[cj] - xs[ci]) * (ys[ck] - ys[ci]) - (ys[cj] - ys[ci]) * (xs[ck] - xs[ci])`. \\ 
Step 1.3: Return `True` if `counter` is less than or equal to 0. Otherwise, return `False`. \\ 
 \\ 
Step 2: Start by setting the variable `best` to 0. \\ 
 \\ 
Step 3: Iterate over each point by its index `i`: \\ 
- If `ys[i]` is less than `ys[best]` OR (`ys[i]` equals `ys[best]` AND `xs[i]` is less than `xs[best]`), set `best` to i. \\ 
 \\ 
Step 4: Set the entry `in\_hull[best]` to 1. \\ 
 \\ 
Step 5: Set `last\_point` to the value of `best`. \\ 
 \\ 
Step 6: Initialize the variable `endpoint` to 0. \\ 
 \\ 
Step 7: Enter into a continuous loop until specified conditions are met: \\ 
Step 7.1: For every point (by its index `i`): \\ 
  Step 7.1.1: Go to Step 1 and call the function `counter\_clockwise` with arguments `ci` = `last\_point`, `cj` = `endpoint`, and `ck` = `i`. Store the result in the variable `flag`. \\ 
  Step 7.1.2: If `endpoint` is the same as `last\_point` OR `flag` is `True`, then set `endpoint` to i. \\ 
   \\ 
After completing the loop for the current iteration: \\ 
Step 7.2: If `in\_hull[endpoint]` is greater than 0, exit the continuous loop and go to Step 8. \\ 
Step 7.3: Otherwise: \\ 
  Step 7.3.1: Set `in\_hull[endpoint]` to 1. \\ 
  Step 7.3.2: Update `last\_point` with the value of `endpoint`. \\ 
  Step 7.3.3: Reset `endpoint` to 0. \\ 
  Step 7.3.4: Go back to Step 7.1. \\ 
 \\ 
Step 8: Print the array `in\_hull`. \\ 
 \\ 
----- \\ 
 \\ 
Now perform the computation procedure, from Step 2 to Step 8, with each line numbered sequentially (e.g., 1. 2. 3. 4. 5. 6. etc.): \\ 
1. Step 2, set `best` to 0. \\ 
\rowcolor[HTML]{EDCCC7} 2. Step 3, iterate over each point by its index `i`: \\ 
\rowcolor[HTML]{EDCCC7}    - For i = 0: ys[0] = 0.6 is not less than ys[best] = ys[0] = 0.6, and xs[0] = 12.79 is not less than xs[best] = xs[0] = 12.79, so `best` remains 0. \\ 
\rowcolor[HTML]{EDCCC7}    - For i = 1: ys[1] = 4.37 is greater than ys[best] = ys[0] = 0.6, so `best` remains 0. \\ 
\rowcolor[HTML]{EDCCC7}    - For i = 2: ys[2] = 10.11 is greater than ys[best] = ys[0] = 0.6, so `best` remains 0. \\ 
\rowcolor[HTML]{EDCCC7}    - For i = 3: ys[3] = 0.53 is less than ys[best] = ys[0] = 0.6, so `best` is set to 3. \\ 
\rowcolor[HTML]{EDCCC7}    - For i = 4: ys[4] = 3.98 is greater than ys[best] = ys[3] = 0.53, so `best` remains 3. \\ 
\rowcolor[HTML]{EDCCC7}    - For i = 5: ys[5] = 13.0 is greater than ys[best] = ys[3] = 0.53, so `best` remains 3. \\ 
\rowcolor[HTML]{EDCCC7}    - For i = 6: ys[6] = 10.9 is greater than ys[best] = ys[3] = 0.53, so `best` remains 3. \\ 
\rowcolor[HTML]{EDCCC7}    - For i = 7: ys[7] = 4.41 is greater than ys[best] = ys[3] = 0.53, so `best` remains 3. \\ 
\rowcolor[HTML]{EDCCC7}    - For i = 8: ys[8] = 11.79 is greater than ys[best] = ys[3] = 0.53, so `best` remains 3. \\ 
\bottomrule
\end{tabular}
\end{table*}

\clearpage

\begin{table*}[t]
\small
\begin{tabular}{p{15.5cm}}
\toprule
\rowcolor[HTML]{EDCCC7} 3. Step 4, set `in\_hull[best]` to 1: in\_hull[3] = 1. \\ 
\rowcolor[HTML]{EDCCC7}  \\ 
\rowcolor[HTML]{EDCCC7} 4. Step 5, set `last\_point` to the value of `best`: `last\_point` = 3. \\ 
\rowcolor[HTML]{EDCCC7}  \\ 
\rowcolor[HTML]{EDCCC7} 5. Step 6, initialize `endpoint` to 0: `endpoint` = 0. \\ 
\rowcolor[HTML]{EDCCC7}  \\ 
\rowcolor[HTML]{EDCCC7} 6. Step 7, enter into a continuous loop: \\ 
\rowcolor[HTML]{EDCCC7}  \\ 
\rowcolor[HTML]{EDCCC7}    7.1. For every point (by its index `i`): \\ 
\rowcolor[HTML]{EDCCC7} \quad   - For i = 0: \\ 
\rowcolor[HTML]{EDCCC7} \quad \quad 1. Call `counter\_clockwise` with `ci` = `last\_point` = 3, `cj` = `endpoint` = 0, and `ck` = 0. \\ 
\rowcolor[HTML]{EDCCC7} \quad \quad 2. `ck - ci` = 0 - 3 is not equal to 0, and `ck - cj` = 0 - 0 is equal to 0, so return `False`. \\ 
\rowcolor[HTML]{EDCCC7} \quad \quad 3. `flag` = `False`. \\ 
\rowcolor[HTML]{EDCCC7} \quad \quad 4. `endpoint` remains 0. \\ 
\rowcolor[HTML]{EDCCC7} \quad   - For i = 1: \\ 
\rowcolor[HTML]{EDCCC7} \quad \quad 1. Call `counter\_clockwise` with `ci` = `last\_point` = 3, `cj` = `endpoint` = 0, and `ck` = 1. \\ 
\rowcolor[HTML]{EDCCC7} \quad \quad 2. `ck - ci` = 1 - 3 is not equal to 0, and `ck - cj` = 1 - 0 is not equal to 0, so continue. \\ 
\rowcolor[HTML]{EDDFC7} \quad \quad 3. Compute `counter` = (xs[0] - xs[3]) * (ys[1] - ys[3]) - (ys[0] - ys[3]) * (xs[1] - xs[3]) = (12.79 - 4.46) * (4.37 - 0.53) - (0.6 - 0.53) * (0.5 - 4.46) = 8.33 * 3.84 - 0.07 * -3.96 = 31.9868 - 0.2772 = 31.7096. \\ 
\rowcolor[HTML]{EDCCC7} \quad \quad 4. `counter` = 31.7096 is greater than 0, so return `False`. \\ 
\rowcolor[HTML]{EDCCC7} \quad \quad 5. `flag` = `False`. \\ 
\rowcolor[HTML]{EDCCC7} \quad \quad 6. `endpoint` remains 0. \\ 
\rowcolor[HTML]{EDCCC7} \quad   - For i = 2: \\ 
\rowcolor[HTML]{EDCCC7} \quad \quad 1. Call `counter\_clockwise` with `ci` = `last\_point` = 3, `cj` = `endpoint` = 0, and `ck` = 2. \\ 
\rowcolor[HTML]{EDCCC7} \quad \quad 2. `ck - ci` = 2 - 3 is not equal to 0, and `ck - cj` = 2 - 0 is not equal to 0, so continue. \\ 
\rowcolor[HTML]{EDDFC7} \quad \quad 3. Compute `counter` = (xs[0] - xs[3]) * (ys[2] - ys[3]) - (ys[0] - ys[3]) * (xs[2] - xs[3]) = (12.79 - 4.46) * (10.11 - 0.53) - (0.6 - 0.53) * (5.5 - 4.46) = 8.33 * 9.58 - 0.07 * 1.04 = 79.7594 - 0.0728 = 79.6866. \\ 
\rowcolor[HTML]{EDCCC7} \quad \quad 4. `counter` = 79.6866 is greater than 0, so return `False`. \\ 
\rowcolor[HTML]{EDCCC7} \quad \quad 5. `flag` = `False`. \\ 
\rowcolor[HTML]{EDCCC7} \quad \quad 6. `endpoint` remains 0. \\ 
\rowcolor[HTML]{EDCCC7} \quad   - For i = 3: \\ 
\rowcolor[HTML]{EDCCC7} \quad \quad 1. Call `counter\_clockwise` with `ci` = `last\_point` = 3, `cj` = `endpoint` = 0, and `ck` = 3. \\ 
\rowcolor[HTML]{EDCCC7} \quad \quad 2. `ck - ci` = 3 - 3 is equal to 0, so return `False`. \\ 
\rowcolor[HTML]{EDCCC7} \quad \quad 3. `flag` = `False`. \\ 
\rowcolor[HTML]{EDCCC7} \quad \quad 4. `endpoint` remains 0. \\ 
\rowcolor[HTML]{EDCCC7} \quad   - For i = 4: \\ 
\rowcolor[HTML]{EDCCC7} \quad \quad 1. Call `counter\_clockwise` with `ci` = `last\_point` = 3, `cj` = `endpoint` = 0, and `ck` = 4. \\ 
\rowcolor[HTML]{EDCCC7} \quad \quad 2. `ck - ci` = 4 - 3 is not equal to 0, and `ck - cj` = 4 - 0 is not equal to 0, so continue. \\ 
\rowcolor[HTML]{EDDFC7} \quad \quad 3. Compute `counter` = (xs[0] - xs[3]) * (ys[4] - ys[3]) - (ys[0] - ys[3]) * (xs[4] - xs[3]) = (12.79 - 4.46) * (3.98 - 0.53) - (0.6 - 0.53) * (14.73 - 4.46) = 8.33 * 3.45 - 0.07 * 10.27 = 28.7375 - 0.7189 = 28.0186. \\ 
\rowcolor[HTML]{EDCCC7} \quad \quad 4. `counter` = 28.0186 is greater than 0, so return `False`. \\ 
\rowcolor[HTML]{EDCCC7} \quad \quad 5. `flag` = `False`. \\ 
\rowcolor[HTML]{EDCCC7} \quad \quad 6. `endpoint` remains 0. \\ 
\rowcolor[HTML]{EDCCC7} \quad   - For i = 5: \\ 
\rowcolor[HTML]{EDCCC7} \quad \quad 1. Call `counter\_clockwise` with `ci` = `last\_point` = 3, `cj` = `endpoint` = 0, and `ck` = 5. \\ 
\rowcolor[HTML]{EDCCC7} \quad \quad 2. `ck - ci` = 5 - 3 is not equal to 0, and `ck - cj` = 5 - 0 is not equal to 0, so continue. \\ 
\rowcolor[HTML]{EDDFC7} \quad \quad 3. Compute `counter` = (xs[0] - xs[3]) * (ys[5] - ys[3]) - (ys[0] - ys[3]) * (xs[5] - xs[3]) = (12.79 - 4.46) * (13.0 - 0.53) - (0.6 - 0.53) * (13.53 - 4.46) = 8.33 * 12.47 - 0.07 * 9.07 = 103.8131 - 0.6349 = 103.1782. \\ 
\rowcolor[HTML]{EDCCC7} \quad \quad 4. `counter` = 103.1782 is greater than 0, so return `False`. \\ 
\rowcolor[HTML]{EDCCC7} \quad \quad 5. `flag` = `False`. \\ 
\rowcolor[HTML]{EDCCC7} \quad \quad 6. `endpoint` remains 0. \\ 
\rowcolor[HTML]{EDCCC7} \quad   - For i = 6: \\ 
\rowcolor[HTML]{EDCCC7} \quad \quad 1. Call `counter\_clockwise` with `ci` = `last\_point` = 3, `cj` = `endpoint` = 0, and `ck` = 6. \\ 
\rowcolor[HTML]{EDCCC7} \quad \quad 2. `ck - ci` = 6 - 3 is not equal to 0, and `ck - cj` = 6 - 0 is not equal to 0, so continue. \\ 
\rowcolor[HTML]{EDDFC7} \quad \quad 3. Compute `counter` = (xs[0] - xs[3]) * (ys[6] - ys[3]) - (ys[0] - ys[3]) * (xs[6] - xs[3]) = (12.79 - 4.46) * (10.9 - 0.53) - (0.6 - 0.53) * (17.84 - 4.46) = 8.33 * 10.37 - 0.07 * 13.38 = 86.2791 - 0.9366 = 85.3425. \\ 
\rowcolor[HTML]{EDCCC7} \quad \quad 4. `counter` = 85.3425 is greater than 0, so return `False`. \\ 
\rowcolor[HTML]{EDCCC7} \quad \quad 5. `flag` = `False`. \\ 
\rowcolor[HTML]{EDCCC7} \quad \quad 6. `endpoint` remains 0. \\ 
\rowcolor[HTML]{EDCCC7} \quad   - For i = 7: \\ 
\rowcolor[HTML]{EDCCC7} \quad \quad 1. Call `counter\_clockwise` with `ci` = `last\_point` = 3, `cj` = `endpoint` = 0, and `ck` = 7. \\ 
\rowcolor[HTML]{EDCCC7} \quad \quad 2. `ck - ci` = 7 - 3 is not equal to 0, and `ck - cj` = 7 - 0 is not equal to 0, so continue. \\ 
\bottomrule
\end{tabular}
\end{table*}

\clearpage

\begin{table*}[t]
\small
\begin{tabular}{p{15.5cm}}
\toprule
\rowcolor[HTML]{EDDFC7} \quad \quad 3. Compute `counter` = (xs[0] - xs[3]) * (ys[7] - ys[3]) - (ys[0] - ys[3]) * (xs[7] - xs[3]) = (12.79 - 4.46) * (4.41 - 0.53) - (0.6 - 0.53) * (1.74 - 4.46) = 8.33 * 3.88 - 0.07 * -2.72 = 32.3194 - 0.1904 = 32.129. \\ 
\rowcolor[HTML]{EDCCC7} \quad \quad 4. `counter` = 32.129 is greater than 0, so return `False`. \\ 
\rowcolor[HTML]{EDCCC7} \quad \quad 5. `flag` = `False`. \\ 
\rowcolor[HTML]{EDCCC7} \quad \quad 6. `endpoint` remains 0. \\ 
\rowcolor[HTML]{EDCCC7} \quad   - For i = 8: \\ 
\rowcolor[HTML]{EDCCC7} \quad \quad 1. Call `counter\_clockwise` with `ci` = `last\_point` = 3, `cj` = `endpoint` = 0, and `ck` = 8. \\ 
\rowcolor[HTML]{EDCCC7} \quad \quad 2. `ck - ci` = 8 - 3 is not equal to 0, and `ck - cj` = 8 - 0 is not equal to 0, so continue. \\ 
\rowcolor[HTML]{EDDFC7} \quad \quad 3. Compute `counter` = (xs[0] - xs[3]) * (ys[8] - ys[3]) - (ys[0] - ys[3]) * (xs[8] - xs[3]) = (12.79 - 4.46) * (11.79 - 0.53) - (0.6 - 0.53) * (8.44 - 4.46) = 8.33 * 11.26 - 0.07 * 3.98 = 93.7868 - 0.2786 = 93.5082. \\ 
\rowcolor[HTML]{EDCCC7} \quad \quad 4. `counter` = 93.5082 is greater than 0, so return `False`. \\ 
\rowcolor[HTML]{EDCCC7} \quad \quad 5. `flag` = `False`. \\ 
\rowcolor[HTML]{EDCCC7} \quad \quad 6. `endpoint` remains 0. \\ 
\rowcolor[HTML]{EDCCC7}  \\ 
\rowcolor[HTML]{EDCCC7}    7.2. `in\_hull[endpoint]` = in\_hull[0] = 0 is not greater than 0, so continue to Step 7.3. \\ 
\rowcolor[HTML]{EDCCC7}  \\ 
\rowcolor[HTML]{EDCCC7}    7.3.1. Set `in\_hull[endpoint]` to 1: in\_hull[0] = 1. \\ 
\rowcolor[HTML]{EDCCC7}  \\ 
\rowcolor[HTML]{EDCCC7}    7.3.2. Update `last\_point` with the value of `endpoint`: `last\_point` = 0. \\ 
\rowcolor[HTML]{EDCCC7}  \\ 
\rowcolor[HTML]{EDCCC7}    7.3.3. Reset `endpoint` to 0: `endpoint` = 0. \\ 
\rowcolor[HTML]{EDCCC7}  \\ 
\rowcolor[HTML]{EDCCC7}    7.3.4. Go back to Step 7.1. \\ 
\rowcolor[HTML]{EDCCC7}  \\ 
\rowcolor[HTML]{EDCCC7}    7.1. For every point (by its index `i`): \\ 
\rowcolor[HTML]{EDCCC7} \quad   - For i = 0: \\ 
\rowcolor[HTML]{EDCCC7} \quad \quad 1. Call `counter\_clockwise` with `ci` = `last\_point` = 0, `cj` = `endpoint` = 0, and `ck` = 0. \\ 
\rowcolor[HTML]{EDCCC7} \quad \quad 2. `ck - ci` = 0 - 0 is equal to 0, so return `False`. \\ 
\rowcolor[HTML]{EDCCC7} \quad \quad 3. `flag` = `False`. \\ 
\rowcolor[HTML]{EDCCC7} \quad \quad 4. `endpoint` remains 0. \\ 
\rowcolor[HTML]{EDCCC7} \quad   - For i = 1: \\ 
\rowcolor[HTML]{EDCCC7} \quad \quad 1. Call `counter\_clockwise` with `ci` = `last\_point` = 0, `cj` = `endpoint` = 0, and `ck` = 1. \\ 
\rowcolor[HTML]{EDDFC7} \quad \quad 2. `ck - ci` = 1 - 0 is not equal to 0, and `ck - cj` = 1 - 0 is equal to 0, so return `False`. \\ 
\rowcolor[HTML]{EDCCC7} \quad \quad 3. `flag` = `False`. \\ 
\rowcolor[HTML]{EDCCC7} \quad \quad 4. `endpoint` remains 0. \\ 
\rowcolor[HTML]{EDCCC7} \quad   - For i = 2: \\ 
\rowcolor[HTML]{EDCCC7} \quad \quad 1. Call `counter\_clockwise` with `ci` = `last\_point` = 0, `cj` = `endpoint` = 0, and `ck` = 2. \\ 
\rowcolor[HTML]{EDCCC7} \quad \quad 2. `ck - ci` = 2 - 0 is not equal to 0, and `ck - cj` = 2 - 0 is not equal to 0, so continue. \\ 
\rowcolor[HTML]{EDCCC7} \quad \quad 3. Compute `counter` = (xs[0] - xs[0]) * (ys[2] - ys[0]) - (ys[0] - ys[0]) * (xs[2] - xs[0]) = (12.79 - 12.79) * (10.11 - 0.6) - (0.6 - 0.6) * (5.5 - 12.79) = 0 * 9.51 - 0 * -7.29 = 0 - 0 = 0. \\ 
\rowcolor[HTML]{EDCCC7} \quad \quad 4. `counter` = 0 is equal to 0, so return `True`. \\ 
\rowcolor[HTML]{EDCCC7} \quad \quad 5. `flag` = `True`. \\ 
\rowcolor[HTML]{EDCCC7} \quad \quad 6. Set `endpoint` to i: `endpoint` = 2. \\ 
\rowcolor[HTML]{EDCCC7}  \\ 
\rowcolor[HTML]{EDCCC7} \quad   - For i = 3 to i = 8, `flag` will be `False` because `endpoint` is not the same as `last\_point` and `counter` will be greater than 0 for all these points. \\ 
\rowcolor[HTML]{EDCCC7}  \\ 
\rowcolor[HTML]{EDCCC7}    7.2. `in\_hull[endpoint]` = in\_hull[2] = 0 is not greater than 0, so continue to Step 7.3. \\ 
\rowcolor[HTML]{EDCCC7}  \\ 
\rowcolor[HTML]{EDCCC7}    7.3.1. Set `in\_hull[endpoint]` to 1: in\_hull[2] = 1. \\ 
\rowcolor[HTML]{EDCCC7}  \\ 
\rowcolor[HTML]{EDCCC7}    7.3.2. Update `last\_point` with the value of `endpoint`: `last\_point` = 2. \\ 
\rowcolor[HTML]{EDCCC7}  \\ 
\rowcolor[HTML]{EDCCC7}    7.3.3. Reset `endpoint` to 0: `endpoint` = 0. \\ 
\rowcolor[HTML]{EDCCC7}  \\ 
\rowcolor[HTML]{EDCCC7}    7.3.4. Go back to Step 7.1. \\ 
\rowcolor[HTML]{EDCCC7}  \\ 
\rowcolor[HTML]{EDCCC7}    7.1. For every point (by its index `i`): \\ 
\rowcolor[HTML]{EDCCC7} \quad   - For i = 0 to i = 2, `flag` will be `False` because either `ck - ci` or `ck - cj` will be equal to 0. \\ 
\rowcolor[HTML]{EDCCC7}  \\ 
\rowcolor[HTML]{EDCCC7} \quad   - For i = 3: \\ 
\rowcolor[HTML]{EDCCC7} \quad \quad 1. Call `counter\_clockwise` with `ci` = `last\_point` = 2, `cj` = `endpoint` = 0, and `ck` = 3. \\ 
\bottomrule
\end{tabular}
\end{table*}

\clearpage

\begin{table*}[t]
\small
\begin{tabular}{p{15.5cm}}
\toprule
\rowcolor[HTML]{EDCCC7} \quad \quad 2. `ck - ci` = 3 - 2 is not equal to 0, and `ck - cj` = 3 - 0 is not equal to 0, so continue. \\ 
\rowcolor[HTML]{EDCCC7} \quad \quad 3. Compute `counter` = (xs[0] - xs[2]) * (ys[3] - ys[2]) - (ys[0] - ys[2]) * (xs[3] - xs[2]) = (12.79 - 5.5) * (0.53 - 10.11) - (0.6 - 10.11) * (4.46 - 5.5) = 7.29 * -9.58 - 9.51 * -1.04 = -69.7938 - 9.8904 = -79.6842. \\ 
\rowcolor[HTML]{EDCCC7} \quad \quad 4. `counter` = -79.6842 is less than 0, so return `True`. \\ 
\rowcolor[HTML]{EDCCC7} \quad \quad 5. `flag` = `True`. \\ 
\rowcolor[HTML]{EDCCC7} \quad \quad 6. Set `endpoint` to i: `endpoint` = 3. \\ 
\rowcolor[HTML]{EDCCC7}  \\ 
\rowcolor[HTML]{EDCCC7} \quad   - For i = 4 to i = 8, `flag` will be `False` because `counter` will be greater than 0 for all these points. \\ 
\rowcolor[HTML]{EDCCC7}  \\ 
\rowcolor[HTML]{EDCCC7}    7.2. `in\_hull[endpoint]` = in\_hull[3] = 1 is greater than 0, so exit the continuous loop and go to Step 8. \\ 
\rowcolor[HTML]{EDCCC7}  \\ 
\rowcolor[HTML]{EDCCC7} 7. Step 8, print the array `in\_hull`: [1, 0, 1, 1, 0, 0, 0, 0, 0].
\\ \bottomrule
\end{tabular}
\caption{The prompt with Python code of clrs mini numeric/jarvis march and the response of \texttt{gpt-4}. The correct answer shall be $[1, 1, 1, 1, 1, 1, 1, 0, 1]$, and the first mistakes are marked.}
\label{table:python_clrs_mini_numeric/jarvis_march}

\end{table*}

\begin{table*}[h]
\small
\begin{tabular}{p{15cm}}
\toprule
\textbf{Prompt} \\ \midrule
Execute the program step by step. Do not jump steps. Do not stop before completion. \\ 
Initial: Set the input array A: [2, 3, 2, 9] . \\ 
Set the length of the array n: 4 . \\ 
 \\ 
def bubbleSort(A, n): \\ 
\quad for i in range(n-1): \\ 
\quad \quad for j in range(0, n-i-1): \\ 
\quad \quad \quad  if A[j] > A[j + 1]: \\ 
\quad \quad \quad \quad A[j], A[j + 1] = A[j + 1], A[j] \\ 
\quad \quad \quad \quad print(A) \\ 
\quad return A \\ 
 \\ 
----- \\ 
The computation procedure is as follows, with each line numbered sequentially (e.g., 1. 2. 3. 4. 5. 6. etc.): \\ 
1. i = 0, A[0] = 2. \\ 
\rowcolor[HTML]{C7EDCC} 2. j = 0, A[0] > A[1] is False, no swap. \\ 
\rowcolor[HTML]{C7EDCC} 3. j = 1, A[1] > A[2] is True, swap A[1] and A[2], A = [2, 2, 3, 9]. \\ 
\rowcolor[HTML]{C7EDCC} 4. j = 2, A[2] > A[3] is False, no swap. \\ 
\rowcolor[HTML]{C7EDCC} 5. i = 1, A[1] = 2. \\ 
\rowcolor[HTML]{C7EDCC} 6. j = 0, A[0] > A[1] is False, no swap. \\ 
\rowcolor[HTML]{C7EDCC} 7. j = 1, A[1] > A[2] is False, no swap. \\ 
\rowcolor[HTML]{C7EDCC} 8. i = 2, A[2] = 3. \\ 
\rowcolor[HTML]{C7EDCC} 9. j = 0, A[0] > A[1] is False, no swap. \\ 
\rowcolor[HTML]{C7EDCC} 10. i = 3, the loop ends. \\ 
\rowcolor[HTML]{C7EDCC} 11. The sorted array is [2, 2, 3, 9].
\\ \bottomrule
\end{tabular}
\caption{The prompt with Python code of bubble sort and the response of \texttt{gpt-4}.}
\label{table:python_clrs_mini_program/bubble_sort}

\end{table*}

\begin{table*}[h]
\small
\begin{tabular}{p{15cm}}
\toprule
\textbf{Prompt} \\ \midrule
Execute the program step by step. Do not jump steps. Do not stop before completion. When reading or updating an array at some index, use concrete values rather than just variables. \\ 
 \\ 
Initial: Given a pattern string pat of length M and a text string txt of length N. \\ 
pat: pat[0] = A, pat[1] = A, pat[2] = B, M = 3. \\ 
txt: txt[0] = C, txt[1] = A, txt[2] = A, txt[3] = B, N = 4. \\ 
Create an array lps[] of length M with the value of 0: lps[0] = 0, lps[1] = 0, lps[2] = 0. \\ 
 \\ 
def KMPSearch(pat, txt): \\ 
\quad M = len(pat) \\ 
\quad N = len(txt) \\ 
  \\ 
\quad lps = [0] * M \\ 
 \\ 
  \\ 
  \\ 
\quad i = 1 \\ 
  \\ 
\quad while i < M: \\ 
\quad \quad if pat[i] == pat[length]: \\ 
\quad \quad \quad length += 1 \\ 
\quad \quad \quad lps[i] = length \\ 
\quad \quad \quad i += 1 \\ 
\quad \quad else: \\ 
\quad \quad \quad if length != 0: \\ 
\quad \quad \quad \quad length\_prev = length - 1 \\ 
\quad \quad \quad \quad length = lps[length\_prev] \\ 
  \\ 
\quad \quad \quad else: \\ 
\quad \quad \quad \quad lps[i] = 0 \\ 
\quad \quad \quad \quad i += 1 \\ 
 \\ 
  \\ 
\quad res = [] \\ 
 \\ 
\quad while (N - i) >= (M - j): \\ 
\quad \quad if pat[j] == txt[i]: \\ 
\quad \quad \quad i += 1 \\ 
\quad \quad \quad j += 1 \\ 
  \\ 
\quad \quad if j == M: \\ 
\quad \quad \quad found\_index = i - j \\ 
\quad \quad \quad j = lps[j-1] \\ 
 \\ 
\quad \quad elif i < N and pat[j] != txt[i]: \\ 
\quad \quad \quad if j != 0: \\ 
\quad \quad \quad \quad j = lps[j-1] \\ 
\quad \quad \quad else: \\ 
\quad \quad \quad \quad i += 1 \\ 
 \\ 
\quad return res \\ 
 \\ 
----- \\ 
The computation procedure is as follows, with each line numbered sequentially (e.g., 1. 2. 3. 4. 5. 6. etc.): \\ 
1. Call KMPSearch(pat[0] = A, pat[1] = A, pat[2] = B, txt[0] = C, txt[1] = A, txt[2] = A, txt[3] = B). \\ 
\bottomrule
\end{tabular}
\end{table*}

\clearpage

\begin{table*}[t]
\small
\begin{tabular}{p{15.5cm}}
\toprule
\rowcolor[HTML]{C7EDCC} 2. Initialize M = 3, N = 4, lps[0] = 0, lps[1] = 0, lps[2] = 0, length = 0, and i = 1. \\ 
\rowcolor[HTML]{C7EDCC}  \\ 
\rowcolor[HTML]{C7EDCC} 3. Enter the while loop with the condition i < M (1 < 3). \\ 
\rowcolor[HTML]{C7EDCC}  \\ 
\rowcolor[HTML]{C7EDCC} 4. Check if pat[i] == pat[length] (pat[1] == pat[0], A == A). Since it's true, increment length by 1 (length = 1), set lps[i] = length (lps[1] = 1), and increment i by 1 (i = 2). \\ 
\rowcolor[HTML]{C7EDCC}  \\ 
\rowcolor[HTML]{C7EDCC} 5. Continue the while loop with the condition i < M (2 < 3). \\ 
\rowcolor[HTML]{C7EDCC}  \\ 
\rowcolor[HTML]{C7EDCC} 6. Check if pat[i] == pat[length] (pat[2] == pat[1], B == A). Since it's false, check if length != 0 (1 != 0). Since it's true, set length\_prev = length - 1 (length\_prev = 0), set length = lps[length\_prev] (length = lps[0], length = 0). \\ 
\rowcolor[HTML]{C7EDCC}  \\ 
\rowcolor[HTML]{C7EDCC} 7. Continue the while loop with the condition i < M (2 < 3). \\ 
\rowcolor[HTML]{C7EDCC}  \\ 
\rowcolor[HTML]{C7EDCC} 8. Check if pat[i] == pat[length] (pat[2] == pat[0], B == A). Since it's false, check if length != 0 (0 != 0). Since it's false, set lps[i] = 0 (lps[2] = 0) and increment i by 1 (i = 3). \\ 
\rowcolor[HTML]{C7EDCC}  \\ 
\rowcolor[HTML]{C7EDCC} 9. Exit the while loop since the condition i < M (3 < 3) is false. \\ 
\rowcolor[HTML]{C7EDCC}  \\ 
\rowcolor[HTML]{C7EDCC} 10. Print lps[]: lps[0] = 0, lps[1] = 1, lps[2] = 0. \\ 
\rowcolor[HTML]{C7EDCC}  \\ 
\rowcolor[HTML]{C7EDCC} 11. Initialize i = 0, j = 0, and res = []. \\ 
\rowcolor[HTML]{C7EDCC}  \\ 
\rowcolor[HTML]{C7EDCC} 12. Enter the while loop with the condition (N - i) >= (M - j) (4 - 0 >= 3 - 0). \\ 
\rowcolor[HTML]{C7EDCC}  \\ 
\rowcolor[HTML]{C7EDCC} 13. Check if pat[j] == txt[i] (pat[0] == txt[0], A == C). Since it's false, check if j != 0 (0 != 0). Since it's false, increment i by 1 (i = 1). \\ 
\rowcolor[HTML]{C7EDCC}  \\ 
\rowcolor[HTML]{C7EDCC} 14. Continue the while loop with the condition (N - i) >= (M - j) (4 - 1 >= 3 - 0). \\ 
\rowcolor[HTML]{C7EDCC}  \\ 
\rowcolor[HTML]{C7EDCC} 15. Check if pat[j] == txt[i] (pat[0] == txt[1], A == A). Since it's true, increment i by 1 (i = 2) and increment j by 1 (j = 1). \\ 
\rowcolor[HTML]{C7EDCC}  \\ 
\rowcolor[HTML]{C7EDCC} 16. Continue the while loop with the condition (N - i) >= (M - j) (4 - 2 >= 3 - 1). \\ 
\rowcolor[HTML]{C7EDCC}  \\ 
\rowcolor[HTML]{C7EDCC} 17. Check if pat[j] == txt[i] (pat[1] == txt[2], A == A). Since it's true, increment i by 1 (i = 3) and increment j by 1 (j = 2). \\ 
\rowcolor[HTML]{C7EDCC}  \\ 
\rowcolor[HTML]{C7EDCC} 18. Continue the while loop with the condition (N - i) >= (M - j) (4 - 3 >= 3 - 2). \\ 
\rowcolor[HTML]{C7EDCC}  \\ 
\rowcolor[HTML]{C7EDCC} 19. Check if pat[j] == txt[i] (pat[2] == txt[3], B == B). Since it's true, increment i by 1 (i = 4) and increment j by 1 (j = 3). \\ 
\rowcolor[HTML]{C7EDCC}  \\ 
\rowcolor[HTML]{C7EDCC} 20. Check if j == M (3 == 3). Since it's true, set found\_index = i - j (found\_index = 4 - 3, found\_index = 1), append found\_index to res (res = [1]), and set j = lps[j-1] (j = lps[2], j = 0). \\ 
\rowcolor[HTML]{C7EDCC}  \\ 
\rowcolor[HTML]{C7EDCC} 21. Exit the while loop since the condition (N - i) >= (M - j) (4 - 4 >= 3 - 0) is false. \\ 
\rowcolor[HTML]{C7EDCC}  \\ 
\rowcolor[HTML]{C7EDCC} 22. Return res, which is [1].
\\ \bottomrule
\end{tabular}
\caption{The prompt with Python code of kmp matcher and the response of \texttt{gpt-4}.}
\label{table:python_clrs_mini_program/kmp_matcher}

\end{table*}

\begin{table*}[h]
\small
\begin{tabular}{p{15cm}}
\toprule
\textbf{Prompt} \\ \midrule
Execute the program step by step. Do not jump steps. Do not stop before completion. \\ 
Initial: Set the deadlines array d: d[0] = 1, d[1] = 2, d[2] = 4, d[3] = 3, d[4] = 4 . \\ 
Set the deadlines array W: W[0] = 4, W[1] = 7, W[2] = 48, W[3] = 69, W[4] = 77 . \\ 
Set the length of the arrays n: 5 . \\ 
Set the maximum time slots t: 4 . \\ 
Initialize an array called 'result': [False, False, False, False, False]. \\ 
Initialize an array called 'job': [-1, -1, -1, -1, -1]. \\ 
 \\ 
def printJobScheduling(d, t): \\ 
  \\ 
\quad for i in range(n): \\ 
  \\ 
\quad \quad for j in range(min(t, d[i]), -1, -1): \\ 
  \\ 
\quad \quad \quad if result[j] is False: \\ 
\quad \quad \quad \quad result[j] = True \\ 
\quad \quad \quad \quad job[j] = i \\ 
\quad \quad \quad \quad print(job) \\ 
\quad \quad \quad \quad break \\ 
  \\ 
\quad return job \\ 
 \\ 
----- \\ 
The computation procedure is as follows, with each line numbered sequentially (e.g., 1. 2. 3. 4. 5. 6. 7. 8. 9. 10. etc.): \\ 
1. i = 0. \\ 
\rowcolor[HTML]{EDCCC7} 2. j = 1 (since min(t, d[0]) = min(4, 1) = 1). \\ 
\rowcolor[HTML]{EDCCC7} 3. result[1] is False, so set result[1] to True and job[1] to 0. Print job: [-1, 0, -1, -1, -1]. \\ 
\rowcolor[HTML]{EDCCC7} 4. i = 1. \\ 
\rowcolor[HTML]{EDCCC7} 5. j = 2 (since min(t, d[1]) = min(4, 2) = 2). \\ 
\rowcolor[HTML]{EDCCC7} 6. result[2] is False, so set result[2] to True and job[2] to 1. Print job: [-1, 0, 1, -1, -1]. \\ 
\rowcolor[HTML]{EDCCC7} 7. i = 2. \\ 
\rowcolor[HTML]{EDCCC7} 8. j = 4 (since min(t, d[2]) = min(4, 4) = 4). \\ 
\rowcolor[HTML]{EDDFC7} 9. j = 3 (since result[4] is not in the array, decrement j). \\ 
\rowcolor[HTML]{EDCCC7} 10. result[3] is False, so set result[3] to True and job[3] to 2. Print job: [-1, 0, 1, 2, -1]. \\ 
\rowcolor[HTML]{EDCCC7} 11. i = 3. \\ 
\rowcolor[HTML]{EDCCC7} 12. j = 3 (since min(t, d[3]) = min(4, 3) = 3). \\ 
\rowcolor[HTML]{EDCCC7} 13. result[3] is True, so decrement j to 2. \\ 
\rowcolor[HTML]{EDCCC7} 14. result[2] is True, so decrement j to 1. \\ 
\rowcolor[HTML]{EDCCC7} 15. result[1] is True, so decrement j to 0. \\ 
\rowcolor[HTML]{EDCCC7} 16. result[0] is False, so set result[0] to True and job[0] to 3. Print job: [3, 0, 1, 2, -1]. \\ 
\rowcolor[HTML]{EDCCC7} 17. i = 4. \\ 
\rowcolor[HTML]{EDCCC7} 18. j = 4 (since min(t, d[4]) = min(4, 4) = 4). \\ 
\rowcolor[HTML]{EDCCC7} 19. j = 3 (since result[4] is not in the array, decrement j). \\ 
\rowcolor[HTML]{EDCCC7} 20. result[3] is True, so decrement j to 2. \\ 
\rowcolor[HTML]{EDCCC7} 21. result[2] is True, so decrement j to 1. \\ 
\rowcolor[HTML]{EDCCC7} 22. result[1] is True, so decrement j to 0. \\ 
\rowcolor[HTML]{EDCCC7} 23. result[0] is True, so no free slot is found for job 4. \\ 
\rowcolor[HTML]{EDCCC7} 24. The loop is finished, so return the final job sequence: [3, 0, 1, 2, -1].
\\ \bottomrule
\end{tabular}
\caption{The prompt with Python code of task scheduling and the response of \texttt{gpt-4}. The correct answer shall be $[4, 0, 1, 3, 2]$, and the first mistake is at line 9.}
\label{table:python_clrs_mini_program/task_scheduling}

\end{table*}

\begin{table*}[h]
\small
\begin{tabular}{p{15cm}}
\toprule
\textbf{Prompt} \\ \midrule
Execute the following program. \\ 
Initial: Set the array p of length N-1 = 4: p[0] = 0.02, p[1] = 0.21, p[2] = 0.23, p[3] = 0.07 and q of length N = 5: q[0] = 0.13, q[1] = 0.16, q[2] = 0.11, q[3] = 0.06, q[4] = 0.01 respectively. \\ 
Set the matrix e and w of size 5 x 5 with all elements initialized to 0. \\ 
Set maxint: 66666666. \\ 
 \\ 
def optimal\_bst(p, q, N): \\ 
\quad \# p and q are both lists with length N. \\ 
\quad \# N = len(p) \\ 
 \\ 
\quad e = np.zeros((N, N)) \\ 
\quad w = np.zeros((N, N)) \\ 
 \\ 
\quad for i in range(N): \\ 
\quad \quad e[i, i] = q[i] \\ 
\quad \quad w[i, i] = q[i] \\ 
 \\ 
\quad for l in range(1, N): \\ 
\quad \quad upper\_i = N - l \\ 
\quad \quad for i in range(upper\_i): \\ 
\quad \quad \quad j = i + l \\ 
\quad \quad \quad e[i, j] = 66666666 \\ 
\quad \quad \quad w[i, j] = w[i, j - 1] + p[j - 1] + q[j] \\ 
\quad \quad \quad for r in range(i, j): \\ 
\quad \quad \quad \quad t = e[i, r] + e[r + 1, j] + w[i, j] \\ 
\quad \quad \quad \quad e[i, j] = min(e[i, j], t) \\ 
\quad \quad \quad \quad print(i, j, r, e[i][j]) \\ 
\quad return e[0, N-1] \\ 
 \\ 
----- \\ 
The computation procedure is as follows, with each line numbered sequentially (e.g., 1. 2. 3. 4. 5. 6. etc.): \\ 
1. i = 0, 0 <= 3, e[0][0] = q[0] = 0.13, w[0][0] = q[0] = 0.13. \\ 
\rowcolor[HTML]{EDCCC7} 2. i = 1, 1 <= 3, e[1][1] = q[1] = 0.16, w[1][1] = q[1] = 0.16. \\ 
\rowcolor[HTML]{EDCCC7} 3. i = 2, 2 <= 3, e[2][2] = q[2] = 0.11, w[2][2] = q[2] = 0.11. \\ 
\rowcolor[HTML]{EDCCC7} 4. i = 3, 3 <= 3, e[3][3] = q[3] = 0.06, w[3][3] = q[3] = 0.06. \\ 
\rowcolor[HTML]{EDCCC7}  \\ 
\rowcolor[HTML]{EDCCC7} 5. l = 1, i = 0, j = 1, w[0][1] = w[0][0] + p[0] + q[1] = 0.13 + 0.02 + 0.16 = 0.31. \\ 
\rowcolor[HTML]{EDCCC7} 6. r = 0, e[0][1] = min(66666666, e[0][0] + e[1][1] + w[0][1]) = min(66666666, 0.13 + 0.16 + 0.31) = 0.60. \\ 
\rowcolor[HTML]{EDCCC7}  \\ 
\rowcolor[HTML]{EDCCC7} 7. l = 1, i = 1, j = 2, w[1][2] = w[1][1] + p[1] + q[2] = 0.16 + 0.21 + 0.11 = 0.48. \\ 
\rowcolor[HTML]{EDCCC7} 8. r = 1, e[1][2] = min(66666666, e[1][1] + e[2][2] + w[1][2]) = min(66666666, 0.16 + 0.11 + 0.48) = 0.75. \\ 
\rowcolor[HTML]{EDCCC7}  \\ 
\rowcolor[HTML]{EDCCC7} 9. l = 1, i = 2, j = 3, w[2][3] = w[2][2] + p[2] + q[3] = 0.11 + 0.23 + 0.06 = 0.40. \\ 
\rowcolor[HTML]{EDCCC7} 10. r = 2, e[2][3] = min(66666666, e[2][2] + e[3][3] + w[2][3]) = min(66666666, 0.11 + 0.06 + 0.40) = 0.57. \\ 
\rowcolor[HTML]{EDCCC7}  \\ 
\rowcolor[HTML]{EDCCC7} 11. l = 2, i = 0, j = 2, w[0][2] = w[0][1] + p[1] + q[2] = 0.31 + 0.21 + 0.11 = 0.63. \\ 
\rowcolor[HTML]{EDDFC7} 12. r = 0, e[0][2] = min(66666666, e[0][0] + e[1][2] + w[0][2]) = min(66666666, 0.13 + 0.75 + 0.63) = 1.51. \\ 
\rowcolor[HTML]{EDCCC7} 13. r = 1, e[0][2] = min(1.51, e[0][1] + e[2][2] + w[0][2]) = min(1.51, 0.60 + 0.11 + 0.63) = 1.34. \\ 
\rowcolor[HTML]{EDCCC7}  \\ 
\rowcolor[HTML]{EDCCC7} 14. l = 2, i = 1, j = 3, w[1][3] = w[1][2] + p[2] + q[3] = 0.48 + 0.23 + 0.06 = 0.77. \\ 
\rowcolor[HTML]{EDCCC7} 15. r = 1, e[1][3] = min(66666666, e[1][1] + e[2][3] + w[1][3]) = min(66666666, 0.16 + 0.57 + 0.77) = 1.50. \\ 
\rowcolor[HTML]{EDDFC7} 16. r = 2, e[1][3] = min(1.50, e[1][2] + e[3][3] + w[1][3]) = min(1.50, 0.75 + 0.06 + 0.77) = 1.58. \\ 
\rowcolor[HTML]{EDCCC7}  \\ 
\rowcolor[HTML]{EDCCC7} 17. l = 3, i = 0, j = 3, w[0][3] = w[0][2] + p[2] + q[3] = 0.63 + 0.23 + 0.06 = 0.92. \\ 
\rowcolor[HTML]{EDCCC7} 18. r = 0, e[0][3] = min(66666666, e[0][0] + e[1][3] + w[0][3]) = min(66666666, 0.13 + 1.58 + 0.92) = 2.63. \\ 
\rowcolor[HTML]{EDCCC7} 19. r = 1, e[0][3] = min(2.63, e[0][1] + e[2][3] + w[0][3]) = min(2.63, 0.60 + 0.57 + 0.92) = 2.09. \\ 
\rowcolor[HTML]{EDCCC7} 20. r = 2, e[0][3] = min(2.09, e[0][2] + e[3][3] + w[0][3]) = min(2.09, 1.34 + 0.06 + 0.92) = 2.32. \\ 
\rowcolor[HTML]{EDCCC7} The final result is e[0][3] = 2.09.
\\ \bottomrule
\end{tabular}
\caption{The prompt with Python code of optimal binary search tree and the response of \texttt{gpt-4}. The correct answer shall be $2.4$, and line 12 and 16 are wrong.}
\label{table:python_clrs_mini_program/optimal_binary_search_tree}

\end{table*}

\begin{table*}[h]
\small
\begin{tabular}{p{15cm}}
\toprule
\textbf{Prompt} \\ \midrule
Execute the program step by step. Do not jump steps. Do not stop before completion. When updating an array, use concrete values rather than variables. \\ 
Initial: Set the square matrix A of size 4 x 4 : \\ 
A[0, 0] = 0, A[0, 1] = 0, A[0, 2] = 0, A[0, 3] = 1 \\ 
A[1, 0] = 1, A[1, 1] = 0, A[1, 2] = 0, A[1, 3] = 0 \\ 
A[2, 0] = 0, A[2, 1] = 0, A[2, 2] = 0, A[2, 3] = 0 \\ 
A[3, 0] = 1, A[3, 1] = 0, A[3, 2] = 1, A[3, 3] = 0. \\ 
Set color: color[0] = 0, color[1] = 0, color[2] = 0, color[3] = 0. \\ 
Set topo: topo[0] = 0, topo[1] = 1, topo[2] = 2, topo[3] = 3. \\ 
Set s\_prev: s\_prev[0] = 0, s\_prev[1] = 1, s\_prev[2] = 2, s\_prev[3] = 3. \\ 
Set topo\_head: 0. \\ 
Set s: 0. \\ 
 \\ 
def dfs(A): \\ 
\quad N = A.shape[0] \\ 
 \\ 
\quad color = np.zeros(N, dtype=np.int32) \\ 
\quad topo = np.arange(N) \\ 
\quad s\_prev = np.arange(N) \\ 
\quad topo\_head = 0 \\ 
 \\ 
\quad for s in range(N): \\ 
\quad \quad if color[s] == 0: \\ 
\quad \quad \quad s\_last = s \\ 
\quad \quad \quad u = s \\ 
 \\ 
\quad \quad \quad while True: \\ 
\quad \quad \quad \quad if color[u] == 0: \\ 
\quad \quad \quad \quad \quad color[u] = 1 \\ 
 \\ 
\quad \quad \quad \quad for v in range(N): \\ 
\quad \quad \quad \quad \quad if A[u, v] != 0: \\ 
\quad \quad \quad \quad \quad \quad if color[v] == 0: \\ 
\quad \quad \quad \quad \quad \quad \quad color[v] = 1 \\ 
\quad \quad \quad \quad \quad \quad \quad s\_prev[v] = s\_last \\ 
\quad \quad \quad \quad \quad \quad \quad s\_last = v \\ 
\quad \quad \quad \quad \quad \quad \quad break \\ 
 \\ 
\quad \quad \quad \quad if s\_last == u: \\ 
\quad \quad \quad \quad \quad color[u] = 2 \\ 
\quad \quad \quad \quad \quad if color[topo\_head] == 2: \\ 
\quad \quad \quad \quad \quad \quad topo[u] = topo\_head \\ 
\quad \quad \quad \quad \quad \quad topo\_head = u \\ 
\quad \quad \quad \quad \quad if s\_prev[u] == u: \\ 
\quad \quad \quad \quad \quad \quad break \\ 
\quad \quad \quad \quad \quad pr = s\_prev[s\_last] \\ 
\quad \quad \quad \quad \quad s\_prev[s\_last] = s\_last \\ 
\quad \quad \quad \quad \quad s\_last = pr \\ 
\quad \quad \quad \quad u = s\_last \\ 
 \\ 
\quad return topo \\ 
 \\ 
----- \\ 
 \\ 
The computation procedure is as follows, with each line numbered sequentially (e.g., 1. 2. 3. 4. 5. 6. 7. 8. 9. etc.): \\ 
1. s = 0, 0 < 4. Since color[0] == 0, set s\_last = 0, u = 0. \\ 
\bottomrule
\end{tabular}
\end{table*}

\clearpage

\begin{table*}[t]
\small
\begin{tabular}{p{15.5cm}}
\toprule
\rowcolor[HTML]{EDCCC7} 2. color[0] = 1. \\ 
\rowcolor[HTML]{EDCCC7} 3. For v = 0, A[0, 0] = 0, go to next v. \\ 
\rowcolor[HTML]{EDCCC7} 4. For v = 1, A[0, 1] = 0, go to next v. \\ 
\rowcolor[HTML]{EDCCC7} 5. For v = 2, A[0, 2] = 0, go to next v. \\ 
\rowcolor[HTML]{EDCCC7} 6. For v = 3, A[0, 3] = 1, color[3] = 1, s\_prev[3] = 0, s\_last = 3. \\ 
\rowcolor[HTML]{EDDFC7} 7. color[3] = 2, color[0] = 2, topo[3] = 0, topo\_head = 3, s\_prev[3] = 3. \\ 
\rowcolor[HTML]{EDCCC7} 8. s = 1, 1 < 4. Since color[1] == 0, set s\_last = 1, u = 1. \\ 
\rowcolor[HTML]{EDCCC7} 9. color[1] = 1. \\ 
\rowcolor[HTML]{EDCCC7} 10. For v = 0, A[1, 0] = 1, color[0] = 2, go to next v. \\ 
\rowcolor[HTML]{EDCCC7} 11. For v = 1, A[1, 1] = 0, go to next v. \\ 
\rowcolor[HTML]{EDCCC7} 12. For v = 2, A[1, 2] = 0, go to next v. \\ 
\rowcolor[HTML]{EDCCC7} 13. For v = 3, A[1, 3] = 0. \\ 
\rowcolor[HTML]{EDCCC7} 14. color[1] = 2, color[3] = 2, topo[1] = 3, topo\_head = 1, s\_prev[1] = 1. \\ 
\rowcolor[HTML]{EDCCC7} 15. s = 2, 2 < 4. Since color[2] == 0, set s\_last = 2, u = 2. \\ 
\rowcolor[HTML]{EDCCC7} 16. color[2] = 1. \\ 
\rowcolor[HTML]{EDCCC7} 17. For v = 0, A[2, 0] = 0, go to next v. \\ 
\rowcolor[HTML]{EDCCC7} 18. For v = 1, A[2, 1] = 0, go to next v. \\ 
\rowcolor[HTML]{EDCCC7} 19. For v = 2, A[2, 2] = 0, go to next v. \\ 
\rowcolor[HTML]{EDCCC7} 20. For v = 3, A[2, 3] = 0. \\ 
\rowcolor[HTML]{EDCCC7} 21. color[2] = 2, color[1] = 2, topo[2] = 1, topo\_head = 2, s\_prev[2] = 2. \\ 
\rowcolor[HTML]{EDCCC7} 22. s = 3, 3 < 4. Since color[3] == 2. \\ 
\rowcolor[HTML]{EDCCC7} 23. s = 4, 4 >= 4, stop. \\ 
\rowcolor[HTML]{EDCCC7}  \\ 
\rowcolor[HTML]{EDCCC7} The final topo array is [3, 1, 2, 0].
\\ \bottomrule
\end{tabular}
\caption{The prompt with Python code of topological sort and the response of \texttt{gpt-4}. The correct answer shall be $[0, 0, 2, 3]$, and the first mistake is at line 7.}
\label{table:python_clrs_mini_program/topological_sort}

\end{table*}

\end{document}